\documentclass[10pt,journal,compsoc]{IEEEtran}

\ifCLASSOPTIONcompsoc
 \usepackage[nocompress]{cite}
\else
 \usepackage{cite}
\fi
\usepackage{alphalph}
\renewcommand{\thetable}{\alphalph{\value{table}}}
\renewcommand{\thefigure}{\alphalph{\value{figure}}}

\usepackage{amsmath}
\usepackage{xspace}
\usepackage{subcaption}
\usepackage{bbm}
\usepackage{nicefrac}
\usepackage{tabularx}
\usepackage{multirow}
\usepackage{tikz}
\usepackage{pgfplots}
\usepackage{amsmath}
\usepackage{amsthm}
\usepackage{amsfonts}
\usepackage{xspace}
\usepackage{mathtools}
\usepackage{algorithm}
\usepackage{algorithmicx}
\usepackage{algpseudocode}
\usepackage{listings}
\usepackage{placeins}
\usepackage{url}
\usepackage{alphalph}

\DeclareFixedFont{\ttb}{T1}{txtt}{bx}{n}{12}
\DeclareFixedFont{\ttm}{T1}{txtt}{m}{n}{12}
\definecolor{deepblue}{rgb}{0,0,0.5}
\definecolor{deepred}{rgb}{0.6,0,0}
\definecolor{deepgreen}{rgb}{0,0.5,0}
\algrenewcommand\algorithmicindent{0.75em}%
\algtext*{EndWhile}
\algtext*{EndIf}
\algtext*{EndFor}
\algtext*{EndProcedure}
\usepackage{caption}
\usepackage{wrapfig}
\usepackage[breakable, theorems, skins]{tcolorbox}
\usepackage{xspace}
\usepackage{array}

\newcolumntype{L}[1]{>{\raggedright\let\newline\\\arraybackslash\hspace{0pt}}m{#1}}
\newcolumntype{C}[1]{>{\centering\let\newline\\\arraybackslash\hspace{0pt}}m{#1}}
\newcolumntype{R}[1]{>{\raggedleft\let\newline\\\arraybackslash\hspace{0pt}}m{#1}}

\def\Vhrulefill{\leavevmode\hrule height 0.7ex depth \dimexpr0.4pt-0.7ex with 1cm\kern0pt}

\usepackage{xcolor}
\definecolor{colorbrewer0}{RGB}{45,45,45}
\definecolor{colorbrewer1}{RGB}{228,26,28}
\definecolor{colorbrewer2}{RGB}{55,126,184}
\definecolor{colorbrewer3}{RGB}{77,175,74}
\definecolor{colorbrewer4}{RGB}{152,78,163}
\definecolor{colorbrewer5}{RGB}{255,127,0}
\definecolor{colorbrewer6}{RGB}{255,255,51}
\definecolor{colorbrewer7}{RGB}{166,86,40}
\definecolor{colorbrewer8}{RGB}{247,129,191}
\definecolor{colorbrewer9}{RGB}{153,153,153}
\definecolor{colorbrewer10}{RGB}{24,167,181}

\newtheorem{proposition}{Proposition}

\definecolor{gray}{RGB}{65, 65, 65}
\definecolor{darkred}{RGB}{255, 0, 0}
\definecolor{darkgreen}{RGB}{25, 100, 25}
\definecolor{darkmagenta}{RGB}{255, 0, 255}
\definecolor{darkblue}{RGB}{0, 0, 255}

\definecolor{colorbrewer0}{RGB}{45,45,45}
\definecolor{colorbrewer1}{RGB}{228,26,28}
\definecolor{colorbrewer2}{RGB}{55,126,184}
\definecolor{colorbrewer3}{RGB}{77,175,74}
\definecolor{colorbrewer4}{RGB}{152,78,163}
\definecolor{colorbrewer5}{RGB}{255,127,0}
\definecolor{colorbrewer6}{RGB}{255,255,51}
\definecolor{colorbrewer7}{RGB}{166,86,40}
\definecolor{colorbrewer8}{RGB}{247,129,191}
\definecolor{colorbrewer9}{RGB}{153,153,153}
\definecolor{colorbrewer10}{RGB}{24,167,181}

\makeatletter
\DeclareRobustCommand\onedot{\futurelet\@let@token\@onedot}
\def\@onedot{\ifx\@let@token.\else.\null\fi\xspace}

\def\eg{\emph{e.g}\onedot} 
\def\ie{\emph{i.e}\onedot} 
\def\cf{\emph{c.f}\onedot} 
\def\etc{\emph{etc}\onedot} 
\def\wrt{w.r.t\onedot} 

\makeatother

\newcommand{\figref}[1]{\Fig~\ref{#1}}
\newcommand{\secref}[1]{\Sec~\ref{#1}}
\newcommand{\appref}[1]{\App~\ref{#1}}

\renewcommand{\algref}[1]{\Alg~\ref{#1}}
\newcommand{\eqnref}[1]{\Eq~\eqref{#1}}
\newcommand{\tabref}[1]{\Tab~\ref{#1}}

\DeclareMathOperator*{\argmax}{argmax}

\makeatletter
\DeclareRobustCommand\onedot{\futurelet\@let@token\@onedot}
\def\@onedot{\ifx\@let@token.\else.\null\fi\xspace}
\def\eg{e.g\onedot} 
\def\ie{i.e\onedot} 
 
\def\cf{cf\onedot} 
\def\etc{etc\onedot}

\def\wrt{wrt\onedot}

\def\Fig{Fig\onedot} \def\Eq{Eq\onedot}
\def\Sec{Sec\onedot} \def\Alg{Alg\onedot}
\def\Tab{Tab\onedot} 
\def\App{App\onedot} 
\makeatother

\DeclareRobustCommand{\RTE}{%
    \ifmmode
    \text{RErr}
    \else
    RErr\xspace
    \fi
}
\DeclareRobustCommand{\TE}{%
    \ifmmode
    \text{Err}
    \else
    Err\xspace
    \fi
}
\DeclareRobustCommand{\wmax}{%
    \ifmmode
    w_{\text{max}}
    \else
    $w_{\text{max}}$\xspace
    \fi
}
\DeclareRobustCommand{\qmax}{%
    \ifmmode
    q_{\text{max}}
    \else
    $q_{\text{max}}$\xspace
    \fi
}
\DeclareRobustCommand{\qmin}{%
    \ifmmode
    q_{\text{min}}
    \else
    $q_{\text{min}}$\xspace
    \fi
}
\DeclareRobustCommand{\Vmin}{%
    \ifmmode
    V_{\text{min}}
    \else
    $V_{\text{min}}$\xspace
    \fi
}
\DeclareRobustCommand{\Vt}{%
    \ifmmode
    V_{\text{th}}
    \else
    $V_{\text{th}}$\xspace
    \fi
}
\DeclareRobustCommand{\biterror}{%
    \ifmmode
    \text{BErr}
    \else
    BErr\xspace
    \fi
}
\DeclareRobustCommand{\pfault}{%
    \ifmmode
    p_{\text{flt}}
    \else
    $p_{\text{{flt}}}$\xspace
    \fi
}
\DeclareRobustCommand{\perror}{%
    \ifmmode
    p_{\text{err}}
    \else
    $p_{\text{err}}$\xspace
    \fi
}

\def\Pr{\mathrm{P}}

\def\Exp{\mathbb{E}}

\newcommand{\Id}{\mathbbm{1}}

\def\min{\mathop{\rm min}\nolimits}
\def\max{\mathop{\rm max}\nolimits}

\def\maxop{\mathop{\rm max}\limits}

\newcommand{\MNIST}{MNIST\xspace}
 
\newcommand{\Cifar}{CIFAR\xspace}
\newcommand{\CifarT}{CIFAR10\xspace}
\newcommand{\CifarH}{CIFAR100\xspace}

\newcommand{\Normal}{\textsc{Normal}\xspace}
\newcommand{\Quant}{\textsc{RQuant}\xspace}
\newcommand{\Clipping}[1][]{\textsc{Clipping}\textsubscript{#1}\xspace}
\newcommand{\PLClipping}[1][]{\textsc{PLClipping}\textsubscript{#1}\xspace}
\newcommand{\Pattern}[1][]{\textsc{PattBET}\textsubscript{#1}\xspace}
\newcommand{\Random}[1][]{\textsc{RandBET}\textsubscript{#1}\xspace}
\newcommand{\PLRandom}[1][]{\textsc{PLRandBET}\textsubscript{#1}\xspace}
\newcommand{\LRandom}[1][]{\textsc{$L_0$RandBET}\textsubscript{#1}\xspace} 
\newcommand{\PLLRandom}[1][]{\textsc{PL$L_0$RandBET}\textsubscript{#1}\xspace} 

\newcommand{\Adv}[1][]{\textsc{AdvBET}\textsubscript{#1}\xspace}

\newcommand{\red}[1]{\noindent{\color{darkred}{#1}}}

\newcommand{\magenta}[1]{\noindent{\color{darkmagenta}{#1}}}
\newcommand{\blue}[1]{\noindent{\color{darkblue}{#1}}}

\newcommand{\revision}[1]{{\color{red}{#1}}}
\renewcommand{\revision}[1]{#1}
\ifCLASSINFOpdf
 \usepackage{graphicx}
\else
\fi
\begin{document}
%

\title{Random and Adversarial Bit Error Robustness:\\Energy-Efficient and Secure DNN Accelerators}

%
%
%
%

\author{David Stutz,
    Nandhini Chandramoorthy,
    Matthias Hein,
    and~Bernt Schiele,~\IEEEmembership{Fellow,~IEEE}%
\IEEEcompsocitemizethanks{ 
\IEEEcompsocthanksitem D. Stutz and B. Schiele are with the Max Planck Institute for Informatics, Saarland Informatics Campus, Germany.\protect\\
E-mail: \{david.stutz,schiele\}@mpi-inf.mpg.de
\IEEEcompsocthanksitem Nandhini Chandramoorthy is with the IBM T. J. Watson Research Center, Yorktown Heights, NY.
E-mail: nandhini.chandramoorthy@ibm.com
\IEEEcompsocthanksitem Matthias Hein is with the University of Tübingen, Germany.\protect\\
E-mail: matthias.hein@uni-tuebingen.de
}
}

%
%

\markboth{Journal of \LaTeX\ Class Files,~Vol.~14, No.~8, August~2015}%
{Stutz \MakeLowercase{\textit{et al.}}: Random and Adversarial Bit Error Robustness: Energy-Efficient and Secure DNN Accelerators}
%



\IEEEtitleabstractindextext{%
\begin{abstract}
    Deep neural network (DNN) accelerators
   	received considerable attention in recent years due to the potential to save energy compared to mainstream hardware.
   	Low-voltage operation of DNN accelerators allows to further reduce energy consumption, however, causes bit-level failures in the memory storing the quantized weights.
   	Furthermore, DNN accelerators are vulnerable to adversarial attacks on voltage controllers or individual bits.
   	In this paper, we show that a combination of \textbf{robust fixed-point quantization}, \textbf{weight clipping}, as well as \textbf{random bit error training (\Random)} or \textbf{adversarial bit error training (\Adv)} improves \textbf{robustness against random or adversarial bit errors in quantized DNN weights significantly}.
   	This leads not only to high energy savings for low-voltage operation \emph{as well as} low-precision quantization, but also improves security of DNN accelerators.
   	\revision{In contrast to related work,} our approach generalizes across operating voltages and accelerators \revision{and does not require hardware changes}.
   	\revision{Moreover, we present a novel adversarial bit error attack and are able to obtain} robustness against both targeted and untargeted bit-level attacks.
   	Without losing more than $0.8\%$/$2\%$ in test accuracy, we can reduce energy consumption on \CifarT by $20\%$/$30\%$ for $8$/$4$-bit quantization. Allowing up to $320$ adversarial bit errors, we reduce test error from above $90\%$ (chance level) to $26.22\%$.
\end{abstract}

\begin{IEEEkeywords}
DNN Accelerators, Bit Error Robustness, Adversarial Bit Errors, Robustness, DNN Quantization
\end{IEEEkeywords}}

\maketitle

\IEEEdisplaynontitleabstractindextext

%
\IEEEpeerreviewmaketitle

\ifCLASSOPTIONcompsoc
\IEEEraisesectionheading{\section{Introduction}\label{sec:introduction}}
\else
\section{Introduction}
\label{sec:introduction}
\fi

\IEEEPARstart{E}{energy}-efficiency is an important goal to lower carbon-dioxide emissions of deep neural network (DNN) driven applications and is a critical prerequisite to enable applications in edge computing.
\emph{DNN accelerators}, \ie, specialized hardware for inference, are used to reduce and limit energy consumption alongside cost and space compared to mainstream hardware, \eg, GPUs.
These accelerators generally feature on-chip SRAM used as scratchpads, \eg, to store DNN weights. Data access/movement
constitutes a dominant component of accelerator energy consumption~\cite{SzeIEEE2017}.
Besides reduced precision \cite{LinICML2016}, DNN accelerators \cite{ReagenISCA2016, KimDATE2018,ChandramoorthyHPCA2019} further lower memory supply voltage to increase energy efficiency since dynamic power varies quadratically with voltage. However, aggressive SRAM supply voltage scaling causes bit-level failures on account of process variation \cite{GanapathyDAC2017,GuoJSSC2009} with direct impact on the stored DNN weights. The rate $p$ of these errors increases exponentially with lowered voltage, causing devastating drops in DNN accuracy. 
Thus, DNN accelerators are also vulnerable to maliciously reducing voltage \cite{TangUSENIX2017} or adversarially inducing individual bit errors \cite{KimISCA2014,MurdockSP2020}. In this paper, we aim to enable very low-voltage operation of DNN accelerators by developing DNNs robust to \textit{random bit errors} in their (quantized) weights.
This also improves security against manipulation of voltage settings~\cite{TangUSENIX2017}. Furthermore, we address robustness against a limited number of \textit{adversarial bit errors}, similar to \cite{RakinICCV2019,RakinARXIV2020,HeCVPR2020}.
In general, DNN robustness to bit errors is a desirable goal to maintain safe operation and should become a standard performance metric in low-power DNN design.

\figref{fig:introduction} shows the average bit error rates of SRAM arrays as supply voltage is scaled below \Vmin, \ie, the measured lowest voltage at which there are no bit errors. Voltage (x-axis) and energy ({\color{colorbrewer1}red}, right y-axis) are normalized \wrt \Vmin and the energy per access at \Vmin, respectively. DNNs robust to a bit error rate ({\color{colorbrewer2}blue}, left y-axis) of, \eg, $p = 1\%$ allow to reduce SRAM energy by roughly $30\%$.
To improve DNN robustness to the induced \textit{random bit errors}, we first consider the impact of fixed-point quantization on robustness. While prior work \cite{MurthyARXIV2019,MerollaARXIV2016,SungARXIV2015} studies robustness \emph{to} quantization, the impact of random bit errors \emph{in} quantized weights has not been considered so far.
\revision{However, bit errors are significantly more severe than quantization errors, as confirmed by substantially worse signal-to-quantization-noise-ratios \cite{LinICML2016}.}
We find that the choice of quantization scheme and its implementation details has tremendous impact on robustness, even though accuracy is not affected.
\revision{Using these insights allows us to use a particularly \textbf{robust quantization} scheme \Quant in \figref{fig:contributions} ({\color{colorbrewer1}red}).}
Additionally, independent of the quantization scheme, we \revision{use} aggressive \textbf{weight clipping} during training. This acts as an explicit regularizer leading to spread out weight distributions, improving
robustness significantly,
\Clipping in \figref{fig:contributions} ({\color{colorbrewer2}blue}). This is in contrast to, \eg, \cite{ZhuangCVPR2018,SungARXIV2015} ignoring weight outliers to reduce quantization range, with sole focus of improving accuracy.

\begin{figure}[t]
    \centering
    \hspace*{0.4cm}
    \begin{subfigure}{0.475\textwidth}
    	\includegraphics[height=4.75cm]{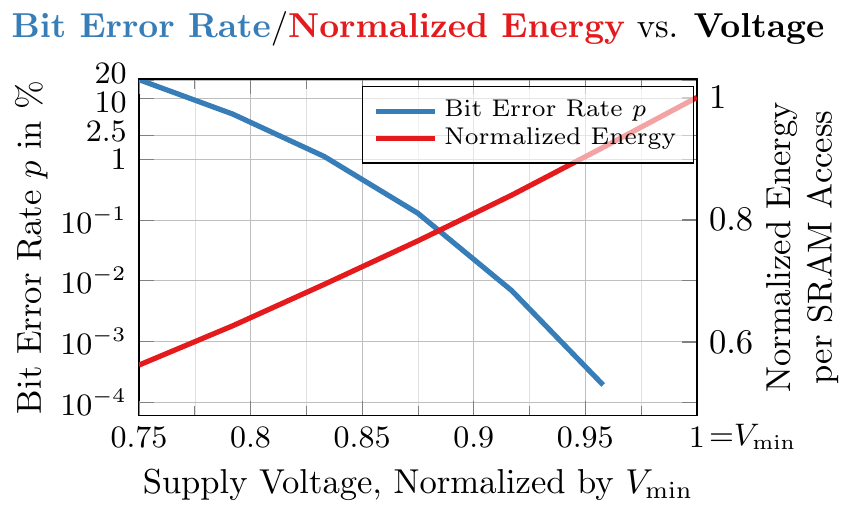}
    \end{subfigure}
    \vspace*{-8px}
    \caption{
    \textbf{Energy and Low-Voltage Operation.} Average bit error rate $p$ ({\color{colorbrewer2}blue}, left y-axis) from $32$ 14nm SRAM arrays of size $512{\times}64$ from \cite{ChandramoorthyHPCA2019} and energy ({\color{colorbrewer1}red}, right y-axis) vs. voltage (x-axis). Voltage is normalized by \Vmin, the minimal measured voltage for error-free operation, and the energy per SRAM access at \Vmin. SRAM accesses have significant impact on the DNN accelerator's energy \cite{ChenISCA2016}. Reducing voltage leads to exponentially increasing bit error rates.
    }
    \label{fig:introduction}
    \vspace*{-0.2cm}
\end{figure}

Conventional error mitigation strategies or circuit techniques are not applicable to mitigate larger rates of bit errors or incur an significant energy/space overhead.
For example, common error correcting codes (ECCs such as SECDED), cannot correct \emph{multiple} bit errors per word (containing multiple DNN weights). However, for $p = 1\%$, the probability of two or more random bit errors in a $64$-bit word is $13.5\%$. Furthermore, an adversary may intentionally target multiple bits per word.
Considering low-voltage induced random bit errors, error detection via redundancy \cite{ReagenISCA2016} or supply voltage boosting \cite{ChandramoorthyHPCA2019} allow error-free operation at the cost of additional energy or space. Therefore, \cite{KimDATE2018} and \cite{KoppulaMICRO2019} propose a co-design approach of training DNNs on \emph{profiled} bit errors (\ie, post-silicon characterization) from SRAM or DRAM, respectively.
These approaches work as long as the spatial bit error patterns can be assumed fixed for a \emph{fixed} accelerator \emph{and} voltage. 
However, the random nature of variation-induced bit errors requires profiling to be carried out for each voltage, memory array and individual chip in order to obtain the corresponding bit error patterns. This makes training DNNs on profiled bit error patterns an expensive process. We demonstrate that the obtained DNNs do \emph{not} generalize across voltages or to unseen bit error patterns, \eg, from other memory arrays, and propose \textbf{random bit error training (\Random)}, in combination with weight clipping and robust quantization, to obtain robustness against completely random bit error patterns, see  \figref{fig:contributions} ({\color{colorbrewer4}violet}). Thereby, it generalizes across chips \emph{and} voltages, without any profiling, hardware-specific data mapping or other circuit-level mitigation strategies. Finally, in contrast to \cite{KimDATE2018,KoppulaMICRO2019}, we also consider bit errors in activations and inputs, as both are temporally stored on the chip's memory and thus subject to bit errors.

Besides low-voltage induced random bit errors, \cite{KimISCA2014,MurdockSP2020} demonstrate the possibility of \emph{adversarially} flipping specific bits. The bit flip attack (BFA) of \cite{RakinICCV2019}, an untargeted search-based attack on (quantized) DNN weights, demonstrates that such attacks can easily degrade DNN accuracy with few bit flips. \cite{HeCVPR2020} proposes a binarization strategy to ``defend'' against BFA. However, the approach was shown to be ineffective
shortly after considering a targeted version of BFA \cite{RakinARXIV2020}, leaving the problem unaddressed. We propose a novel attack based on projected gradient descent, inspired by recent work on adversarial examples \cite{MadryICLR2018,StutzICML2020}. We demonstrate that our attack is both more effective and more efficient. Moreover, in contrast to BFA, our adversarial bit attack enables \textbf{adversarial bit error training} (\textbf{\Adv}).
As shown in \figref{fig:contributions} (right), \Adv ({\color{magenta}magenta}) improves robustness against adversarial bit errors considerably, outperforming \Clipping ({\color{colorbrewer2}blue}) and \Random ({\color{colorbrewer4}violet}) which, surprisingly, provide very strong baselines. As a result, we are able to obtain robustness to both random and adversarial bit errors, enabling energy-efficient \emph{and} secure DNN accelerators.

\begin{figure}[t]
	\centering
	\hspace*{-0.45cm}
	\begin{subfigure}[t]{0.36\textwidth}
		\vspace*{0px}
		\centering
		\bfseries \CifarT: Random Bit Error Robustness
		\vskip 2px
		
        \includegraphics[height=4cm]{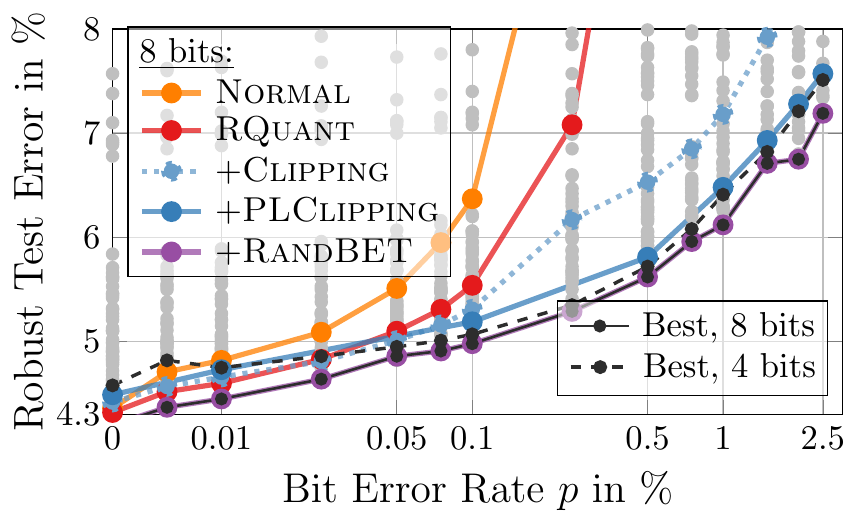}
    \end{subfigure}
    \begin{subfigure}[t]{0.005\textwidth}
    	\vspace*{0px}
    	\centering
    
    	{\color{gray}\rule{0.5px}{4.25cm}}
    \end{subfigure}
    \begin{subfigure}[t]{0.12\textwidth}
    	\vspace*{0px}
    	\centering
    	\bfseries Adversarial
    	\vskip 8px
    	
    	\small
    	\begin{tabular}{|@{\hskip 4px}c@{\hskip 4px}|@{\hskip 4px}c@{\hskip 4px}|}
    		\hline
    		& \bfseries\RTE $\downarrow$\\[1px]
    		\hline
    		\multirow{8}{*}{\rotatebox{90}{\bfseries 320 \emph{adv.} bit errors\hspace*{0.25cm}}} &\footnotesize\Quant:\\[1px]
    		&\bfseries{\color{colorbrewer1}91.18}\\[1px]
    		&\footnotesize\Clipping:\\[1px]
    		&\bfseries{\color{colorbrewer2}60.76}\\[1px]
    		&\footnotesize\Random:\\[1px]
    		&\bfseries{\color{colorbrewer4}33.86}\\[1px]
    		&\footnotesize\Adv:\\[1px]
    		&\bfseries{\color{magenta}26.22}\\[1px]
    		\hline
    	\end{tabular}
    \end{subfigure}
    \vspace*{-8px}
    \caption{
    \textbf{Robustness to Random Bit Errors.} \emph{Left:} Robust test error \RTE after injecting \emph{random} bit errors (lower is better $\downarrow$, y-axis) plotted against bit error rate $p$ (x-axis). For $8$ bit, robust quantization (\Quant, {\color{colorbrewer1}red}), additionally weight clipping (\Clipping, {\color{colorbrewer2}dotted blue}) or \emph{per-layer} weight clipping (\PLClipping, {\color{colorbrewer2}solid blue}) and finally adding random bit error training (\Random, {\color{colorbrewer4}violet}) robustness improves significantly. 
    Robustness to higher bit error rates allows more energy efficient operation, \cf \figref{fig:introduction}.
    The Pareto optimal frontier is shown for $8$ bit (black solid) and $4$ bit (dashed) quantization. \emph{Right:} \RTE against up to $320$ \emph{adversarial} bit errors, showing that \Clipping combined with \Random or \Adv also allow secure operation. 
    }
    \label{fig:contributions}
    \vspace*{-0.2cm}
\end{figure} 
\begin{figure*}
	\centering
	\begin{tikzpicture}
		\node[anchor=north west] at (0,0){\includegraphics[width=4cm]{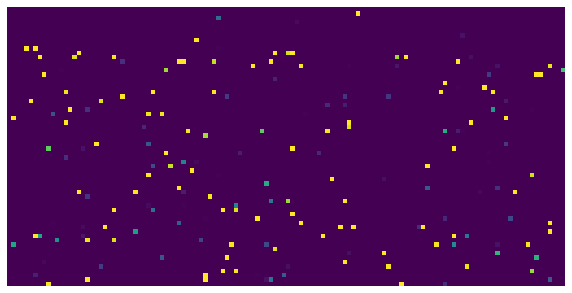}};
		\node[anchor=north west] at (4,0){\includegraphics[width=4cm]{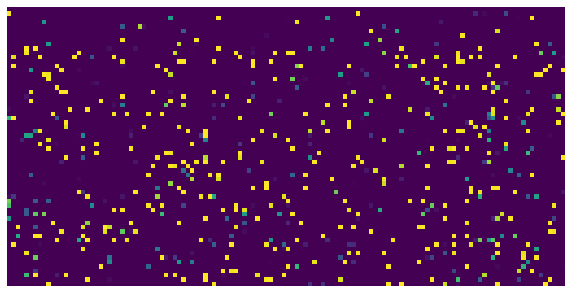}};
		\node[anchor=north west] at (8,0){\includegraphics[width=4cm]{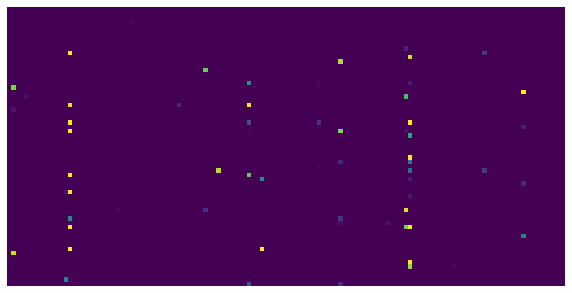}};
		\node[anchor=north west] at (12,0){\includegraphics[width=4cm]{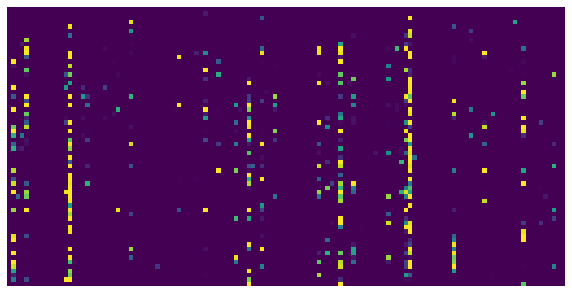}};
		
		\node[anchor=south east,fill opacity=0.75,fill=white] at (4, -2){$p{\approx}0.86\%$};
		\node[anchor=south east,fill opacity=0.75,fill=white] at (8, -2){$p{\approx}2.75\%$};
		\node[anchor=south east,fill opacity=0.75,fill=white] at (12, -2){$p{\approx}0.14\%$};
		\node[anchor=south east,fill opacity=0.75,fill=white] at (16, -2){$p{\approx}1.08\%$};
		
		\draw[white!30!black,-, line width=0.75pt] (8.125,-2.4) -- (8.125,0.4);
		\draw[thick,->] (0.05,0.05) -- (1,0.05);
		\draw[thick,->] (-0.05,-0.05) -- (-0.05,-1);
		\node[anchor=south] at (0.75,0.05){128 columns};
		\node[rotate=90,anchor=south] at (-0.05,-0.6){64 rows};
		
		\node[anchor=south] at (4,-0.15) {\bfseries Chip 1};
		\node[anchor=south] at (12,-0.15) {\bfseries Chip 2};
		
		\draw[thick,->] (2.5,-2.8) -- (4.5,-2.8) node[anchor=west] {bit error rate $p$ increases};
		\draw[thick,->] (2.5,-2.4) -- (4.5,-2.4) node[anchor=west] {voltage decreases};
		
		\draw[thick,-] (9.75, -2.2) -- (9.75,-2.4) -- (10.75,-2.4);
		\node[anchor=west] at (10.75, -2.4){bit errors subset of};
		\draw[thick,->] (13.75,-2.4) -- (14.75,-2.4) -- (14.75, -2.2);
	\end{tikzpicture}
	\vspace*{-8px}
	\caption{\textbf{Exemplary SRAM Bit Error Patterns.} Measured bit errors from two chips with on-chip SRAM (left and right), showing bit flip probability for a segment of size $64 \times 128$ bits: {\color{yellow!75!black}yellow} indicates a bit flip probability of one, {\color{violet}violet} indicates zero probability. We show measurements corresponding to two supply voltages.
	With lower voltage, bit error rate increases. Also, the bit errors for higher voltage (= lower bit error rate) are a subset of those for lower voltage (= higher rate), \cf \secref{sec:errors}. Our error model randomly distributes bit errors across space. However, as example, we also show SRAM chip 2 which has a different spatial distribution with bit errors distributed along columns. We aim to obtain robustness across different memory arrays, voltages \emph{and} allowing arbitrary DNN weight to memory mappings.}
	\label{fig:errors}
	\vspace*{-0.1cm} 
\end{figure*}
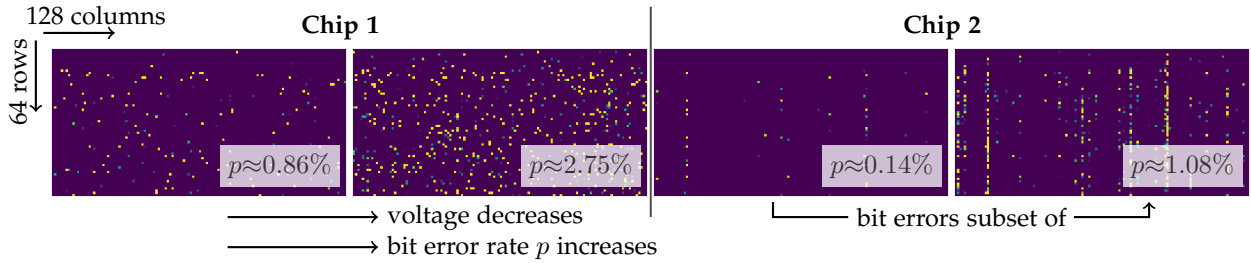

\textbf{Contributions:}
We combine \textbf{robust quantization \Quant} with \textbf{weight clipping} and \textbf{random bit error training (\Random)} or \textbf{adversarial bit error training (\Adv)} in order to obtain high robustness against low-voltage induced, \textit{random bit errors} or maliciously crafted, \textit{adversarial bit errors}. We consider fixed-point quantization schemes in terms of robustness \emph{and} accuracy, instead of \emph{solely} focusing on accuracy as related work. Furthermore, we show that aggressive weight clipping, as regularization during training, is an effective strategy to improve robustness through redundancy. In contrast to \cite{KimDATE2018,KoppulaMICRO2019}, the robustness obtained through \Random generalizes across chips \emph{and} voltages, as evaluated on profiled SRAM bit error patterns from \cite{ChandramoorthyHPCA2019}. In contrast to \cite{RakinICCV2019,RakinARXIV2020}, our (untargeted or targeted) adversarial bit error attack is based on gradient descent, improving effectiveness and efficiency, and our \Adv improves robustness against targeted and untargeted attacks, outperforming the recently broken binarization approach of \cite{HeCVPR2020}.
Finally, we discuss the involved trade-offs regarding robustness (against random or adversarial bit errors) and accuracy and make our code publicly available to facilitate research in this practically important area of DNN robustness. \figref{fig:contributions} (left) highlights key results for \Random on \CifarT: with $8$/$4$ bit quantization and an increase in test error of less than $0.8\%$/$2\%$, roughly $20\%$/$30\%$ energy savings are possible -- on top of energy savings from using low-precision quantization. Similarly, \Adv, \cf \figref{fig:contributions} (right), obtains $26.22\%$ (robust) test error against up to $320$ adversarial bit errors in the weights.

A preliminary version of this work has been accepted at MLSys'21 \cite{StutzMLSYS2021}. We further improved robustness against low-voltage induced random bit errors using \emph{per-layer} weight clipping (\PLClipping, {\color{colorbrewer2}solid blue} in \figref{fig:contributions}). Furthermore, we consider random bit errors in activations and inputs which are also (temporally) stored on the SRAM and thus subject to bit errors. In both cases, the negative impact can be limited using approaches similar to \Random. Beyond our work on low-voltage induced, random bit errors in \cite{StutzMLSYS2021}, we tackle the more challenging task of \textit{adversarial bit errors}, \cf \figref{fig:contributions} (right). We devise a flexible adversarial bit error attack based on projected gradient descent that can be used in an untargeted or targeted setting and is more effective and efficient compared to related work \cite{RakinICCV2019}. Moreover, our attack enables adversarial bit error training (\Adv, {\color{magenta}magenta}), which improves robustness significantly. We also show that \Random ({\color{colorbrewer4}violet}) provides surprisingly good robustness against adversarial bit errors, thereby enabling both energy-efficient and secure DNN accelerators.
\section{Related Work}
\label{sec:related-work}

\textbf{Quantization:} DNN Quantization \cite{GuoARXIV2018b} is usually motivated by faster
DNN inference, \eg, through fixed-point quantization and arithmetic \cite{ShinICASSP2017,LinICML2016,LiNIPS2017}, and energy savings. To avoid reduced accuracy, quantization is considered during training \cite{JacobCVPR2018,KrishnamoorthiARXIV2018,HouICLR2018} instead of post-training or with fine-tuning \cite{BannerNIPS2019,nvrt,nervana}, enabling low-bit quantization such as binary DNNs \cite{RastegariECCV2016,CourbariauxNIPS2015}. Some works also consider quantizing activations \cite{RastegariECCV2016,ChoiARXIV2018,HubaraJMLR2017} or gradients \cite{SeideINTERSPEECH2014,AlistarhARXIV2016,ZhouARXIV2016}.
\revision{In contrast to \cite{ZhouARXIV2016,LiCVPR2019,JacobCVPR2018}, we quantize \emph{all} weights, including batch normalization parameters and biases, instead of ``folding'' batch normalization into the preceding convolution.}

\revision{\textbf{Quantization Errors:}
Several works \cite{MurthyARXIV2019,MerollaARXIV2016,SungARXIV2015,AlizadehICLR2020} study the robustness of DNNs to \emph{quantization errors}.
To improve performance of quantized models, \cite{MishchenkoICMLA2019,StockICLR2021,HouICLR2018} explicitly integrate such quantization errors into training.
While this is implicitly already done implicitly in quantization-aware training, \cite{MishchenkoICMLA2019} additionally performs on-device arithmetic during training.
Moreover, \cite{DongBMVC2017,StockICLR2021} apply (multiple) quantization schemes to random layers during training to improve gradient flow.
In concurrent work, \cite{BaskinTCS2021} replaces the straight-through estimator in quantization-aware training by differentially injecting uniform noise.
However, we found that quantization errors are significantly less severe than bit errors after quantization.
Unfortunately, the robustness of quantization methods against bit errors has not been studied, despite our findings that quantization impacts bit error robustness significantly.}

\revision{\textbf{Clipping in Quantization:}
Works such as \cite{ZhuangCVPR2018,SungARXIV2015,ParkISCA2018} clip weight outliers to reduce quantization error of inliers, improving accuracy.
Similar, \cite{ChoiMLSYS2019}, learns clipping for quantized activations during training.
In contrast, we consider \emph{weight clipping} independent of quantization \emph{as regularization during training} which spreads out the weight distribution and improves robustness to bit errors.
This is similar to weight clipping used for training generative adversarial networks \cite{GulrajaniNIPS2017,ArjovskyICML2017}.
However, robustness is not explored.}

\begin{figure*}[t]
	\centering
	\begin{tikzpicture}
		\node[anchor=north west] at (-0.25,0){\includegraphics[height=3cm]{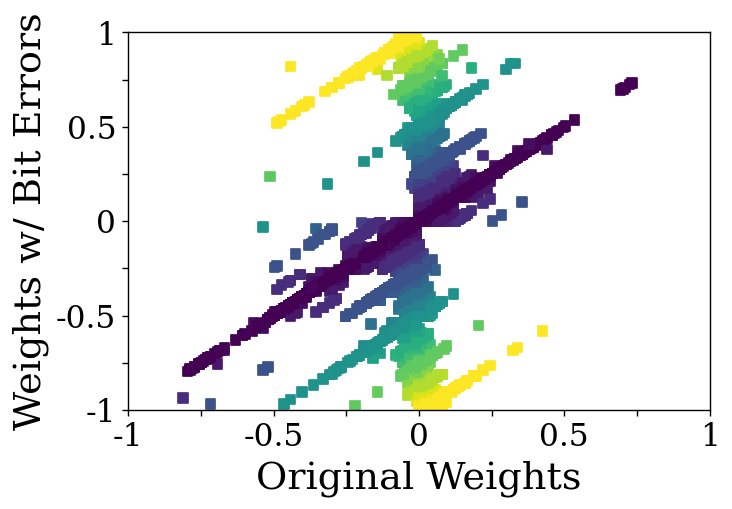}};
		\node[anchor=north west] at (4,0){\includegraphics[height=3cm,trim=1cm 0 0 0,clip]{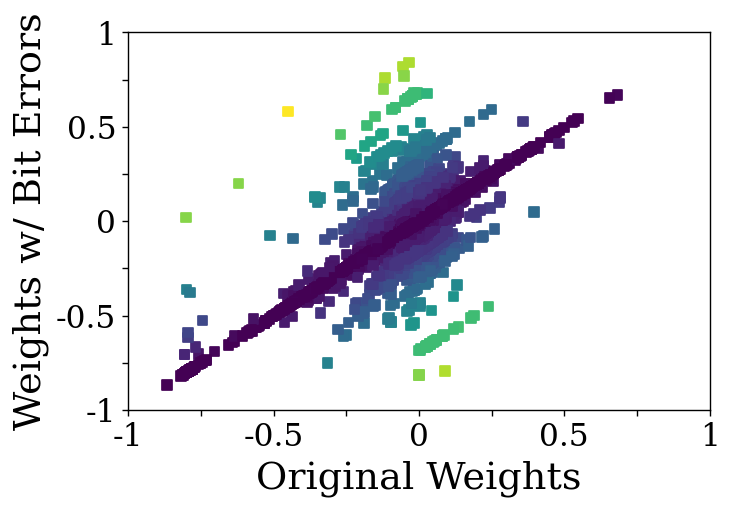}};
		\node[anchor=north west] at (8,0){\includegraphics[height=3cm,trim=1cm 0 0 0,clip]{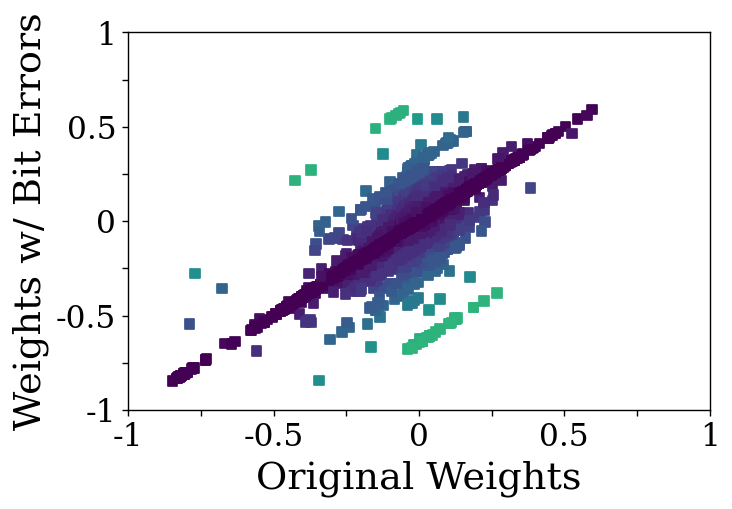}};
		\node[anchor=north west] at (12,0){\includegraphics[height=3cm,trim=1cm 0 0 0,clip]{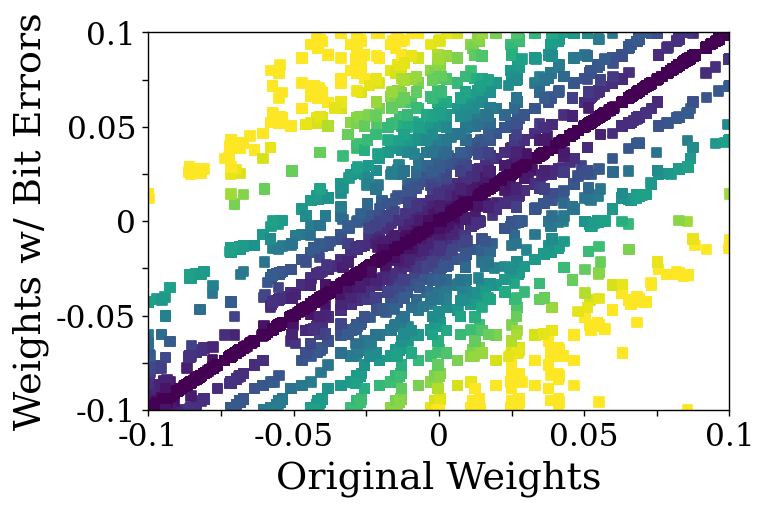}};
		
		\draw[-,white!50!black] (12.125, 0.25) -- (12.125, -3.25);
		\node[anchor=south,xshift=0.25cm,yshift=-0.25cm] at (2, 0){\textbf{Global}, $\qmax=1$, $m = 8$};
		\node[anchor=south,xshift=0.25cm,yshift=-0.25cm] at (6, 0){\textbf{Per-Layer} (=\Normal)};
		\node[anchor=south,xshift=0.25cm,yshift=-0.25cm] at (10, 0){\textbf{{\color{red}+}Asymmetric}};
		\node[anchor=south,xshift=0.25cm,yshift=-0.25cm] at (14, 0){{\color{red}+}\textbf{\Clipping[$0.1$]}, $m = 4$};
	\end{tikzpicture} 
	\vspace*{-8px}
	\caption{\textbf{\revision{Impact of} Random Bit Errors.} Original weights (x-axis) plotted against perturbed weights with bit errors (y-axis), for different fixed-point quantization schemes with $m = 8$ bit (left) and $p=2.5\%$. We also show the $m = 4$ bit case with \Clipping at $\wmax = 0.1$, \cf \secref{subsec:robustness-clipping}. Color indicates absolute error: from zero ({\color{violet}violet}) to the maximal possible error ({\color{yellow!75!black}yellow}) of $1$ (left) and $0.1$ (right). Asymmetric per-layer quantization reduces the impact of bit errors compared to the symmetric per-layer/global quantization. Clipping reduces absolute error, but the errors \emph{relative} to $\wmax$ increase. \revision{As discussed in \secref{subsec:robustness-quantization}, bit errors are substantially more severe than quantization errors.}}
	\label{fig:quantization}
	\vspace*{-0.1cm}
\end{figure*}
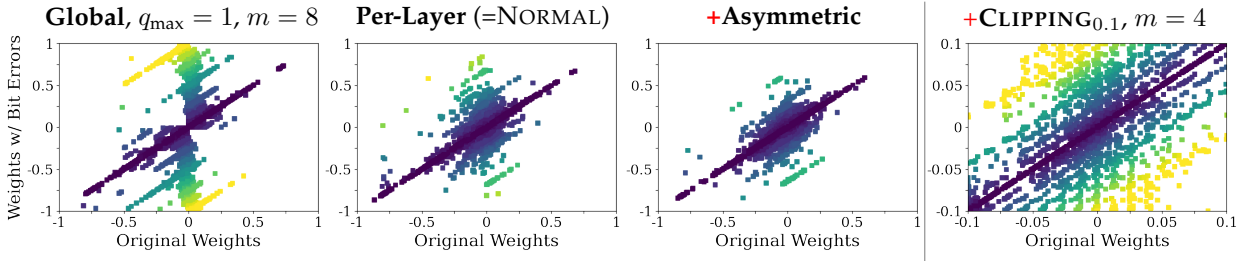

\textbf{Low-Voltage, Random Bit Errors in DNN Accelerators:} Recent work \cite{GanapathyDAC2017,GanapathyHPCA2019} demonstrates that bit flips in SRAMs increase exponentially when reducing voltage below \Vmin. The authors of~\cite{ChandramoorthyHPCA2019} study the impact of bit flips in different layers of DNNs, showing severe accuracy degradation. Similar observations hold for DRAM \cite{ChangPOMACS2017}. To prevent accuracy drops at low voltages, \cite{ReagenISCA2016} combines SRAM fault detection with logic to set faulty data reads to zero. \cite{ChandramoorthyHPCA2019} uses supply voltage boosting for SRAMs to ensure error-free, low-voltage operation, while \cite{SrinivasanDATE2016} proposes storing critical bits in specifically robust SRAM cells. However, such methods incur power and area overhead. Thus, \cite{KimDATE2018} and \cite{KoppulaMICRO2019} propose co-design approaches combining training on profiled SRAM/DRAM bit errors with hardware mitigation strategies and clever weight to memory mapping. Besides low-voltage operation for energy efficiency, recent work \cite{TangUSENIX2017} shows that an attacker can reduce voltage maliciously. In contrast to \cite{KimDATE2018,KoppulaMICRO2019}, our \emph{random bit error training (\Random)} obtains robustness that generalizes across chips and voltages without expensive chip-specific profiling or hardware mitigation strategies. Furthermore, \cite{KimDATE2018,KoppulaMICRO2019} do not address the role of quantization, and we demonstrate that these approaches can benefit from our weight clipping, as well. We show that energy savings from low-voltage operation and low-precision \cite{ParkISCA2018} can be combined. Finally, in contrast to existing work \cite{KimDATE2018,KoppulaMICRO2019}, we also study low-voltage induced bit errors in DNN activations and inputs.

\textbf{Adversarial Bit Errors in DNN Accelerators:} Works such as \cite{KimISCA2014,MurdockSP2020} demonstrate software-based approaches to induce few, but targeted, bit flips in DRAM. The impact of such attacks on (quantized) DNN weights has recently been studied in \cite{RakinICCV2019}: The proposed bit flip attack (BFA) is a search-based strategy to find as few bit errors as possible such that accuracy reduces to chance level. However, the binarization approach of \cite{HeCVPR2020}, improving robustness against \emph{untargeted} BFA, has been shown to be ineffective against a \emph{targeted} version of BFA \cite{RakinARXIV2020}. Moreover, the authors of \cite{HeCVPR2020} conclude that training on adversarial bit errors is \emph{not} a promising defense. In contrast, we propose a more effective and efficient, gradient-based adversarial bit error attack and demonstrate that \textit{adversarial bit error training (\Adv)} using our attack improves robustness against both untargeted \emph{and} targeted attacks, including BFA. \Adv is similar in spirit to training on adversarial examples which received considerable attention recently \cite{MadryICLR2018,ZhangICML2019,StutzICML2020,WuARXIV2020c}.

\textbf{Weight Robustness:} Only few works consider weight robustness: \cite{WengAAAI2020} certify the robustness of weights with respect to $L_\infty$ perturbations and \cite{CheneyARXIV2017} study Gaussian noise on weights. \cite{RakinICCV2019,HeCVPR2020} consider identifying and (adversarially) flipping few vulnerable bits in quantized weights.
Fault tolerance, in contrast, describes structural changes such as removed units, and is rooted in early work such as \cite{NetiTNN1992,Chiu1994}.
Finally, \cite{JiCCS2018,DumfordARXIV2018} explicitly manipulate weights in order to integrate backdoors. We study robustness against \emph{random bit errors}, which exhibit a quite special noise pattern compared to $L_\infty$ or Gaussian noise, \cf \figref{fig:quantization}.
\section{Bit Errors in Quantized DNN Weights}
\label{sec:errors}

In the following, we introduce the bit error models considered in this paper: random bit errors (\secref{subsec:random-errors}), induced through low-voltage operation of accelerator memory, and adversarial bit errors (\secref{subsec:adversarial-errors}), maliciously crafted and injected by an adversary to degrade DNN accuracy.

\textbf{Notation:} Let $f(x; w)$ be a DNN taking an example $x \in [0, 1]^D$, \eg, an image, and weights $w \in \mathbb{R}^W$ as input. The DNN is trained by minimizing the cross-entropy loss $\mathcal{L}$ on a training set $\{(x_n, y_n)\}_{n = 1}^N$ consisting of examples $x_n$ and corresponding labels $y_n \in \{1, \ldots, K\}$, $K$ denoting the number of classes. We assume a weight $w_i \in [\qmin, \qmax]$, \ie, within the \textit{quantization range}, to be quantized using a function $Q$. As we will detail in \secref{subsec:robustness-quantization}, $Q$ maps floating-point values to $m$-bit (signed or unsigned) integers. With $v_i = Q(w_i)$, we denote the integer corresponding to the quantized value of $w_i$, \ie, $v_i$ is the bit representation of $w_i$ after quantization represented as integer. Finally, $d_H(v, v')$ denotes the bit-level Hamming distance between the integers $v$ and $v'$.

\subsection{Low-Voltage Induced Random Bit Errors}
\label{subsec:random-errors}

We assume the quantized DNN weights to be stored  on multiple memory banks, \eg, SRAM in the case of on-chip scratchpads or DRAM for off-chip memory. As shown in \cite{GanapathyDAC2017,KimDATE2018,ChandramoorthyHPCA2019}, the probability of memory bit cell failures increases exponentially as operating voltage is scaled below $\Vmin$, \ie, the minimal voltage required for reliable operation, see \figref{fig:introduction}. This is done intentionally to reduce energy consumption \cite{ChandramoorthyHPCA2019,KimDATE2018,KoppulaMICRO2019} or adversarially by an attacker \cite{TangUSENIX2017}. Process variation during fabrication causes a variation in the vulnerability of individual bit cells. As shown in \figref{fig:errors} (left), for a specific memory array, bit cell failures are typically approximately random and independent of each other \cite{GanapathyDAC2017} even so chips showing patterns with stronger dependencies are possible, \cf \figref{fig:errors} (right). Nevertheless, there is generally an ``inherited'' distribution of bit cell failures across voltages: as described in \cite{GanapathyHPCA2019}, if a bit error occurred at a given voltage, it is likely to occur at lower voltages, as made explicit in \figref{fig:errors}. However, across different SRAM arrays in a chip or different chips, the patterns or spatial distributions of bit errors is usually different and can be assumed random \cite{ChandramoorthyHPCA2019}. Throughout the paper, we use the following bit error model:

\textbf{Random Bit Error Model:}
\textit{The probability of a bit error is $p$ (in \%) for all weight values and bits. For a fixed memory array, bit errors are persistent across supply voltages, \ie, bit errors at probability $p'{\leq}p$ also occur at probability $p$. A bit error flips the currently stored bit. Random bit error injection is denoted $\biterror_p$.}

\begin{algorithm}[t]
\caption{\textbf{Adversarial Bit Errors.} We maximize cross-entropy loss using projected gradient ascent while ensuring that at most $\epsilon$ bits are flipped. Line \ref{line:attack-update} may include backtracking and Line \ref{line:attack-delta} may include momentum or gradient normalization. \red{Quantized weights} in \red{red}; \blue{de-quantized weights} in \blue{blue}; and \magenta{floating-point operations} in \magenta{magenta}. For coherence with \algref{alg:training}, \textsc{AdvBitErrors} takes the quantized weights $\red{v} = Q(\magenta{w})$ as input.}
\label{alg:attack}
\begin{algorithmic}[1]
\small
\Procedure{AdvBitErrors}{\red{$v$}, $\epsilon$}
    \State {\color{darkgreen}\# perturb quantized weights by flipping at most $\epsilon$ bits:}
    \State initialize {\color{red}$\tilde{v}^{(0)}$} subject to $d_H(\red{\tilde{v}^{(0)}}, \red{v}){\,\leq\,}\epsilon$, $d_H(\red{\tilde{v}^{(0)}_i}, \red{v_i}){\,\leq\,}1$
    \State ${\color{blue}\tilde{w}_q^{(0)}} = Q^{-1}({\color{red}\tilde{v}^{(0)}})$ {\color{darkgreen}\# de-quantize perturbed weights}
    \State $\magenta{\tilde{w}^{(0)}} = \blue{\tilde{w}_q^{(0)}}$ {\color{darkgreen}\# floating-point weights to acc. updates}
	\For{$t = 0,\ldots,T - 1$}
	    \State {\color{darkgreen}\# fixed batch $\{(x_b, y_b)\}_{b=1}^B$}
    	\State {\color{darkgreen}\# forward${\,+\,}$backward pass w/ de-quantized weights:}
    	\State $\Delta^{(t)} = \nabla_w \sum_{b = 1}^B \mathcal{L}(f(x_b; \blue{\tilde{w}_q^{(t)}}), y_b)$\label{line:attack-delta}
    	\State $\magenta{\tilde{w}^{(t + 1)}} = \magenta{\tilde{w}^{(t)}} + \gamma\Delta^{(t)}$ \label{line:attack-update} {\color{darkgreen}\# update w/o quantization:}
    	\State $\red{\tilde{v}^{(t + 1)}} = Q(\magenta{\tilde{w}^{(t + 1)}})$ {\color{darkgreen}\# quantization for projection}
    	\State project onto $d_H(\red{\tilde{v}^{(t + 1)}}, \red{v}){\,\leq\,}\epsilon$, $d_H(\red{\tilde{v}^{(t + 1)}_i}, \red{v_i}){\,\leq\,}1$
    	\State $\blue{\tilde{w}_q^{(t + 1)}} = Q^{-1}(\red{\tilde{v}^{(t + 1)}})$ {\color{darkgreen}\# de-quantization}
	\EndFor
	\State \textbf{return} $\blue{\tilde{w}_q^{(T)}}$ {\color{darkgreen}\# de-quantized weights after projection}
\EndProcedure
\end{algorithmic}
\end{algorithm}

This error model realistically captures the nature of low-voltage induced bit errors, from both SRAM and DRAM as confirmed in \cite{ChandramoorthyHPCA2019,KimDATE2018,KoppulaMICRO2019}. However, our approach in \secref{sec:robustness} is model-agnostic: the error model can be refined if extensive memory characterization results are available for individual chips. For example, faulty bit cells with $1$-to-$0$ or $0$-to-$1$ flips might not be equally likely. Similarly, as in \cite{KoppulaMICRO2019}, bit errors might be biased towards alignment along rows or columns of the memory array. The latter case is illustrated in \figref{fig:errors} (right). However, estimating these specifics requires testing infrastructure and detailed characterization of individual chips. 
More importantly, it introduces the risk of overfitting to few specific memories/chips. 
Furthermore, we demonstrate that the robustness obtained using our uniform error model generalizes to bit error distributions with strong spatial biases as in \figref{fig:errors} (right).

We assume the quantized weights to be mapped linearly to the memory. This is the most direct approach and, in contrast to \cite{KoppulaMICRO2019}, does not require knowledge of the exact spatial distribution of bit errors. This also means that we do not map particularly vulnerable weights to more reliable memory cells, and therefore no changes to the hardware or the application are required. Thus, in practice, for $W$ weights and $m$ bits per weight value, we sample uniformly $u \sim U(0, 1)^{W \times m}$. Then, the $j$-th bit in the quantized weight $v_i = Q(w_i)$ is flipped iff $u_{ij} \leq p$.
Our model assumes that the flipped bits at lower probability $p' \leq p$ are a subset of the flipped bits at probability $p$ and that bit flips to $1$ and $0$ are equally likely. The noise pattern of random bit errors is illustrated in \figref{fig:quantization}: 
for example, a bit flip in the most-significant bit (MSB) of the signed integer $v_i$ results in a change of half of the quantized range (also \cf \secref{subsec:robustness-quantization}).

In the case of on-chip SRAM, inputs and activations will also be subject to low-voltage induced bit errors. This is because the SRAM memory banks are used as scratchpads to temporally store intermediate computations such as inputs and activations. As described in detail in \secref{subsec:experiments-activations-inputs}, inputs are subject to random bit errors \emph{once} before being fed to the DNN. Activations, \ie, the result of intermediate layers of the DNN, are subject to random bit errors multiple times throughout a forward pass. This is modeled by (independently) injecting random bit errors in the activations after each ``block'' consisting of convolution, normalization and ReLU layers. This assumes that activations are temporally stored on the SRAM scratchpads after each such block. In practice, the data flow of a DNN accelerator is manually tailored to the DNN architecture \emph{and} chip design, which is also why energy estimation for DNN accelerators is very difficult \cite{WuICCAD2019,YangACSSC2017}. Furthermore, normalization schemes (group \cite{WuECCV2018} or batch normalization \cite{IoffeICML2015}) and ReLU activations can be ``folded into'' the preceding convolutional layer \cite{LiCVPR2019,JacobCVPR2018}. Thus, considering the activations to go through temporal storage on the SRAM after each block is a realistic approximation of the actual data flow.

\begin{algorithm}[t]
\caption{\textbf{Random Bit Error Training (\Random).} The forward passes are performed using \blue{de-quantized weights (blue)}. Perturbed weights are obtained by injecting bit errors in the \red{quantized weights (in red)}. The update, averaging gradients from both forward passes, is performed in \magenta{floating-point (magenta)}. Also see \figref{fig:flowchart}.}
\label{alg:training}
\begin{algorithmic}[1]
\small
\Procedure{RandBET}{$p$}
    \State initialize \magenta{$w^{(0)}$}
	\For{$t = 0, \ldots, T - 1$}
    	\State sample batch $\{(x_b, y_b)\}_{b = 1}^B$
    	\State $\magenta{w^{(t)}} = \min(\wmax, \max(-\wmax, \magenta{w^{(t)}}))$ {\color{darkgreen}\# clipping}
    	\State $\red{v^{(t)}} = Q(\magenta{w^{(t)}})$ {\color{darkgreen}\# quantization}
        \State $\blue{w_q^{(t)}} = Q^{-1}(\red{v^{(t)}})$ {\color{darkgreen}\# de-quantization}
        \State {\color{darkgreen}\# clean forward and backward pass:}
        \State $\Delta^{(t)} = \nabla_w \sum_{b = 1}^B \mathcal{L}(f(x_b; \blue{w_q^{(t)}}), y_b)$
        \State {\color{darkgreen}\# \emph{perturbed} forward and backward pass:}
        \State $\blue{\tilde{w}_q^{(t)}}{\hskip 1px=\hskip 1px}Q^{-1}(\biterror_p(\red{v^{(t)}}))$ (or $\textsc{AdvBitErrors}(\red{v^{(t)}}, \epsilon)$)\label{line:attack}
        \State $\tilde{\Delta}^{(t)} = \nabla_w \sum_{b = 1}^B \mathcal{L}(f(x_b; \blue{\tilde{w}_q^{(t)}}), y_b)$
        \State {\color{darkgreen}\# average gradients and weight update:}
        \State $\magenta{w^{(t + 1)}} = \magenta{w^{(t)}} - \gamma(\Delta^{(t)} + \tilde{\Delta}^{(t)})$
    \EndFor
    \State \textbf{return} $\blue{w_q^{(T)}} = Q^{-1}(Q(\magenta{w^{(T)}}))$
\EndProcedure
\end{algorithmic}
\end{algorithm}

\subsection{Adversarial Bit Errors}
\label{subsec:adversarial-errors}

Following recent attacks on memory \cite{KimISCA2014,BreierCCS2018,MurdockSP2020,RakinICCV2019} and complementing our work on random bit errors \cite{StutzMLSYS2021}, we also consider adversarial bit errors. We constrain the number of induced bit errors by $\epsilon$, similar to the $L_p$-constrained adversarial inputs \cite{SzegedyICLR2014}.
Furthermore, we consider only one bit flip per weight value to simplify the projection onto the discrete constraint set.
Then, given knowledge of memory layout and addressing schemes, an adversary can use, \eg, RowHammer \cite{KimISCA2014}, in order to flip as many of the adversarially selected bits. Note that, in practice, not all of these bits will be vulnerable to an end-to-end RowHammer attack on memory, which we do not focus on. However, from a robustness viewpoint, it makes sense to consider a slightly stronger threat model than actually realistic. 
Overall, our white-box threat model is defined as follows:

\begin{figure*}
    \vspace*{-0.15cm}
    \centering
    \includegraphics[width=0.85\textwidth]{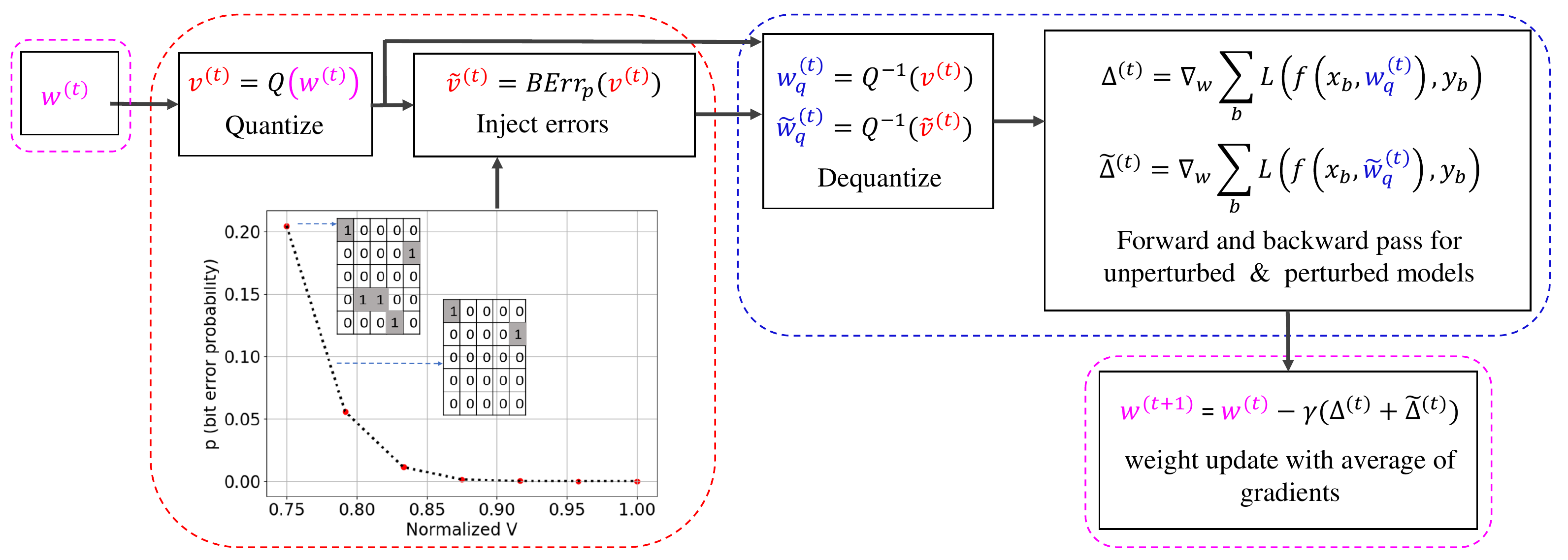}
    \vspace*{-6px}
    \caption{\textbf{Random Bit Error Training (\Random).} We illustrate the data-flow for \Random as in \algref{alg:training}. Here, $\biterror_p$ injects random bit errors in the \red{quantized weights} $\red{v^{(t)}} = Q(\magenta{w^{(t)}})$, resulting in $\red{\tilde{v}^{(t)}}$, while the forward pass is performed on the \blue{de-quantized perturbed weights} $\blue{\tilde{w}_q^{(t)}} = Q^{-1}(\red{\tilde{v}^{(t)}})$, \ie, fixed-point arithmetic is not emulated. The weight update during training is not affected by bit errors and computed in \magenta{floating point}.}
    \label{fig:flowchart}
    \vspace*{-0.1cm}
\end{figure*}

\textbf{Adversarial Bit Error Model:}
\textit{An adversary can flip up to $\epsilon$ bits, at most one bit per (quantized) weight, to reduce accuracy and has full access to the DNN, its weights and gradients.}

Note that we do \emph{not} consider adversarial bit errors in inputs or activations.
\revision{We also emphasize that this assumes a white-box settings where the adversary can not only access the DNNs weights, but also knows about the used quantization scheme.}
Following the projected gradient ascent approach of \cite{MadryICLR2018} and letting $d_H$ be the (bit-level) Hamming distance, we intend to maximize cross-entropy loss $\mathcal{L}$ on a mini-batch $\{(x_b, y_b)\}_{b = 1}^B$ of examples as \emph{untargeted} attack:
\begin{align}
	\begin{split}
	    &\max_{\tilde{v}} \sum_{b = 1}^B \mathcal{L}(f(x_b; Q^{-1}(\tilde{v})), y_b)\\
	    \text{s.t.}&\quad d_H(\tilde{v}, v) \leq \epsilon,\quad d_H(\tilde{v}_i, v_i) \leq 1\label{eq:attack}
    \end{split}
\end{align}
Note that $y_b$ are the ground truth labels. We also consider a targeted version of \eqnref{eq:attack}, similar to \cite{RakinARXIV2020}, where we minimize the cross-entropy loss between predictions and an arbitrary but fixed target label: $\min_{\tilde{v}} \sum_{b = 1}^B \mathcal{L}(f(x_b; Q^{-1}(\tilde{v})), y_t)$ where $y_t$ is the same target label across all examples~$x_b$.
As made explicit in \eqnref{eq:attack}, we work on bit-level, \ie, optimize over the two's complement signed integer representation $\tilde{v}_i \in \{-2^{m - 1}-1,\ldots,2^{m - 1} - 1\}$ corresponding to the underlying bits of the perturbed weights $\tilde{w} = Q(\tilde{v})$. 
We will adversarially inject bit errors based on the gradient of \eqnref{eq:attack} and perform a projection onto the Hamming constraints $d_H(\tilde{v}, v) \leq \epsilon$ and $d_H(\tilde{v}_i, v_i) \leq 1$ with respect to the quantized, clean weights $v = Q(w)$. This means that we maximize \eqnref{eq:attack} through projected gradient ascent where the forward and backward pass are performed in floating point:
\begin{align}
	\begin{split}
	    &\tilde{w}^{(t + 1)} = \tilde{w}^{(t)} + \gamma \Delta^{(t)}\quad\text{with}\\
	    &\Delta^{(t)} = \sum_{b = 1}^B \nabla_w \mathcal{L}(f(x_b; \tilde{w}_q^{(t)}), y_b),\text{ }\tilde{w}_q^{(t)} = Q^{-1}(Q(\tilde{w}^{(t)}))\label{eq:attack-iterates}
    \end{split}
\end{align}
followed by the projection of $\tilde{v}^{(t + 1)} = Q(\tilde{w}^{(t + 1)})$ onto the (bit-level) Hamming constraints of \eqnref{eq:attack}. Here, $\gamma$ is the step size. The updates are performed in floating point, while the forward pass is performed using the de-quantized weights $\tilde{w}_q^{(t)}$. The perturbed weights $\tilde{w}^{(0)} = Q^{-1}(\tilde{v}^{(0)})$ are initialized by uniformly picking $k \in [0, \epsilon]$ bits to be flipped in $v = Q(w)$ in order to obtain $\tilde{v}^{(0)}$.
Our adversarial bit attack is summarized in pseudocode in \algref{alg:attack}.

The Hamming-projection is similar to the $L_0$ projection used for adversarial inputs, \eg, in \cite{CroceICCV2019}. Dropping the superscript $t$ for brevity, in each iteration, we solve the following projection problem:
\begin{align}
    \begin{split}
	    &\min_{\tilde{v}'} \| Q^{-1}(\tilde{v}) - Q^{-1}(\tilde{v}')\|_2^2\\
	    \text{s.t.}&\quad d_H(v_i, \tilde{v}'_i) \leq 1,\quad d_H(v, \tilde{v}') \leq \epsilon
    \end{split}
    \label{eq:projection}
\end{align}
where $\tilde{v} = Q(\tilde{w})$ are the quantized, perturbed weights after \eqnref{eq:attack-iterates} and $v = Q(w)$ are the quantized, clean weights. We optimize over $\tilde{v}'$ which will be the perturbed weights after the projection, \ie, as close as possible to $\tilde{v}$ while fulfilling the constraints above.
This can be solved in two steps as the objective and the constraint set are separable: The first step involves keeping only the top-$\epsilon$ changed values, \ie, the top-$\epsilon$ weights with the largest difference $|w_i - \tilde{w}_i|$. The second step
can be solved by keeping only the most significant bit changed in $\tilde{v}_i$ compared to $v_i$ as detailed in our supplementary material.
The optimization problem in \eqnref{eq:attack} is challenging due to the projection onto the non-convex set of Hamming constraints. We adopt best practices from computing adversarial inputs: normalizing the gradient \cite{CroceICCV2019} (per-layer using the $L_\infty$ norm) and momentum \cite{DongCVPR2018}.

\section{Robustness Against Bit Errors}
\label{sec:robustness}

We address robustness against random and/or adversarial bit errors in three steps: First, we analyze the impact of fixed-point quantization schemes on bit error robustness. This has been neglected both in prior work on low-voltage DNN accelerators \cite{KimDATE2018,KoppulaMICRO2019} and in work on quantization robustness \cite{MurthyARXIV2019,MerollaARXIV2016,SungARXIV2015}. This yields our \textbf{robust quantization} (\secref{subsec:robustness-quantization}). On top, we propose aggressive \textbf{weight clipping} as regularization during training (\secref{subsec:robustness-clipping}).
Weight clipping enforces a more uniformly distributed, \ie, redundant, weight distribution, improving robustness. We show that this is due to minimizing the cross-entropy loss, enforcing large logit differences.
Finally, in addition to robust quantization and weight clipping, we perform \textbf{random bit error training (\Random)} (\secref{subsec:robustness-training}) or \textbf{adversarial bit error training (\Adv)} (\secref{subsec:adversarial-robustness-training}). For \Random, in contrast to the fixed bit error patterns in \cite{KimDATE2018,KoppulaMICRO2019}, we train on completely \emph{random} bit errors and, thus, generalize across chips and voltages. Regarding \Adv, we train on \emph{adversarial} bit errors, computed as outlined in \secref{subsec:adversarial-errors}. Generalization of bit error robustness is measured using \emph{robust test error (\RTE)}, the test error after injecting bit errors (lower is more robust).

\subsection{Robust Fixed-Point Quantization}
\label{subsec:robustness-quantization}

We consider quantization-aware training \cite{JacobCVPR2018,KrishnamoorthiARXIV2018} using a generic, deterministic fixed-point quantization scheme commonly used in DNN accelerators \cite{ChandramoorthyHPCA2019}. However, we focus on the impact of quantization schemes on robustness against random bit errors, mostly neglected so far \cite{MurthyARXIV2019,MerollaARXIV2016,SungARXIV2015}. We find that quantization affects robustness significantly, even if accuracy is largely unaffected.

\textbf{Fixed-Point Quantization:} Quantization determines how weights are represented in memory, \eg, on SRAM. In a \emph{fixed-point quantization} scheme, $m$ bits allow representing $2^m$ distinct values. 
A weight $w_i \in [-\qmax, \qmax]$
is represented by a \textit{signed} $m$-bit integer $v_i = Q(w_i)$ corresponding to the underlying bits. Here, $[-\qmax, \qmax]$ is the \emph{symmetric} quantization range and signed integers use two's complement representation. Then, $Q: [-\qmax, \qmax] \mapsto \{-2^{m - 1} - 1, \ldots, 2^{m - 1} - 1\}$ is defined as 
\begin{align}
    Q(w_i) = \left\lfloor \frac{w_i}{\Delta}\right\rfloor,\text{  }
    Q^{-1}(v_i) = \Delta v_i,\text{  }
    \Delta = \frac{\qmax}{2^{m - 1} - 1}.
    \label{eq:quantization}
\end{align}
This quantization is symmetric around zero and zero is represented exactly.
By default, we only quantize weights, not activations or gradients. However, in contrast to related work \cite{ZhouARXIV2016,McDonnellICLR2018,LiCVPR2019,JacobCVPR2018}, we quantize \emph{all} layers, including biases and batch normalization parameters \cite{IoffeICML2015} (commonly ``folded'' into preceding convolutional layers).
Flipping the most significant bit (MSB, \ie, sign bit) leads to an absolute error of half the quantization range, \ie, $\qmax$ ({\color{yellow!75!black!}yellow} in \figref{fig:quantization}).
Flipping the least significant bit (LSB) incurs an error of $\Delta$. Thus, the impact of bit errors ``scales with'' $\qmax$.

\textbf{Global and Per-Layer Quantization:} $\qmax$ can be chosen to accommodate all weights, \ie, $\qmax = \max_i |w_i|$. This is called \emph{global} quantization. However, it has become standard to apply quantization \textit{per-layer} allowing to adapt $\qmax$ to each layer. As in PyTorch \cite{PaszkeNIPSWORK2017}, we consider weights and biases of each layer separately. By reducing the quantization range for each layer individually, the errors incurred by bit flips are automatically minimized, \cf \figref{fig:quantization}. The
\textbf{per-layer, symmetric quantization is our default reference}, referred to as \Normal. However, it turns out that it is further beneficial to consider arbitrary quantization ranges $[\qmin, \qmax]$ (allowing $\qmin > 0$). In practice, we
first map $[\qmin, \qmax]$ to $[-1,1]$ and then quantize $[-1,1]$ using \eqnref{eq:quantization}.
Overall, per-layer asymmetric quantization has the finest granularity, \ie, lowest $\Delta$ and approximation error. Nevertheless, it is not the most robust
quantization.

\textbf{Robust Quantization:} \eqnref{eq:quantization} does \emph{not} provide optimal robustness against bit errors. First, the floor operation $\lfloor \nicefrac{w_i}{\Delta}\rfloor$ is commonly implemented as float-to-integer conversion. Using proper rounding $\lceil\nicefrac{w_i}{\Delta}\rfloor$ instead has negligible impact on accuracy, even though quantization error improves slightly. In stark contrast, bit error robustness is improved considerably. During training, DNNs can compensate the differences in approximation errors, even for small precision $m < 8$. However, at test time,
rounding decreases the impact of bit errors considerably. Second, \eqnref{eq:quantization} uses signed integers for symmetric quantization. For asymmetric quantization, with arbitrary $[\qmin, \qmax]$, we found quantization into \emph{unsigned} integers to improve robustness, \ie, $Q : [\qmin, \qmax] \mapsto \{{\color{red}0}, \ldots, {\color{red}2^m - 1}\}$. This is implemented using an additive term of $2^{m-1}-1$ in \eqnref{eq:quantization}. While accuracy is not affected, the effect of bit errors in the sign bit changes: in symmetric quantization, the sign bit mirrors the sign of the weight value. For asymmetric quantization, an unsigned integer representation is more meaningful. Overall, our \textbf{robust fixed-point quantization (\Quant)}  uses per-layer, asymmetric quantization into unsigned integers with rounding.
These seemingly \emph{small differences} have little to no impact on accuracy but tremendous impact on robustness against bit errors, see \secref{subsec:experiments-quantization}.

\revision{\textbf{Quantization Errors vs. Bit Errors:} Tackling the impact of \emph{bit errors} on quantized weights, as shown in \figref{fig:quantization}, is very different from considering \emph{quantization errors}. The latter are essentially approximation errors and are, for the above fixed-point quantization scheme, deterministic and fixed once the model is trained. Low-voltage induced bit errors, in contrast, are entirely random at test time. Moreover, bit errors can induce absolute changes significantly above commonly observed quantization errors, even for low $m$. Measured as signal-to-noise-ratio, following \cite{LinICML2016}, we obtain roughly $33.6$dB for a $8$-bit quantized model using our robust quantization. In contrast, $p = 1\%$ random bit errors result in a \emph{negative} ratio of $-0.65$dB, indicating that the bit errors actually dominate the  ``signal'' (\ie, weights). See our supplementary material for a thorough discussion.}

\subsection{Training with Weight Clipping as Regularization}
\label{subsec:robustness-clipping}

Simple \textbf{weight clipping} refers to constraining the weights to $[-\wmax, \wmax]$ \emph{during training}, where $\wmax$ is a hyper-parameter. Generally, $\wmax$ is independent of the quantization range(s) which always adapt(s) to the weight range(s) at hand. However, weight clipping limits the maximum possible quantization range (\cf \secref{subsec:robustness-quantization}), \ie, $\qmax \leq \wmax$.
It might seem that weight clipping with small $\wmax$ automatically improves robustness against bit errors as the absolute errors are reduced. However, the \emph{relative} errors are not influenced by re-scaling. As the DNN's decision is usually invariant to re-scaling, reducing the scale of the weights does not impact robustness. In fact, the mean relative error of the weights in \figref{fig:quantization} (right) increased with clipping at $\wmax{=} 0.1$. Thus, weight clipping does \emph{not} ``trivially'' improve robustness by reducing the scale of weights. Nevertheless, we found that weight clipping actually improves robustness considerably on top of our robust quantization.

\begin{figure}[t]
	\centering
	\vspace*{-0.1cm}
	\hspace*{-0.3cm}
	\begin{subfigure}{0.24\textwidth}
		\vspace*{3px}
		
		\includegraphics[height=1.35cm]{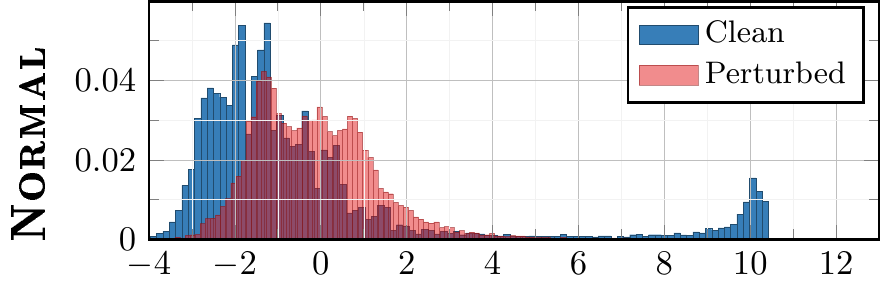}
	\end{subfigure}
	\begin{subfigure}{0.12\textwidth}
		\vspace*{0px}
		
		\includegraphics[height=1.425cm]{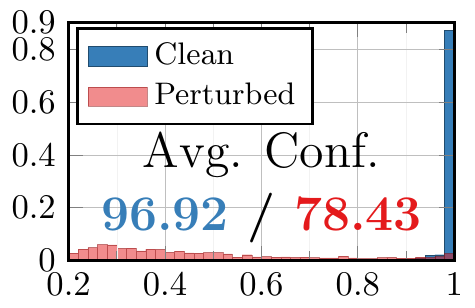}
	\end{subfigure}
	\begin{subfigure}{0.12\textwidth}
		\vspace*{0px}
		
		\includegraphics[height=1.425cm]{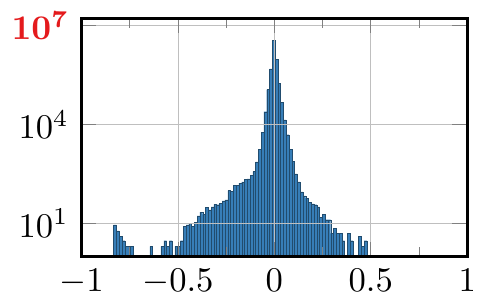}
	\end{subfigure}
	
	\hspace*{-0.4cm}
	\begin{subfigure}{0.24\textwidth}
		\vspace*{3px}
		
		\includegraphics[height=1.35cm]{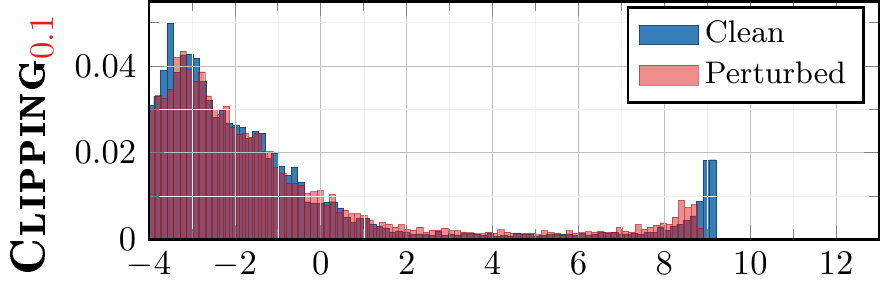}
	\end{subfigure}
	\begin{subfigure}{0.12\textwidth}
		\vspace*{0px}
		
		\includegraphics[height=1.425cm]{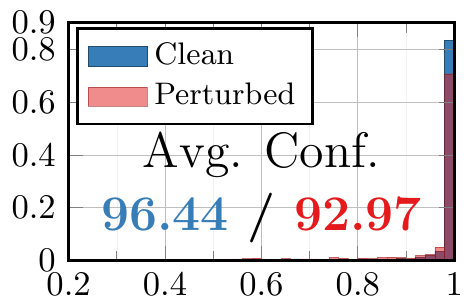}
	\end{subfigure}
	\begin{subfigure}{0.12\textwidth}
		\vspace*{3px}
		
		\includegraphics[height=1.425cm]{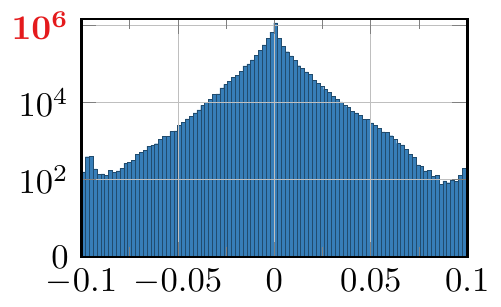}
	\end{subfigure}
	
	\hspace*{-0.3cm}
	{\color{black!25!white}\rule{0.5\textwidth}{0.5px}}
	
	\hspace*{-0.4cm}
	\begin{subfigure}{0.24\textwidth}
		\vspace*{0px}
		
		\includegraphics[height=1.825cm]{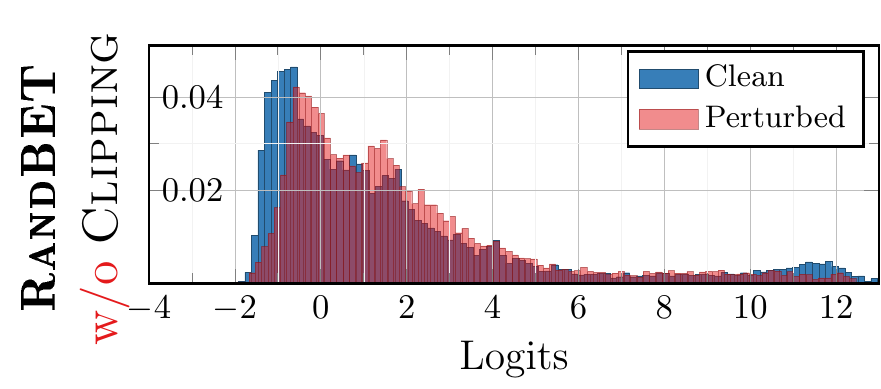}
	\end{subfigure}
	\begin{subfigure}{0.12\textwidth}
		\vspace*{0px}
		
		\includegraphics[height=1.7cm]{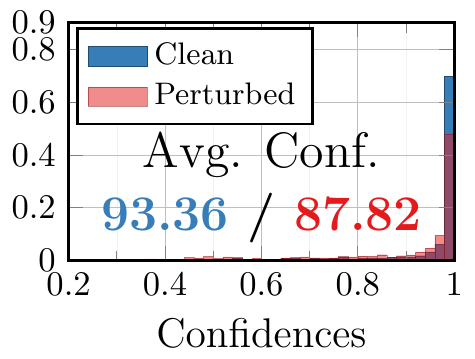}
	\end{subfigure}
	\begin{subfigure}{0.12\textwidth}
		\vspace*{3px}
		
		\includegraphics[height=1.7cm]{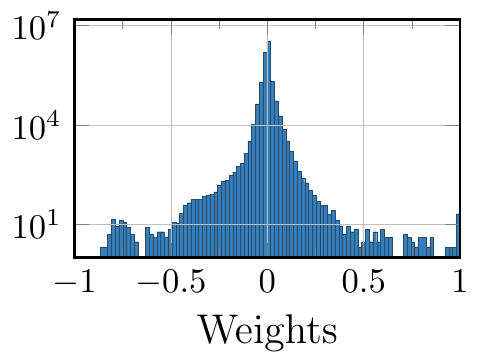}
	\end{subfigure}
	
	\vspace*{-8px}
	\caption{\textbf{Effect of Weight Clipping.} On \CifarT, weight clipping constraints the weights (right), thereby implicitly limiting the possible range for logits (left, {\color{colorbrewer2}blue}).
	However, even for $\wmax{=}0.1$ the DNN is able to produce high confidences (middle, {\color{colorbrewer2}blue}), suggesting that more weights are used to obtain these logits. Furthermore, the impact of random bit errors, $p = 1\%$, on the logits/confidences ({\color{colorbrewer1}red}) is reduced significantly. \Random (trained with $p = 1\%$, w/o weight clipping), increases the range of weights and is less effective at preserving logit/confidence distribution.
	}
	\label{fig:clipping}
	\vspace*{-0.2cm}
\end{figure}

The interplay of weight clipping and minimizing the cross-entropy loss during training is the key. High confidences can only be achieved by large differences in the logits. Because the weights are limited to $[-\wmax, \wmax]$, large logits can only be achieved using more weights in each layer to produce larger outputs. This is illustrated in \figref{fig:clipping} (right): using $\wmax{=}0.1$, the weights are (depending on the layer) up to $5$ times smaller. Considering deep NNs, the ``effective'' scale factor for the logits is significantly larger, scaling exponentially with the number of layers. Thus, using $\wmax{=}0.1$ is a significant constraint on the DNNs ability to produce large logits. As result, weight clipping produces a much more uniform weight distribution.
\figref{fig:clipping} (left and middle) shows that a DNN constrained at $\wmax{=}0.1$ can produce similar logit and confidence distributions (in {\color{colorbrewer2}blue}) as the unclipped DNN. And random bit errors have a significantly smaller impact on the logits and confidences (in {\color{colorbrewer1}red}). \figref{fig:clipping} (right column) also shows the induced redundancy in the weight distribution. Weight clipping leads to more weights being utilized, \ie, less weights are zero (note log-scale, marked in {\color{colorbrewer1}red}, on the y-axis). Also, more weights reach large values. We found weight clipping to be an easy-to-use but effective measure to improve weight robustness.

Building on our preliminary work \cite{StutzMLSYS2021}, \textbf{per-layer weight clipping} extends ``global'' weight clipping by allowing per-layer weight constraints $w_{\text{max},l}$. This is based on the observation that weights in different layers can have radically different ranges. Clipping weights globally to $[-\wmax,\wmax]$ may result in only few layers being actually constrained and regularized. The regularization effect is less pronounced for the remaining layers, reducing the potential impact in terms of robustness. Thus, \emph{per-layer} weight clipping constraints each layer $l$ individually to $[-w_{\text{max},l}, w_{\text{max},l}]$. Here, weights and biases are treated individually as biases exhibit significantly different ranges. The per-layer constraints $w_{\text{max},l}$ are derived from the relative weight ranges of a DNN without weight clipping. For example, we found that the first convolutional layer as well as the logit layer usually have significantly larger range. Letting $w_{l,i}$ be the weights of layer $l$ with the largest absolute weight value, we define $\kappa_{l'} = \nicefrac{\max_i |w_{l',i}|}{\max_i |w_{l,i}|} \leq 1$ for all other layers $l'$. Then, for each $l'$, we define $w_{\text{max},l}$ as $\max(0.2, \kappa_{l'})\wmax$. \Clipping[$\wmax{=}0.1$] to refer to \emph{global} weight clipping with, \eg, $\wmax = 0.1$, and \PLClipping[$\wmax{=}0.25$] to denote \emph{per-layer} weight clipping with, \eg, $\wmax = 0.25$. For more results supporting the regularization effect of (per-layer) weight clipping, see our supplementary material.

\begin{table}[t]
	\centering
	\caption{\textbf{Quantization Robustness.} \RTE for random bit errors at $p = 0.05\%$ and $p = 0.5\%$
		for normal training with different quantization schemes discussed in \secref{subsec:robustness-quantization}. Minor differences
		can have large impact on \RTE while clean test error is largely unaffected. For $8$ bit the second row shows \Normal quantization (symmetric/per-layer) whereas the last row is our \Quant.
		For $4$ bits we show \Clipping[$0.1$]+\Quant with and without rounding.
	}
	\label{tab:quantization-robustness}
	\vspace*{-0.2cm}
	\begin{tabular}{| c | l | c | c | c |}
		\hline
		\multicolumn{2}{|c|}{Quantization Schemes} & \multirow{2}{*}{\begin{tabular}{@{}c@{}}\TE\\in \%\end{tabular}}& \multicolumn{2}{c|}{\RTE in \%}\\
		\cline{4-5} 
		\multicolumn{2}{|c|}{(
			\CifarT)} && $p{=}0.05$ & $p{=}0.5$\\
		\hline
		\hline
		\multirow{5}{*}{\rotatebox{90}{$8$ bit}} & \eqnref{eq:quantization}, global & 4.63 & 86.01 {\color{gray}\scriptsize ${\pm}$3.65} & 90.71 {\color{gray}\scriptsize ${\pm}$0.49}\\
		& \eqnref{eq:quantization}, per-layer & 4.36 & 5.51 {\color{gray}\scriptsize ${\pm}$0.19} & 24.76 {\color{gray}\scriptsize ${\pm}$4.71}\\
		& +asymmetric & 4.36 & 6.47 {\color{gray}\scriptsize ${\pm}$0.22} & {\color{colorbrewer1}40.78} {\color{gray}\scriptsize ${\pm}$7.56}\\
		& +unsigned & 4.42 & 6.97 {\color{gray}\scriptsize ${\pm}$0.28} & 17.00 {\color{gray}\scriptsize ${\pm}$2.77}\\
		& +rounding (=\Quant) & 4.32 & \bfseries 5.10 {\color{gray}\scriptsize ${\pm}$0.13} & \bfseries 11.28 {\color{gray}\scriptsize ${\pm}$1.47}\\
		\hline
		\hline
		\multirow{2}{*}{\rotatebox{90}{$4$ bit}} & w/o rounding* & 5.81 & 90.40 {\color{gray}\tiny ${\pm}$0.21} & 90.36 {\color{gray}\tiny ${\pm}$0.2}\\
		& w/ rounding* & \bfseries 5.29 & \bfseries 5.75 {\color{gray}\tiny ${\pm}$0.06} & \bfseries 7.71 {\color{gray}\tiny ${\pm}$0.36}\\
		\hline
	\end{tabular}
	\vspace*{-0.1cm}
\end{table}

\subsection{Random Bit Error Training (\Random)}
\label{subsec:robustness-training}

In \emph{addition to} weight clipping and robust quantization, we inject random bit errors with probability $p$ during training to further improve robustness. This results in the following learning problem, which we optimize as illustrated in \figref{fig:flowchart}:
\begin{align}
	\begin{split}
    	&\min_w \mathbb{E}[\mathcal{L}(f(x; \tilde{w}), y) + \mathcal{L}(f(x; w), y)]\\
    	\text{s.t.}&\quad v = Q(w),\, \tilde{v} = \biterror_p(v),\, \tilde{w} = Q^{-1}(\tilde{v}).
   	\end{split}\label{eq:random-training-average}
\end{align}
where $(x, y)$ are labeled examples, $\mathcal{L}$ is the cross-entropy loss and $v = Q(w)$ denotes the (element-wise) quantized weights $w$ which are to be learned. $\biterror_p(v)$ injects random bit errors with rate $p$ in $v$. Note that we consider both the loss on clean weights and weights with bit errors. This is desirable to avoid an increase in (clean) test error and stabilizes training compared to training only on bit errors in the weights. Note that bit error rate $p$ implies, in expectation, $pmW$ bit errors.

Following \algref{alg:training}, we use stochastic gradient descent to optimize \eqnref{eq:random-training-average}, by performing the gradient computation using the perturbed weights $\tilde{w} = Q^{-1}(\tilde{v})$ with $\tilde{v} = \biterror_p(v)$, while applying the gradient update on the (floating-point) clean weights $w$. In spirit, this is similar to data augmentation, however, the perturbation is applied on the weights instead of the inputs. As we found that introducing bit errors right from the start may prevent the DNN from converging, we apply bit errors as soon as the (clean) cross-entropy loss is below $1.75$.
\revision{\Random is different from training with quantization errors: The injected bit errors are entirely random while quantization errors are highly correlated throughout training. That is, our quantization errors are deterministic given fixed weights and weights tend to change only slightly in later epochs. Also, bit errors are injected in \emph{all} layers, in contrast to gradual quantization or quantizing random layers in each iteration \cite{StockICLR2021,ZhouICLR2017}.}

Interestingly, weight clipping and \Random have somewhat orthogonal effects, which allows combining them easily in practice: While weight clipping encourages redundancy in weights by constraining them to $[-\wmax,\wmax]$, 
\Random (w/o weight clipping) causes the DNN to have larger tails in the weight distribution, as shown in \figref{fig:clipping} (bottom). However, considering logits and confidences, especially with random bit errors (in {\color{colorbrewer1}red}), \Random alone performs slightly worse than \Clipping[$0.1$]. Thus, \Random becomes particularly effective when combined with weight clipping, as we make explicit using the notation \Random[$\wmax$] and in \algref{alg:training}.

\begin{table}
	\centering
	\caption{\textbf{Weight Clipping Robustness.}
	Clean \TE and \RTE as well as clean confidence and confidence at $p{=}1\%$ bit errors (in \%, higher is better, $\uparrow$) for \Clipping, \Clipping with label smoothing (+LS) and \PLClipping. \TE increases for \Clipping[$\wmax = 0.025$] where the DNN is not able to produce large (clean) confidences. LS consistently reduces robustness, indicating that robustness is due to enforcing high confidence during training \emph{and} weight clipping. Per-layer weight constraints are beneficial in terms of both robustness and clean performance, \ie, \RTE and \TE. Finally, using batch normalization (BN) worsens robustness significantly.}
	\label{tab:clipping-robustness}
	\vspace*{-0.2cm} 
	\begin{tabular}{| l | c | c | c | c | c |}
		\hline
		Model & \multirow{2}{*}{\begin{tabular}{@{}c@{}}\TE\\in \%\end{tabular}} & \multirow{2}{*}{\begin{tabular}{@{}c@{}}Conf\\in \%\end{tabular}} & \multirow{2}{*}{\begin{tabular}{@{}c@{}}Conf\\$p{=}1$\end{tabular}} & \multicolumn{2}{c|}{\RTE in \%}\\
		\cline{5-6}
		(\CifarT) & & & & $p{=}0.1$ & $p{=}1$\\
		\hline 
		\hline
		\Quant & \bfseries 4.32 & \bfseries 97.42 & 78.43 & 5.54 & 32.05\\
		\hline
		\Clipping[$0.15$] & 4.42 & 96.90 & 88.41 & 5.31 & 13.08\\
		\Clipping[$0.05$] & 5.44 & 95.90 & 94.73 & 5.90 & 7.18\\
		\Clipping[$0.025$] & 7.10 & {\color{colorbrewer1}84.69} & 83.28 & 7.40 & 8.18\\
		\hline
		\revision{\PLClipping[$0.2$]} & \revision{4.71} & \revision{96.68} & \revision{95.83} & \revision{5.20} & \bfseries \revision{6.53}\\
		\PLClipping[$0.1$] & 5.62 & 94.57 & 93.98 & 5.91 & 6.65\\
		\hline
		\hline
		\Clipping[$0.1$] ({\color{red}BN}) & 4.46 & 97.09 & 84.86 & 5.32 & 18.32\\
		\Clipping[$0.15$]+LS & 4.67 & 88.22 & 47.55 & 5.83 & {\color{colorbrewer2}29.40}\\
		\hline
	\end{tabular}
	\vspace*{-0.1cm}
\end{table}

\subsection{Adversarial Bit Error Training (\Adv)}
\label{subsec:adversarial-robustness-training}

In order to specifically address adversarial bit errors (\cf \secref{subsec:adversarial-errors}), \Random can be re-formulated to train with adversarial bit errors. Essentially, this results in a min-max formulation similar to \cite{MadryICLR2018}:
\begin{align} 
    \begin{split}
        &\min_w \mathbb{E}[\max_{\tilde{v}}\mathcal{L}(f(x; Q^{-1}(\tilde{v})), y)]\\
        \text{s.t.}&\quad d_H(\tilde{v}, v) \leq \epsilon, d_H(\tilde{v}_i, v_i) \leq 1
    \end{split}\label{eq:adversarial-training-average}
\end{align}
where the inner maximization problem, \ie, the attack is solved following \algref{alg:attack}.
In addition to not training on adversarial bit errors for a (clean) cross-entropy above $1.75$, we clip gradients to $[-0.05, 0.05]$. This is required as the cross-entropy loss on adversarially perturbed weights $\tilde{w}$ can easily be one or two magnitudes larger than on the clean weights.
Unfortunately, training is very sensitive to the hyper-parameters of the attack, including the step size, gradient normalization and momentum. This holds both for convergence during training and for the obtained robustness after training.
\section{Experiments}
\label{sec:experiments}

We present experiments on \Cifar \cite{Krizhevsky2009} \revision{and TinyImageNet \cite{tinyimagenet}}, considering \emph{random} bit error robustness first, followed by discussing \emph{adversarial} bit errors. To this end, we first analyze the impact of fixed-point quantization schemes on robustness (\secref{subsec:experiments-quantization}). Subsequently, we discuss weight clipping (\Clipping, \secref{subsec:experiments-clipping}), showing that improved robustness originates from increased redundancy in the weight distribution. Then, we focus on random bit error training (\Random, \secref{subsec:experiments-randbet}). We show that related work \cite{KimDATE2018,KoppulaMICRO2019} does not generalize, while \Random generalizes across chips and voltages, as demonstrated on profiled bit errors. We further consider random bit errors in inputs and activations (\secref{subsec:experiments-activations-inputs}). Finally, we discuss our adversarial bit error attack in comparison to BFA \cite{RakinARXIV2020} (\secref{subsec:experiments-advbet}) and show that \Clipping as well as \Random or \Adv increase robustness against adversarial bit errors significantly.

\begin{table}[t]
	\centering
	\caption{\textbf{Fixed Pattern Bit Error Training.} \RTE for training on an entirely fixed bit error pattern (\Pattern). \emph{Top:} Evaluation on the same pattern; \Pattern trained on $p = 2.5\%$ does not generalize to $p = 1\%$ even though the bit errors for $p = 1\%$ are a subset of those seen during training for $p = 2.5\%$ (in {\color{colorbrewer1}red}). \emph{Bottom:} \Pattern also fails to generalize to completely random bit errors.
	}
	\label{tab:randbet-baselines}
	\vspace*{-0.2cm}
	\begin{tabular}{| l | c | c |}
		\hline
		Model (\CifarT) & \multicolumn{2}{c|}{\RTE in \%, $p$ in \%}\\
		\hline
		\hline
		\textbf{Evaluation on Fixed Pattern} & $p{=}1$ & $p{=}2.5$\\
		\hline
		\Pattern $p{=}2.5$ & {\color{colorbrewer1}14.14} & 7.87\\
		\Pattern[$0.15$] $p{=}2.5$ & {\color{colorbrewer1}8.50} & 7.41\\
		\hline\hline
		\textbf{Evaluation on \emph{Random} Patterns} & $p{=}1$ & $p{=}2.5$\\
		\hline
		\Pattern[$0.15$] $p{=}2.5$ & 12.09 & 61.59\\
		\hline
	\end{tabular}
	\vspace*{-0.1cm} 
\end{table}

\textbf{Metrics:} We report (clean) test error \TE (lower is better, $\downarrow$), corresponding to \emph{clean} weights, and \textbf{robust test error \RTE} ($\downarrow$) which is the
\textbf{test error after injecting bit errors into the weights}. For random
bit errors we report the \emph{average} \RTE and its standard deviation for $50$ samples of random bit errors with rate $p$ as detailed in \secref{sec:errors}.
For adversarial bit errors, we report \emph{max} (\ie, worst-case) \RTE across a total of $80$ restarts as described in detail in \secref{subsec:experiments-advbet}. Evaluation is performed on $9000$ test examples.

\textbf{Architecture:} We use SimpleNet \cite{HasanpourARXIV2016}, providing comparable performance to ResNets \cite{HeCVPR2016} with only $W{=}5.5\text{Mio}$ weights on \CifarT.
On \CifarH, we use a Wide ResNet (WRN) \cite{ZagoruykoBMVC2016} and on TinyImageNet a ResNet-18 \cite{HeCVPR2016}.
In all cases, we use group normalization (GN) \cite{WuECCV2018}. Batch normalization (BN) \cite{IoffeICML2015} works as well but models using BN yield
consistently worse robustness against bit errors, see \secref{subsec:experiments-advbet} or \tabref{tab:clipping-robustness}.

\textbf{Training:} We use stochastic gradient descent with an initial learning rate of $0.05$, multiplied by $0.1$ after $\nicefrac{2}{5}$, $\nicefrac{3}{5}$ and $\nicefrac{4}{5}$ of $100$/$250$ epochs on \revision{TinyImageNet}/\Cifar. We whiten the input images and use AutoAugment \cite{CubukARXIV2018} with Cutout \cite{DevriesARXIV2017}. For \Random, random bit error injection starts when the loss is below \revision{1.75/3.5/6 on \CifarT/\CifarH/TinyImageNet}. Normal training with the standard and our robust quantization are denoted \Normal and \Quant, respectively. Weight clipping with $\wmax$ is referred to as \Clipping[\wmax], corresponding to results from \cite{StutzMLSYS2021},
and its per-layer variant is denoted \PLClipping[\wmax]. Similarly, we refer to \Random/\Adv with (global) weight clipping as \Random[\wmax]/\Adv[\wmax] and with per-layer weight clipping as \PLRandom[\wmax].
For \Quant, $m = 8$, we obtain $4.3\%$ \TE on \CifarT, $18.5\%$ \TE on \CifarH \revision{ and $36.5\%$ on TinyImageNet}.

Our \textbf{supplementary material} includes implementation details, more information on our experimental setup, and complementary experiments: \revision{on \MNIST \cite{LecunIEEE1998},} robustness of BN, other architectures, qualitative results for \Clipping and complete results for $m = 4,3,2$ bits precision. Also, we discuss a simple guarantee how the average \RTE relates
to the true expected robust error. Our \textbf{code} will be made available.

\begin{table}[t]
	\centering
	\caption{\textbf{Random Bit Error Training (\Random).}
	Average \RTE (and standard deviation) of \Random evaluated at various bit error rates $p$ and using $m = 8$ or $4$ bit precision. For low $p$, weight clipping provides sufficient robustness, especially considering \PLClipping. However for $p \geq 0.5$, \Random increases robustness significantly, both based on \Clipping and \PLClipping. This is particularly pronounced for low-precision, \eg, $m = 4$bits.}
	\label{tab:randbet-robustness}
	\vspace*{-0.2cm}
	\begin{tabular}{|@{\hskip 3px}c@{\hskip 3px}|@{\hskip 3px}l@{\hskip 3px}|@{\hskip 3px}c@{\hskip 3px}|@{\hskip 3px}c@{\hskip 3px}|@{\hskip 3px}c@{\hskip 3px}|@{\hskip 3px}c@{\hskip 3px}|}
		\hline
		& Model (\CifarT) & \multirow{2}{*}{\begin{tabular}{@{}c@{}}\TE\\in \%\end{tabular}} &\multicolumn{3}{c|}{\RTE in \%}\\
		\cline{4-6}
		& $p$ in \% && $p{=}0.5$ & $p{=}1$ & $p{=}1.5$\\
		\hline
		\hline
		\multirow{7}{*}{\rotatebox{90}{$8$bit}} & \Quant & \bfseries 4.32 & 11.28 {\color{gray}\tiny ${\pm}$1.47} & 32.05 {\color{gray}\tiny ${\pm}$6} & 68.65 {\color{gray}\tiny ${\pm}$9.23}\\
		& \Clipping[$0.1$] & 4.82 & 6.95 {\color{gray}\tiny ${\pm}$0.24} & 8.93 {\color{gray}\tiny ${\pm}$0.46} & 12.22 {\color{gray}\tiny ${\pm}$1.29}\\
		& \PLClipping[{\color{colorbrewer1}$0.25$}] & 4.96 & 6.21 {\color{gray}\tiny ${\pm}$0.16} & 7.04 {\color{gray}\tiny ${\pm}$0.28} & 8.14 {\color{gray}\tiny ${\pm}$0.49}\\
		& \Random[$0.1$] $p{=}0.1$ & 4.72 & 6.74 {\color{gray}\tiny ${\pm}$0.29} & 8.53 {\color{gray}\tiny ${\pm}$0.58} & 11.40 {\color{gray}\tiny ${\pm}$1.27}\\
		& \Random[$0.1$] $p{=}1$ & 4.90 & 6.36 {\color{gray}\tiny ${\pm}$0.17} & 7.41 {\color{gray}\tiny ${\pm}$0.29} & 8.65 {\color{gray}\tiny ${\pm}$0.37}\\
		& \PLRandom[{\color{colorbrewer1}$0.25$}] $p{=}0.1$ & 4.49 & 5.80 {\color{gray}\tiny ${\pm}$0.16} & 6.65 {\color{gray}\tiny ${\pm}$0.22} & 7.59 {\color{gray}\tiny ${\pm}$0.34}\\
		& \PLRandom[{\color{colorbrewer1}$0.25$}] $p{=}1$ & 4.62 & \bfseries 5.62 {\color{gray}\tiny ${\pm}$0.13} & \bfseries 6.36 {\color{gray}\tiny ${\pm}$0.2} & \bfseries 7.02 {\color{gray}\tiny ${\pm}$0.27}\\
		
		\hline
		\hline
		\multirow{4}{*}{\rotatebox{90}{$4$bit}} & \Clipping[$0.1$] & 5.29 & 7.71 {\color{gray}\tiny ${\pm}$0.36} & 10.62 {\color{gray}\tiny ${\pm}$1.08} & 15.79 {\color{gray}\tiny ${\pm}$2.54}\\
		& \PLClipping[{\color{colorbrewer1}$0.25$}] & \bfseries 4.63 & 6.15 {\color{gray}\tiny ${\pm}$0.16} & 7.34 {\color{gray}\tiny ${\pm}$0.33} & 8.70 {\color{gray}\tiny ${\pm}$0.62}\\
		& \Random[$0.1$] $p{=}1$ & 5.39 & 7.04 {\color{gray}\tiny ${\pm}$0.21} & 8.34 {\color{gray}\tiny ${\pm}$0.42} & 9.77 {\color{gray}\tiny ${\pm}$0.81}\\
		
		& \PLRandom[{\color{colorbrewer1}$0.25$}] $p{=}1$ & 4.83 & \bfseries 5.95 {\color{gray}\tiny ${\pm}$0.12} & \bfseries 6.65 {\color{gray}\tiny ${\pm}$0.19} & \bfseries 7.48 {\color{gray}\tiny ${\pm}$0.32}\\
		\hline
	\end{tabular}
	\vspace*{-0.1cm}
\end{table}

\subsection{Quantization Choice Impacts Robustness}
\label{subsec:experiments-quantization}

Quantization schemes affect robustness significantly, even when not affecting accuracy.
For example, \tabref{tab:quantization-robustness} shows that per-layer quantization reduces \RTE significantly for small bit error rates, \eg, $p = 0.05\%$. While asymmetric quantization further reduces the quantization range, \RTE increases, especially for large bit error rates, \eg, $p = 0.5\%$ (marked in {\color{colorbrewer1}red}). This is despite \figref{fig:quantization} showing a slightly smaller impact of bit errors. This is caused by an asymmetric quantization into \emph{signed} integers: Bit flips in the most significant bit (MSB, \ie, sign bit) are not meaningful if the quantized range is not symmetric as the sign bit does not reflect the sign of the represented weight value. Similarly, replacing integer conversion of $\nicefrac{w_i}{\Delta}$ by proper rounding, $\lceil\nicefrac{w_i}{\Delta}\rfloor$, reduces \RTE significantly (resulting in our \Quant).
This becomes particularly important for $m = 4$. Here, rounding also improves clean \TE slightly, but the effect is significantly less pronounced. Proper rounding generally reduces the approximation error of the quantization scheme. These errors are magnified when considering bit errors at test time, even though DNNs can compensate such differences during training to achieve good accuracy, \ie, low \TE. For $m = 4$ or lower, we also found weight clipping to help training, obtaining lower \TE.
Overall, we show that random bit errors induce unique error distributions in DNN weights, heavily dependent on details of the employed fixed-point quantization scheme. We think that robustness against bit errors should become an important criterion for the design of DNN quantization. While our \Quant performs fairly well, finding an ``optimal'' robust quantization scheme is an interesting open problem.

\subsection{Weight Clipping Improves Robustness}
\label{subsec:experiments-clipping}

While the quantization range adapts to the weight range
after every update during training, weight clipping explicitly constraints the weights to $[-\wmax, \wmax]$.
\tabref{tab:clipping-robustness} shows the effect of
different $\wmax$ for \CifarT with 8 bit precision. The clean test error is not affected for \Clipping[$\mathbf{\wmax{=}0.15}$] but
one has already strong robustness improvements for $p=1\%$
compared to \Quant (\RTE of 13.18\% vs 32.05\%). Further reducing $\wmax$ leads to a slow increase in clean \TE and decrease in average clean confidence, while significantly
improving \RTE to $7.18\%$ for $p=1\%$ at $\wmax=0.05$. For $\wmax=0.025$
the DNN is no longer able to achieve high confidence (marked in {\color{colorbrewer1}red}) which leads to stronger loss of clean \TE. Interestingly, the gap between clean and perturbed confidences under bit errors for $p=1\%$ is (almost) monotonically decreasing. These findings generalize to other datasets and precisions. However, for low precision $m\leq 4$ the effects are stronger
as \Quant alone does not yield any robust models and weight clipping is essential for achieving robustness. 

\begin{table}[t]
	\centering
	\caption{\textbf{Generalization to Profiled Bit Errors.} \RTE for \Random and \PLRandom on two different profiled chips. The bit error rates differ across chips due to measurements at different voltages, also see \figref{fig:errors}. Chip 2 exhibits a bit error distribution significantly different from uniform random bit errors: bit errors are strongly aligned along columns and biased towards $0$-to-$1$ flips, \cf \figref{fig:errors}. Nevertheless, \Random generalizes surprisingly well.}
	\label{tab:randbet-generalization}
	\vspace*{-0.2cm}
	\begin{tabular}{| l | l | c | c |}
		\hline
		Chip (\figref{fig:errors}) & Model (\CifarT)& \multicolumn{2}{c|}{\RTE in \%}\\
		\hline
		\hline
		\bfseries Chip 1 && $p{\approx}0.86$ & {\color{colorbrewer1}$p{\approx}2.75$}\\
		\hline
		& \Random[$0.05$] $p{=}1.5$ & 7.04 & 9.37\\
		&\PLRandom[$0.15$], $p{=}2$ & 6.14 & 7.58\\
		\hline
		\hline
		\bfseries Chip 2 && $p{\approx}0.14$ & {\color{colorbrewer1}$p{\approx}1.08$}\\
		\hline
		& \Random[$0.05$] $p{=}1.5$ & 6.00 & 9.00\\
		&\PLRandom[$0.15$], $p{=}2$ & 5.34 & 7.34\\
		\hline
	\end{tabular}
	\vspace*{-0.1cm}
\end{table}

As discussed in \secref{subsec:robustness-clipping} the robustness of the DNN originates in the cross-entropy loss enforcing high confidences on the training set and, thus, large logits while weight clipping works against having large logits. Therefore, the network has to utilize more weights with larger absolute values (compared to $\wmax$).
In order to test this hypothesis, we limit the confidences that need to be achieved via label smoothing \cite{SzegedyCVPR2016}, targeting $0.9$ for the true class and $\nicefrac{0.1}{9}$ for the other classes. According to \secref{subsec:robustness-clipping}, this should lead to less robustness, as the DNN has to use ``less'' weights. Indeed, in \tabref{tab:clipping-robustness}, \RTE at $p=1\%$ increases from $13.08\%$ for \Clipping[$0.15$] to $29.4\%$ when using label smoothing (marked in {\color{colorbrewer2}blue}). Moreover, the difference between average clean and perturbed confidence is significantly larger for DNNs trained with label smoothing. \revision{This can be confirmed with label \emph{noise} which is equivalent to label smoothing in expectation. In the supplementary material we also show that weight clipping outperforms several regularization baselines, including \cite{BuschjaegerDATE2021}.}

Per-layer weight clipping, \ie, \PLClipping, further improves robustness and at the same time lowers test error compared to \Clipping. For example, in \tabref{tab:clipping-robustness}, \PLClipping[$0.2$] reduces \RTE for $p{=}1\%$ to $6.48\%$ compared to $7.18$ for \Clipping[$0.05$]. Simultaneously, clean \TE improves from $5.44\%$ to $4.84$. This emphasizes that layers can have radically different weight ranges and, thus, regularization through weight clipping needs to be layer-specific. In our supplementary material we also show that weight clipping also leads to robustness against $L_\infty$ perturbations which generally affect all weights in contrast to random bit errors, and provide more qualitative results about the change of the weight distribution induced by clipping.

\subsection{\Random Yields Generalizable Robustness}
\label{subsec:experiments-randbet}

\begin{figure*}[t]
	\centering
	\hspace*{-0.3cm}
	\begin{subfigure}{0.32\textwidth}
		\centering
		\includegraphics[width=5.75cm]{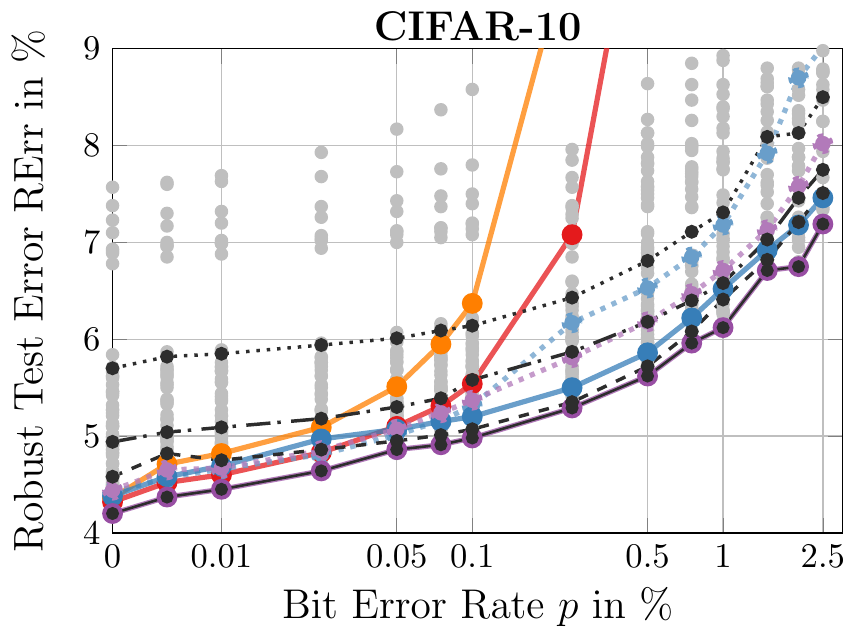}
	\end{subfigure}
	\hfill
	\begin{subfigure}{0.32\textwidth}
		\centering
		\includegraphics[width=5.75cm]{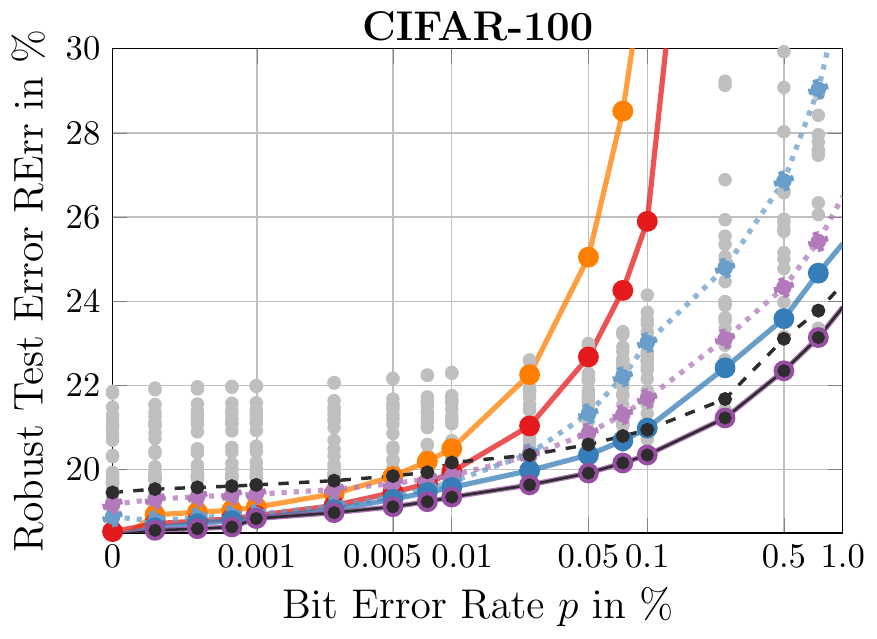}
	\end{subfigure}
	\hfill
	\begin{subfigure}{0.32\textwidth}
		\centering
		\includegraphics[width=5.75cm]{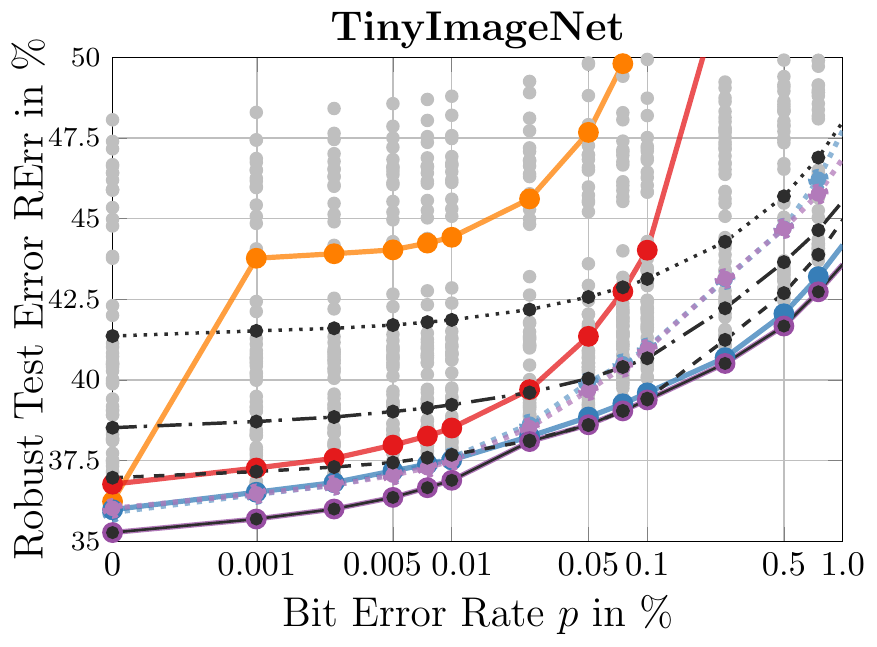}
	\end{subfigure}
	
	\hspace*{-0.1cm}
	\fbox{
	\begin{subfigure}{0.98\textwidth}
		\centering
		\includegraphics[width=1\textwidth]{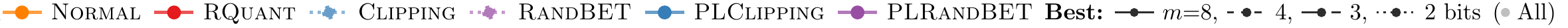}
	\end{subfigure}
	}
	\vspace*{-14px}
	\caption{\textbf{Bit Error Robustness on \CifarT, \CifarH and \revision{TinyImageNet}.} Average \RTE plotted against bit error rate $p$, both in \%. We considered various models (in {\color{gray}$\bullet$ gray}), corresponding to different $\wmax$ and $p$ during training. We explicitly plot the best model for each bit error rate: for \Normal ({\color{colorbrewer5}orange}), \Quant ({\color{colorbrewer1}red}), \Clipping ({\color{colorbrewer2}blue}) and \Random ({\color{colorbrewer4}violet}). Note that these might correspond to different $\wmax$ and $p$ (also across datasets). Across all approaches, we plot the per-error-rate best model in black: for $m = 8,4,3,2$ bits, depending on dataset. For $8$ bit and low bit error rates, \Clipping is often sufficient. However, for $4$ bit or higher bit error rates, \Random is crucial to keep \RTE low.}
	\label{fig:summary}
	\vspace*{-0.1cm}
\end{figure*}

In the following, we present experiments on \Random, showing that training on fixed, profiled bit errors patterns is not sufficient to generalize across voltages and chips. Thus, training on random bit errors in \Random is essential, and further improves robustness when applied on top of \Quant and \Clipping. Finally, we present results when evaluating \Random on real, profiled bit errors corresponding to three different chips.
Furthermore, both \Clipping and \Random can also be applied in a post-training quantization setting by replacing random bit errors during \Random with $L_0$ errors in weights.

\textbf{Training on Profiled Errors Does Not Generalize:}
Co-design approaches such as \cite{KimDATE2018,KoppulaMICRO2019} combine training DNNs on profiled SRAM or DRAM bit errors with hardware-approaches to limit the errors' impact.
However, profiling SRAM or DRAM requires expensive infrastructure, expert knowledge and time.
More importantly, training on profiled bit errors does not generalize to previously unseen bit error distributions (\eg, other chips or voltages): \tabref{tab:randbet-baselines} (top) shows \RTE of \Pattern, \ie, pattern-specific bit error training. The main problem is that \Pattern does not even generalize to lower bit error rates (\ie, higher voltages) of the same pattern as trained on (marked in {\color{colorbrewer1}red}). This is striking as, following \figref{fig:errors}, the bit errors form a subset of the bit errors seen during training: training with $p = 2.5\%$ bit errors does not provide robustness for $p = 1\%$, \RTE increases $7.9\%$ to $14.1\%$. It is not surprising, that \tabref{tab:randbet-baselines} (bottom) also demonstrates that \Pattern does not generalize to random bit error patterns: \RTE increases from $7.4\%$ to $61.6\%$ at $p = 2.5\%$. The same observations can be made when training on real, profiled bit errors corresponding to the chips in \figref{fig:errors}.
Overall, obtaining robustness that generalizes across voltages \emph{and} chips is crucial for low-voltage operation to become practical.

\textbf{\Random Improves Robustness:}
Our \Random, combined with weight clipping, further improves robustness and additionally generalizes across chips and voltages. \tabref{tab:randbet-robustness} shows results for weight clipping and \Random with $\wmax = 0.1$ and $m = 8,4$ bits precision. \Random is particularly effective against large bit error rates, \eg, $p = 1.5\%$, reducing \RTE from $12.22\%$ to $8.65\%$ ($m = 8$ bits) with global weight clipping and even further to $7.02\%$ with per-layer clipping, \ie \PLRandom. The effect is pronounced for $4$ bits or even lower precision, where models are generally less robust. The optimal combination of weight clipping and \Random depends on the bit error rate. However, we note that \Random consistently improves over \Clipping or \PLClipping. For example, in \tabref{tab:clipping-robustness}, lowering $\wmax$ to $0.05$ reduces \RTE below \Random[$0.1$] with $p{=}1\%$ for some bit error rates. Similar observations hold for \PLClipping.
We also emphasize that \Random generalizes to lower bit errors than trained on, in stark contrast to the fixed-pattern training \Pattern.

\begin{table*}[t]
	\centering
	\caption{\textbf{Bit Errors in Inputs and Activations.} Average \RTE against bit errors in weights, inputs and activations. We use \PLRandom to inject bit errors in weights (rate $p_w$), bit errors in inputs (rate $p_i$, {\color{colorbrewer5}orange}) and/or bit errors in activations (rate $p_a$, {\color{colorbrewer4}violet}) during training. Bit errors in inputs and activations are difficult to tolerate. Extreme \Clipping (\eg, $\wmax$) might worsen robustness against bit errors in inputs/activations. \PLRandom against bit errors in weights, inputs and activations is significantly harder, resulting in higher \TE, while improving robustness considerably.
	}
	\label{tab:input-activation-robustness}
	\vspace*{-0.2cm}
	\begin{tabularx}{\textwidth}{|X||c||c|c||c|c||c|c|c|}
		\hline
		Model (\CifarT) && \multicolumn{2}{c||}{bit errors in weights} & \multicolumn{2}{c||}{\color{colorbrewer5}bit errors in inp.} && \multicolumn{2}{c|}{\color{colorbrewer4}bit errors in act.}\\
		\cline{3-9}
		$\mathbf{\wmax{=}0.25}$, $m{=}8$ bit weight/{\color{colorbrewer5}inp.}/{\color{colorbrewer4}act.} quantization & \multirow{2}{*}{\begin{tabular}{@{}c@{}}\TE\\in \%\end{tabular}} & \multicolumn{2}{c||}{\RTE in \%} & \multicolumn{2}{c||}{\RTE in \%} & \multirow{2}{*}{\begin{tabular}{@{}c@{}}\TE in \%\\{\scriptsize(act. quant.)}\end{tabular}} & \multicolumn{2}{c|}{\RTE in \%}\\
		\cline{3-6}\cline{8-9}
		bit errors in weights/{\color{colorbrewer5}inp.}/{\color{colorbrewer4}act.}, $p$ in \% && $p{=}0.1$ & $p{=}1$ & $p{=}0.1$ & $p{=}0.5$ && $p{=}0.1$ & $p{=}0.5$\\
		\hline
		\hline		
		\PLClipping & 4.96 & 5.39 & 7.04 & 10.80 & 22.80 & 5.16 & 7.38 & 21.58\\
		\PLClipping[{\color{colorbrewer1}$0.1$}] & 5.62 & 5.91 & 6.65 & 12.80 & 26.50 & 5.84 & 8.72 & 27.36\\
		\hline
		\PLRandom, $p_w{=}0.1$ & \bfseries 4.49 & \bfseries 4.98 & 6.65 & 11.00 & 22.80 & \bfseries 4.71 & 7.25 & 24.94\\
		\PLRandom, $p_w{=}1$ & 4.62 & 5.02 & \bfseries 6.36 & 11.30 & 22.40 & 4.83 & \bfseries 6.92 & 19.83\\
		\hline
		\PLRandom, $p_w{=}1$, {\color{colorbrewer5}$p_i{=}0.1$} & 5.50 & 5.99 & 7.49 & \bfseries 7.70 & \bfseries 9.10 & 5.71 & 8.37 & 25.83\\
		\PLRandom, $p_w{=}1$, {\color{colorbrewer5}$p_i{=}0.1$}, {\color{colorbrewer4}$p_a{=}0.1$} & 9.16 & 9.60 & 11.09 & 11.50 & 13.80 & 9.31 & 10.54 & 13.51\\
		\hline
		\PLRandom, {\color{colorbrewer4}$p_a{=}0.5$} & 5.43 & 5.91 & 7.96 & 10.90 & 21.90 & 5.68 & 6.74 & \bfseries 10.16\\
		\PLRandom, $p_w{=}1$, {\color{colorbrewer4}$p_a{=}0.1$} & 7.66 & 8.27 & 10.47 & 13.80 & 24.70 & 7.89 & 9.09 & 12.17\\
		\hline
	\end{tabularx}
\end{table*}

\textbf{\Random Generalizes to Profiled Bit Errors:}
\Random also generalizes to bit errors profiled from real chips, corresponding to \figref{fig:errors}. \tabref{tab:randbet-generalization} shows results on the two profiled chips of \figref{fig:errors}. Profiling was done at various voltage levels, resulting in different bit error rates for each chip. To simulate various weights to memory mappings, we apply various offsets before linearly mapping weights to the profiled SRAM arrays. \tabref{tab:randbet-generalization} reports average \RTE, showing that \Random generalizes quite well to these profiled bit errors. Regarding chip 1, \Random performs very well, even for large $p\approx 2.75$, as the bit error distribution of chip 1 largely matches our error model in \secref{sec:errors}, \cf \figref{fig:errors} (left). In contrast, with chip 2 we picked a more difficult bit error distribution which is strongly aligned along columns,  potentially hitting many MSBs simultaneously. Thus, \RTE is similar for chip 2 even for a lower bit error rate $p \approx 1.08$ (marked in {\color{colorbrewer1}red})
but energy savings are still possible without degrading prediction performance. 

\revision{
\subsection{Summary and End-to-End Ablation}
}

Our final experiments are summarized in \figref{fig:summary}. We consider \Normal quantization vs. our robust quantization \Quant, various \Clipping and \Random models with different $\wmax$ and $p$ during training (indicated in {\color{gray}$\bullet$ gray}) and plot \RTE against bit error rate $p$ at test time. On all datasets \Quant outperforms \Normal. On \CifarT (left), \RTE increases significantly for \Quant ({\color{colorbrewer1}red}) starting at $p \approx 0.25\%$ bit error rate. While \Clipping ({\color{colorbrewer2}blue}) generally reduces \RTE, only \Random ({\color{colorbrewer4}violet}) can keep \RTE around $6\%$ or lower for a bit error rate of $p \approx 0.5\%$. The best model for each bit error rate $p$ (black and solid for $m = 8$) might vary.
\revision{This confirms our observations in the ablation experiments of previous sections: \Quant and \Clipping are necessary for reasonable robustness, but \Random gets more important for higher bit error rates, even if its absolute improvement is smaller.}
On \CifarT, \RTE increases slightly for $m = 4$. However, for $m = 3,2$ \RTE increases more significantly as clean \TE increases by $1-2\%$.
Nevertheless, \RTE only increases slightly for larger bit error rates $p$.
In all cases, \RTE increases monotonically, ensuring safe operation at higher voltages. The best trade-off depends on the application: higher energy savings require a larger ``sacrifice'' in terms of \RTE.
\revision{These observations can be confirmed on \CifarH and TinyImageNet, which are generally more difficult, resulting in a slightly quicker increase in \RTE for lower bit error rates.
However, while \Normal performs still reasonable on \CifarH, even bit error rates of $p{=}0.001\%$ already cause an increase of more than $7\%$ in \RTE on TinyImageNet. Moreover, the advantage of \Random over \Clipping reduces on TinyImageNet, especially without per-layer weight clipping. This indicates that training with bit errors gets more difficult.}
Our supplementary material includes a confidence-interval based guarantee showing that \RTE will not deviate strongly from the results in \figref{fig:summary} \revision{as well as results on MNIST.}

\vspace*{16px}
\subsection{Robustness to Bit Errors in Inputs and Activations}
\label{subsec:experiments-activations-inputs}

While \Random successfully improves robustness against low-voltage induced bit errors in the \emph{weights}, both inputs and activations might also be subject to random bit errors when (temporarily) stored on the SRAM scratchpad. Thus, we also consider injecting bit errors in inputs and activations, making first steps towards a ``fully'' robust DNN. First, we take a closer look at the impact of bit errors in inputs and activations. Then, we adapt \Random to improve robustness. For Clarity, in text and \tabref{tab:input-activation-robustness}, we use $p_w$, $p_i$ and $p_a$ to denote the bit error rate in weights, inputs and activations, respectively. We further color-code bit errors in inputs as {\color{colorbrewer5}orange} and activations as {\color{colorbrewer4}violet}.

\textbf{Bit Error Model in Inputs and Activations:} Following our description in \secref{subsec:random-errors}, we inject bit errors in both inputs and activations. Inputs are quantized using $m = 8$ bit with $[\qmin, \qmax] = [0,1]$. Note that this does not introduce errors as images are typically provided in $8$ bit quantization per channel. Activations are also quantized using $m = 8$ bit using our robust fixed-point quantization scheme. Note that we do not employ any advanced activation quantization schemes such as activation clipping \cite{ChoiARXIV2018}. Bit errors are injected \emph{once} into inputs before being fed to the DNN and \emph{once} into the activations after \emph{each} block consisting of convolutional layer, normalization layer (\ie, GN) and ReLU activation. This assumes that activations after each such block are temporally stored on the SRAM scratchpads. As detailed in \secref{subsec:random-errors}, while the actual data flow is highly specific to both chip and DNN architecture, this is a realistic assumption. As with bit errors in the weights, we evaluate using $50$ random bit error patterns and make sure that for rate $p' \leq p$ the bit errors introduced in inputs/activations are a subset of those for rate $p$. We refer to our supplementary material for additional details.

\textbf{Input and Activation Bit Error Robustness:} Bit errors have severe impact on accuracy not only when occurring in weights but also in inputs and activations. \tabref{tab:input-activation-robustness} shows robustness, \ie, average \RTE, of various models on \CifarT against bit errors in weights, inputs (in {\color{colorbrewer5}orange}) or activations (in {\color{colorbrewer4}violet}). For activation quantization, we additionally report the (clean) \TE \emph{after} activation quantization (without bit errors). While being simplistic, our activation quantization has negligible impact on \TE. We found bit errors in inputs and activations to be challenging in terms of robustness. Even for small bit error rates, \eg, $p = 0.1\%$, \RTE increases significantly, to at least $7.7\%$ and $6.92\%$ \RTE for inputs and activations, respectively. While \PLRandom (training on random bit errors \emph{in weights}) helps against bit errors in activations, it has no impact on robustness against bit errors in inputs. Extreme \PLClipping, in contrast, \eg, using $\wmax = 0.1$ tends to reduce robustness in both cases. These results show that low-voltage operation is complicated when taking inputs and activations into account. While separate SRAM arrays for weights, inputs and activations can be used, allowing varying levels of bit errors, this is potentially undesirable from a design perspective.

\textbf{\Random for Inputs and Activations:} In order to obtain robustness against random bit errors in inputs and/or activations, we adapt \Random to allow bit error injection in inputs and/or activations (in addition to weights) during training. \tabref{tab:input-activation-robustness} shows that injecting either input bit errors (bit error rate $p_i$ in {\color{colorbrewer5}orange}) or activation bit errors (bit error rate $p_a$ in {\color{colorbrewer4}violet}) helps robustness, but also makes training significantly more difficult. Indeed, injecting bit errors in weights, inputs \emph{and} activations increases (clean) \TE significantly, to $9.16\%$ from $4.62\%$ (for \Random with bit errors in weights only). We found that this difficulty mainly stems from injecting bit errors in activations during training: While \Random (activations only) with $p_a{=}0.5\%$ affects (clean) \TE only slightly ($5.43\%$), bit errors in weights \emph{and} activations (\ie, $p_w{=}1\%$ and $p_a{=}0.1\%$) results in an increase to $7.66\%$.
This increase in \TE also translates to an increase in \RTE against bit errors in weights or activations.
As a result, injecting bit errors only in weights and inputs (\eg, $p_w{=}1\%$ and $p_i = 0.1\%$) might be beneficial as it avoids a significant increase in (clean) \TE, while still providing some robustness against bit errors in activations. Overall, we made a significant step towards DNNs ``fully'' robust against low-voltage induced random bit errors, but the problem remains difficult.

\begin{table}[t]
	\centering
	\caption{\textbf{Bit Flip Attack (BFA) \cite{RakinICCV2019}.} Worst \RTE 6against the bit flip attack (BFA) of \cite{RakinICCV2019} for various allowed budgets $\epsilon$ of bit errors.
	On \CifarT, we also present results when using batch normalization (BN). Surprisingly, \Clipping is quite successful in ``defending'' BFA as long as BN is avoided, even for very large $\epsilon$. In \tabref{tab:adversarial-ablation}, our adversarial bit error attack outperforms BFA significantly.}
	\label{tab:bfa-robustness}
	\vspace*{-0.2cm}
	\begin{tabular}{|l|c|c|c|c|}
		\hline
		Model / Dataset & \multicolumn{4}{c|}{\RTE in \% for $\epsilon$ Bit Errors}\\
		\hline
		\hline
		\CifarT & $\epsilon{=}160$ & $\epsilon{=}320$ & $\epsilon{=}640$ & $\epsilon{=}960$\\
		\hline
		\Quant \emph{(GN)} & 34.42 & 89.01 & 90.10 & 90.02\\
		\Clipping[$0.05$] {\color{red}BN} & 89.48 & 89.16 & 89.19 & 89.40\\
		\Clipping[$0.05$] \emph{(GN)} & 14.84 & 24.96 & 49.56 & 51.88\\
		\Adv[$0.05$], $\epsilon{=}160$ \emph{(GN)} & \bfseries 13.72 & \bfseries 15.53 & \bfseries 25.91 & \bfseries 44.47\\
		\hline
	\end{tabular}
\end{table}

\subsection{Robustness Against Adversarial Bit Errors} 
\label{subsec:experiments-advbet}

In this section, we switch focus and consider \emph{adversarial} bit error robustness. To this end, we consider both the BFA attack from related work \cite{RakinICCV2019} and our own adversarial bit error attack from \secref{subsec:adversarial-errors}. As ``defense'', we consider \Clipping, \Random and our adversarial bit error training (\Adv) which are able to improve robustness considerably -- both against BFA and our adversarial bit level attack.

We train \Adv using $T = 10$ iterations of our adversarial bit error attack, with learning rate $0.5$ (no momentum) after normalizing the gradient by its $L_\infty$ norm, \cf \algref{alg:attack}. For evaluation, we use the official BFA implementation and run it for $5$ restarts. Each run, we allow $5$ bit flips per iteration, resulting in a total of $5\cdot T$ allowed bit flips with $T$ iterations, \ie, $\epsilon := 5\cdot T$ in our adversarial bit error model. We run our adversarial bit error attack for $T = 100$ iterations.
For comprehensive evaluation, we consider a total of $80$ random restarts for various combinations of hyper-parameters. We compute our adversarial bit error attack, \ie, solve \eqnref{eq:attack}, on $100$ held-out test examples and evaluate, as described before, on $9000$ test examples.

\textbf{Limitations of BFA:} We start by considering BFA, showing that it is not as effective (and efficient) against our DNNs, compared to the results in \cite{RakinICCV2019}. \tabref{tab:bfa-robustness} reports worst (\ie, max) \RTE on \CifarT for various models.
\revision{BFA is effective in attacking our \Quant model starting with $\epsilon{=}320$ bit errors, increasing \RTE to $89.01\%$.}
\revision{However, \Clipping is already very robust, reducing \RTE to $24.96\%$.}
Here, we also show results considering batch normalization (BN, marked in {\color{red}red}), as used in \cite{RakinICCV2019}. When training \Clipping with BN, the DNN is significantly less robust. In fact, BFA is suddenly able to increase \RTE to ${\sim}90\%$ even for $\epsilon{=} 160$.
However, we found that BFA does \emph{not} attack the batch normalization parameters (\ie, scale and bias). Instead, as shown in \tabref{tab:clipping-robustness} against random bit errors, we found that BN is generally less robust.
\revision{Finally, BFA tends to increase \RTE by consecutively changing the weights so that it finally predicts a single class for all inputs.
While this causes the loss to increase monotonically,}
BFA needs between 1 and 2 seconds per iteration and the number of bit flips is (indirectly) tied to the number of iterations.
This makes BFA unfit to be used for \Adv. We address some of these limitations using our adversarial bit error attack from \secref{subsec:adversarial-errors}.

\begin{table}[t]
	\centering
	\caption{\textbf{Adversarial Bit Error Ablation.} Worst \RTE for \Quant and \Clipping, considering different $\epsilon$ and settings  on \CifarT: attacking only the \textbf{log}it layer (\ie, last layer), or only first \textbf{conv}olutional layer, using untargeted (``\textbf{U}'') or targeted (``\textbf{T}'') attacks.
		Targeted attacks are usually easier to optimizer and more effective. The first convolutional or logit layer is particularly vulnerable.}
	\label{tab:adversarial-ablation}
	\vspace*{-0.2cm}
	\begin{tabular}{|@{\hskip 4.5px}l@{\hskip 4.5px}|@{\hskip 4.5px}c@{\hskip 4.5px}|@{\hskip 4.5px}c@{\hskip 4.5px}|@{\hskip 4.5px}c@{\hskip 4.5px}|@{\hskip 4.5px}c@{\hskip 4.5px}|@{\hskip 4.5px}c@{\hskip 4.5px}|}
		\hline
		& \TE \% & \multicolumn{4}{c|}{Worst \RTE in \%}\\
		\hline
		&& U {\color{gray}\footnotesize(all)} & T {\color{gray}\footnotesize(all)} & log & conv\\
		\hline
		\hline
		\CifarT ($W{\approx}5.5\text{Mio}$) && \multicolumn{4}{c|}{$\epsilon{=}320$}\\
		\hline
		\Quant & 4.89 & 8.54 & 91.18 & 91.18 & 89.06\\
		\Clipping[$0.05$] & 5.34 & 24.04 & 35.20 & 35.86 & 60.76\\
		\Adv[$0.05$], $\epsilon{=}160$ & 5.54 & 10.01 & 20.20 & 26.22 & 12.33\\
		\hline
	\end{tabular}
\end{table}
\begin{figure}[t]
	\vspace*{-0.1cm}
	\begin{minipage}{0.46\textwidth}
		\centering
		\footnotesize\bfseries \CifarT: Adversarial Bit Errors on \Clipping[$0.05$], $\epsilon{=}640$
	\end{minipage}\\
	\hspace*{-0.35cm}
	\begin{minipage}[t]{0.185\textwidth}
		\centering
		\includegraphics[height=2.3cm]{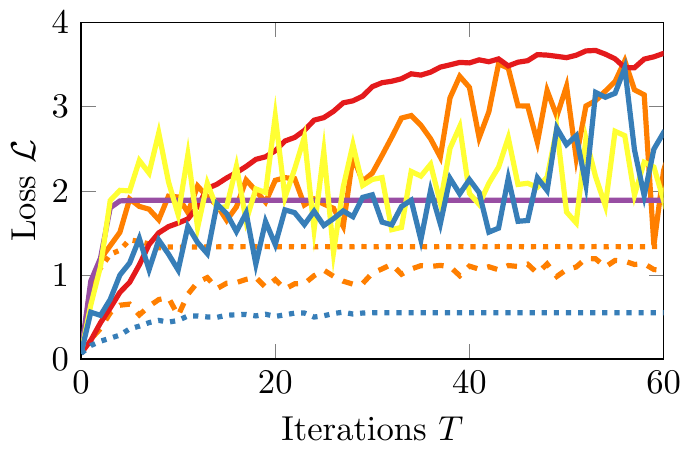}
	\end{minipage}
	\begin{minipage}[t]{0.29\textwidth}
		\centering
		\includegraphics[height=2.3cm]{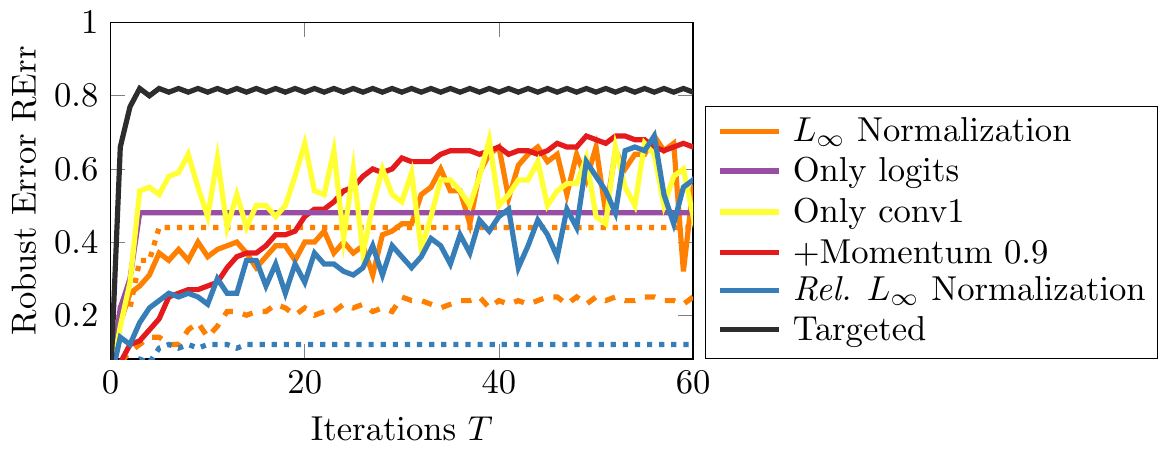}
	\end{minipage}
	\vspace*{-6px}
	\caption{\textbf{Adversarial Bit Error Iterations.} We plot loss, \cf \eqnref{eq:attack}, and robust error \RTE against iterations, both \emph{measured on the $100$ held-out test examples used to find, \ie, train, adversarial bit errors}. Clearly, $L_\infty$ gradient normalization \emph{with} momentum (in {\color{colorbrewer1}red}), targeting first convolutional ({\color{colorbrewer6}yellow}) or logit layer ({\color{colorbrewer4}violet}) is most effective for untargeted attacks. Unfortunately, the attack gets easily stuck in bad optima if learning rate is not optimal. Targeted attacks simplify optimization and are often more effective (black, right).}
	\label{fig:adversarial-ablation}
	\vspace*{-0.1cm}
\end{figure}

\textbf{More Effective Adversarial Bit Errors:} Using appropriate hyper-parameters and considering \emph{both} untargeted and targeted attacks, our adversarial bit error attack is more effective and efficient compared to BFA. \tabref{tab:adversarial-ablation} shows (worst) \RTE on \CifarT, showing that our adversarial bit errors achieve higher \RTE compared to BFA, \eg, for $\epsilon{=}320$. As also reported in \cite{RakinARXIV2020}, we found that targeted attacks are generally more effective. This is because the ``easiest'' way to increase \RTE is to force the DNN to predict a constant label, which the targeted attacks explicitly do. Similarly, targeting \emph{only} the logit layer is usually sufficient for high \RTE. Interestingly, attacking only the first convolutional layer is quite effective, as well. We also emphasize that we are considering more adversarial bit errors (\ie, larger $\epsilon$) on \CifarT, even though \CifarT is considerably more difficult. This is due to the increased number of weights (roughly $5.5\text{Mio}$) on \CifarT. \figref{fig:adversarial-ablation} also shows that gradient normalization and momentum are essential for the untargeted attack to be successful. This is important as running the targeted attack for each target label during \Adv is prohibitively expensive. Nevertheless, the attack remains sensitive to, \eg, the learning rate, but $\gamma{=}1$ works reasonably well across models, given enough random restarts to avoid poor optima. Thus, in our evaluation, we run both targeted and untargeted attacks, attacking all weights, only the first convolutional and/or logit layer and consider the worst-case across a total of $80$ random restarts. Overall, our attack provides a much more realistic estimate of adversarial bit error robustness. Furthermore, our attack requires only between 0.15 and 0.2 seconds per iteration and runtime is independent of $\epsilon$.

\begin{table}[t]
	\centering
	\caption{\textbf{Adversarial Bit Error Robustness and \Adv.} \RTE against adversarial bit errors for \Clipping, \Random and \Adv on \MNIST, \CifarT and \revision{TinyImageNet}. We consider the worst-case across multiple random restarts, including targeted and untargeted attacks as well as attacks on all or particularly vulnerable (\eg, logit) layers. While \Adv improves robustness considerably on \MNIST, \Clipping and \Random are very strong baselines on \CifarT. As we generally consider larger $\epsilon$ on \CifarT, this makes it hard for \Adv to further improve results.}
	\label{tab:advbet}
	\vspace*{-0.2cm}
	\hspace*{-0.1cm}
	\begin{tabular}{|@{\hskip 4px}L{2.85cm}@{\hskip 4px}|@{\hskip 4.5px}C{0.85cm}@{\hskip 4.5px}|@{\hskip 4.5px}C{0.85cm}@{\hskip 4.5px}|@{\hskip 4.5px}C{0.85cm}@{\hskip 4.5px}|@{\hskip 4.5px}C{0.85cm}@{\hskip 4.5px}|@{\hskip 4.5px}C{0.85cm}@{\hskip 4.5px}|}
		\hline
		& \TE \% & \multicolumn{4}{c|}{\RTE in \%}\\
		\hline
		\hline
		\MNIST && $\epsilon{=}80$ & $\epsilon{=}160$ & $\epsilon{=}240$ & $\epsilon{=}320$\\
		\hline
		\Quant & 0.37 & 91.08 & 91.08 & 91.08 & 91.08\\
		\Clipping[$0.05$] & 0.38 & 85.09 & 88.81 & 90.11 & 90.26\\
		\Random[$0.05$], $p{=}20$ & 0.39 & 10.13 & 69.90 & 81.16 & 81.94\\
		\Adv[$0.05$], $\epsilon{=}240$ & \bfseries 0.31 & 11.58 & 28.34 & 41.66 & 71.16\\
		\Adv[$0.05$] T, $\epsilon{=}240$ & 0.36 & \bfseries 10.10 & \bfseries 19.44 & \bfseries 31.23 & \bfseries 51.01\\
		\hline
		\CifarT && $\epsilon{=}160$ & $\epsilon{=}320$ & $\epsilon{=}480$ & $\epsilon{=}640$\\
		\hline
		\Quant &\bfseries  4.89 & 91.18 & 91.18 & 91.18 & 91.18\\
		\Clipping[$0.05$] & 5.34 & 20.48 & 60.76 & 79.12 & 83.93\\
		\Random[$0.05$], $p{=}2$ & 5.42 & \bfseries 14.66 & 33.86 & \bfseries 54.24 & 80.36\\
		\Adv[$0.05$], $\epsilon{=}160$ & 5.54 & 15.20 & \bfseries 26.22 & 55.06 & \bfseries 77.43\\
		\hline
	\end{tabular}

	\revision{
	\hspace*{-0.1cm}
	\begin{tabular}{|@{\hskip 4.5px}L{2.85cm}@{\hskip 4.5px}|@{\hskip 4.5px}C{0.85cm}@{\hskip 4.5px}|@{\hskip 4.5px}C{0.85cm}@{\hskip 4.5px}|@{\hskip 4.5px}C{0.85cm}@{\hskip 4.5px}|@{\hskip 4.5px}C{0.85cm}@{\hskip 4.5px}|@{\hskip 4.5px}C{0.85cm}@{\hskip 4.5px}|}
		\hline
		TinyImageNet && $\epsilon{=}80$ & $\epsilon{=}160$ & $\epsilon{=}240$ & $\epsilon{=}320$\\
		\hline
		\Quant & 36.77 & 99.70 & 99.78 & 99.78 & 99.78\\
		\Clipping[$0.1$] & 37.42 & 54.47 & 82.94 & 96.37 & 99.47\\
		\Random[$0.1$], $p{=}1$ & 42.30 & 58.11 & 76.74 & 99.94 & 99.51\\
		\Adv[$0.1$], $\epsilon{=}160$ & 37.83 & 52.91 & 61.06 & 97.73 & 99.58\\
		\hline
	\end{tabular}
	}
	\vspace*{-0.1cm}
\end{table}

\textbf{\Clipping and \Random Improve Adversarial Bit Error Robustness:} As shown for BFA in \tabref{tab:bfa-robustness}, we find that \Clipping and \Random are surprisingly robust against adversarial bit errors. Specifically, \tabref{tab:advbet} reports \RTE on \MNIST and \CifarT. While \Clipping does not perform well on \MNIST, \Random reduces \RTE against $\epsilon{=}80$ from $85.09\%$ to $10.13\%$. On \CifarT, in contrast, considering larger $\epsilon$, \Clipping alone is quite effective, with $20.48\%$ \RTE against $\epsilon{=}160$. Nevertheless, \Random further improves over \Clipping. This is counter-intuitive considering, \eg, robustness against adversarial examples where training against \emph{random} perturbations does generally not provide \emph{adversarial robustness}. However, \Random is trained against large bit error rates, \eg, $p{=}2\%$ on \CifarT, with an expected $\epsilon{=}8{\cdot}W{\approx} 880\text{k}$ bit errors, $110\text{k}$ in the most significant bits (MSBs). \revision{This also holds for TinyImageNet.} For adversarial bit error, in contrast, we consider up to $\epsilon{=}640$ on \CifarT. In terms of BFA, complementing the results in \tabref{tab:bfa-robustness}, we need on average $2253$ bit errors to increase \RTE above $90\%$ for \Random on \CifarT. In contrast, \cite{HeCVPR2020} report $541$ required bit flips (ResNet-20, $W{\approx}4.3\text{Mio}$) to ``break'' their proposed binarized DNN, which has been reduced to $35$ in \cite{RakinARXIV2020} using targeted BFA. Overall, these results show that random and adversarial bit error robustness are aligned well, allowing to \emph{secure} low-voltage operation of DNN accelerators.

\textbf{\Adv Improves Adversarial Bit Error Robustness:} Using \Adv, we can further boost robustness against adversarial bit errors, \cf \tabref{tab:advbet}. On \MNIST, in particular, \Adv is able to reduce \RTE from above $80\%$ for \Random or \Clipping, to $41.66\%$ against up to $\epsilon{=}240$ adversarial bit errors. As \tabref{tab:adversarial-ablation} illustrates, targeted attacks are generally considered stronger. Thus, training with a targeted attack, selecting a random target label in each iteration, further boosts robustness to $31.23\%$ \RTE. However, these improvements do not easily generalize to \CifarT. We suspect this is due to two reasons: First, we found that training with too large $\epsilon$ is difficult (also \cf increased \TE in \tabref{tab:advbet}), \ie, \Adv with larger $\epsilon$ does not improve robustness because training becomes too hard. This is why we report results for \Adv trained on $\epsilon{=}160$. Second, \Clipping alone is significantly more robust on \CifarT than on \MNIST, resulting in a particularly strong baseline. We suspect that architectural differences have a significant impact on how effective \Clipping is against adversarial bit errors. For example, DNNs on \CifarT have inherently more weights in the first convolutional layer (relative to $W$, due to larger input dimensionality,) which \tabref{tab:adversarial-ablation} shows to be particularly vulnerable.
Overall, \Adv can be used to further boost robustness against \emph{adversarial} bit errors, beyond \Clipping.

\section{Conclusion}
\label{sec:conclusion}

We propose a combination of \textbf{robust quantization}, \textbf{weight clipping} and \textbf{random bit error training (\Random)} or \textbf{adversarial bit error training (\Adv)} to train DNNs robust against random and adversarial bit errors in their (quantized) weights. This enables secure low-voltage operation of DNN accelerators. Specifically, we consider robustness against random bit errors induced by operating the accelerator memory far below its rated voltage \cite{ChandramoorthyHPCA2019}. We show that quantization details have tremendous impact on robustness, even though we use a very simple fixed-point quantization scheme without any outlier treatment \cite{ZhuangCVPR2018,SungARXIV2015,ParkISCA2018}. By encouraging redundancy in the weights, clipping is another simple but effective strategy to improve robustness. In contrast to related work, \Random does \emph{not} require expert knowledge or profiling infrastructure \cite{KimDATE2018,KoppulaMICRO2019} and generalizes across chips, with different bit error patterns, and voltages. As a result, we also avoid expensive circuit techniques \cite{ReagenISCA2016,ChandramoorthyHPCA2019}. Furthermore, complementing existing research, we discuss low-voltage induced random bit errors in inputs and activations. Finally, we propose a novel \emph{adversarial} bit error attack that is more effective and efficient compared to existing attacks \cite{RakinICCV2019} and can be utilized for \Adv. Surprisingly, we find that \Clipping and \Random also improve robustness against adversarial bit errors. However, \Adv further improves robustness specifically against adversarial bit errors. Altogether, by improving DNN robustness against random and adversarial bit errors, we enable both energy-efficient and secure DNN accelerators.

\bibliographystyle{IEEEtran}
\bibliography{bibliography}

\clearpage
{
	\setcounter{section}{0}
	\renewcommand{\thesection}{\Alph{section}}
	\setcounter{figure}{0}
	\renewcommand{\thefigure}{\Alph{figure}}
	\setcounter{table}{0}
	\renewcommand{\floatpagefraction}{.8}
	\renewcommand{\thetable}{\AlphAlph{\value{table}}}
	\raggedbottom
	\section{Overview}

\IEEEPARstart{I}{n} the main paper, we consider low-voltage operation of deep neural network (DNN) accelerators. This allows to reduce energy consumption significantly, while causing bit errors in the memory storing the (quantized) DNN weights. We show that robust fixed-point quantization (\Quant), weight clipping (\Clipping) and random bit error training (\Random) improve robustness to the induced \emph{random} bit errors significantly, even for low-precision quantization. In contrast to related work, our approach generalizes across accelerators and operating voltages. Furthermore, this improves security of DNN accelerators against attacks on voltage controllers. Furthermore, we consider DNN robustness against \emph{adversarial} bit errors which have recently been shown to degrade accuracy significantly. To this end, we propose to combine \Clipping with adversarial bit error training (\Adv) to achieve robustness against both targeted and untargeted bit-level attacks. In this supplementary material, we provide complementary experimental results and discussion corresponding to the experiments presented in the main paper.

\subsection{Outline}
\label{subsec:supp-outline}

This supplementary material is organized as follows:
\begin{itemize}
	\item \secref{sec:supp-introduction}: additional \textbf{details on Fig. 1} from the main paper (corresponding to \figref{fig:introduction}).
	\item \secref{sec:supp-related-work}: discussion of \textbf{related work} considering adversarial and corruption robustness, backdooring and fault tolerance.
	\item \secref{sec:supp-main}: background for our  \textbf{random bit error models} from the hardware perspective, including details on the profiled bit errors in \secref{subsec:supp-errors-profiled} (\figref{fig:supp-errors}) and a simple probabilistic bound on DNN performance under random bit errors in \secref{subsec:supp-bound}.
	\item \secref{sec:supp-main-adversarial} details on our \textbf{adversarial bit error attack}.
	\item \secref{sec:supp-implementation}: implementation details for quantization and \textbf{bit manipulation in PyTorch} \cite{ParkARXIV2020}.
	\item \secref{sec:supp-clipping}: how to use \textbf{weight clipping with group/batch normalization}.
	\item \secref{subsec:supp-experiments-setup}: further details on our \textbf{experimental setup} (\tabref{tab:supp-architectures}, \ref{tab:supp-accuracy}).
	\item \secref{sec:supp-experiments}: complementary \textbf{experiments} as outlined below.
	\begin{itemize}
		\item \secref{subsec:supp-experiments-bn}: a discussion of bit error robustness and batch normalization (\tabref{tab:supp-bn}).
		\item \secref{subsec:supp-experiments-quantization}: ablation for our robust fixed-point quantization (\Quant, \tabref{tab:supp-quantization}).
		\item \secref{subsec:supp-experiments-clipping}: details and ablation for weight clipping (\Clipping, \figref{fig:supp-clipping-inf} and \ref{fig:supp-clipping}, \tabref{tab:supp-clipping-scaling} and \ref{tab:supp-clipping}).
		\item \secref{subsec:supp-experiments-randbet}: ablation for random bit error training (\Random, \tabref{tab:supp-randbet-symmetric}, \ref{tab:supp-randbet-variants}, \ref{tab:supp-randbet-resnet}).
		\item \secref{subsec:supp-plc}: ablation and details for per-layer clipping (\PLClipping, \figref{fig:supp-plclipping} and \tabref{tab:supp-plclipping}).
		\item \secref{subsec:supp-randbet-baselines}: additional results on profiled bit errors (\tabref{tab:supp-randbet-generalization} and \ref{tab:supp-randbet-baselines}).
		\item \secref{subsec:experiments-stress}: computation of the guarantees from \secref{subsec:supp-bound}, Prop. \ref{prop:bound} (\tabref{tab:supp-stress}).
		\item \secref{subsec:supp-experiments-summary}: summary results for all datasets and precisions individually (\figref{fig:supp-summary}).
		\item \revision{\secref{subsec:supp-experiments-post}: results for post-training quantization (\tabref{tab:supp-post-robustness}).}
		\item \secref{subsec:supp-experiments-activations-inputs}: additional results and discussion of bit errors in activations and inputs (\tabref{tab:supp-inputs} and \ref{tab:supp-activations}).
		\item \secref{subsec:supp-experiments-adversarial}: more results for adversarial bit errors and adversarial bit error training (\Adv, \figref{fig:supp-adversarial-ablation} and \tabref{tab:supp-bfa}, \ref{tab:supp-adversarial-ablation}, \ref{tab:supp-advbet}).
	\end{itemize}
	\item \tabref{tab:supp-summary-cifar10}, \ref{tab:supp-summary-cifar10-plc} \revision{to \ref{tab:supp-summary-tinyimagenet-2}}: more results on \MNIST, \CifarT, \CifarH and TinyImageNet in tabular form corresponding to \revision{\figref{fig:supp-summary} and} Fig. 7 in the main paper.
\end{itemize}

\begin{figure}[t]
	\centering
	\hspace*{0.4cm}
	\begin{subfigure}{0.475\textwidth}
		\includegraphics[height=4.75cm]{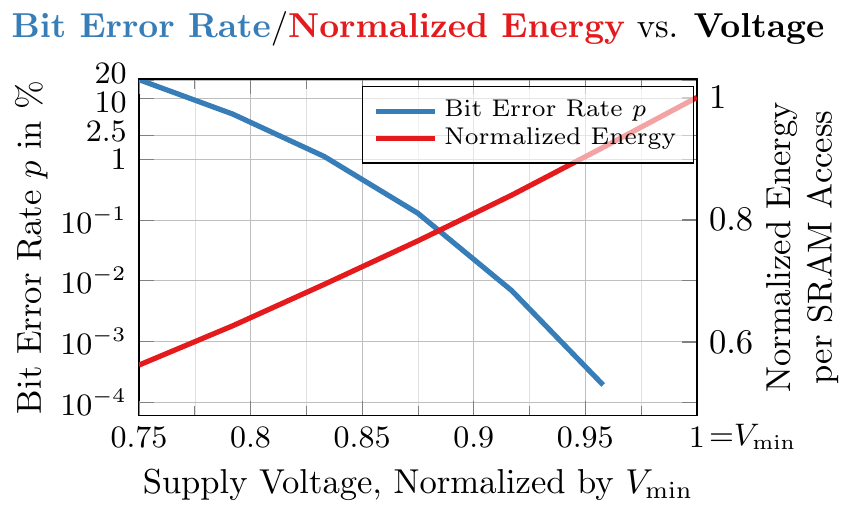}
	\end{subfigure}
	\vspace*{-8px}
	\caption{
		\textbf{Energy and Low-Voltage Operation.} Average bit error rate $p$ ({\color{colorbrewer2}blue}, left y-axis) from $32$ 14nm SRAM arrays of size $512{\times}64$ from \cite{ChandramoorthyHPCA2019} and energy ({\color{colorbrewer1}red}, right y-axis) vs. voltage (x-axis). Voltage is normalized by \Vmin, the minimal measured voltage for error-free operation, as well as the energy per SRAM access at \Vmin. SRAM accesses have significant impact on the DNN accelerator's energy \cite{ChenISCA2016}. Reducing voltage leads to exponentially increasing bit error rates.
	}
	\label{fig:introduction}
\end{figure}
\begin{figure*}[t]
	\setlength{\linewidth}{\textwidth}
	\setlength{\hsize}{\textwidth}
	\centering
	\centering
	\begin{tikzpicture}
		\node[anchor=north west] at (0,0){\includegraphics[width=4cm]{errors_18_2}};
		\node[anchor=north west] at (4,0){\includegraphics[width=4cm]{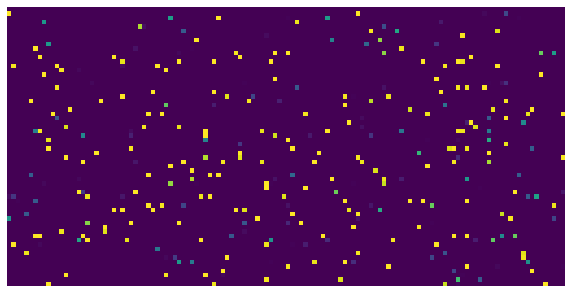}};
		\node[anchor=north west] at (8,0){\includegraphics[width=4cm]{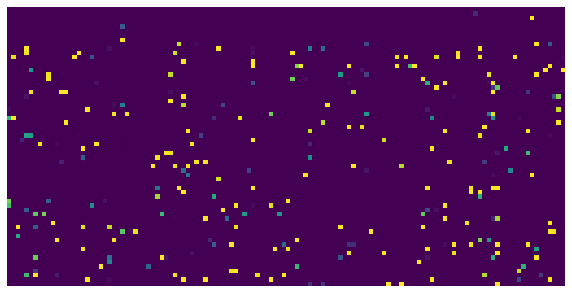}};
		\node[anchor=north west] at (12,0){\includegraphics[width=4cm]{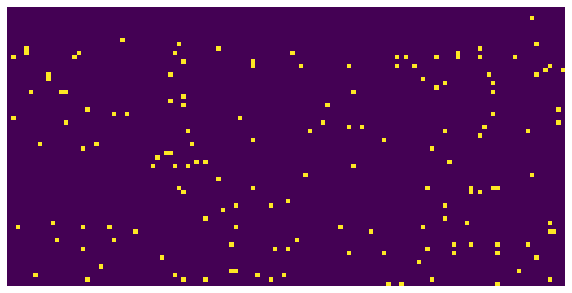}};
		
		\node[anchor=north west] at (0,-2.5){\includegraphics[width=4cm]{errors_n_2}};
		\node[anchor=north west] at (4,-2.5){\includegraphics[width=4cm]{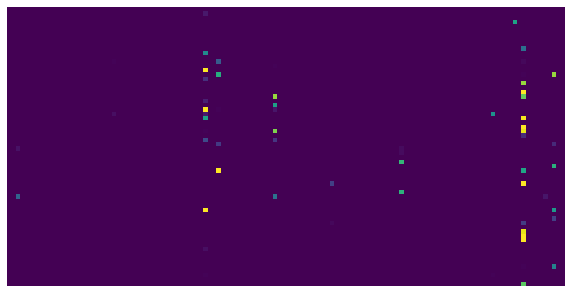}};
		\node[anchor=north west] at (8,-2.5){\includegraphics[width=4cm]{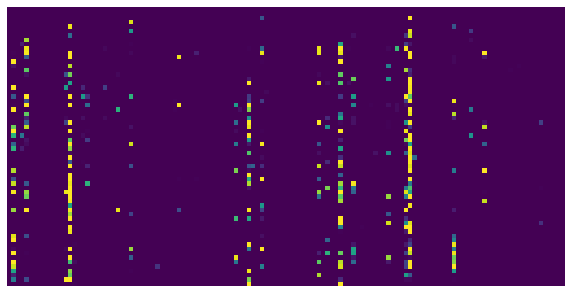}};
		\node[anchor=north west] at (12,-2.5){\includegraphics[width=4cm]{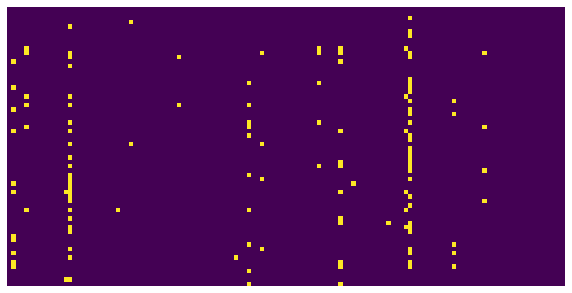}};
		
		\node[anchor=south east,fill opacity=0.75,fill=white] at (4, -2){$p{\approx}2.75\%$};
		\node[anchor=south east,fill opacity=0.75,fill=white] at (4, -4.5){$p{\approx}1.08\%$};
		
		\draw[white!30!black,-, line width=0.75pt] (12.125,-5.1) -- (12.125,0.4);
		\draw[thick,->] (0.05,0.05) -- (1,0.05);
		\draw[thick,->] (-0.05,-0.05) -- (-0.05,-1);
		\node[anchor=south] at (0.75,0.05){128 columns};
		\node[rotate=90,anchor=south] at (-0.05,-0.6){64 rows};
		
		\node[anchor=south] at (6,-0.2) {\bfseries Chip 1};
		\node[anchor=north] at (6,-2.2) {\bfseries Chip 2};
		
		\node at (2, -5){Overall bit flips};
		\node at (4, -5){=};
		\node at (6, -5){$1$-to-$0$ flips};
		\node at (8, -5){+};
		\node at (10, -5){$0$-to-$1$ flips};
		\node at (14, -5){persistent errors};
	\end{tikzpicture}
	\vspace*{-6px}
	\caption{\textbf{Low-Voltage Induced Bit Errors on Profiled Chips.} We break the the bit error distribution of chips 1 and 2 down into $1$-to-$0$ and $0$-to-$1$ bit flips. Additionally, we show that most of the bit errors are actually persistent and, thus, not subject to randomness. As before, we show a sub-array of size $64 \times 128$ from all profiled bit cells (\ie, across all SRAM arrays). \secref{sec:supp-introduction} includes details on profiling.}
	\label{fig:supp-errors}
\end{figure*}

\section{Energy Savings in \figref{fig:introduction}}
\label{sec:supp-introduction}

\figref{fig:introduction} shows bit error rate characterization results of SRAMs in the DNN accelerator chip described in \cite{ChandramoorthyHPCA2019}, fabricated using 14nm FinFET technology. The average bit error rate is measured from 32 SRAMs, each SRAM array of size 4KB (512 $\times$ 64 bit), as supply voltage is scaled down. \textit{Bit error rate} $p$ (in \%) at a given supply voltage is measured as the count of read or write bit cell failures averaged over the total number of bit cells in the SRAM. A bit cell failure refers to reading 1 on writing 0 or reading 0 on writing 1.  For a more comprehensive characterization of SRAMs in 14nm technology, the reader is referred to \cite{GanapathyDAC2017}. \figref{fig:introduction} also shows the energy per write and read access of a 4KB (512 $\times$ 64 bit) SRAM, obtained from Cadence Spectre simulations. Energy is obtained at the same constant clock frequency at all supply voltages. The voltage (x-axis) shown is normalized over \Vmin which is the lowest measured voltage at which there are no bit cell failures.  Energy shown in the graph (secondary axis on the right) is also normalized over the energy per access at \Vmin.

Accelerators such as \cite{ChenISCA2016,ChenASPLOS2014,ChandramoorthyHPCA2019, ReagenISCA2016, nvdla, DuISCA2015,SharmaISCA2018} have a large amount of on-chip SRAM to store weights and intermediate computations. Total dynamic energy of accelerator SRAMs can be obtained as the total number of SRAM accesses times the energy of a single SRAM access. Optimized dataflow in accelerators leads to better re-use of weights read from memories in computation, reducing the number of such memory accesses ~\cite{ChenISCA2016,ChenASPLOS2014, nvdla}. Low voltage operation focuses on reducing the memory access energy, leading to significant energy savings as shown.
	\section{Related Work}
\label{sec:supp-related-work}

\begin{table*}[t]
    \centering
   	\caption{\textbf{Architectures, Number of Weights $\mathbf{W}$, Expected Number of Bit Errors.} \textit{Left and Middle:} SimpleNet architectures used for \MNIST and \CifarT with the corresponding output sizes, channels $N_C$, height $N_H$ and width $N_W$, and the total number of weights $W$. We use group normalization \emph{with} learnable scale/bias, but reparameterized as outlined in \appref{sec:supp-clipping}. \textit{Right:} The number of expected bit errors for random bit errors, \ie, $pmW$. With {\color{colorbrewer4}$\ast$} we mark ``blocks'' of convolutional, normalization and ReLU layer after which we inject bit errors in activations for the experiments in \secref{subsec:supp-experiments-activations-inputs}. \revision{For details on the employed ResNet and Wide ResNet architectures used on \CifarH and TinyImageNet, we refer to \cite{HeCVPR2016,ZagoruykoBMVC2016}.}}
    \label{tab:supp-architectures}
    \vspace*{-0.25cm}
    \begin{subfigure}[t]{.30\textwidth}
        \vspace*{0px}
    
        \footnotesize
\begin{tabular}{|l|c|}
    \hline
    \multicolumn{2}{|c|}{\textbf{SimpleNet} on \textbf{MNIST}}\\
    \hline
    Layer & Output Size\\
    & $N_C, N_H, N_W$\\
    \hline
    \hline
    Conv+GN+ReLU{\color{colorbrewer4}$\ast$} & $32, 28, 28$\\
    Conv+GN+ReLU{\color{colorbrewer4}$\ast$} & $64, 28, 28$\\
    Conv+GN+ReLU{\color{colorbrewer4}$\ast$} & $64, 28, 28$\\
    Conv+GN+ReLU{\color{colorbrewer4}$\ast$} & $64, 28, 28$\\
    Pool & $64, 14, 14$\\
    Conv+GN+ReLU{\color{colorbrewer4}$\ast$} & $64, 14, 14$\\
    Conv+GN+ReLU{\color{colorbrewer4}$\ast$} & $64, 14, 14$\\
    Conv+GN+ReLU{\color{colorbrewer4}$\ast$} & $128, 14, 14$\\
    Pool & $128, 7, 7$\\
    Conv+GN+ReLU{\color{colorbrewer4}$\ast$} & $256, 7, 7$\\
    Conv+GN+ReLU{\color{colorbrewer4}$\ast$} & $1024, 7, 7$\\
    Conv+GN+ReLU{\color{colorbrewer4}$\ast$} & $128, 7, 7$\\
    Pool & $128, 3, 3$\\
    Conv+GN+ReLU{\color{colorbrewer4}$\ast$} & $128, 3, 3$\\
    Pool & $128, 1, 1$\\
    FC & $10$\\
    \hline
    \hline
    $W$ & 1,082,826\\
    \hline
\end{tabular}
    \end{subfigure}
    \begin{subfigure}[t]{.30\textwidth}
        \vspace*{0px} 
        
        \footnotesize
\begin{tabular}{|l|c|}
    \hline
    \multicolumn{2}{|c|}{\textbf{SimpleNet} on \textbf{\CifarT}}\\
    \hline
    Layer & Output Size\\
    & $N_C, N_H, N_W$\\
    \hline
    \hline
    Conv+GN+ReLU{\color{colorbrewer4}$\ast$} & $64, 32, 32$\\
    Conv+GN+ReLU{\color{colorbrewer4}$\ast$} & $128, 32, 32$\\
    Conv+GN+ReLU{\color{colorbrewer4}$\ast$} & $128, 32, 32$\\
    Conv+GN+ReLU{\color{colorbrewer4}$\ast$} & $128, 32, 32$\\
    Pool & $128, 16, 16$\\
    Conv+GN+ReLU{\color{colorbrewer4}$\ast$} & $128, 16, 16$\\
    Conv+GN+ReLU{\color{colorbrewer4}$\ast$} & $128, 16, 16$\\
    Conv+GN+ReLU{\color{colorbrewer4}$\ast$} & $256, 16, 16$\\
    Pool & $256, 8, 8$\\
    Conv+GN+ReLU{\color{colorbrewer4}$\ast$} & $256, 8, 8$\\
    Conv+GN+ReLU{\color{colorbrewer4}$\ast$} & $256, 8, 8$\\
    Pool & $256, 4, 4$\\
    Conv+GN+ReLU{\color{colorbrewer4}$\ast$} & $512, 4, 4$\\
    Pool & $512, 2, 2$\\
    Conv+GN+ReLU{\color{colorbrewer4}$\ast$} & $2048, 2, 2$\\
    Conv+GN+ReLU{\color{colorbrewer4}$\ast$} & $256, 2, 2$\\
    Pool & $256, 1, 1$\\
    Conv+GN+ReLU{\color{colorbrewer4}$\ast$} & $256, 1, 1$\\
    Pool & $256, 1, 1$\\
    FC & $10$\\
    \hline
    \hline
    $W$ & 5,498,378\\
    \hline
\end{tabular}
    \end{subfigure}
    \begin{subfigure}[t]{.20\textwidth}
        \vspace*{0px}
        
        \footnotesize
\begin{tabular}{|l|c|}
    \hline
    \multicolumn{2}{|c|}{$\mathbf{p}$ on \textbf{MNIST}}\\
    \hline
    $p$ in \% & $pmW$, $m = 8$\\
    \hline
    \hline
    \multicolumn{2}{|c|}{\textit{Random} Bit Errors}\\
    \hline
    $10$ & 866260\\
    $5$ & 433130\\
    $1.5$ & 129939\\
    $1$ & 86626\\
    $0.5$ & 43313\\
    \hline
\end{tabular}
        
        \footnotesize
\begin{tabular}{|l|c|}
    \hline
    \multicolumn{2}{|c|}{$\mathbf{p}$ on \textbf{\Cifar}}\\
    \hline
    $p$ in \% & $pmW$, $m = 8$\\
    \hline
    \hline
    \multicolumn{2}{|c|}{\textit{Random} Bit Errors}\\
    \hline
    $1$ & 439870\\
    $0.5$ & 219935\\
    $0.01$ & 43987\\
    \hline
\end{tabular}
    \end{subfigure}
\end{table*}

In the following, we briefly review work on adversarial robustness, fault tolerance and backdooring. These areas are broadly related to the topic of the main paper.

\revision{\textbf{Adversarial Robustness:} Robustness of DNNs against adversarially perturbed or randomly corrupted inputs received considerable attention in recent years, see, \eg, relevant surveys \cite{BiggioARXIV2018,XuARXIV2019}. Adversarial examples \cite{SzegedyARXIV2013}
have been shown to be possible in a white-box setting, with full access to the DNN, \eg, \cite{MadryICLR2018,MoosaviCVPR2016,CarliniSP2017,DongARXIV2017,ChiangARXIV2019,CroceARXIV2020,RozsaBMVC2017,UesatoICML2018,LiuARXIV2018,RonyICCV2021,GowalARXIV2019b,ZhengAAAI2019,BrendelNIPS2019}, as well as in a black-box setting, without access to DNN weights and gradients, \eg, \cite{ChenAISEC2017,IlyasICML2018,AndriushchenkoARXIV2019,BrunnerARXIV2019,CroceICML2020b,LiICCV2021b,GuoICML2019,LiCVPR2020,ShiCVPR2020,ChenSP2020,ChengICLR2019,ZhaoAAAI2020,ChenKDD2020,AldujailiICLR2020,ChengICLR2020,ChengICLR2020,ChenECCV2020,ShuklaKDD2021,BhagojiECCV2018,TuAAAI2019,SuyaUSENIX2020,MahoCVPR2021,DongPAMI2021}. Such attacks are also transferable between models \cite{LiuARXIV2016,PapernotARXIV2016,PapernotASIACCS2017,PapernotSP2016b,DongCVPR2019,DemontisUSENIX2019} and can be applied in the physical world \cite{LuARXIV2017,KurakinARXIV2016b,LiICML2019,XuECCV2020}.
Recent benchmarks generally combine white- and black-box attacks for reliable evaluation \cite{CroceARXIV2020,YaoARXIV2021}.
Adversarial inputs have also been considered for quantized DNNs \cite{KhalilICLR2019}.}

\revision{Obtaining robustness against adversarial inputs is challenging and many approaches have been proposed, including regularization schemes \cite{LyuICDM2015,HeinARXIV2017,JakubovitzARXIV2018,RossARXIV2017,HoffmanARXIV2019,YuARXIV2018,YangARXIV2020b,ChanARXIV2019b,RahnamaARXIV2019,LiuARXIV2020c},
ensemble methods \cite{ZhouARXIV2018,LiuARXIV2017,StraussARXIV2017,ZahavyARXIV2016,HeARXIV2017,SenARXIV2020},
distillation \cite{PapernotSP2016,GoldblumARXIV2019,WangICML2018},
pre-processing or dimensionality-reduction approaches \cite{XuARXIV2017,BhagojiARXIV2017,BuckmanICLR2018,PrakashARXIV2018b,MiyazatoICIP2019,QiuARXIV2020,ShahafiOPENREVIEW2018,MoosaviARXIV2018},
or detection schemes 
\cite{LiICCV2017,BhagojiARXIV2017,FeinmanARXIV2017,GongARXIV2017,GrosseARXIV2017,HendrycksICLR2017,MetzenARXIV2017,ZhengNIPS2018,SmithUAI2018,LiaoCVPR2018,MaARXIV2018,AmsalegWIFS2017,LeeNIPS2018,RothICML2019,LiuCVPR2019,ShanCCS2020,SperlEUROSP2020,CohenCVPR2020,TianAAAI2021}, to name just a selection.
Unfortunately, a significant part can be broken using adaptive attacks, \eg, in \cite{CarliniARXIV2017b,CarliniARXIV2016,CarliniARXIV2019b,AthalyeARXIV2018b,CarliniARXIV2017}.
Recent work focuses on achieving certified/provable robustness \cite{CohenARXIV2019,YangARXIV2020,KumarARXIV2020b,ZhangNIPS2018,ZhangARXIV2019,WongICML2018,GowalARXIV2019,GehrSP2018,MirmanICML2018,SinghNIPS2018,LeeARXIV2019} and adversarial training \cite{MadryICLR2018}, \ie, training on adversarial inputs generated on-the-fly. Adversarial training has been shown to work well empirically, and flaws such as reduced accuracy \cite{StutzCVPR2019,TsiprasARXIV2018} or generalization to attacks not seen during training has been addressed repeatedly \cite{CarmonNIPS2019,UesatoARXIV2019,StutzICML2020,TramerARXIV2019,MainiARXIV2019}
Since, many variants of adversarial training have been proposed, using instance-aware threat models \cite{BalajiARXIV2019,DingARXIV2018}, curriculum training \cite{YuARXIV2019,CaiIJCAI2018}, or regularizers \cite{PangNIPS2020,ZhangNIPS2019,WanECCV2020,BuiECCV2020,RakinCVPR2019,LiICCV2021}, and many more \cite{LambARXIV2018b,WangNIPS2020,ChengARXIV2020,XieARXIV2020,ZhangARXIV2020,ZiICCV2021}.}

\revision{\textbf{Corruption Robustness:}} Corrupted inputs, in contrast, consider ``naturally'' occurring corruptions to which robustness/invariance is desirable for practical applications. Popular benchmarks such as MNIST-C \cite{MuICMLWORK2019}, Cifar10-C or ImageNet-C \cite{HendrycksARXIV2019} promote research on corruption robustness by extending standard datasets with common corruptions, \eg, blur, noise, saturation changes \etc. It is argued that adversarial robustness, and robustness to random corruptions is related. Approaches are often similar, \eg, based on adversarial training \cite{StutzICML2020,LopesICMLWORK2019,KangARXIV2019}. In contrast, we mainly consider random bit errors in the weights, not the inputs.

\textbf{Fault Tolerance:} Fault tolerance, describes structural changes such as removed units, and has been studied in early works such as \cite{AlippiISCAS1994,NetiTNN1992,Chiu1994}. These approaches obtain fault tolerant NNs using approaches similar to adversarial training \cite{DeodhareTNN1998,LeeICASSP2014}.
Recently, hardware mitigation strategies \cite{MarquesSIPS2017}, weight regularization \cite{RahmanICIP2018,DeyTSMCS2018,LeungTNN2010}, fault detection \cite{XiaDAC2017} or GAN-based training \cite{DudduARXIV2019} has been explored. Generally, a wide range of different faults/errors are considered, including node faults \cite{LeeICASSP2014,DeodhareTNN1998}, hardware soft errors \cite{AziziMazreahNAS2018}, timing errors \cite{DengDATE2015} or transient errors in general \cite{SalamiSBACPAD2018}. However, to the best of our knowledge, large rates of non-transient bit errors provoked through low-voltage operation has not been considered. Nevertheless, some of these approaches are related to ours in spirit: \cite{DuASPDAC2014} consider inexact computation for energy-efficiency and \cite{CavalieriOJINN1999,KlachkoIJCNN2019,HoangDATE2020} constrain weights and/or activations to limit the impact of various errors -- similar to our weight clipping.
Additionally, fault tolerance of adversarially robust models has been considered in \cite{DudduARXIV2019b}. We refer to \cite{TorreshuitzilIEEEACCESS2017} for a comprehensive survey. In contrast, we do \emph{not} consider structural changes/errors in DNNs.

\textbf{Backdooring:} The goal of backdooring is to introduce a backdoor into a DNN, allowing to control the classification result by fixed input perturbations at test time. This is usually achieved through data poisoning  \cite{LiuNDSS2018,LiaoARXIV2018,ZhangASIACCS2018}. However, some works also consider directly manipulating the weights \cite{JiCCS2018,DumfordARXIV2018}. However, such weight perturbations are explicitly constructed not to affect accuracy on test examples without backdoor. In contrast, we consider random bit errors (\ie, weight perturbations) that degrade accuracy significantly.
	\section{Low-Voltage Induced Random Bit Errors in Quantized DNN Weights}
\label{sec:supp-main}

We provide a more detailed discussion of the considered error model: random bit errors, induced through low-voltage operation of SRAM or DRAM commonly used on DNN accelerators \cite{KimDATE2018,KoppulaMICRO2019}. Work such as \cite{ChandramoorthyHPCA2019,KoppulaMICRO2019} model the effect of low-voltage induced bit errors using two parameters: the probability $\pfault$ of bit cells in accelerator memory, being faulty and the probability $\perror$ that a faulty bit cell results in a bit error on access.
Following measurements in works such as \cite{GanapathyHPCA2019,KimDATE2018}, we assume that these errors are \emph{not} transient errors by setting $\perror = 100\%$ such that the overall probability of bit errors is $p := \pfault \cdot \perror = \pfault$. In doing so, we consider the worst-case where faulty bit cells \emph{always} induce bit errors. However, the noise model from the main paper remains valid for any arbitrary but fixed $\perror \neq 100\%$. For the reminder of this document, we assume the probability of bit error $p = \pfault$, with $\perror = 100\%$, as in the main paper. In the following, we describe the two parameters, $\pfault$ and $\perror$, in more details.

\textbf{Faulty Bit Cells.} Due to variations in the fabrication process, SRAM bit cells become more or less vulnerable to low-voltage operation. For a specific voltage, the resulting bit cell failures can be assumed to be random and independent of each other. We assume a bit to be faulty with probability $\pfault$ increasing exponentially with decreased voltage \cite{GanapathyDAC2017,GanapathyHPCA2019,KimDATE2018,ChandramoorthyHPCA2019}. Furthermore, the faulty bits for $\pfault' \leq \pfault$ can be assumed to be a subset of those for $\pfault$. For a fixed chip, consisting of multiple memory arrays, the pattern (spatial distribution) of faulty cells is fixed for a specific supply voltage. Across chips/memory arrays, however, faulty cells are assumed to be random and independent of each other.

\textbf{Bit Errors in Faulty Bit Cells:} Faulty cells may cause bit errors with probability $\perror$ upon read/write access.
We note that bit errors read from memory affect \emph{all} computations performed on the read weight value. 
We assume that a bit error flips the currently stored bit, where flips $0$-to-$1$ and $1$-to-$0$ are assumed equally likely.

\subsection{Profiled Bit Errors}
\label{subsec:supp-errors-profiled}

\figref{fig:supp-errors} splits the bit error distributions of Fig. 3 in the main paper into a $0$-to-$1$ flip and a $1$-to-$0$ bit flip map. The obtained maps, $p_{\text{1t0}}$ and $p_{\text{0t1}}$, contain per-bit flip probabilities for $1$-to-$0$ and $0$-to-$1$ bit flips. In this particular profiled chip, \figref{fig:supp-errors} (bottom), $0$-to-$1$ flips are more likely. Similarly, \figref{fig:supp-errors} (right) shows that most  $0$-to-$1$ flips are actually persistent across time \ie, not random transient errors. 
The following table summarizing the key statistics of the profiled chips: the overall bit error rate $p$, the rate of $1$-to-$0$ and $0$-to-$1$ flips $p_{\text{1t0}}$ and $p_{\text{0t1}}$, and the rate of persistent errors $p_{\text{sa}}$, all in \%:

\begin{table}[t]
    \centering
    \caption{\revision{\textbf{Signal-to-Noise Ratios (SNRs) for Quantization and Bit Errors:} The impact of random bit errors is more severe than quantization errors, which are fixed after training. This results in substantially lower SNRs in dB.}}
    \label{tab:supp-sqnr}
    \vspace*{-0.1cm}
    \begin{tabular}{|l|c|c|c|c|}
        \hline
        Model & SQNR & \multicolumn{3}{c|}{SNR}\\
        \cline{2-5}
        && $p = 0.01\%$ & $p = 0.1\%$ & $p = 1\%$\\
        \hline
        \hline
        \Normal & 33.60 & 19.18 & 9.28 & \red{-0.69}\\
        \Clipping[$0.1$] & 36.19 & 22.19 & 12.21 & 2.27\\
        \hline
    \end{tabular}
\end{table}

\begin{center}
\footnotesize
\begin{tabular}{| l | c | c | c | c |}
	\hline
	Chip & $p$ & $p_{\text{0t1}}$ & $p_{\text{1t0}}$ & $p_{\text{sa}}$ \\
	\hline
	\multirow{2}{*}{1} & 2.744 & 1.27 & 1.47 & 1.223\\ 
	& 0.866 & 0.38 & 0.49 & 0.393\\ 
	\hline
	\multirow{3}{*}{2} & 4.707 &  3.443 & 1.091 & 0.627\\ 
	& 1.01 &  0.82 & 0.19 & 0.105\\ 
	& 0.136 & 0.115 & 0.021 & 0.01 \\ 
	\hline
	\multirow{2}{*}{3} & 2.297 & 1.81 & 0.48 & 0.204 \\ 
	& 0.597 & 0.496 & 0.0995 & 0.206 \\ 
	\hline
\end{tabular}
\end{center}

For evaluation, we assume that the DNN weights are mapped linearly onto the memory of these chips. The bit error maps are of size $8192 \times 128$ bits for chips 2 and 3 and $2048 \times 128$ bits for chip 1. Furthermore, to simulate various different mappings, we repeat this procedure with various offsets and compute average \RTE across all mappings. For results, we refer to \appref{subsec:supp-randbet-baselines}.

\subsection{Bounding Generalization to Random Bit Errors}
\label{subsec:supp-bound}

Let $w$ denote the final weights of a trained DNN $f$. We test $f$ using $n$ i.i.d. test examples, \ie, $(x_i,y_i)_{i=1}^n$. We denote by $w'$ the weights where each bit of $w$ is flipped with probability $p$ uniformly at random, corresponding to the error model from the main paper.
The expected \emph{clean} error  of $f$ is given by
\begin{align*}
	\Exp[\Id_{f(x;w)\neq y}] = \Pr(f(x;w)\neq y).
\end{align*}
The expected \emph{robust} error (regarding i.i.d. test examples drawn from the data distribution) with random bit errors in the (quantized) weights is
\begin{align*}
	\Exp[ \Id_{f(x;w')\neq y}] = \Pr( f(x;w')\neq y).
\end{align*}
Here, the weights of the neural network are themselves random variables. Therefore, with $x, y, w$, and $w'$ we denote the random variables corresponding to test example, test label, weights and weights with random bit errors. With $x_j, y_j, w_i$ and $w'_i$ we denote actual examples. Then, the following proposition derives a simple, probabilistic bound on the deviation of expected robust error from the empirically measured one (\ie, \RTE in our experiments):

\begin{table}[t]
	\centering
	\caption{\textbf{Quantization-Aware Training Accuracies.} Clean \TE for $m = 8$ bits or lower using our robust fixed-point quantization. We obtain competitive performance for $m = 8$ and $m = 4$ bits. On \CifarH, a Wide ResNet (WRN) and, \revision{on TinyImageNet, a ResNet-18} clearly outperform our standard SimpleNet model. \revision{Batch normalization (BN), while reducing \TE  is significantly less robust than group normalization (GN)}, \cf \tabref{tab:supp-bn}. * For $m \leq 4$, we report results with weight clipping, \Clipping[$0.1$].}
	\label{tab:supp-accuracy}
	\vspace*{-0.25cm}
	\hspace*{-0.25cm}
	\begin{subfigure}[t]{0.19\textwidth}
		\vspace*{0px}
		\begin{tabular}{|@{\hskip 4px}l@{\hskip 4px}|@{\hskip 4px}c@{\hskip 4px}|}
			\multicolumn{2}{c}{\bfseries \CifarT}\\
			\multicolumn{2}{c}{\bfseries SimpleNet+GN}\\
			\hline
			Quant. $m$ & \TE in \%\\
			\hline
			-- & 4.34\\
			8 & \bfseries 4.32\\
			4* & 5.29\\
			3* & 5.71\\
			\hline
		\end{tabular}
	\end{subfigure}
	\begin{subfigure}[t]{0.28\textwidth}
		\vspace*{0px}
		\begin{tabular}{|@{\hskip 4px}l@{\hskip 4px}|@{\hskip 4px}c@{\hskip 4px}|@{\hskip 4px}c@{\hskip 4px}|}
			\multicolumn{3}{c}{\bfseries \CifarT}\\
			\multicolumn{3}{c}{\bfseries Arch. Comparison}\\
			\hline
			Model & no Quant. & $m = 8$\\
			\hline
			SimpleNet+GN & 4.34 & 4.32\\
			SimpleNet+BN & 4.04 & 3.83\\
			ResNet-50+GN & 5.88 & 6.81\\
			ResNet-50+BN & \bfseries 3.91 & \bfseries 3.67\\
			\hline
		\end{tabular}
	\end{subfigure}\\[4px]
	
	\hspace*{-0.5cm}
	\begin{subfigure}[t]{0.25\textwidth}
		\vspace*{0px}
		\small
		\begin{tabular}{| l | c |}
			\multicolumn{2}{c}{\bfseries \CifarH}\\
			\multicolumn{2}{c}{\bfseries Quant. + Arch. Comparison}\\
			\hline
			Quant. $m$, Model & \TE in \%\\
			\hline
			8, SimpleNet & 23.68\\
			8, WRN & 18.53\\
			\hline
		\end{tabular}
	\end{subfigure}
	\begin{subfigure}[t]{0.21\textwidth}
		\vspace*{0px}
		\small
		\revision{
			\begin{tabular}{| l | c |}
				\multicolumn{2}{c}{\bfseries TinyImageNet}\\
				\multicolumn{2}{c}{\bfseries Arch. Comparison}\\
				\hline
				Model & \TE in \%\\
				\hline
				SimpleNet+BN & 35.91\\
				SimpleNet+GN & 41.24\\
				ResNet-18+BN & 32.72\\
				ResNet-18+GN & 36.39\\
				\hline
			\end{tabular}
		}
	\end{subfigure}
	\vspace*{-0.1cm}
\end{table}

\begin{proposition}
	\label{prop:bound}
	Let $w'_i$, $i = 1,\ldots,l$ be $l$ examples of weights with bit errors (each bit flipped with probability $p$). Then it holds
	\begin{align*}
		\Pr\Big( \frac{1}{nl}&\sum_{j=1}^n \sum_{i=1}^l \Id_{f(x_j;w'_i)\neq y_j} - \Pr(f(x;w')\neq y)\geq \epsilon\Big)\\
		&\quad\leq (n+1) e^{-n\epsilon^2 \frac{l}{(\sqrt{l}+\sqrt{n})^2}}.
	\end{align*}
	As alternative formulation, with probability $1-\delta$ it holds
	\begin{align*}
		\Pr( f(x; w'_i)\neq y) < & \frac{1}{nl}\sum_{j=1}^n \sum_{i=1}^l \Id_{f(x_j; w'_i)\neq y_j}\\ &+ \quad \leq\sqrt{\frac{\log\Big(\frac{n+1}{\delta}\Big)}{n}} \frac{\sqrt{l}+\sqrt{n}}{\sqrt{l}}.
	\end{align*} 
\end{proposition}
{\small
\begin{proof}
Let $0<\alpha<1$. Using the Hoeffding inequality and union bound, we have:
\begin{align*}
	&\Pr\Big(\maxop_{j=1,\ldots,n} \frac{1}{l}\sum_{i=1}^l \Id_{f(x_j; w'_i)\neq y_j} - \Exp_{w'}[\Id_{f(x_j; w')\neq y_j}] > \alpha\epsilon\Big) \\
	=& \Pr\Big(\bigcup_{j=1,\ldots,n}\big\{ \frac{1}{l}\sum_{i=1}^l \Id_{f(x_j; w'_i)\neq y_j} - \Exp_{w'}[\Id_{f(x_j; w')\neq y_j}] > \alpha\epsilon\big\}\Big)\\
	&\leq \; n\, e^{-l\alpha^2 \epsilon^2}.
\end{align*}
Then, again by Hoeffding's inequality, it holds:
\begin{align*}
	&\Pr\Big( \frac{1}{n}\sum_{j=1}^n \Exp_{w'}[\Id_{f(x_j; w')\neq y_j}]
	- \Exp_{x,y}[\Exp_{w'}[\Id_{f(x; w')\neq y}]] > (1-\alpha) \epsilon \Big)\\
	&\leq \; e^{-n\epsilon^2 (1-\alpha)^2}.
\end{align*}
Thus, using
\begin{align*}
	a + b>\epsilon \Longrightarrow \{a > \alpha \epsilon\} \cup \{b > (1-\alpha)\epsilon\}
\end{align*}
gives us:
\begin{align*}
	&\Pr\Big( \frac{1}{nl}\sum_{j=1}^n \sum_{i=1}^l \Id_{f(x_j; w'_i)\neq y_j} - \Pr( f(x; w')\neq y)\geq \epsilon\Big)\\
	=& \Pr\Big( \frac{1}{n}\sum_{j=1}^n \big(\frac{1}{l}\sum_{i=1}^l \Id_{f_{w'_i}(x_j)\neq y_j} - \Exp_{w'}[\Id_{f(x_j; w'_i)\neq y_j}]\big)\\
	 +& \frac{1}{n}\sum_{j=1}^n \Exp_{w'}[\Id_{f(x_j; w'_i)\neq y_j}] - \Pr( f(x; w')\neq y)\geq \epsilon\Big)\\
	\leq & \Pr\Big( \frac{1}{n}\sum_{j=1}^n \big(\frac{1}{l}\sum_{i=1}^l \Id_{f(x_j; w'_i)\neq y_j} - \Exp_{w'}[\Id_{f(x_j; w')\neq y_j}]\big)>\alpha \epsilon\Big)\\
	+& \Pr\Big( \frac{1}{n}\sum_{j=1}^n \Exp_{w'}[\Id_{f(x_j; w')\neq y_j}] - \Pr( f(x; w')\neq y)\geq (1-\alpha)\epsilon\Big)\\
	\leq &  n\, e^{-l\alpha^2 \epsilon^2} + e^{-n\epsilon^2 (1-\alpha)^2} 
\end{align*}
Having both exponential terms have the same exponent yields $\alpha=\frac{\sqrt{n}}{\sqrt{l}+\sqrt{n}}$, and we get the upper bound of the proposition.
\end{proof}
}

\begin{table}[t]
	\centering
	\caption{\textbf{Batch Normalization not Robust.} \RTE with group normalization (GN) or batch normalization (BN). \RTE increases when using BN even though clean \TE improves slightly compared GN. However, using batch statistics at test time (\ie, ``training mode'' in PyTorch) improves \RTE significantly indicating that the statistics accumulated throughout training do not account for random bit errors. We use \textbf{group normalization as default.}}
	\label{tab:supp-bn}
	\vspace*{-0.25cm}
	\hspace*{-0.25cm}
	\begin{tabular}{| c | l | c | c | c |}
		\hline
		\multicolumn{5}{|c|}{\bfseries \CifarT ($\mathbf{m = 8}$ bit): robustness of BN}\\
		\hline
		&& \multirow{2}{*}{\begin{tabular}{@{}c@{}}\TE\\in \%\end{tabular}} & \multicolumn{2}{c|}{\RTE in \%}\\
		\hline
		&& & $p{=}0.1$ & $p{=}0.5$\\
		\hline
		\hline
		\multirow{2}{*}{GN} & \Normal & 4.32 & 5.54 & 11.28\\
		& \Clipping[$0.1$] & 4.82 & 5.58 & 6.95\\
		\hline
		\hline
		\multicolumn{5}{|c|}{\textbf{BN w/ \emph{Accumulated} Statistics}}\\
		\hline
		\multirow{2}{*}{BN} & \Normal & 3.83 & 6.36 & 52.52\\
		& \Clipping[$0.1$] & 4.46 & 5.32 & 8.25\\
		\hline
		\hline
		\multicolumn{5}{|c|}{\textbf{BN w/ Batch Statistics at Test Time}}\\
		\hline
		\multirow{2}{*}{BN} & \Normal & 3.83 & 6.65 & 9.63\\
		& \Clipping[$0.1$] & 4.46 & 6.57 & 7.29\\
		\hline
	\end{tabular}
	\vspace*{-0.1cm}
\end{table}
\begin{table*}[t]
	\centering
	\caption{\textbf{Impact of Quantization Scheme on Robustness.} We report \TE and \RTE for various bit error rates $p$ for the quantization scheme in \eqnref{eq:supp-quantization} with global, per-layer and asymmetric quantization, $m = 8$ bits. Instead of quantizing into signed integer, using unsigned integers works better for asymmetric quantization. Furthermore, proper rounding instead of integer conversion also improves robustness. Note that influence on clean \TE is negligible, \ie, the DNN can ``learn around'' these difference in quantization-aware training. Especially for $m = 4$ bit, the latter makes a significant difference in terms of robustness.}
	\label{tab:supp-quantization}
	\vspace*{-0.25cm}
	\begin{tabular}{| c | l | c | c | c | c | c | c | c |}
		\hline
		\multicolumn{9}{|c|}{\bfseries \CifarT: quantization robustness}\\
		\hline
		& Model & \multirow{2}{*}{\begin{tabular}{c}\TE\\in \%\end{tabular}} & \multicolumn{6}{c|}{\RTE in \%, $p$ in \% p=0.01}\\
		\cline{4-9}
		& (see text) && $0.01$ & $0.05$ & $0.1$ & $0.5$ & $1$ & $1.5$\\
		\hline
		\hline
		\multirow{5}{*}{\rotatebox{90}{$m = 8$ bit}} & \eqnref{eq:supp-quantization}, global & 4.63 & 10.70 {\color{gray}\scriptsize ${\pm}$1.37} & 86.01 {\color{gray}\scriptsize ${\pm}$3.65} & 90.36 {\color{gray}\scriptsize ${\pm}$0.66} & 90.71 {\color{gray}\scriptsize ${\pm}$0.49} & 90.57 {\color{gray}\scriptsize ${\pm}$0.43} & --\\
		& \eqnref{eq:supp-quantization}, per-layer (= \Normal) & 4.36 & 4.82 {\color{gray}\scriptsize ${\pm}$0.07} & 5.51 {\color{gray}\scriptsize ${\pm}$0.19} & 6.37 {\color{gray}\scriptsize ${\pm}$0.32} & 24.76 {\color{gray}\scriptsize ${\pm}$4.71} & 72.65 {\color{gray}\scriptsize ${\pm}$6.35} & 87.40 {\color{gray}\scriptsize ${\pm}$2.47}\\
		& +asymmetric & 4.36 & 5.76 {\color{gray}\scriptsize ${\pm}$0.09} & 6.47 {\color{gray}\scriptsize ${\pm}$0.22} & 7.85 {\color{gray}\scriptsize ${\pm}$0.46} & 40.78 {\color{gray}\scriptsize ${\pm}$7.56} & 76.72 {\color{gray}\scriptsize ${\pm}$7.01} & 85.83 {\color{gray}\scriptsize ${\pm}$2.58}\\
		& +unsigned & 4.42 & 6.58 {\color{gray}\scriptsize ${\pm}$0.13} & 6.97 {\color{gray}\scriptsize ${\pm}$0.28} & 7.49 {\color{gray}\scriptsize ${\pm}$0.41} & 17.00 {\color{gray}\scriptsize ${\pm}$2.77} & 54.57 {\color{gray}\scriptsize ${\pm}$8.58} & 83.18 {\color{gray}\scriptsize ${\pm}$3.94}\\
		& +rounded (= \Quant) & 4.32 & 4.60 {\color{gray}\scriptsize ${\pm}$0.08} & 5.10 {\color{gray}\scriptsize ${\pm}$0.13} & 5.54 {\color{gray}\scriptsize ${\pm}$0.2} & 11.28 {\color{gray}\scriptsize ${\pm}$1.47} & 32.05 {\color{gray}\scriptsize ${\pm}$6} & 68.65 {\color{gray}\scriptsize ${\pm}$9.23}\\
		\hline
		\hline
		\multirow{2}{*}{\rotatebox{90}{$4$ bit}} & integer conversion & 5.81 & 90.46 {\color{gray}\scriptsize ${\pm}$0.2} & 90.40 {\color{gray}\scriptsize ${\pm}$0.21} & 90.39 {\color{gray}\scriptsize ${\pm}$0.22} & 90.36 {\color{gray}\scriptsize ${\pm}$0.2} & 90.36 {\color{gray}\scriptsize ${\pm}$0.22} & 90.39 {\color{gray}\scriptsize ${\pm}$0.22}\\
		& proper rounding & 5.29 & 5.49 {\color{gray}\scriptsize ${\pm}$0.04} & 5.75 {\color{gray}\scriptsize ${\pm}$0.06} & 5.99 {\color{gray}\scriptsize ${\pm}$0.09} & 7.71 {\color{gray}\scriptsize ${\pm}$0.36} & 10.62 {\color{gray}\scriptsize ${\pm}$1.08} & 15.79 {\color{gray}\scriptsize ${\pm}$2.54}\\
		\hline
	\end{tabular}
	\vspace*{-0.1cm} 
\end{table*}

\textbf{Remarks:}
The samples of bit error injected weights $\{w'_i\}_{i=1}^l$ can actually be different for any test example $(x_j, y_j)$, even though this is not the case in our evaluation. Thus, the above bound involves a stronger result: for any test example, the empirical test error with random bit errors (\ie, robust test error \RTE) and the expected one have to be similar with the same margin. Note also that this bound holds for any fixed bit error distribution as the only requirement is that the bit error patterns we draw are i.i.d. but not the bit errors on the pattern.
In \appref{subsec:experiments-stress}, we consider results with $l = 10^6$, \ie, $l \gg n$ with $n = 10^4$ on \CifarT such that $\nicefrac{l}{(\sqrt{l}+\sqrt{n})^2}$ tends towards one. With $\delta=0.99$ the excess term $\sqrt{\frac{\log\Big(\frac{n+1}{\delta}\Big)}{n}} \frac{\sqrt{l}+\sqrt{n}}{\sqrt{l}}$ in the Proposition is equal to $4.1\%$. Thus larger test sets would be required to get stronger guarantees, \eg, for $n=10^5$ one would get $1.7\%$.

\section{Adversarial Bit Errors}
\label{sec:supp-main-adversarial}

As introduced in the main paper, our adversarial bit error attack can be formulated as the following optimization problem on a fixed mini-batch of examples $\{(x_b,y_b)\}_{b = 1}^B$:
\begin{align}
	\begin{split}
	    &\max_{\tilde{v}} \sum_{b = 1}^B \mathcal{L}(f(x_b; Q^{-1}(\tilde{v})), y_b)\\
	    \text{s.t. }& d_H(\tilde{v}, v) \leq \epsilon,\quad d_H(\tilde{v}_i, v_i) \leq 1
 \end{split}\label{eq:supp-attack}
\end{align}
where $\tilde{v}$ are the quantized weights (signed or unsigned $m$-bit integers) and $d_H$ denotes the (bit-level) Hamming distance. The total number of bit errors $d_H(\tilde{v}, v)$ is constrained by $\epsilon$, and we allow at most one bit error per weight value, \ie, $d_H(\tilde{v}_i, v_i) \leq 1$. These constraints are enforced through projection, after iteratively computing:
\begin{align}
	\begin{split}
	    &\tilde{w}^{(t + 1)} = \tilde{w}^{(t)} + \gamma \Delta^{(t)}\quad\text{ with }\\
	    &\Delta^{(t)} = \sum_{b = 1}^B \mathcal{L}(f(x_b; \tilde{w}_q^{(t)}), y_b),\text{ }\tilde{w}_q^{(t)} = Q^{-1}(Q(\tilde{w}^{(t)}))
    \end{split}\label{eq:supp-attack-iterates}
\end{align}
where $\gamma$ is the step size. We note that the forward pass is performed on the de-quantized weights $\tilde{w}_q^{(t)} = Q^{-1}(Q(\tilde{w}^{(t)}))$, while the update is performed in floating point.

\begin{figure}[t]
	\centering
	\includegraphics[width=0.4\textwidth]{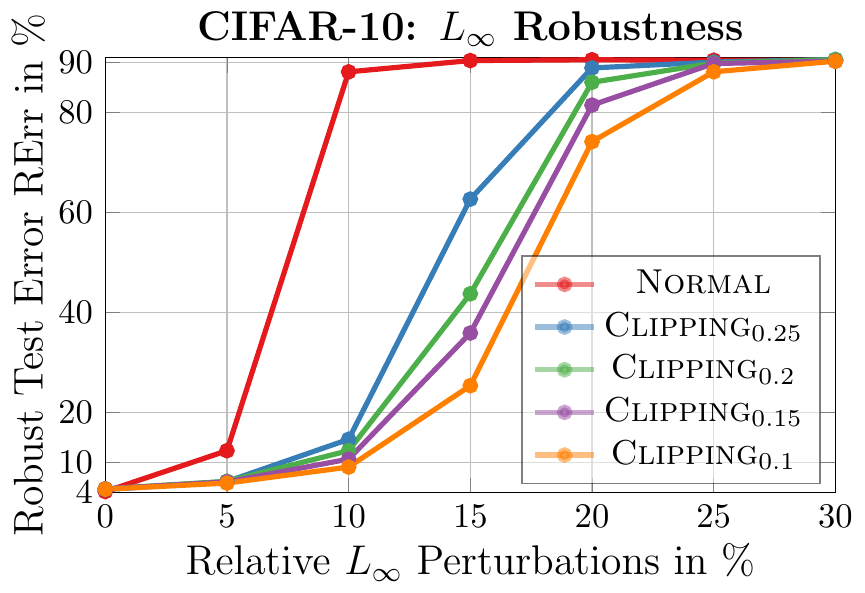}
	\vspace*{-8px}
	\caption{\textbf{Weight Clipping Improves $L_\infty$ Robustness.} On \CifarT, we plot \RTE for \emph{relative} $L_\infty$ perturbations on weights: Random noise with $L_\infty$-norm smaller than or equal to $x\%$ of the weight range is applied. \Clipping clearly improves robustness. Again, the relative magnitude of noise is not affected by weight clipping. Note that $L_\infty$ noise usually affects all weights, while random bit errors affect only a portion of the weights.}
	\label{fig:supp-clipping-inf}
	\vspace*{-0.2cm}
\end{figure}

The projection after the update of \eqnref{eq:supp-attack-iterates} requires solving the following optimization problem:
\begin{align}
	\begin{split}
	    &\min_{\tilde{v}'} \| Q^{-1}(\tilde{v}) - Q^{-1}(\tilde{v}')\|_2^2\\
	    \text{s.t. }& d_H(v_i, \tilde{v}'_i) \leq 1,\quad d_H(v, \tilde{v}') \leq \epsilon
 \end{split}\label{eq:supp-projection}
\end{align}
where we dropped the superscript $t$ for simplicity. Here, $\tilde{v} = Q(\tilde{w})$ are the quantized, perturbed weights after \eqnref{eq:supp-attack-iterates} and $\tilde{w}' = Q^{-1}(\tilde{v}')$ will be the projected weights. As the objective and the constraint set are separable, this problem can be divided into the following two problems: First, we rank the weights by their corresponding changes
\begin{align}
    \left|Q^{-1}(Q(w_i)) - Q^{-1}(\tilde{v}_i)\right| = |w_{q,i} - \tilde{w}_{q,i}|
\end{align}
where $w$ are the original, clean weights and $w_q$ the corresponding de-quantized weights. Then, only the top-$\epsilon$ changes are kept. All other perturbed weights $\tilde{w}_{q,i}$ are reset to the original, clean weights $w_{q,i}$. For the selected weights, only the most significant changed bit is kept. In practice, considering $\tilde{v}_i$ and $v_i$ from \eqnref{eq:supp-projection} corresponding to one of the top-$\epsilon$ changes, if $d_H(\tilde{v}_i, v_i) > 1$, only the highest changed bit is kept. In practice, this can be implemented (and parallelized) easily on the $m$-bit integers $\tilde{v}_i$ and $v_i$ while computing the (bit-level) Hamming distance $d_H$.

The optimization problem \eqnref{eq:supp-attack} is challenging due to the non-convex constraint set that we project onto after each iteration. Therefore, we use several random restarts, each initialized by randomly selecting $k \in [0, \epsilon]$ bits to be flipped in $v$ to obtain $\tilde{v}^{(0)}$. We note that initialization by randomly flipping bits is important as, without initialization, \ie, $\tilde{w}^{(0)} := w$, the loss $\mathcal{L}$ in \eqnref{eq:supp-attack} will be close to zero. We also found that initializing with $k = \epsilon$ leads to difficulties in the first few iterations, which is why we sample $k \in [0, \epsilon]$ uniformly. Additionally, we normalize the gradient $\Delta^{(t)}$ in \eqnref{eq:supp-attack-iterates}:
\begin{align}
    \hat{\Delta}^{(t,l)} = \frac{\Delta^{(t,l)}}{\|\Delta^{(t,l)}\|_\infty}
\end{align}
for each layer $l$ individually (considering biases as separate layer),
before applying the update, \ie, $\tilde{w}^{(t + 1)} = \tilde{w}^{(t)} + \gamma \hat{\Delta}^{(t)}$.
Instead of considering $\tilde{w}^{(T)}$, \ie, the perturbed weights after exactly $T$ iterations, we use
\begin{align}
    \tilde{w}^{(t^*)}\quad\text{with}\quad t^* = \argmax_t \sum_{b = 1}^B \mathcal{L}(f(x_b; \tilde{w}_q^{(t)}), y_b)
\end{align}
instead. Finally, we also use momentum. Nevertheless, despite these optimization tricks, the attack remains very sensitive to hyper-parameters, especially regarding the step-size. Thus, running multiple random restarts are key.

\section{Quantization and Bit Manipulation in PyTorch}
\label{sec:supp-implementation}

Our fixed-point quantization $Q$, as introduced in the main paper, is defined as
\begin{align}
	Q(w_i) = \left\lfloor \frac{w_i}{\Delta}\right\rfloor,\text{  }
	Q^{-1}(v_i) = \Delta v_i,\text{  }
	\Delta = \frac{\qmax}{2^{m - 1} - 1}.
	\label{eq:supp-quantization}
\end{align}
This quantizes weights $w_i \in [-\qmax, \qmax] \subset \mathbb{R}$ into signed integers $\{-2^{m-1}-1,\ldots,2^{m-1}-1\}$. Here, the quantization range $[-\qmax,\qmax]$ is symmetric around zero. Note that zero is represented exactly. To implement asymmetric quantization, the same scheme can be used to quantize weights $w_i \in [\qmin, \qmax]$ within any arbitrary, potentially asymmetric, interval. To this end, \eqnref{eq:supp-quantization} with $\qmax = 1$ is used and the weights in $[\qmin,\qmax]$ are mapped linearly to $[-1, 1]$ using the transformation $N$:
\begin{align}
	N(w_i) = \left(\frac{w_i - \qmin}{\qmax - \qmin}\right)\cdot 2 - 1.\label{eq:asymmetric-quantization}
\end{align}
Generally, $\qmin$ and $\qmax$ are chosen to reflect minimum and maximum weight value -- either from all weights (global quantization) or per-layer.
Furthermore, we argue that asymmetric quantization becomes more robust when using \emph{unsigned} integers as representation. In this case, \eqnref{eq:supp-quantization} can be adapted using a simple additive term:
\begin{align}
	\begin{split}
		Q(w_i) &= \left\lceil \frac{w_i}{\Delta}\right\rfloor + (2^{m - 1} - 1)\\
		Q^{-1}(v_i) &= \Delta (v_i - (2^{m - 1} - 1))
	\end{split}\label{eq:unsigned-quantization}
\end{align}
We use asymmetric quantization using $N$ in \eqnref{eq:asymmetric-quantization} with \eqnref{eq:unsigned-quantization} as our \emph{robust} fixed-point quantization.

We implement ``fake'' fixed-point quantization for quantization-aware training and bit error injection directly in PyTorch \cite{PaszkeNIPSWORK2017}. Here, fake quantization means that computation is performed in floating point, but before doing a forward pass, the DNN is quantized and de-quantized, \ie, $w_q = Q^{-1}(Q(w))$. Note that we quantize into \emph{unsigned} $8$ bit integers, irrespective of the target precision $m \leq 8$. To later induce random bit errors, the $8 - m$ most significant bits (MSBs) are masked for $m < 8$. Bit manipulation of unsigned $8$ bit integers is then implemented in C/CUDA and interfaced to Python using CuPy \cite{cupy} or CFFI \cite{cffi}. These functions can directly operate on PyTorch tensors, allowing bit manipulation on the CPU as well as the GPU. We will make our code publicly available to facilitate research into DNN robustness against random bit errors.

\revision{
\section{Quantization Errors vs. Bit Errors}

To demonstrate that bit errors are significantly more severe than quantization errors, we follow \cite{LinICML2016} and compute the corresponding signal-to-noise-ratios (SNRs).
For quantization, this is generally called signal-to-quantization-noise-ratio (SQNR) which we compute in dB as
\begin{align}
    10 \log_{10} \left(\frac{\mathbb{E}[w_i^2]}{\mathbb{E}[(\tilde{w}_i - w_i)^2]}\right)
\end{align}
where $w_i$ denote weight values and $\tilde{w}_i$ the corresponding \emph{perturbed} weights. For measuring SQNR, the perturbed weights $\tilde{w} = w_q = Q^{-1}(Q(w))$ are the de-quantized ones.
A higher value generally indicates fewer or smaller quantization errors, \ie, better for quantization. Similarly, for random bit errors, we can compute the SNR by additionally injecting bit errors before de-quantization: $\tilde{w} = Q^{-1}(\biterror_p(Q(w)))$. Note that quantization errors are deterministic in our case, \ie, if the weights $w$ are fixed after training, the errors $\tilde{w} - w$ are also fixed.
In contrast, we consider bit errors to be random at test time. In both cases, $\tilde{w}_i - w_i = 0$ is possible. For quantization this means that the value $w_i$ can be quantized without loss. In the case of bit errors, this indicates that the corresponding quantized value $v_i$ was not ``hit'' by bit errors. In practice, for simplicity, we compute the SNRs by approximating the expectations. While quantization is applied on a per-layer basis, we compute SNRs across all layers (computing SNR per-layer leads to the same conclusions).

\begin{table}[t]
	\centering
	\caption{\textbf{Weight Clipping with Weight Scaling.} For group normalization (GN) without the re-parameterization in \secref{sec:supp-clipping}, our DNNs are scale-invariant. Scaling \Quant down to the weight range of \Clipping[$0.25$], however, does not improve robustness. More importantly, scaling \Clipping[$0.05$] up to have the same weight range as \Quant preserves robustness. Thus, the robustness benefit of \Clipping is \emph{not} due to reduced quantization range or smaller absolute errors.}
	\label{tab:supp-clipping-scaling}
	\vspace*{-0.25cm}
	\begin{tabular}{| l | c | c | c |}
		\hline
		\multicolumn{4}{| c |}{\bfseries \CifarT ($\mathbf{m = 8}$ bit): scaling w/o re-parameterized GN}\\
		\hline
		Model & \multirow{2}{*}{\begin{tabular}{@{}c@{}}\TE\\in \%\end{tabular}} & \multicolumn{2}{c|}{\RTE in \%, $p$ in \%}\\
		\hline
		(see text) & & $p{=}0.1$ & $p{=}1$\\
		\hline
		\hline
		\Normal & 4.32 & 5.54 {\color{gray}\scriptsize ${\pm}$0.2} & 32.05 {\color{gray}\scriptsize ${\pm}$6}\\
		\Clipping[$0.1$] & 4.82 & 5.58 {\color{gray}\scriptsize ${\pm}$0.1} & 8.93 {\color{gray}\scriptsize ${\pm}$0.46}\\
		\hline
		\Normal $\rightarrow$ \Clipping[$0.1$] & 4.32 & 5.55 {\color{gray}\scriptsize ${\pm}$0.2} & 88.71 {\color{gray}\scriptsize ${\pm}$2.42}\\
		\Clipping[$0.1$] $\rightarrow$ \Normal & 4.82 & 78.47 {\color{gray}\scriptsize ${\pm}$0.49} & 9.20 {\color{gray}\scriptsize ${\pm}$0.67}\\
		\hline
	\end{tabular}
\end{table}
\begin{table*}[t]
	\centering
	\caption{\textbf{Weight Clipping Improves Robustness.} We report \TE and \RTE for various experiments on the robustness of weight clipping with $\wmax$, \ie, \Clipping[$\wmax$]. First, we show that the robustness benefit of \Clipping is independent of quantization-aware training, robustness also improves when applying post-training quantization. Then, we show results for both symmetric and asymmetric quantization. For the latter we demonstrate that label smoothing \cite{SzegedyCVPR2016} reduces the obtained robustness. This supports our hypothesis that weight clipping, driven by minimizing cross-entropy loss during training, improves robustness through redundancy.}
	\label{tab:supp-clipping}
	\vspace*{-0.25cm}
	\begin{tabular}{| c | l | c | c | c | c | c | c | c |}
		\hline
		\multicolumn{9}{|c|}{\bfseries \CifarT ($\mathbf{m = 8}$ bit): clipping robustness for post- and during-training quantization}\\
		\hline
		&Model & \multirow{2}{*}{\begin{tabular}{c}\TE\\in \%\end{tabular}} & \multicolumn{6}{c|}{\RTE in \%, $p$ in \% p=0.01}\\
		\cline{4-9}
		&&& $0.01$ & $0.05$ & $0.1$ & $0.5$ & $1$ & $1.5$\\
		\hline
		\hline
		\multirow{6}{*}{\rotatebox{90}{\begin{tabular}{@{}c@{}}Post-Training\\Asymmetric\end{tabular}}} & \Normal & 4.37 & 4.95 {\color{gray}\scriptsize ${\pm}$0.11} & 5.47 {\color{gray}\scriptsize ${\pm}$0.17} & 6.03 {\color{gray}\scriptsize ${\pm}$0.22} & 15.42 {\color{gray}\scriptsize ${\pm}$3.4} & 51.83 {\color{gray}\scriptsize ${\pm}$9.92} & 81.74 {\color{gray}\scriptsize ${\pm}$5.14}\\
		& \Quant & 4.27 & 4.59 {\color{gray}\scriptsize ${\pm}$0.08} & 5.10 {\color{gray}\scriptsize ${\pm}$0.13} & 5.54 {\color{gray}\scriptsize ${\pm}$0.15} & 10.59 {\color{gray}\scriptsize ${\pm}$1.11} & 30.58 {\color{gray}\scriptsize ${\pm}$6.05} & 63.72 {\color{gray}\scriptsize ${\pm}$6.89}\\
		& \Clipping[$0.25$] & 4.96 & 5.24 {\color{gray}\scriptsize ${\pm}$0.07} & 5.73 {\color{gray}\scriptsize ${\pm}$0.14} & 6.16 {\color{gray}\scriptsize ${\pm}$0.21} & 10.51 {\color{gray}\scriptsize ${\pm}$0.91} & 26.27 {\color{gray}\scriptsize ${\pm}$5.65} & 61.49 {\color{gray}\scriptsize ${\pm}$9.03}\\
		& \Clipping[$0.2$] & 5.24 & 5.48 {\color{gray}\scriptsize ${\pm}$0.05} & 5.87 {\color{gray}\scriptsize ${\pm}$0.09} & 6.23 {\color{gray}\scriptsize ${\pm}$0.13} & 9.47 {\color{gray}\scriptsize ${\pm}$0.7} & 19.78 {\color{gray}\scriptsize ${\pm}$3.58} & 43.64 {\color{gray}\scriptsize ${\pm}$8.2}\\
		& \Clipping[$0.15$] & 5.38 & 5.63 {\color{gray}\scriptsize ${\pm}$0.05} & 6.03 {\color{gray}\scriptsize ${\pm}$0.09} & 6.38 {\color{gray}\scriptsize ${\pm}$0.13} & 8.80 {\color{gray}\scriptsize ${\pm}$0.41} & 15.74 {\color{gray}\scriptsize ${\pm}$2.24} & 36.29 {\color{gray}\scriptsize ${\pm}$7.34}\\
		& \Clipping[$0.1$] & 5.32 & 5.52 {\color{gray}\scriptsize ${\pm}$0.04} & 5.82 {\color{gray}\scriptsize ${\pm}$0.06} & 6.05 {\color{gray}\scriptsize ${\pm}$0.07} & 7.45 {\color{gray}\scriptsize ${\pm}$0.26} & 9.80 {\color{gray}\scriptsize ${\pm}$0.62} & 17.56 {\color{gray}\scriptsize ${\pm}$3.08}\\
		\hline
		\hline
		\multirow{7}{*}{\rotatebox{90}{\begin{tabular}{@{}c@{}}Symmetric\\(during training)\end{tabular}}} & \Normal & 4.36 & 4.82 {\color{gray}\scriptsize ${\pm}$0.07} & 5.51 {\color{gray}\scriptsize ${\pm}$0.19} & 6.37 {\color{gray}\scriptsize ${\pm}$0.32} & 24.76 {\color{gray}\scriptsize ${\pm}$4.71} & 72.65 {\color{gray}\scriptsize ${\pm}$6.35} & 87.40 {\color{gray}\scriptsize ${\pm}$2.47}\\
		& \Quant & 4.39 & 4.77 {\color{gray}\scriptsize ${\pm}$0.08} & 5.43 {\color{gray}\scriptsize ${\pm}$0.21} & 6.10 {\color{gray}\scriptsize ${\pm}$0.32} & 17.11 {\color{gray}\scriptsize ${\pm}$3.07} & 55.35 {\color{gray}\scriptsize ${\pm}$9.4} & 82.84 {\color{gray}\scriptsize ${\pm}$4.52}\\
		& \Clipping[$0.25$] & 4.63 & 4.99 {\color{gray}\scriptsize ${\pm}$0.07} & 5.53 {\color{gray}\scriptsize ${\pm}$0.1} & 6.06 {\color{gray}\scriptsize ${\pm}$0.16} & 13.55 {\color{gray}\scriptsize ${\pm}$1.42} & 41.64 {\color{gray}\scriptsize ${\pm}$7.35} & 73.39 {\color{gray}\scriptsize ${\pm}$7.15}\\
		& \Clipping[$0.2$] & 4.50 & 4.79 {\color{gray}\scriptsize ${\pm}$0.06} & 5.25 {\color{gray}\scriptsize ${\pm}$0.09} & 5.65 {\color{gray}\scriptsize ${\pm}$0.16} & 9.64 {\color{gray}\scriptsize ${\pm}$0.99} & 21.37 {\color{gray}\scriptsize ${\pm}$4.23} & 45.68 {\color{gray}\scriptsize ${\pm}$7.9}\\
		& \Clipping[$0.15$] & 5.18 & 5.42 {\color{gray}\scriptsize ${\pm}$0.05} & 5.76 {\color{gray}\scriptsize ${\pm}$0.08} & 6.07 {\color{gray}\scriptsize ${\pm}$0.09} & 8.36 {\color{gray}\scriptsize ${\pm}$0.43} & 13.80 {\color{gray}\scriptsize ${\pm}$1.45} & 24.70 {\color{gray}\scriptsize ${\pm}$3.77}\\
		& \Clipping[$0.1$] & 4.86 & 5.07 {\color{gray}\scriptsize ${\pm}$0.04} & 5.34 {\color{gray}\scriptsize ${\pm}$0.06} & 5.59 {\color{gray}\scriptsize ${\pm}$0.1} & 7.12 {\color{gray}\scriptsize ${\pm}$0.3} & 9.44 {\color{gray}\scriptsize ${\pm}$0.7} & 13.14 {\color{gray}\scriptsize ${\pm}$1.79}\\
		& \Clipping[$0.05$] & 5.56 & 5.70 {\color{gray}\scriptsize ${\pm}$0.03} & 5.89 {\color{gray}\scriptsize ${\pm}$0.06} & 6.03 {\color{gray}\scriptsize ${\pm}$0.08} & 6.68 {\color{gray}\scriptsize ${\pm}$0.14} & 7.31 {\color{gray}\scriptsize ${\pm}$0.2} & 8.06 {\color{gray}\scriptsize ${\pm}$0.36}\\
		\hline
		\hline
		\multirow{11}{*}{\rotatebox{90}{\begin{tabular}{@{}c@{}}{\color{red}\textbf{A}}symmetric (default) quant.\\(during training)\end{tabular}}} & \Normal & 4.36 & 4.82 {\color{gray}\scriptsize ${\pm}$0.07} & 5.51 {\color{gray}\scriptsize ${\pm}$0.19} & 6.37 {\color{gray}\scriptsize ${\pm}$0.32} & 24.76 {\color{gray}\scriptsize ${\pm}$4.71} & 72.65 {\color{gray}\scriptsize ${\pm}$6.35} & 87.40 {\color{gray}\scriptsize ${\pm}$2.47}\\
		& \Quant & 4.32 & 4.60 {\color{gray}\scriptsize ${\pm}$0.08} & 5.10 {\color{gray}\scriptsize ${\pm}$0.13} & 5.54 {\color{gray}\scriptsize ${\pm}$0.2} & 11.28 {\color{gray}\scriptsize ${\pm}$1.47} & 32.05 {\color{gray}\scriptsize ${\pm}$6} & 68.65 {\color{gray}\scriptsize ${\pm}$9.23}\\
		& \Clipping[$0.25$] & 4.58 & 4.84 {\color{gray}\scriptsize ${\pm}$0.05} & 5.29 {\color{gray}\scriptsize ${\pm}$0.12} & 5.71 {\color{gray}\scriptsize ${\pm}$0.16} & 10.52 {\color{gray}\scriptsize ${\pm}$1.14} & 27.95 {\color{gray}\scriptsize ${\pm}$4.16} & 62.46 {\color{gray}\scriptsize ${\pm}$8.89}\\
		& \Clipping[$0.2$] & 4.63 & 4.91 {\color{gray}\scriptsize ${\pm}$0.05} & 5.28 {\color{gray}\scriptsize ${\pm}$0.08} & 5.62 {\color{gray}\scriptsize ${\pm}$0.11} & 8.27 {\color{gray}\scriptsize ${\pm}$0.35} & 18.00 {\color{gray}\scriptsize ${\pm}$2.84} & 53.74 {\color{gray}\scriptsize ${\pm}$8.89}\\
		& \Clipping[$0.15$] & 4.42 & 4.66 {\color{gray}\scriptsize ${\pm}$0.05} & 5.01 {\color{gray}\scriptsize ${\pm}$0.09} & 5.31 {\color{gray}\scriptsize ${\pm}$0.12} & 7.81 {\color{gray}\scriptsize ${\pm}$0.6} & 13.08 {\color{gray}\scriptsize ${\pm}$2.21} & 23.85 {\color{gray}\scriptsize ${\pm}$5.07}\\
		& \Clipping[$0.1$] & 4.82 & 5.04 {\color{gray}\scriptsize ${\pm}$0.04} & 5.33 {\color{gray}\scriptsize ${\pm}$0.07} & 5.58 {\color{gray}\scriptsize ${\pm}$0.1} & 6.95 {\color{gray}\scriptsize ${\pm}$0.24} & 8.93 {\color{gray}\scriptsize ${\pm}$0.46} & 12.22 {\color{gray}\scriptsize ${\pm}$1.29}\\
		& \Clipping[$0.05$] & 5.44 & 5.59 {\color{gray}\scriptsize ${\pm}$0.04} & 5.76 {\color{gray}\scriptsize ${\pm}$0.07} & 5.90 {\color{gray}\scriptsize ${\pm}$0.07} & 6.53 {\color{gray}\scriptsize ${\pm}$0.13} & 7.18 {\color{gray}\scriptsize ${\pm}$0.16} & 7.92 {\color{gray}\scriptsize ${\pm}$0.25}\\
		\cline{2-9} 
		& \Clipping[$0.2$]+LS & 4.48 & 4.77 {\color{gray}\scriptsize ${\pm}$0.05} & 5.19 {\color{gray}\scriptsize ${\pm}$0.1} & 5.55 {\color{gray}\scriptsize ${\pm}$0.12} & 9.46 {\color{gray}\scriptsize ${\pm}$0.82} & 32.49 {\color{gray}\scriptsize ${\pm}$5.07} & 68.60 {\color{gray}\scriptsize ${\pm}$7.33}\\
		& \Clipping[$0.15$]+LS & 4.67 & 4.86 {\color{gray}\scriptsize ${\pm}$0.05} & 5.23 {\color{gray}\scriptsize ${\pm}$0.08} & 5.83 {\color{gray}\scriptsize ${\pm}$0.12} & 7.99 {\color{gray}\scriptsize ${\pm}$0.43} & 29.40 {\color{gray}\scriptsize ${\pm}$6.99} & 68.99 {\color{gray}\scriptsize ${\pm}$8.48}\\
		& \Clipping[$0.1$]+LS & 4.82 & 5.05 {\color{gray}\scriptsize ${\pm}$0.04} & 5.37 {\color{gray}\scriptsize ${\pm}$0.08} & 6.10 {\color{gray}\scriptsize ${\pm}$0.11} & 7.36 {\color{gray}\scriptsize ${\pm}$0.4} & 10.59 {\color{gray}\scriptsize ${\pm}$1.01} & 18.31 {\color{gray}\scriptsize ${\pm}$2.84}\\
		& \Clipping[$0.05$]+LS & 5.30 & 5.43 {\color{gray}\scriptsize ${\pm}$0.03} & 5.63 {\color{gray}\scriptsize ${\pm}$0.06} & 6.43 {\color{gray}\scriptsize ${\pm}$0.07} & 6.51 {\color{gray}\scriptsize ${\pm}$0.15} & 7.30 {\color{gray}\scriptsize ${\pm}$0.23} & 8.06 {\color{gray}\scriptsize ${\pm}$0.38}\\
		\hline
	\end{tabular}
\end{table*}

In \tabref{tab:supp-sqnr}, we show SQNR and SNR, respectively, for our robust quantization with $m = 8$ bits. We apply this quantization \emph{post-training} to two models: \Normal and \Clipping[$0.1$]. This means that we do \emph{not} perform quantization-aware training since this would result in zero quantization errors as the de-quantized model is returned. We obtain SQNRs between 33 and 36 which is to be expected for $m = 8$ according to \cite{LinICML2016}. Bit errors, in contrast reduce SNR significantly, even for small bit error rates such as $p = 0.01\%$. For $p = 1\%$, SNRs are close to zero or even negative. This indicates that the noise injected through bit errors actually dominates the ``signal'', \ie, is too strong to properly recognize the underlying weights. This confirms our observations in the main paper that bit errors cause very different and challenging error distributions in weights. It also justifies that existing approaches of tackling quantization errors, \eg, \cite{MishchenkoICMLA2019,StockICLR2021,HouICLR2018} which mostly rely on randomized, quantization-aware training, are not applicable.
}

\section{Weight Clipping with Group/Batch Normalization}
\label{sec:supp-clipping}

While weight clipping, \ie, globally constraining weights to $[-\wmax, \wmax]$ during training, is easy to implement, we make a simple adjustment to group and batch normalization layers: we re-parameterize the scale parameter $\alpha$ of batch/ group normalization, which usually defaults to $\alpha = 1$ and may cause problems when clipped, \eg, to $[-0.1, 0.1]$. In particular with aggressive weight clipping, $\alpha \leq \wmax < 1$, the normalization layers lose their ability to represent the identity function, considered important for batch normalization in \cite{IoffeICML2015}. Our re-parameterization introduces a learnable, auxiliary parameter $\alpha'$ such that $\alpha$ is $\alpha = 1 + \alpha'$ to solve this problem.
	\section{Experimental Setup}
\label{subsec:supp-experiments-setup}

\textbf{Datasets:} We conduct experiments on \MNIST\footnote{\url{http://yann.lecun.com/exdb/mnist/}} \cite{LecunIEEE1998} and \Cifar\footnote{\url{https://www.cs.toronto.edu/~kriz/cifar.html}} \cite{Krizhevsky2009}. \MNIST consists of $60\text{k}$ training and $10\text{k}$ test images from $10$ classes. These are gray-scale and of size $28 \times 28$ pixels. \Cifar consists of $50\text{k}$ training and $10\text{k}$ test images of size $32\times 32\times 3$ (\ie, color images). \CifarT has images corresponding to $10$ classes, \CifarH contains images from $100$ classes.

\textbf{Architecture:} The used SimpleNet architectures \cite{HasanpourARXIV2016} are summarized in \tabref{tab:supp-architectures}, including the total number of weights $W$. On \Cifar, this results in a total of roughly $W \approx 5.5\text{M}$ weights. Due to the lower resolution on \MNIST, channel width in each convolutional layer is halved, and one stage of convolutional layers including a pooling layer is skipped. This results in a total of roughly $W \approx 1\text{M}$ weights. In both cases, we replaced batch normalization (BN) \cite{IoffeICML2015} with group normalization (GN) \cite{WuECCV2018}. The GN layers are re-parameterized as in \appref{sec:supp-clipping} to facilitate weight clipping. \tabref{tab:supp-architectures} also includes the expected number of bit errors given various rates $p$ for random bit errors. Regarding the number of weights $W$, SimpleNet compares favorably to, \eg, VGG \cite{SimonyanICLR2015}: VGG-16 has $14\text{M}$ weights on \Cifar. Additionally, we found SimpleNet to be easier to train without BN, which is desirable as BN reduces robustness to bit errors significantly, \cf \appref{subsec:supp-experiments-bn}. The ResNet-50 \cite{HeCVPR2016} used for experiments in \appref{subsec:supp-experiments-randbet} \revision{and the ResNet-18 used on TinyImageNet} follow the official PyTorch \cite{PaszkeNIPSWORK2017} implementation. \revision{On TinyImageNet, this results in roughly $11.7\text{Mio}$ weights.} The Wide ResNet (WRN) \cite{ZagoruykoBMVC2016} used on \CifarH is adapted from\footnote{\url{https://github.com/meliketoy/wide-resnet.pytorch}}, but we use $12$ base channels, instead of $16$, reducing $W$ from roughly $36.5\text{Mio}$ to $20.5\text{Mio}$.

\begin{figure*}[t]
	\centering
	\begin{subfigure}{0.35\textwidth}
		\vspace*{0px}
		\centering
		\Quant
		
		\includegraphics[width=1\textwidth]{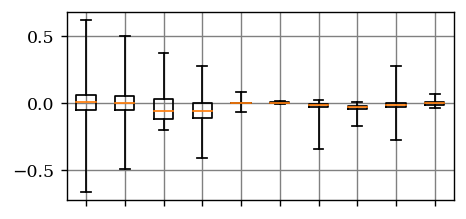}
	\end{subfigure}
	\begin{subfigure}{0.35\textwidth}
		\vspace*{0px}
		\centering
		\Random (w/o weight clipping)
		
		\includegraphics[width=1\textwidth]{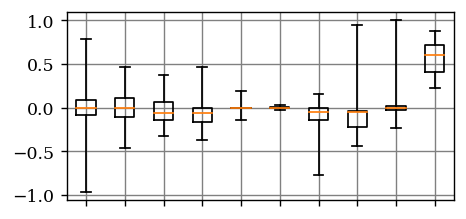}
	\end{subfigure}
	\\
	\begin{subfigure}{0.35\textwidth}
		\vspace*{0px}
		\centering
		\Clipping[$0.1$]
			
		\includegraphics[width=1\textwidth]{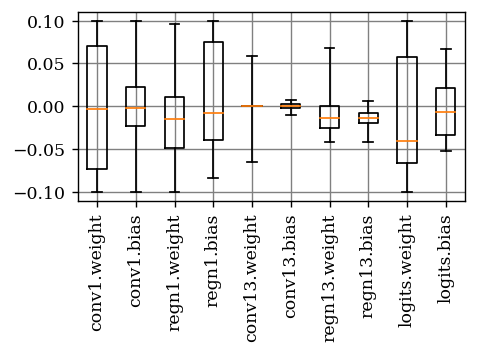}
	\end{subfigure}
	\begin{subfigure}{0.35\textwidth}
		\vspace*{0px}
		\centering
		\Clipping[$0.05$]
			
		\includegraphics[width=1\textwidth]{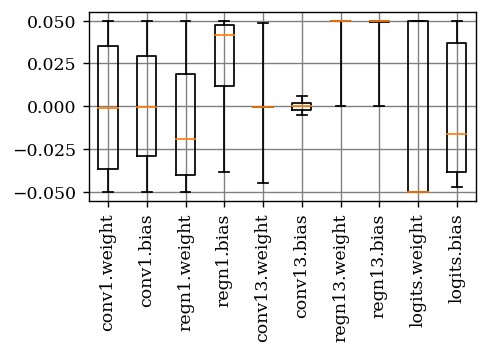}
	\end{subfigure}
	\\
	\begin{subfigure}{0.16\textwidth}
		\vspace*{0px}
		\centering
		\Quant
		\includegraphics[width=1\textwidth]{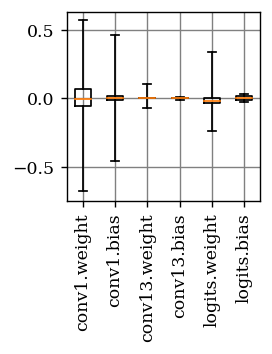}
	\end{subfigure}
	\begin{subfigure}{0.16\textwidth}
		\vspace*{0px}
		\centering
		\Clipping[$0.25$] $\uparrow$
			
		\includegraphics[width=1\textwidth]{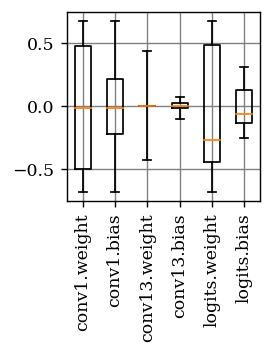}
	\end{subfigure}
	\begin{subfigure}{0.35\textwidth}
		\includegraphics[width=1\textwidth]{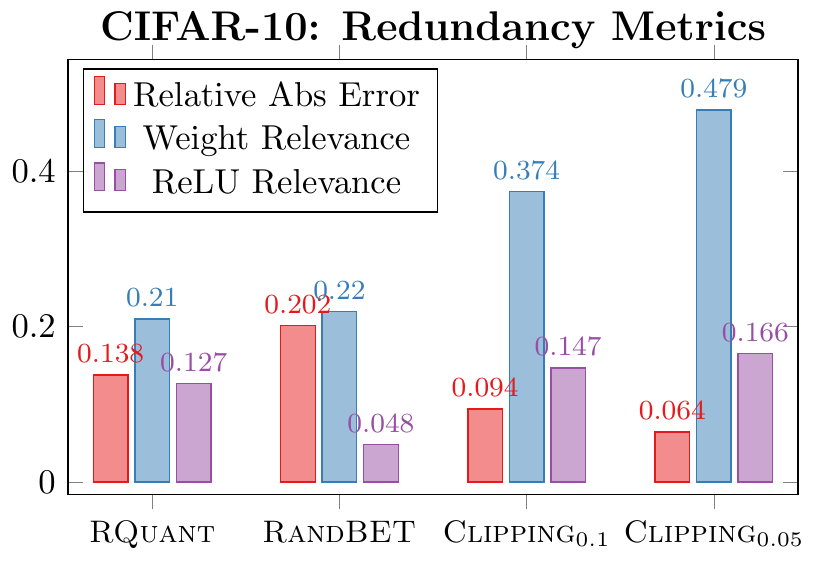}
	\end{subfigure}
	\vspace*{-6px}
	\caption{\textbf{Weight Clipping Increases Redundancy.} We show weight distributions of selected layers (top) for \Quant, \Random (without weight clipping) as well as \Clipping[$0.1$] and \Clipping[$0.05$]. We show weights and biases for the logit layer as well as the first and last (13th) convolutional layer. Scale/Bias parameters of GN are also included. Below (left), this is shown for the scaling experiment from \tabref{tab:supp-clipping-scaling}. Note that \Random only affects the logit layer, while \Clipping increases the used (relative) weight range significantly. On the bottom (right), we plot various measures of redundancy, see the text for discussion and details. The relative absolute error is computed considering random bit errors with probability $p = 1\%$.}
	\label{fig:supp-clipping}
\end{figure*}

\textbf{Training:} We use stochastic gradient descent to minimize cross-entropy loss. We use an initial learning rate of $0.05$, multiplied by $0.1$ after $\nicefrac{2}{5}$, $\nicefrac{3}{5}$ and $\nicefrac{4}{5}$ of \revision{$100$/$250$/$100$ epochs on \MNIST/\Cifar/TinyImageNet}. Our batch size is $128$ and momentum of $0.9$ is used together with weight decay of $5\cdot10^{-4}$. The batch size was reduced to $64$ on TinyImageNet. On \Cifar and TinyImageNet, we whiten the input images and use AutoAugment\footnote{\url{https://github.com/DeepVoltaire/AutoAugment}} \cite{CubukARXIV2018} with Cutout \cite{DevriesARXIV2017}. Cutout is applied with a window size of \revision{$16 \times 16$/$32\times 32$ on \Cifar/TinyImageNet}, and independent of AutoAugment, we apply random cropping with up to $4$/$8$ pixels. Created black spaces are filled using the mean image color (grayish). \revision{As \cite{CubukARXIV2018} does not provide an augmentation policy for TinyImageNet, we used the same as for \Cifar and found it work reasonably well.} Initialization follows \cite{HeICCV2015}. The full training set is used for training, and we do \emph{not} rely on early stopping. For \Random, we use $\lambda = 1$ and start injecting bit errors when the loss is below 1.75 on \MNIST/\CifarT, 3.5 on \CifarH \revision{ or 6 on TinyImageNet}. \tabref{tab:supp-accuracy} highlights clean test error (\TE) obtained for various precision $m$ and compared to other architectures, \eg, ResNet-50, on \CifarT, which performs worse when using GN. For \Adv, we use $\lambda = 0$ and also start injecting adversarial bit errors when the loss is 1.75 or smaller (3.5 on \CifarH, \revision{6 on TinyImageNet}). We use $T = 10$ iterations of our adversarial bit error attack, with learning rate $0.5$, no backtracking \cite{StutzICML2020} or momentum \cite{DongCVPR2018}, and per-layer $L_\infty$ gradient normalization as outlined in the main paper.

\textbf{Random Bit Errors:} We simulate $50$ different chips with enough memory arrays to accommodate all weights by drawing uniform samples $u^{(c)} \sim U(0, 1)^{W \times m}$ for each chip $c$ and all $m$ bits for a total of $W$ weights. Then, for chip $c$, bit $j$ in weight $w_i$ is flipped iff $u^{(c)}_{ij} \leq p$. This assumes a linear memory layout of all $W$ weights. The pattern, \ie, spatial distribution, of bit errors for chip $c$ is fixed by $u^{(c)}$, while across all $50$ chips, bit errors are uniformly distributed. We emphasize that we pre-determine $u^{(c)}$, $c = 1,\ldots,50$, once for all our experiments using fixed random seeds. Thus, our robustness results are entirely comparable across all models as well as bit error rates $p$. Also note that the bit errors for a fixed chip $c$ at probability $p' < p$ are a subset of those for bit error rate $p$. The expected number of bit errors for various rates $p$ is summarized in \tabref{tab:supp-architectures}. For random bit errors in activations and inputs we follow a similar strategy as outlined in \secref{subsec:supp-experiments-activations-inputs}.

\begin{table*}[t]
	\centering
	\caption{\textbf{\Random Robustness with \emph{Symmetric} Quantization.} Average \RTE and standard deviation for \Clipping and \Random with $\wmax = 0.1$ and \textbf{s}ymmetric quantization, \ie, larger quantization range than {\color{red}a}symmetric quantization. Also \cf \tabref{tab:supp-clipping} and \tabref{tab:supp-summary-cifar10}. Robustness decreases slightly compared to asymmetric quantization, however, \Clipping and \Random are still effective in reducing \RTE against high bit error rates $p$.}
	\label{tab:supp-randbet-symmetric}
	\vspace*{-0.25cm}
	\begin{tabular}{|l | c | c | c | c | c | c | c |}
		\hline
		\multicolumn{8}{|c|}{\bfseries \CifarT ($\mathbf{m = 8}$ bit): \Random with symmetric quantization}\\
		\hline
		Model & \multirow{2}{*}{\begin{tabular}{c}\TE\\in \%\end{tabular}} & \multicolumn{6}{c|}{\RTE in \%, $p$ in \%}\\
		\cline{3-8}
		&& $0.01$ & $0.05$ & $0.1$ & $0.5$ & $1$ & $1.5$\\
		\hline
		\hline
		\Normal & 4.36 & 4.82 {\color{gray}\scriptsize ${\pm}$0.07} & 5.51 {\color{gray}\scriptsize ${\pm}$0.19} & 6.37 {\color{gray}\scriptsize ${\pm}$0.32} & 24.76 {\color{gray}\scriptsize ${\pm}$4.71} & 72.65 {\color{gray}\scriptsize ${\pm}$6.35} & 87.40 {\color{gray}\scriptsize ${\pm}$2.47}\\
		\Quant & 4.39 & 4.77 {\color{gray}\scriptsize ${\pm}$0.08} & 5.43 {\color{gray}\scriptsize ${\pm}$0.21} & 6.10 {\color{gray}\scriptsize ${\pm}$0.32} & 17.11 {\color{gray}\scriptsize ${\pm}$3.07} & 55.35 {\color{gray}\scriptsize ${\pm}$9.4} & 82.84 {\color{gray}\scriptsize ${\pm}$4.52}\\
		\Clipping[$0.1$] & 4.86 & 5.07 {\color{gray}\scriptsize ${\pm}$0.04} & 5.34 {\color{gray}\scriptsize ${\pm}$0.06} & 5.59 {\color{gray}\scriptsize ${\pm}$0.1} & 7.12 {\color{gray}\scriptsize ${\pm}$0.3} & 9.44 {\color{gray}\scriptsize ${\pm}$0.7} & 13.14 {\color{gray}\scriptsize ${\pm}$1.79}\\
		\hline
		\Random[$0.1$] $p{=}0.01$ & 5.07 & 5.27 {\color{gray}\scriptsize ${\pm}$0.04} & 5.54 {\color{gray}\scriptsize ${\pm}$0.07} & 5.73 {\color{gray}\scriptsize ${\pm}$0.11} & 7.18 {\color{gray}\scriptsize ${\pm}$0.29} & 9.63 {\color{gray}\scriptsize ${\pm}$0.9} & 13.81 {\color{gray}\scriptsize ${\pm}$2.2}\\
		\Random[$0.1$] $p{=}0.1$ & 4.62 & 4.83 {\color{gray}\scriptsize ${\pm}$0.04} & 5.09 {\color{gray}\scriptsize ${\pm}$0.08} & 5.31 {\color{gray}\scriptsize ${\pm}$0.08} & 6.70 {\color{gray}\scriptsize ${\pm}$0.28} & 8.89 {\color{gray}\scriptsize ${\pm}$0.59} & 12.20 {\color{gray}\scriptsize ${\pm}$1.33}\\
		\Random[$0.1$] $p{=}1$ & 5.03 & 5.22 {\color{gray}\scriptsize ${\pm}$0.04} & 5.43 {\color{gray}\scriptsize ${\pm}$0.06} & 5.61 {\color{gray}\scriptsize ${\pm}$0.07} & 6.56 {\color{gray}\scriptsize ${\pm}$0.13} & 7.70 {\color{gray}\scriptsize ${\pm}$0.26} & 8.99 {\color{gray}\scriptsize ${\pm}$0.42}\\
		\Random[$0.1$] $p{=}1.5$ & 5.24 & 5.37 {\color{gray}\scriptsize ${\pm}$0.03} & 5.57 {\color{gray}\scriptsize ${\pm}$0.06} & 5.76 {\color{gray}\scriptsize ${\pm}$0.07} & 6.66 {\color{gray}\scriptsize ${\pm}$0.14} & 7.62 {\color{gray}\scriptsize ${\pm}$0.25} & 8.71 {\color{gray}\scriptsize ${\pm}$0.42}\\
		\Random[$0.1$] $p{=}2$ & 5.82 & 5.97 {\color{gray}\scriptsize ${\pm}$0.04} & 6.19 {\color{gray}\scriptsize ${\pm}$0.07} & 6.37 {\color{gray}\scriptsize ${\pm}$0.09} & 7.22 {\color{gray}\scriptsize ${\pm}$0.19} & 8.03 {\color{gray}\scriptsize ${\pm}$0.23} & 8.96 {\color{gray}\scriptsize ${\pm}$0.38}\\
		\hline
	\end{tabular}
\end{table*}
\begin{table*}[t]
	\centering
	\caption{\revision{\textbf{Weight Clipping Comparison to Other Regularization Techniques:} \textbf{Left:} We compare weight clipping to several related regularization baselines, including weight decay, dropout, a modified hinge loss (MHL) following \cite{BuschjaegerDATE2021}, label noise (LN). We report results after hyper-parameter optimization and observe that none of these techniques improves bit error robustness as significantly as weight clipping; in fact, weight decay, label noise and MHL reduce robustness. \textbf{Right:} Using additional pseudo-labeled training examples following \cite{CarmonNIPS2019} as regularization consistently improves robustness slightly, but only when combined with weight clipping.}}
	\label{tab:supp-regularization}
	\vspace*{-0.1cm}
	\begin{subfigure}[t]{0.49\textwidth}
		\vspace*{0px}
		\begin{tabular}{|l|c|c|c|}
			\hline
			\multicolumn{4}{|c|}{\bfseries\CifarT: regularization baselines}\\
			\hline
			Model & \TE & \multicolumn{2}{c|}{\RTE}\\
			\cline{2-4}
			&& $p = 0.1\%$ & $p = 1\%$\\
			\hline
			\hline
			\Quant & \bfseries 4.22 & \bfseries 5.54 & \bfseries 32.05 \\
			\Quant + 0.01 weight decay (best) & 5.80 & 7.26 & 37.58\\
			\Quant + dropout & 4.55 & 5.79 & 26.92\\
			\Quant + MHL margin $1$ (best) & 6.62 & 7.62 & 28.65\\
			\Quant + MHL margin $8$ & 5.92 & 7.48 & 34.90\\
			\hline
			\Clipping[$0.1$] & \bfseries 4.81 & \bfseries 5.58 & 8.93\\
			\Clipping[$0.1$] + 0.1 LN & 5.01 & 5.68 & 10.88\\
			\Clipping[$0.1$] +dropout & 5.30 & 5.95 & \bfseries 8.83\\
			\Clipping[$0.1$] + MHL margin $1$ & 6.78 & 7.84 & 16.28\\
			\Clipping[$0.1$] + MHL margin $8$ (best) & 7.27 & 7.96 & 10.17\\
			\hline
		\end{tabular}
	\end{subfigure}
	\begin{subfigure}[t]{0.49\textwidth}
		\vspace*{0px}
		\begin{tabular}{|l|c|c|c|}
			\hline
			\multicolumn{4}{|c|}{\bfseries\CifarT: additional unlabeled data \cite{CarmonNIPS2019}}\\
			\hline
			Model & \TE & \multicolumn{2}{c|}{\RTE}\\
			\cline{2-4}
			&& $p = 0.1\%$ & $p = 1\%$\\
			\hline
			\hline
			\Quant & 4.22 & 5.54 & \bfseries 32.05 \\
			\Quant + pseudo-labeled data & \bfseries 4.15 & \bfseries 5.25 & 35.66\\
			\hline
			\Clipping[$0.25$] & \bfseries 4.53 & 5.71 & 27.95\\
			\Clipping[$0.25$] + pseudo-labeled data & 4.54 & \bfseries 5.46 & \bfseries 20.02\\
			\hline
			\Clipping[$0.1$] & \bfseries 4.81 & 5.58 & 8.93\\
			\Clipping[$0.1$] + pseudo-labeled data & 4.96 & \bfseries 5.63 & \bfseries 8.15\\
			\hline
			\Clipping[$0.05$] & \bfseries 5.42 & 5.90 & 7.18\\
			\Clipping[$0.05$] + pseudo-labeled data & \bfseries 5.42 & \bfseries 5.74 & \bfseries 6.75\\
			\hline
		\end{tabular}
	\end{subfigure}
\end{table*}

\textbf{Adversarial Bit Errors:} For evaluation, and fixed $\epsilon$, we run our adversarial bit error attack for a total of $80$ random restarts using the following settings: $5$ restarts for the untargeted attack with learning rate $1$ with and without momentum $0.9$ ($10$ restarts in total) and $10$ restarts for the targeted attack ($1$ for each potential target label on \MNIST and \CifarT). This is done attacking all layers. Furthermore, we use $5$ untargeted restarts with momentum, and $10$ targeted restarts for: attacking only the logit layer, only the first convolutional layer, both the logit and the first convolutional layer, or all layers \emph{except} the logit or first convolutional layer. In total, this makes $20 + 4\cdot15 = 80$ restarts of our adversarial bit error attack.

\textbf{Bit Flip Attack (BFA) \cite{RakinICCV2019}:} We follow the official PyTorch code\footnote{\url{https://github.com/elliothe/BFA}}. Specifically, we use the provided implementation of BFA to attack our models by integrating our SimpleNet models into the provided attack/evaluation code. This means that we also use the quantization scheme of \cite{RakinICCV2019}, not our robust fixed-point quantization scheme. However, our models still used \Quant (and optionally \Clipping or \Random) during training. We use $m = 8$ bits. We allow $5$ bit flips per iteration, for a total of $T$ iterations, equaling $\epsilon = 5\cdot T$ bit flips. For comparability, we adapted the code to compute adversarial bit flips on the last $100$ test examples and evaluate on the first $9000$ test examples. We allow $5$ restarts for each $\epsilon$.

\section{Experiments}
\label{sec:supp-experiments}

\subsection{Batch Normalization}
\label{subsec:supp-experiments-bn}

We deliberately replace batch normalization (BN) \cite{IoffeICML2015} by group normalization (GN) \cite{WuECCV2018} in our experiments. \tabref{tab:supp-bn} demonstrates that \RTE increases significantly when using BN compared to GN indicating that BN is more vulnerable to bit errors in DNN weights. For example, without clipping, \RTE increases from $11.28\%$ to staggering $52.52\%$ when replacing GN with BN. Note that, following \appref{sec:supp-clipping}, the BN/GN parameters (\ie, scale/bias) are re-parameterized to account for weight clipping. The observations in \tabref{tab:supp-bn} can also be confirmed without quantization, \eg, considering random $L_\infty$ noise in the weights. We suspect that the running statistics accumulated during training do not account for the random bit errors at test time, even for \Random. This is confirmed in \tabref{tab:supp-bn} (bottom) showing that \RTE reduces significantly when using the batch statistics at test time. Generally, BN improves accuracy, but might not be beneficial in terms of robustness, as also discussed for adversarial examples \cite{GallowayARXIV2019}. Using GN also motivates our use of SimpleNet instead of, \eg, ResNet-50, which generally performs worse with GN, \cf \tabref{tab:supp-accuracy}.

\subsection{Robust Quantization}
\label{subsec:supp-experiments-quantization}

\tabref{tab:supp-quantization} shows results complementary to the main paper, considering additional bit error rates $p$. Note that, for $m = 8$ bit, changes in the quantization has negligible impact on clean \TE. Only the change from global to per-layer quantization makes a difference. However, considering \RTE for larger bit error rates, reducing the quantization range, \eg, through per-layer and asymmetric quantization, improves robustness significantly. Other aspects of the quantization scheme also play an important role, especially for low-precision such as $m = 4$ bit, \cf \tabref{tab:supp-quantization}, as outlined in the following.

For example, using asymmetric quantization into \emph{signed} integers actually increases \RTE for larger $p$ compared to ``just'' using symmetric per-layer quantization (rows 2 and 3). Using \emph{unsigned} integers instead reduces \RTE significantly. We believe this to be due to the two's complement representation of signed integers being used with an \textbf{a}symmetric quantization range. In symmetric quantization (around $0$, \ie, $[-\qmax, \qmax]$), bit errors in the sign bit incur not only a change of the integer's sign, but also the corresponding change in the weights sign\footnote{An \emph{un}signed integer of value $127$ is represented as $0111 1111$. Flipping the most (left-most) significant bit results in $1111 1111$ corresponding to $255$, \ie, the value increases. For a signed integer in two's complement representation, the same bit flip changes the value from $127$ to $-1$, while $0$-to-$1$ not affecting the sign bit generally increase value (also for negative integers).}. Assuming an asymmetric quantization of $[\qmin, \qmax]$ with $0 < \qmin < \qmax$, bit errors in sign bits are less meaningful. For example, flipping any bit $0$-to-$1$ usually increases the value of the integer. However, a $0$-to-$1$ flip in the sign bit actually decreases the value and produces a \emph{negative} integer. However, this change from positive to negative is not reflected in the corresponding weight value (as $\qmin > 0$). For high bit error rates $p\%$, this happens more and more frequently and these changes seem to have larger impact on DNN performance, \ie, \RTE.

Additionally, we considered the difference between using integer conversion for $\nicefrac{w_i}{\Delta}$ and using proper rounding, \ie, $\lceil\nicefrac{w_i}{\Delta}\rfloor$. We emphasize that, for $m = 8$ bit, there is \emph{no} significant difference in terms of clean \TE. However, using proper rounding reduces the approximation error slightly. For $m = 8$ bit, using $p = 2.5\%$ bit error rate, the average absolute error (in the weights) across $10$ random bit error patterns reduces by $2\%$. Nevertheless, it has significantly larger impact on \RTE. For $m = 4$, this is more pronounced: rounding reduces the average absolute error by roughly $67\%$. Surprisingly, this is not at all reflected in the clean \TE, which only decreases from $5.81\%$ to $5.29\%$. It seems that the DNN learns to compensate these errors during training. At test time, however, \RTE reflects this difference in terms of robustness.

Overall, we found that robust quantization plays a key role. While both weight clipping (\Clipping) and random bit error training (\Random) can improve robustness further, robust quantization lays the foundation for these improvements to be possible. Thus, we encourage authors to consider robustness in the design of future DNN quantization schemes. Even simple improvements over our basic fixed-point quantization scheme may have significant impact in terms of robustness. For example, proper handling of outliers \cite{ZhuangCVPR2018,SungARXIV2015}, learned quantization \cite{ZhangECCV2018}, or adaptive/non-uniform quantization \cite{ZhouAAAI2018,ParkECCV2018,NagelICCV2019} are promising directions to further improve robustness. Finally, we believe that this also poses new theoretical challenges, \ie, studying (fixed-point) quantization with respect to robustness \emph{and} quantization error.

\subsection{(Global) Weight Clipping (\Clipping)}
\label{subsec:supp-experiments-clipping}

In \tabref{tab:supp-clipping} we present robustness results, \ie, \RTE, for weight clipping. Note that weight clipping constraints the weights during training to $[-\wmax, \wmax]$ through projection. We demonstrate that weight clipping can also be used independent of quantization. To this end, we train DNNs with weight clipping, but without quantization. We apply post-training quantization and evaluate bit error robustness. While the robustness is reduced slightly compared to quantization-aware training \emph{and} weight clipping, the robustness benefits of weight clipping are clearly visible. For example, clipping at $\wmax = 0.1$ improves \RTE from $30.58\%$ to $9.8\%$ against $p = 1\%$ bit error rate when performing post-training quantization. With symmetric quantization-aware training, \Clipping[$0.1$] improves slightly to $7.31\%$. Below (middle), we show results for weight clipping and symmetric quantization. These results are complemented in \tabref{tab:supp-randbet-symmetric} with \Random. Symmetric quantization might be preferable due to reduced computation and energy cost compared to asymmetric quantization. However, this also increases \RTE slightly. Nevertheless, \Clipping consistently improves robustness, independent of the difference in quantization. Finally, on the bottom, we show results confirming the adverse effect of label smoothing \cite{SzegedyCVPR2016} on \RTE. \figref{fig:supp-clipping-inf} also shows that the obtained robustness generalizes to other error models such as $L_\infty$ weight perturbations, see caption for details.

We hypothesize that weight clipping improves robustness as it encourages redundancy in weights and activations during training. This is because cross-entropy loss encourages large logits and weight clipping forces the DNN to ``utilize'' many different weights to produce large logits. \tabref{tab:supp-clipping-scaling} presents a simple experiment in support of our hypothesis. We already emphasized that, relatively, weight clipping does \emph{not} reduce the impact of bit errors. Nevertheless, when using group normalization (GN), the trained DNNs are scale-invariant in their weights (\eg, as discussed in \cite{DinhICML2017}). Note that the GN parameters (scale and bias) are \emph{not} scaled. We down-scale the weights of \Quant to have the same maximum absolute weight value as \Clipping[$0.1$] (\Quant $\rightarrow$ \Clipping[$0.1$]).
This scaling is applied globally, not per layer. Similarly, we up-scale the weights of \Clipping[$0.1$] to the same maximum absolute weight value as \Quant (\Clipping[$0.1$] $\rightarrow$ \Quant). \tabref{tab:supp-clipping-scaling} shows that ``just'' down-scaling does not induce robustness, as expected. More importantly, up-scaling the weights after training with weight clipping, $\wmax = 0.1$, preservers robustness. This simple experiment supports our argument that \Clipping does not improve robustness due to the reduced quantization range but acts as a regularizer as described in the main paper.

\begin{table}[t]
	\centering
	\caption{\textbf{\Random Variants.} \TE and \RTE for \Random and two variants: curricular \Random, with $p$ being increased slowly from $\nicefrac{p}{20}$ to $p$ during the first half of training; and ``alternating'' \Random where weight updates increasing quantization range, \ie, increasing the maximum absolute weight per layer, are not possible based on gradients from perturbed weights, see \secref{subsec:supp-experiments-randbet} for details. Both variants decrease robustness slightly. This is in contrast to, \eg, \cite{KoppulaMICRO2019}, using curricular training on profiled bit errors with success.}
	\label{tab:supp-randbet-variants}
	\vspace*{-0.25cm}
	\begin{tabular}{| l | c | c | c |}
		\hline
		\multicolumn{4}{|c|}{\bfseries \CifarT ($\mathbf{m = 8}$ bit): \Random variants}\\
		\hline
		& \multirow{2}{*}{\begin{tabular}{@{}c@{}}\TE\\in \%\end{tabular}} & \multicolumn{2}{c|}{\RTE in \%}\\
		\hline
		& & $p{=}0.1$ & $p{=}1$\\
		\hline
		\hline
		\Random $p{=}0.1$, $\wmax = 0.1$ & 4.93 & 5.67 & 8.65\\
		\Random $p{=}1$, $\wmax = 0.1$ & 5.06 & 5.87 & 7.60\\
		Curriculum \Random $p{=}1$, $\wmax = 0.1$ & 4.89 & 5.78 & 8.51\\
		Curriculum \Random $p{=}1$, $\wmax = 0.1$ & 5.32 & 6.13 & 7.98\\
		Alternating \Random $p{=}1$, $\wmax = 0.1$ & 5.07 & 5.91 & 8.93\\
		Alternating \Random $p{=}1$, $\wmax = 0.1$ & 5.24 & 6.25 & 8.02\\
		\hline
	\end{tabular}
\end{table}

\figref{fig:supp-clipping} presents further supporting evidence for our hypothesis: While \Random mainly affects the logits layer, \Clipping clearly increases the weight range used by the DNN. Here, the weight range is understood relative to $\wmax$ (or the maximum absolute weight value for \Normal). This is pronounced in particular when up-scaling the clipped model (bottom left). Finally, \figref{fig:supp-clipping} (bottom right) also considers three attempts to measure redundancy in weights and activations. The relative absolute error is computed with respect to $p = 1\%$ bit error rate and decreases for \Clipping, meaning that random bit errors have less impact. \emph{Weight relevance} is computed as the sum of absolute weights, \ie, $\sum_i |w_i|$, normalized by the maximum absolute weight: $\nicefrac{\sum_i |w_i|}{\max_i |w_i|}$. This metric measures how many weights are, considering their absolute value, relevant. Finally, We also measure activation redundancy using ReLU relevance, computing the fraction of non-zero activations after the final ReLU activation. \Clipping increases redundancy in the final layer significantly. Finally, \figref{fig:supp-clipping} (bottom left) shows the difference in weight distributions by up-scaling \Clipping[$0.25$] to the same weight range as \Normal. Clearly, \Clipping causes more non-zero weights be learned by the DNN. This can be observed across all types of parameters, \ie, weights or biases as well as convolutional or fully connected layers.

\begin{table}[t]
	\centering
	\caption{\textbf{\Random with ResNets.} We report \RTE for \Quant, \Clipping and \Random using ResNet-20 and ResNet-50. According to \tabref{tab:supp-accuracy}, \TE increases significantly when using group normalization for ResNet-50, explaining the generally higher \RTE. However, using ResNets, \Clipping and \Random continue to improve robustness significantly, despite a ResNet-50 having roughly $23.5\text{Mio}$ weights.}
	\label{tab:supp-randbet-resnet}
	\vspace*{-0.25cm}
	\begin{tabular}{| l | c | c | c |}
		\hline
		\multicolumn{4}{|c|}{\bfseries \CifarT ($\mathbf{m = 8}$ bit): ResNet architectures}\\
		\hline
		& \multirow{2}{*}{\begin{tabular}{@{}c@{}}\TE\\in \%\end{tabular}} & \multicolumn{2}{c|}{\RTE in \%}\\
		\hline
		& & $p{=}0.5$ & $p{=}1.5$\\
		\hline
		\hline
		\multicolumn{4}{|c|}{\bfseries ResNet-20}\\
		\hline
		\Quant & 4.34 & 13.89 {\color{gray}\scriptsize ${\pm}$2.45} & 81.25 {\color{gray}\scriptsize ${\pm}$5.08}\\
		\Clipping[$0.1$] & 4.83 & 6.76 {\color{gray}\scriptsize ${\pm}$0.16} & 11.23 {\color{gray}\scriptsize ${\pm}$0.97}\\
		\Random[$0.1$], $p{=}1$ & 5.28 & 6.72 {\color{gray}\scriptsize ${\pm}$0.19} & 8.96 {\color{gray}\scriptsize ${\pm}$0.49}\\
		\hline
		\hline
		\multicolumn{4}{|c|}{\bfseries ResNet-50}\\
		\hline
		\Quant & 6.81 & 32.94 {\color{gray}\scriptsize ${\pm}$5.51} & 90.98 {\color{gray}\scriptsize ${\pm}$0.67}\\
		\Clipping[$0.1$] & 5.99 & 9.27 {\color{gray}\scriptsize ${\pm}$0.44} & 36.39 {\color{gray}\scriptsize ${\pm}$7.03}\\
		\Random[$0.1$], $p{=}1$ & 6.04 & 7.87 {\color{gray}\scriptsize ${\pm}$0.22} & 11.27 {\color{gray}\scriptsize ${\pm}$0.6}\\
		\hline
	\end{tabular}
\end{table}
\begin{figure*}
	\centering
	\includegraphics[width=0.85\textwidth]{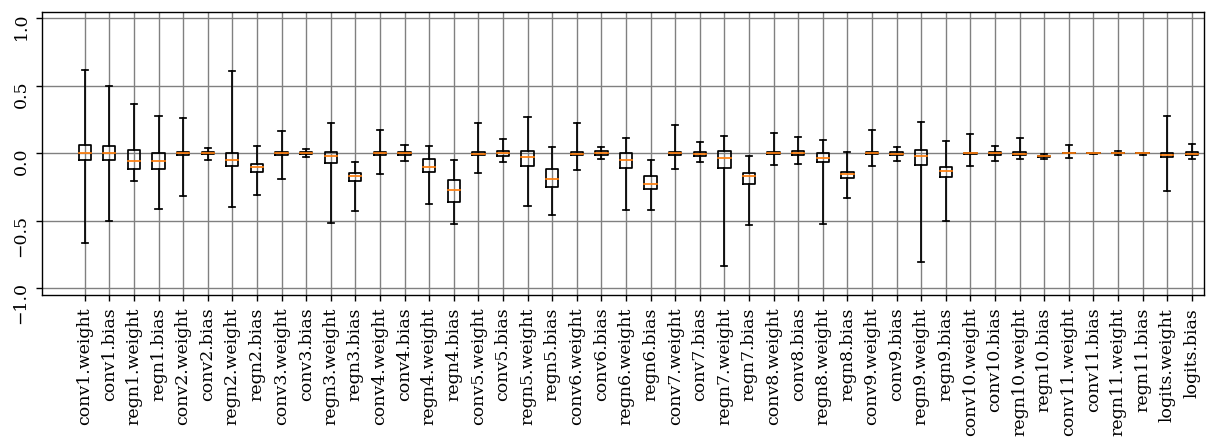}
	\vspace*{-6px}
	\caption{\textbf{Per-Layer Weight Clipping.} On \Cifar, we show the weight ranges, per layer, corresponding to \Quant (\ie, without weight clipping) and the architecture from \tabref{tab:supp-architectures}. As can be seen, few layers exhibit large ranges. Thus, per-layer clipping computes a layer-specific weight constraint $[-w_{\text{max},l},w_{\text{max},l}]$, derived from a global $\wmax$ and based on the weight range relative to the layer with largest range. As not to ``over-constraint'' layers with small weight range, a minimum range of $[-0.2\wmax, 0.2\wmax]$ is enforced.}
	\label{fig:supp-plclipping}
\end{figure*}
\begin{table*}[t]
	\centering
	\caption{\textbf{\PLClipping and \PLRandom Further Push Robustness.} \TE and \RTE for \PLClipping and \PLRandom with various configurations of $\wmax$ and $p$ used for training. Compared to \tabref{tab:supp-randbet-symmetric}, reporting results for \Clipping and \Random, \ie, without per-layer weight clipping, robustness can be improved significantly, while simultaneously reducing (clean) \TE.}
	\label{tab:supp-plclipping}
	\vspace*{-0.25cm}
	\begin{tabular}{|l | c | c | c | c | c | c | c |}
		\hline
		\multicolumn{8}{|c|}{\bfseries \CifarT ($\mathbf{m = 8}$ bit): \emph{per-layer} clipping and \Random robustness}\\
		\hline
		Model & \multirow{2}{*}{\begin{tabular}{c}\TE\\in \%\end{tabular}} & \multicolumn{6}{c|}{\RTE in \%, $p$ in \%}\\
		\cline{3-8}
		&& $0.01$ & $0.05$ & $0.1$ & $0.5$ & $1$ & $1.5$\\
		\hline
		\hline
		\PLClipping[$1$] & 4.68 & 4.98 {\color{gray}\scriptsize ${\pm}$0.1} & 5.52 {\color{gray}\scriptsize ${\pm}$0.16} & 6.06 {\color{gray}\scriptsize ${\pm}$0.2} & 13.56 {\color{gray}\scriptsize ${\pm}$2.74} & 40.72 {\color{gray}\scriptsize ${\pm}$10.6} & 73.66 {\color{gray}\scriptsize ${\pm}$7.74}\\
		\PLClipping[$0.5$] & 4.49 & 4.73 {\color{gray}\scriptsize ${\pm}$0.06} & 5.07 {\color{gray}\scriptsize ${\pm}$0.11} & 5.36 {\color{gray}\scriptsize ${\pm}$0.12} & 7.76 {\color{gray}\scriptsize ${\pm}$0.48} & 13.04 {\color{gray}\scriptsize ${\pm}$1.61} & 23.54 {\color{gray}\scriptsize ${\pm}$4.67}\\
		\PLClipping[$0.25$] & 4.78 & 4.93 {\color{gray}\scriptsize ${\pm}$0.04} & 5.15 {\color{gray}\scriptsize ${\pm}$0.07} & 5.32 {\color{gray}\scriptsize ${\pm}$0.1} & 6.17 {\color{gray}\scriptsize ${\pm}$0.17} & 7.16 {\color{gray}\scriptsize ${\pm}$0.33} & 8.25 {\color{gray}\scriptsize ${\pm}$0.56}\\
		\PLClipping[$0.2$] & 4.84 & 4.95 {\color{gray}\scriptsize ${\pm}$0.04} & 5.08 {\color{gray}\scriptsize ${\pm}$0.05} & 5.19 {\color{gray}\scriptsize ${\pm}$0.07} & 5.81 {\color{gray}\scriptsize ${\pm}$0.13} & 6.48 {\color{gray}\scriptsize ${\pm}$0.2} & 7.18 {\color{gray}\scriptsize ${\pm}$0.32}\\
		\PLClipping[$0.15$] & 5.31 & 5.41 {\color{gray}\scriptsize ${\pm}$0.03} & 5.54 {\color{gray}\scriptsize ${\pm}$0.05} & 5.63 {\color{gray}\scriptsize ${\pm}$0.06} & 6.06 {\color{gray}\scriptsize ${\pm}$0.08} & 6.53 {\color{gray}\scriptsize ${\pm}$0.14} & 6.93 {\color{gray}\scriptsize ${\pm}$0.2}\\
		\PLClipping[$0.1$] & 5.62 & 5.69 {\color{gray}\scriptsize ${\pm}$0.03} & 5.78 {\color{gray}\scriptsize ${\pm}$0.05} & 5.86 {\color{gray}\scriptsize ${\pm}$0.05} & 6.27 {\color{gray}\scriptsize ${\pm}$0.09} & 6.66 {\color{gray}\scriptsize ${\pm}$0.13} & 6.96 {\color{gray}\scriptsize ${\pm}$0.16}\\
		\hline
		\hline
		\PLRandom[$0.25$] $p{=}0.1$ & 4.49 & 4.62 {\color{gray}\scriptsize ${\pm}$0.04} & 4.82 {\color{gray}\scriptsize ${\pm}$0.07} & 4.98 {\color{gray}\scriptsize ${\pm}$0.08} & 5.80 {\color{gray}\scriptsize ${\pm}$0.16} & 6.65 {\color{gray}\scriptsize ${\pm}$0.22} & 7.59 {\color{gray}\scriptsize ${\pm}$0.34}\\
		\hline
		\PLRandom[$0.25$] $p{=}1$ & 4.62 & 4.73 {\color{gray}\scriptsize ${\pm}$0.03} & 4.90 {\color{gray}\scriptsize ${\pm}$0.06} & 5.02 {\color{gray}\scriptsize ${\pm}$0.06} & 5.62 {\color{gray}\scriptsize ${\pm}$0.13} & 6.36 {\color{gray}\scriptsize ${\pm}$0.2} & 7.02 {\color{gray}\scriptsize ${\pm}$0.27}\\
		\PLRandom[$0.1$] $p{=}1$ & 5.66 & 5.76 {\color{gray}\scriptsize ${\pm}$0.03} & 5.88 {\color{gray}\scriptsize ${\pm}$0.06} & 5.96 {\color{gray}\scriptsize ${\pm}$0.05} & 6.29 {\color{gray}\scriptsize ${\pm}$0.09} & 6.59 {\color{gray}\scriptsize ${\pm}$0.11} & 6.87 {\color{gray}\scriptsize ${\pm}$0.12}\\
		\hline
		\PLRandom[$0.25$] $p{=}2$ & 4.94 & 5.06 {\color{gray}\scriptsize ${\pm}$0.04} & 5.22 {\color{gray}\scriptsize ${\pm}$0.06} & 5.33 {\color{gray}\scriptsize ${\pm}$0.06} & 5.92 {\color{gray}\scriptsize ${\pm}$0.13} & 6.48 {\color{gray}\scriptsize ${\pm}$0.19} & 7.04 {\color{gray}\scriptsize ${\pm}$0.25}\\
		\PLRandom[$0.1$] $p{=}2$ & 5.60 & 5.67 {\color{gray}\scriptsize ${\pm}$0.02} & 5.77 {\color{gray}\scriptsize ${\pm}$0.05} & 5.84 {\color{gray}\scriptsize ${\pm}$0.05} & 6.20 {\color{gray}\scriptsize ${\pm}$0.09} & 6.49 {\color{gray}\scriptsize ${\pm}$0.09} & 6.72 {\color{gray}\scriptsize ${\pm}$0.13}\\
		\hline
	\end{tabular}
\end{table*}

As \Clipping adds an additional hyper-parameter, \tabref{tab:supp-clipping} also illustrates that $\wmax$ can easily be tuned based on clean performance. Specifically, lower $\wmax$ will eventually increase \TE and reduce confidences (alongside increasing cross-entropy loss). This increase in \TE is usually not desirable except when optimizing for robust performance, \ie, considering \RTE. Also, we found that weight clipping does not (negatively) interact with any other hyper-parameters or regularizers. For example, as described in \secref{subsec:supp-experiments-setup}, we use weight clipping in combination with AutoAugment/Cutout and weight decay without problems. Furthermore, it was not necessary to adjust our training setup (\ie, optimizer, learning rate, epochs, \etc), even for low $\wmax$.

\revision{
\subsubsection{Comparison to Other Regularization Methods}

We compare weight clipping to several simple regularization baselines, supporting our observations of the significant robustness improvements that weight clipping obtains.
Specifically, we consider regular weight decay,
dropout \cite{SrivastavaJMLR2014},
additional unlabeled/pseudo-labeled data following \cite{CarmonNIPS2019}, and the modified hinge loss (MHL) of \cite{BuschjaegerDATE2021}.
We emphasize that the latter is concurrent work designed for bit error robustness of \emph{binary} neural networks.

We found that none of these methods improves bit error robustness to the extent weight clipping does.
In fact, \tabref{tab:supp-regularization} shows that weight decay increases \RTE, which we also observed when combined with weight clipping.
Dropout, in contrast improves robustness over \Quant, but does not improve consistently over \Clipping, as exemplarily shown for \Clipping[$0.1$].
Similarly, the modified hinge loss (replacing our cross-entropy loss) suggested in \cite{BuschjaegerDATE2021} improves over \Quant with margin $1$, but reduces robustness for larger margins.
This is counter-intuitive as \cite{BuschjaegerDATE2021} advocate for large margins. However, the margin is significantly more meaningful and interpretable for binary neural networks.
We also emphasize that clean \TE actually reduces significantly when using the modified hinge loss and robustness cannot be improved over \Clipping[$0.1$].
Finally, while using additional unlabeled data does not improve over \Quant, it does consistently improve robustness of \Clipping for $\wmax \in \{0.25, 0.1, 0.05\}$.
}

\subsection{Random Bit Error Training (\Random)}
\label{subsec:supp-experiments-randbet}

\tabref{tab:supp-randbet-symmetric} shows complementary results for \Random using \emph{symmetric} quantization. Symmetric quantization generally tends to reduce robustness, \ie, increase \RTE, across all bit error rates $p$. Thus, the positive impact of \Random is pronounced, \ie, \Random becomes more important to obtain high robustness when less robust fixed-point quantization is used. These experiments also demonstrate the utility of \Random independent of the quantization scheme at hand. 

\begin{table}[t]
	\centering
	\caption{\textbf{Generalization to Profiled Bit Errors.} We show \RTE on profiled bit errors, chips 1-3, for \Random as well as \Clipping. Note that for chip 3, \Clipping[$0.05$] performs slightly better than \Random. However, using per-layer clipping, \PLRandom performs best.}
	\label{tab:supp-randbet-generalization}
	\vspace*{-0.25cm}
	\hspace*{-0.15cm}
	\begin{tabular}{| c | l | c | c | c |}
		\hline
		\multicolumn{5}{|c|}{\bfseries\CifarT: Generalization to Profiled Bit Errors}\\
		\hline
		Chip & Model & \multirow{2}{*}{\begin{tabular}{@{}c@{}}\TE\\in \%\end{tabular}} & \multicolumn{2}{c|}{\RTE in \%}\\
		\cline{4-5}
		& (\CifarT) && $p{\approx}0.86$ & $p{\approx}2.7$\\
		\hline
		\hline
		\multirow{6}{*}{1} & \Quant & 4.32 & 23.57 & 89.84\\
		& \Clipping[$0.05$] & 5.44 & 7.17 & 10.50\\
		& \Random[$0.05$] $p{=}1.5$ & 5.62 & 7.04 & 9.37\\
		\cline{2-5}
		& \PLClipping[$0.15$] & 5.27 & 6.52 & 8.48\\
		& \PLRandom[$0.25$], $p{=}1$ & 4.62 & 6.22 & 9.81\\
		& \PLRandom[$0.15$], $p{=}2$ & 4.99 & 6.14 & 7.58\\
		\hline
		\hline
		&&& $p{\approx}0.14$ & $p{\approx}1$\\
		\hline
		\multirow{6}{*}{2} & \Quant & 4.32 & 6.00 & 74.00\\
		& \Clipping[$0.05$] & 5.44 & 5.98 & 10.02\\
		& \Random[$0.05$] $p{=}1.5$ & 5.62 & 6.00 & 9.00\\
		\cline{2-5}
		& \PLClipping[$0.15$] & 5.27 & 5.64 & 8.97\\
		& \PLRandom[$0.25$], $p{=}1$ & 4.62 & 5.13 & 8.86\\
		& \PLRandom[$0.15$], $p{=}2$ & 4.99 & 5.34 & 7.34\\
		\hline
		\hline
		&&& $p{\approx}0.03$ & $p{\approx}0.5$\\
		\hline
		\multirow{6}{*}{3} & \Quant & 4.32 & 5.47 & 80.49\\
		& \Clipping[$0.05$] & 5.44 & 5.78 & 11.88\\
		& \Random[$0.05$] $p{=}1.5$ & 5.62 & 5.85 & 12.44\\
		\cline{2-5}
		& \PLClipping[$0.15$] & 5.27 & 5.52 & 16.14\\
		& \PLRandom[$0.25$], $p{=}1$ & 4.62 & 4.91 & 11.94\\
		& \PLRandom[$0.15$], $p{=}2$ & 4.99 & 5.19 & 8.63\\
		\hline
	\end{tabular}
\end{table}

As ablation for \Random, we consider two variants of \Random motivated by related work \cite{KoppulaMICRO2019}. Specifically, in \cite{KoppulaMICRO2019}, the bit error rate seen during training is increased slowly during training. Note that \cite{KoppulaMICRO2019} trains on fixed bit error patterns. Thus, increasing the bit error rate during training is essential to ensure that the DNN is robust to any bit error rate $p' < p$ smaller than the target bit error rate. While this is generally the case using our \Random, \tabref{tab:supp-randbet-variants} shows that slowly increasing the random bit error rate during training, called ``curricular'' \Random, has no significant benefit over standard \Random. In fact, \RTE increases slightly. Similarly, we found that \Random tends to increase the range of weights: the weights are ``spread out'', \cf \figref{fig:supp-clipping} (top right). This also increases the quantization range, which has negative impact on robustness. Thus, we experimented with \Random using two weight updates per iteration: one using clean weights, one on weights with bit errors. This is in contrast to averaging both updates as described in the main paper. Updates computed from perturbed weights are limited to the current quantization ranges, \ie, the maximum absolute error cannot change. This is ensured through projection. This makes sure that \Random does not increase the quantization range during training as changes in the quantization range are limited to updates from clean weights. Again, \tabref{tab:supp-randbet-variants} shows this variant to perform slightly worse.

\tabref{tab:supp-randbet-resnet} also shows results on \CifarT using ResNet-20 and ResNet-50. We note that, in both cases, we use group normalization (GN) instead of batch normalization (BN) as outlined in \secref{subsec:supp-experiments-bn}. ResNet-50, in particular, suffers from using GN due to the significant depth: the clean \TE reduces from $3.67\%$ to $6.81\%$ in \tabref{tab:supp-accuracy}. Nevertheless, \Clipping and \Random remain effective against random bit errors, even for higher bit error rates of $p = 1.5\%$. This is striking as ResNet-50 consists of roughly $23.5\text{Mio}$ weights, compared to $5.5\text{Mio}$ of the used SimpleNet in the main paper.

Following the \Random algorithm outlined in the main paper, \Random adds an additional forward and backward pass during training, increasing training complexity roughly by a factor of two. In practice, however, we found that training time for \Random (in comparison with \Clipping) roughly triples. This is due to our custom implementation of bit error injection, which was not optimized for speed. However, we believe that training time can be reduced significantly using an efficient CUDA implementation of bit error injection. We also note that inference time remains unchanged. In this respect, bit error mitigation strategies in hardware are clearly less desirable due to increased inference time, space and energy consumption.

\subsection{Per-Layer \Clipping and \Random}
\label{subsec:supp-plc}

\begin{table}[t]
	\centering
	\caption{\textbf{Fixed Pattern Bit Error Training.} We report \RTE for training on fixed, profiled bit error patterns (\Pattern). Note that for \Pattern on chip 1/2 we used only the stuck-at-errors shown in \figref{fig:supp-errors}, which is why the bit error rates deviate from those reported in the main paper, \cf \figref{fig:supp-errors}.}
	\label{tab:supp-randbet-baselines}
	\vspace*{-0.25cm}
	\begin{tabular}{| L{4cm} | C{1cm} | C{1cm} |}
		\hline
		\multicolumn{3}{|c|}{\bfseries \CifarT: Fixed Pattern Training}\\
		\hline
		Model (\CifarT) & \multicolumn{2}{c|}{\RTE in \%, $p$ in \%}\\
		\hline
		\hline
		\textbf{\emph{Profiled} Bit Errors (Chip 1)} & $p{\approx}0.39$ & $p{\approx}1.22$\\
		\hline
		\Pattern, $p{\approx}1.22$ & 9.52 & 7.20\\
		\Pattern, $p{\approx}0.39$ & 5.77 & 67.87\\
		\Pattern[$0.15$], $p{\approx}1.22$ & 7.67 & 6.52\\
		\Pattern[$0.15$], $p{\approx}0.39$ & 5.94 & 30.96\\
		\hline
		\hline
		\textbf{\emph{Profiled} Bit Errors (Chip 2)} & $p{\approx}0.1$ & $p{\approx}0.63$\\
		\hline
		\Pattern $p{\approx}0.63$ &  85.84 & 10.76\\
		\Pattern, $p{\approx}0.1$ & 90.56 & 5.93\\
		\Pattern[$0.15$] $p{\approx}0.63$ & 12.02 & 8.70\\
		\Pattern[$0.15$] $p{\approx}0.1$ & 90.68 & 6.51\\
		\hline
	\end{tabular}
\end{table}
\begin{table}[t]
	\centering
	\caption{\textbf{Results for Probabilistic Guarantees.}. Average \RTE and standard deviation for $l = 1\text{Mio}$ random bit error patterns. In comparison with the results for $l = 50$ from the main paper, there are no significant changes in \RTE.}
	\label{tab:supp-stress}
	\vspace*{-0.25cm}
	\hspace*{-0.15cm}
	\begin{tabular}{|l | c | c | c |}
		\hline
		\multicolumn{4}{|c|}{\bfseries\CifarT: Stress Test for Guarantees}\\
		\hline
		Model & \multirow{2}{*}{\begin{tabular}{@{}c@{}}\TE\\in \%\end{tabular}} & \multicolumn{2}{c|}{\RTE in \%, $p = 1\%$}\\
		\hline
		(\CifarT) && $l=50$ & {\color{red}$\mathbf{l=1\text{Mio}}$}\\
		\hline
		\hline
		\Quant & 4.32 & 32.05 {\color{gray}\scriptsize ${\pm}$6} & 31.97 {\color{gray}\scriptsize ${\pm}$6.35}\\
		\Clipping[$0.05$] & 5.44 & 7.18 {\color{gray}\scriptsize ${\pm}$0.16} & 7.19 {\color{gray}\scriptsize ${\pm}$0.2}\\
		\Random[$0.05$] $p{=}2$ & 5.42 & 6.71 {\color{gray}\scriptsize ${\pm}$0.11} & 6.73 {\color{gray}\scriptsize ${\pm}$0.15}\\
		\hline
		\PLClipping[$0.15$] & 5.31 & 6.53 {\color{gray}\scriptsize ${\pm}$0.14} & 6.52 {\color{gray}\scriptsize ${\pm}$0.14}\\
		\PLRandom[$0.25$], $p{=}1$ & 4.62 & 6.36 {\color{gray}\scriptsize ${\pm}$0.2} & 6.29 {\color{gray}\scriptsize ${\pm}$0.2}\\
		\PLRandom[$0.15$], $p{=}2$ & 4.99 & \bfseries 6.12 {\color{gray}\scriptsize ${\pm}$0.13} & \bfseries 6.12 {\color{gray}\scriptsize ${\pm}$0.14}\\
		\hline
	\end{tabular}
\end{table}
\begin{figure*}[t]
	\centering
	\begin{subfigure}{0.26\textwidth}
		\includegraphics[height=3.4cm]{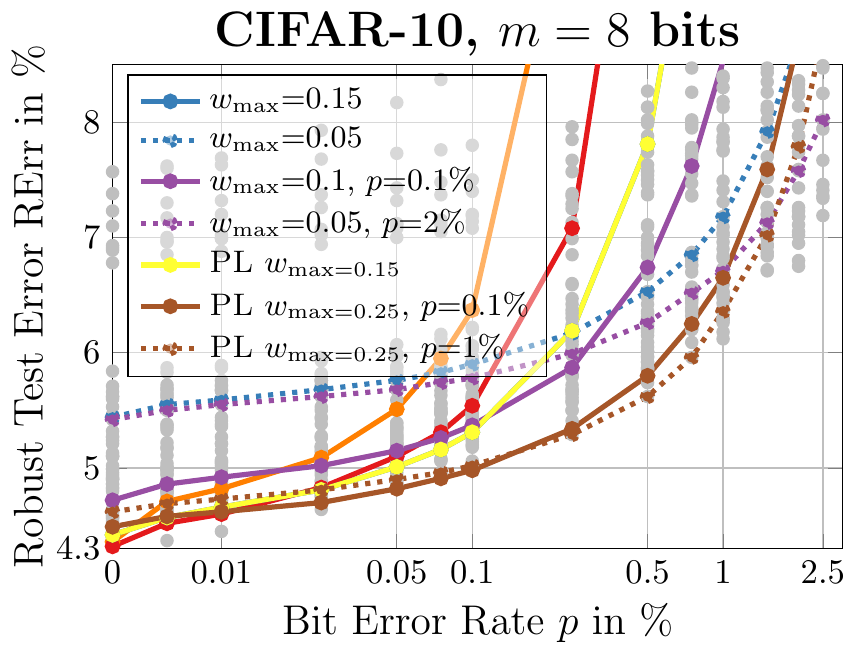}
	\end{subfigure}
	\begin{subfigure}{0.24\textwidth}
		\includegraphics[height=3.4cm]{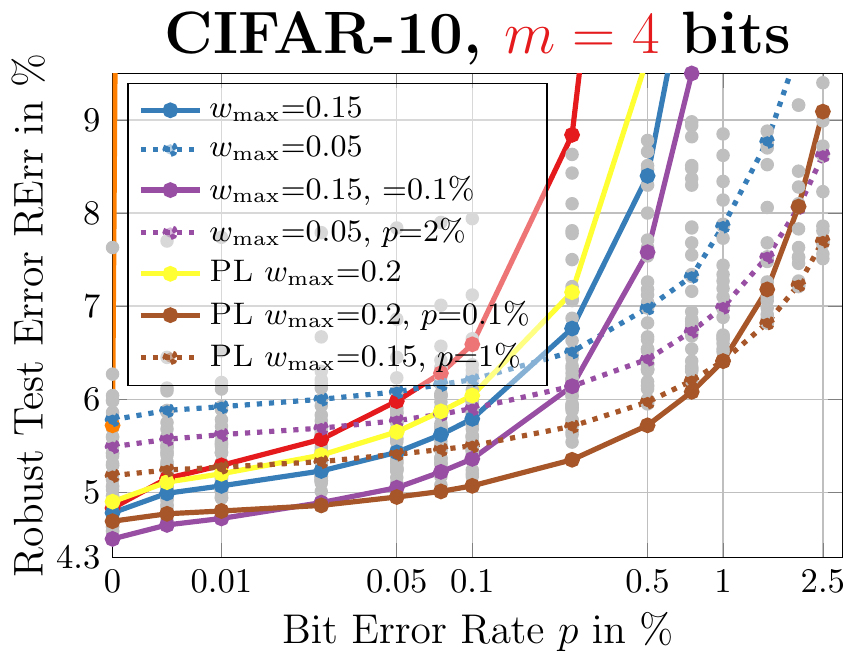}
	\end{subfigure}
	\begin{subfigure}{0.24\textwidth}
		\includegraphics[height=3.4cm]{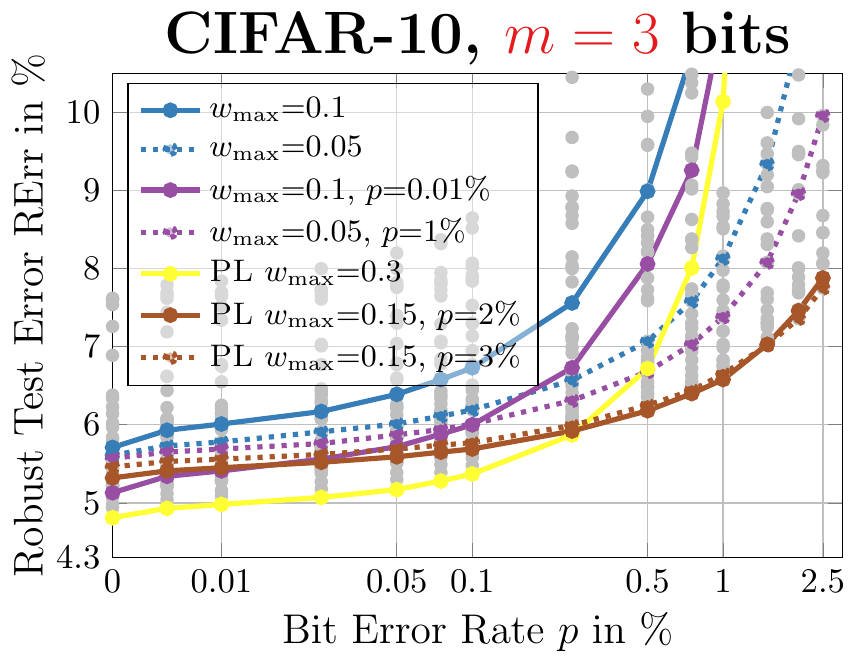}
	\end{subfigure}
	\begin{subfigure}{0.24\textwidth}
		\includegraphics[height=3.4cm]{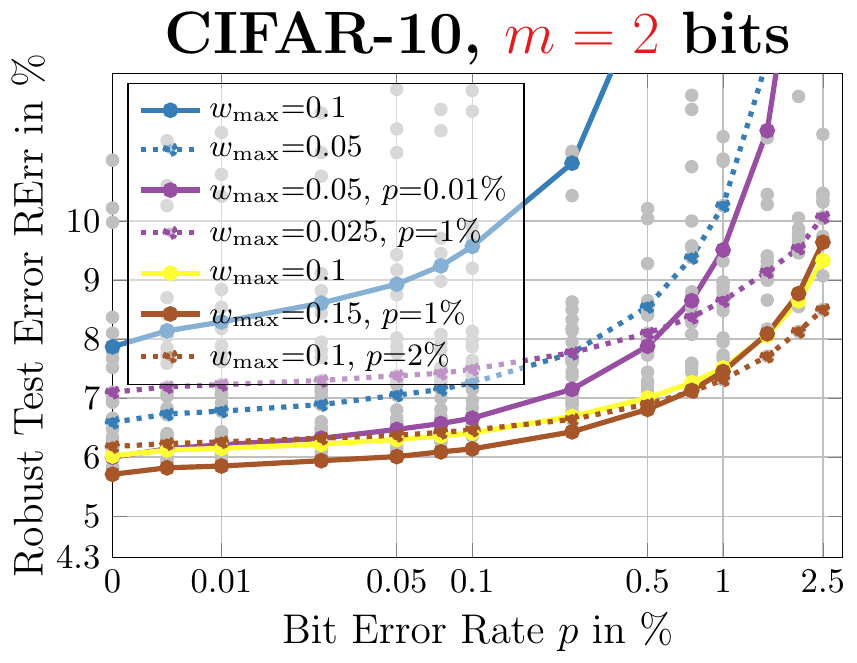}
	\end{subfigure}
	\\[2px]
	
	\begin{subfigure}{0.26\textwidth}
		\includegraphics[height=3.4cm]{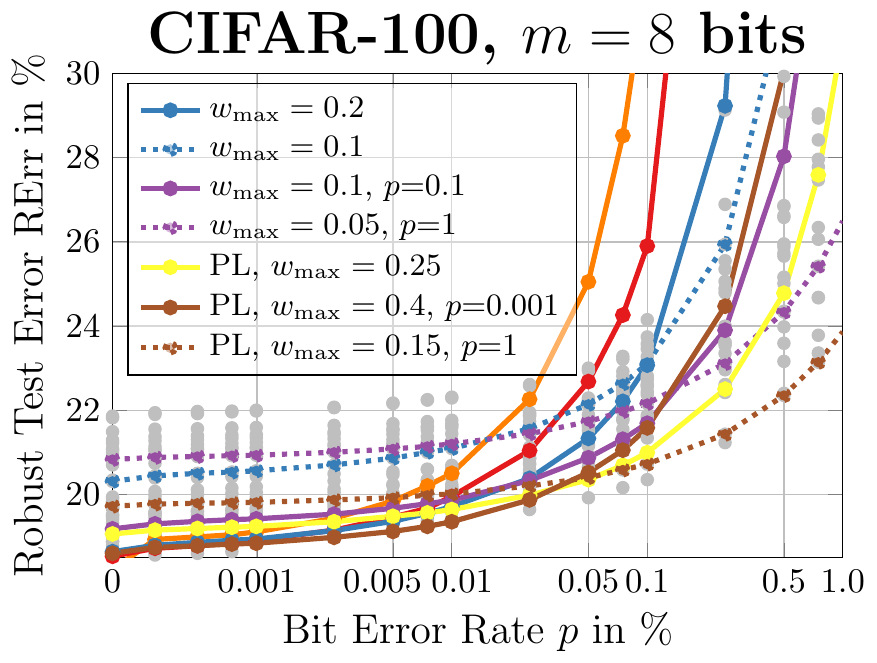}
	\end{subfigure}
	\begin{subfigure}{0.24\textwidth}
		\includegraphics[height=3.4cm]{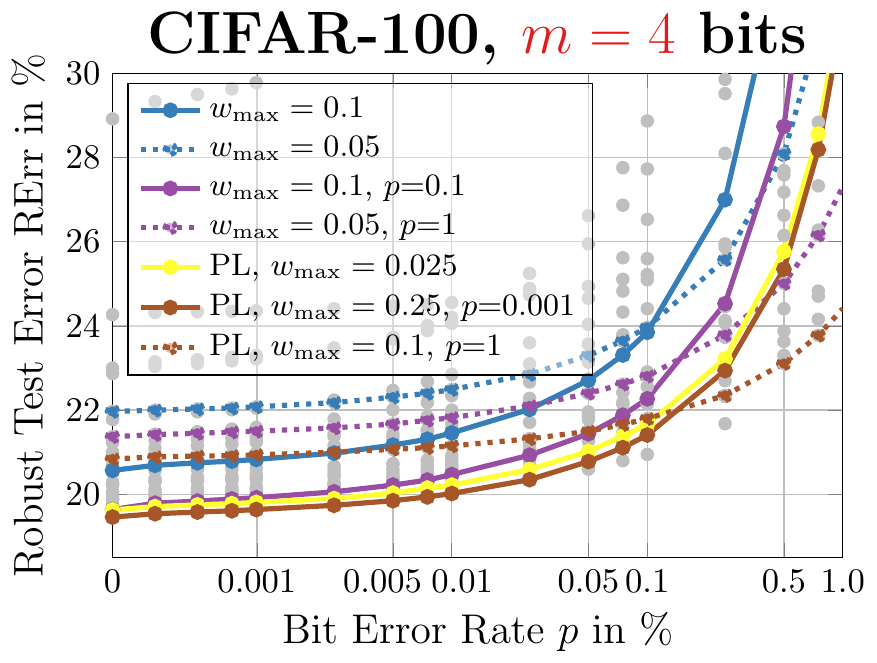}
	\end{subfigure}
	\begin{subfigure}{0.24\textwidth}
		\hphantom{\includegraphics[height=3.4cm]{cifar100_summary_4bit.pdf}}
	\end{subfigure}
	\begin{subfigure}{0.24\textwidth}
		\hphantom{\includegraphics[height=3.4cm]{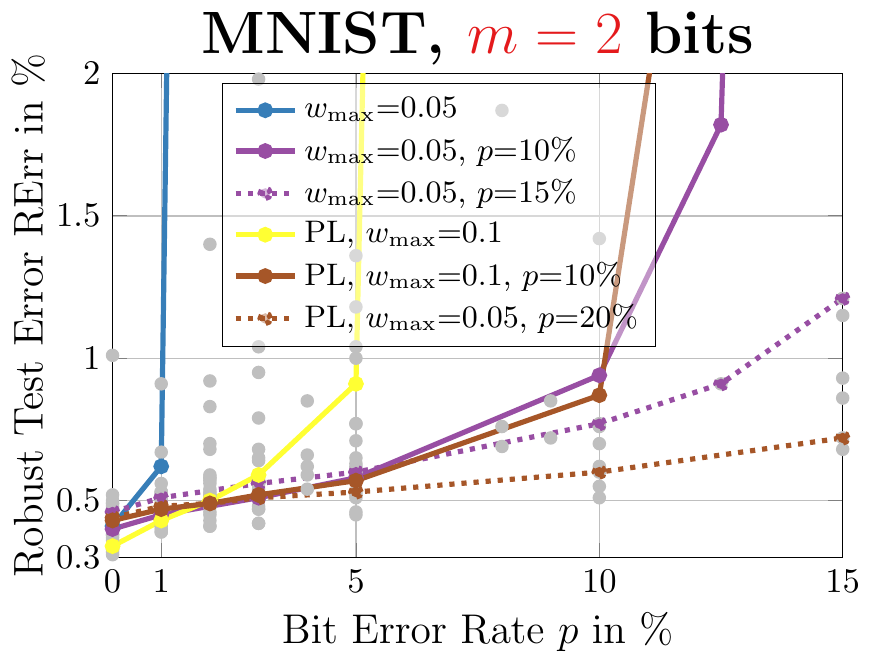}}
	\end{subfigure}
	\\[2px]
	\begin{subfigure}{0.24\textwidth}
		\includegraphics[height=3.4cm]{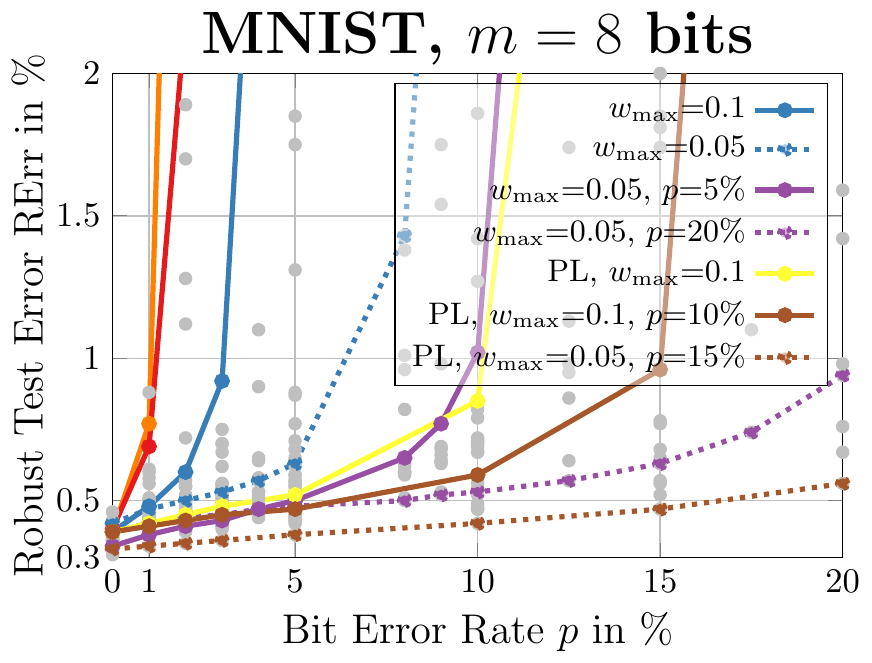}
	\end{subfigure}
	\begin{subfigure}{0.24\textwidth}
		\includegraphics[height=3.4cm]{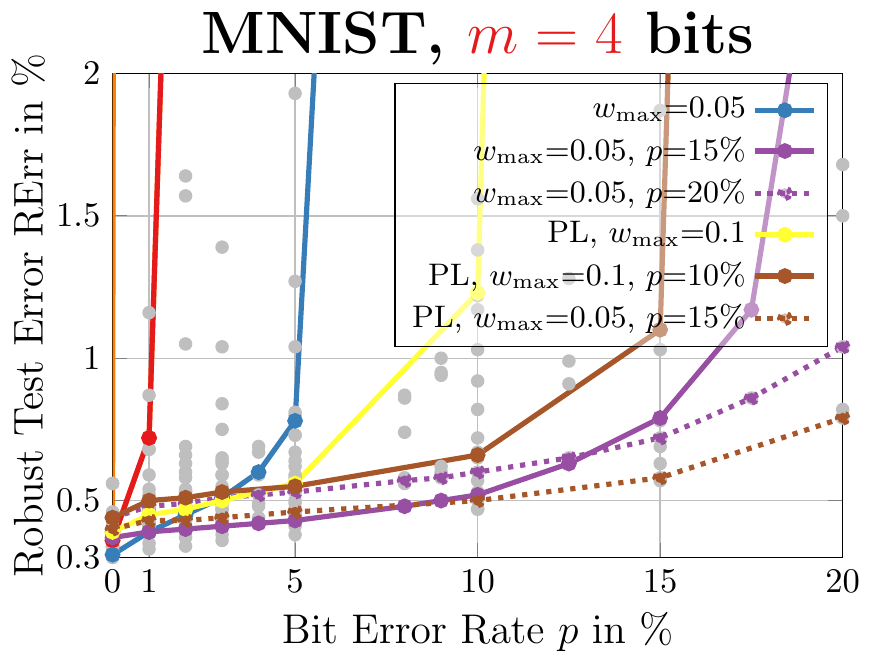}
	\end{subfigure}
	\begin{subfigure}{0.24\textwidth}
		\includegraphics[height=3.4cm]{mnist_summary_2bit.pdf}
	\end{subfigure}
	\begin{subfigure}{0.24\textwidth}
		\hphantom{\includegraphics[height=3.4cm]{mnist_summary_2bit.pdf}}
	\end{subfigure}
	\\[2px]
	\begin{subfigure}{0.24\textwidth}
		\includegraphics[height=3.4cm]{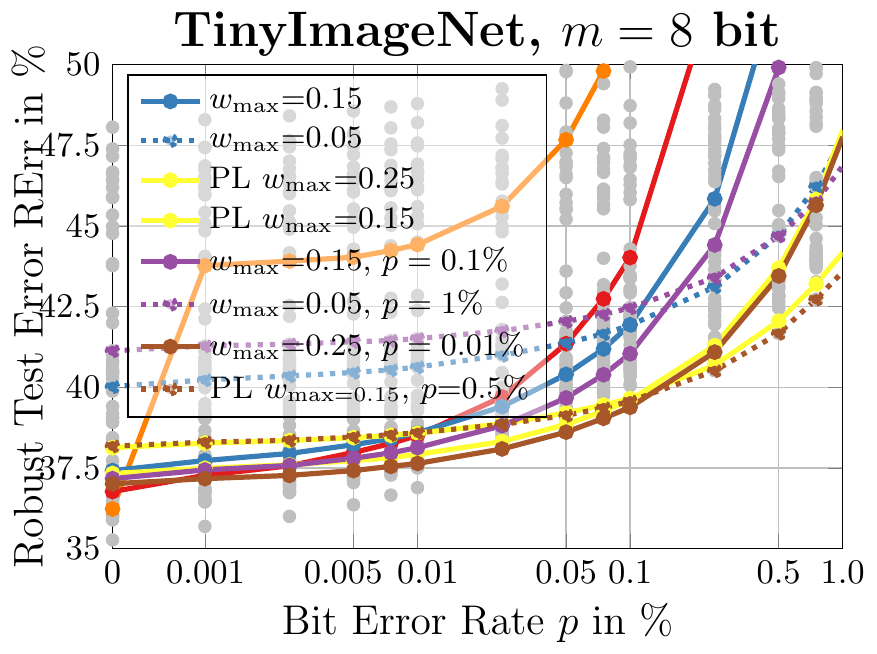}
	\end{subfigure}
	\begin{subfigure}{0.24\textwidth}
		\includegraphics[height=3.4cm]{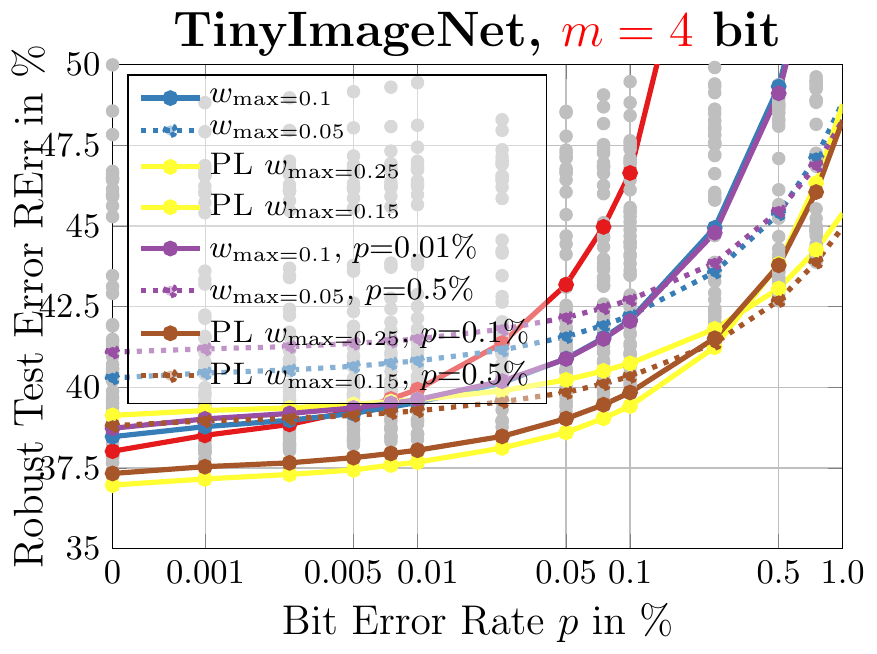}
	\end{subfigure}
	\begin{subfigure}{0.24\textwidth}
		\includegraphics[height=3.4cm]{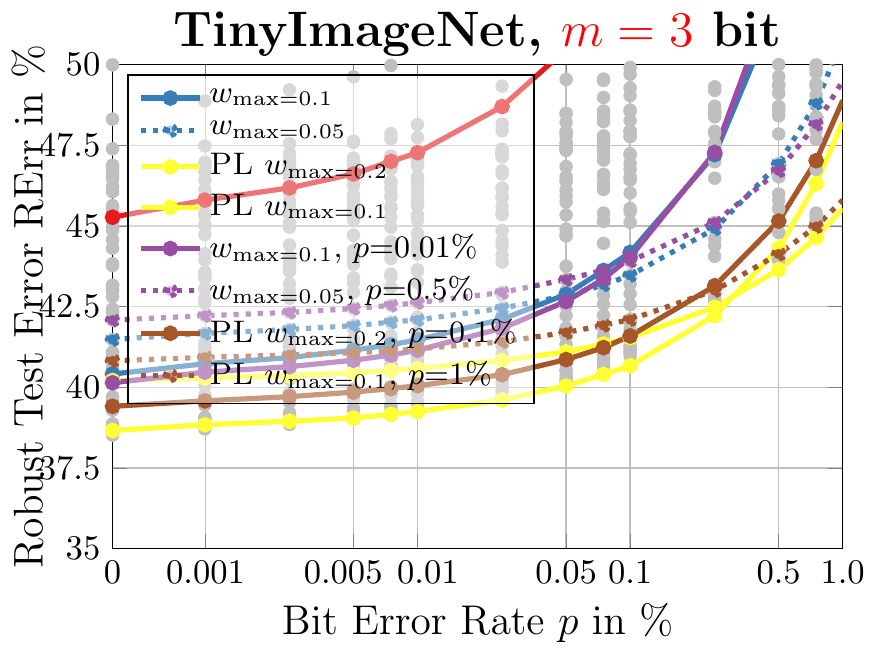}
	\end{subfigure}
	\begin{subfigure}{0.24\textwidth}
		\includegraphics[height=3.4cm]{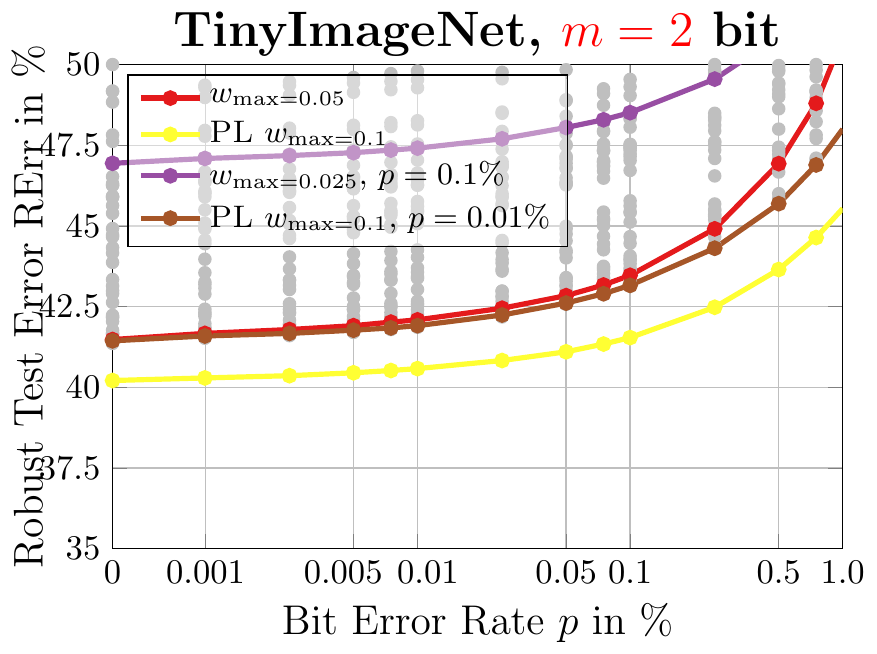}
	\end{subfigure}
	\\[2px]
	
	\hspace*{-0.1cm}
	\fbox{
		\begin{subfigure}{0.98\textwidth}
			\centering 
			\includegraphics[width=0.8\textwidth]{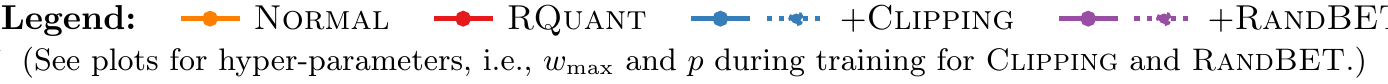}
		\end{subfigure}
	}
	\vspace*{-8px}
	\caption{\textbf{Summary Results on \CifarT, \CifarH, \MNIST \revision{ and TinyImageNet}.} We plot \RTE against bit error rate, highlighting individual \Clipping, \Random, \PLClipping and \PLRandom models. Note that the main paper, in contrast, presents the best, \ie, lowest \RTE, model for each bit error rate $p$ individually. Instead, individual models help to illustrate the involved trade-offs: \Clipping with small $\wmax$ or \Random with high bit error rate $p$ increases the clean \TE, thereby also increasing \RTE for very small bit error rates. However, \RTE against large bit error rates can be reduced.
	}
	\label{fig:supp-summary}
\end{figure*}

\figref{fig:supp-plclipping} shows the per-layer weight ranges of \Quant trained without weight clipping that we used to determine the constraints for \emph{per-layer} weight clipping (\PLClipping) on \CifarT. Specifically,  we found that only few layers exhibit large weight ranges, \eg, the first few convolutional layers (``conv1'' ``and conv2'' in \figref{fig:supp-plclipping}), few group normalization layers (``regn7'' or ``regn9'') and the final logit layer (``logits''). Thus, these layers are affected significantly when reducing $\wmax$ and lead to poor (clean) \TE in \tabref{tab:supp-randbet-baselines}, \eg, for $\wmax = 0.05$. Therefore, we also consider per-layer weight clipping, \ie, using individual values $w_{\text{max},l}$ for each layer, as described in detail in the main paper. We use the \Quant model from \figref{fig:supp-plclipping} as reference for all models on \CifarT. On \MNIST and \CifarH, we use the corresponding \Quant models as reference. When additionally combined with \Random, we use \PLRandom do denote random bit error training \emph{with} per-layer weight clipping.

\tabref{tab:supp-plclipping} shows results for \PLClipping and \PLRandom considering various bit error rates $p$. In comparison to the results in \tabref{tab:supp-randbet-symmetric}, robustness can be improved considerably, while also improving clean performance. For example, \Clipping[$0.1$] (\ie, global) obtains $13.14\%$ \RTE a bit error rate of $p = 1.5\%$. Using \PLClipping, this can be improved to $6.96\%$ for \PLClipping[$0.1$]. Similarly, \Random benefits from per-layer clipping. However, the difference between \PLClipping and \PLRandom, considering \RTE, is significantly smaller than before. This illustrates that \emph{per-layer} weight clipping can have tremendous impact on robustness.

\subsection{Profiled Bit Errors}
\label{subsec:supp-randbet-baselines}

\begin{table*}[t]
	\centering
	\caption{\textbf{Input Robustness for \Clipping and \PLRandom.} Average \RTE for bit errors in inputs. Images are quantized using $m = 8$ bit quantization (per channel) in $[0, 1]$, which does \emph{not} introduce errors as the images are already provided in $8$-bit (per channel). Note that \PLRandom considers only bit errors in weights during training. We note that \TE is reported on the first $1\text{k}$ test images, to be comparable with \RTE for bit errors in inputs. Extreme \Clipping generally reduces robustness to such input perturbations. Similarly, \PLRandom (bit errors in weights during training) may reduce robustness, indicating that robustness against bit errors in weights is in conflict with robustness against bit errors in the inputs. Nevertheless, \PLRandom (training on bit errors in inputs) can improve robustness significantly.}
	\label{tab:supp-inputs}
	\vspace*{-0.2cm}
	\begin{tabularx}{\textwidth}{|X|c|c|c|}
		\hline
		\multicolumn{4}{|c|}{\bfseries \CifarT: Bit Error Robustness in Inputs (\Random)}\\
		\hline
		Model (\CifarT) & \multirow{2}{*}{\begin{tabular}{@{}c@{}}\TE\\in \%\end{tabular}} & \multicolumn{2}{c|}{\RTE in \%}\\
		\cline{3-4} 
		bit errors in {\color{colorbrewer5}inputs}, $p$ in \% && $p{=}0.1$ & $p{=}0.5$ \\
		\hline
		\hline
		\Quant & 4.22 & 11.10 & 23.70\\
		\Clipping[$0.25$] & 4.53 & 11.30 & 22.70\\
		\Clipping[$0.1$] & 4.81 & 12.50 & 25.40\\
		\hline
		\PLClipping[$0.25$] & 4.92 & 10.80 & 22.80\\
		\PLClipping[$0.1$] & 5.73 & 12.80 & 26.50\\
		\hline
		\hline
		\PLRandom[$0.75$] (weights only), $p_w{=}0.1$ & 4.59 & 10.20 & 22.20\\
		\PLRandom[$0.5$] (weights only), $p_w{=}0.1$ & 4.48 & 10.90 & 21.00\\
		\PLRandom[$0.25$] (weights only), $p_w{=}0.1$ & 4.46 & 11.00 & 22.80\\
		\hline
		\PLRandom[$0.5$] (weights only), $p_w{=}1$ & 5.04 & 11.00 & 24.20\\
		\PLRandom[$0.25$] (weights only), $p_w{=}1$ & 4.64 & 11.30 & 22.40\\
		\PLRandom[$0.1$] (weights only), $p_w{=}1$ & 5.63 & 13.00 & 25.80\\
		\hline
		\hline
		\PLRandom[$0.25$] weights+inputs, $p_w{=}0.1$, {\color{colorbrewer5}$p_i{=}0.1$} & 4.94 & 7.60 & 11.70\\
		\PLRandom[$0.25$] weights+inputs, $p_w{=}0.1$, {\color{colorbrewer5}$p_i{=}0.5$} & 4.74 & 8.00 & 12.90\\
		\hline
		\PLRandom[$0.1$] weights+inputs, $p_w{=}0.1$, {\color{colorbrewer5}$p_i{=}0.1$} & 5.94 & 9.00 & 15.10\\
		\PLRandom[$0.1$] weights+inputs, $p_w{=}0.1$, {\color{colorbrewer5}$p_i{=}0.5$} & 5.72 & 8.80 & 17.50\\
		\hline
		\PLRandom[$0.25$] weights+inputs, $p_w{=}1$, {\color{colorbrewer5}$p_i{=}0.1$} & 5.57 & 7.70 & 9.10\\
		\PLRandom[$0.25$] weights+inputs, $p_w{=}1$, {\color{colorbrewer5}$p_i{=}0.5$} & 5.39 & 7.60 & 8.90\\
		\hline
		\hline
		\PLRandom[$0.25$] weights+inputs+activations, $p_w{=}0.1$, {\color{colorbrewer5}$p_i{=}0.1$}, {\color{colorbrewer4}$p_a{=}0.1$} & 5.16 & 7.90 & 12.20\\
		\PLRandom[$0.25$] weights+inputs+activations, $p_w{=}0.1$, {\color{colorbrewer5}$p_i{=}0.1$}, {\color{colorbrewer4}$p_a{=}0.5$} & 5.02 & 7.60 & 12.00\\
		\hline
		\PLRandom[$0.25$] weights+inputs+activations, $p_w{=}1$, {\color{colorbrewer5}$p_i{=}0.1$}, {\color{colorbrewer4}$p_a{=}0.1$} & 9.09 & 11.50 & 13.80\\
		\PLRandom[$0.25$] weights+inputs+activations, $p_w{=}1$, {\color{colorbrewer5}$p_i{=}0.1$}, {\color{colorbrewer4}$p_a{=}0.5$} & 8.97 & 11.10 & 14.10\\
		\hline
	\end{tabularx}
\end{table*}
\begin{table*}[t]
	\centering
	\caption{\textbf{Activation Robustness for \Clipping and \PLRandom.} Average \RTE for bit errors in activations; the same $m = 8$ bit quantization is used for weights and activations. \PLRandom considers only bit errors in weights during training. We note that activation quantization without bit errors has negligible impact on \TE. However, when training on bit errors in activations, \TE may increase similarly as with bit errors in the weights. Regarding robustness against bit errors in activations, \Clipping and \PLRandom improve robustness, however, extreme clipping or \Random with large bit error rates in the weights reduces robustness.}
	\label{tab:supp-activations}
	\vspace*{-0.2cm}
	\begin{tabularx}{\textwidth}{|X|c|c|c|c|}
		\hline
		\multicolumn{5}{|c|}{\bfseries \CifarT: Bit Error Robustness in Activations}\\
		\hline
		Model (\CifarT) & \multirow{2}{*}{\begin{tabular}{@{}c@{}}\TE in \%\\(\emph{w/o} quant. act.)\end{tabular}} & \multirow{2}{*}{\begin{tabular}{@{}c@{}}\TE in \%\\(quant. act.)\end{tabular}} &\multicolumn{2}{c|}{\RTE in \%}\\
		\cline{4-5} 
		$8$-bit activation quantization, bit errors in {\color{colorbrewer4}activations}, $p$ in \% &&& $p{=}0.1$ & $p{=}0.5$ \\
		\hline
		\hline
		\Quant & 4.32 & 4.53 & 8.93 & 47.12\\
		\Clipping[$0.25$] & 4.58 & 4.67 & 7.97 & 31.98\\
		\Clipping[$0.1$] & 4.82 & 5.13 & 7.86 & 24.38\\
		\hline
		\PLClipping[$0.25$] & 4.96 & 5.16 & 7.38 & 21.58\\
		\PLClipping[$0.1$] & 5.62 & 5.84 & 8.72 & 27.36\\
		\hline
		\hline
		\PLRandom[$0.75$] (weights only), $p_w{=}0.1$ & 4.57 & 4.78 & 8.74 & 38.99\\
		\PLRandom[$0.5$] (weights only), $p_w{=}0.1$ & 4.48 & 4.79 & 7.46 & 27.67\\
		\PLRandom[$0.25$] (weights only), $p_w{=}0.1$ & 4.49 & 4.71 & 7.25 & 24.94\\
		\hline
		\PLRandom[$0.5$] (weights only), $p_w{=}1$ & 5.11 & 5.33 & 8.10 & 26.90\\
		\PLRandom[$0.25$] (weights only), $p_w{=}1$ & 4.62 & 4.83 & 6.92 & 19.83\\
		\PLRandom[$0.1$] (weights only), $p_w{=}1$ & 5.66 & 5.92 & 9.31 & 35.79\\
		\hline
		\hline
		\PLRandom[$0.25$] activations only, {\color{colorbrewer4}$p_a{=}0.1$} & 4.73 & 4.84 & 6.40 & 12.10\\
		\PLRandom[$0.25$] activations only, {\color{colorbrewer4}$p_a{=}0.5$} & 5.43 & 5.68 & 6.74 & 10.16\\
		\hline
		\PLRandom[$0.1$] activations only, {\color{colorbrewer4}$p_a{=}0.1$} & 5.92 & 6.09 & 7.56 & 12.62\\
		\PLRandom[$0.1$] activations only, {\color{colorbrewer4}$p_a{=}0.5$} & 6.52 & 6.63 & 7.77 & 10.81\\
		\hline
		\hline
		\PLRandom[$0.25$] weights+activations, $p_w{=}1$, {\color{colorbrewer4}$p_a{=}0.1$} & 7.59 & 7.80 & 8.88 & 11.79\\
		\PLRandom[$0.25$] weights+activations, $p_w{=}1$, {\color{colorbrewer4}$p_a{=}0.5$} & 7.66 & 7.89 & 9.09 & 12.17\\
		\PLRandom[$0.25$] weights+activations, $p_w{=}1$, {\color{colorbrewer4}$p_a{=}1$} & 7.68 & 7.83 & 9.05 & 12.07\\
		\hline
		\hline
		\PLRandom[$0.25$] weights+inputs+activations, $p_w{=}1$, {\color{colorbrewer5}$p_i{=}0.1$}, {\color{colorbrewer4}$p_a{=}0.1$} & 9.16 & 9.31 & 10.54 & 13.51\\
		\PLRandom[$0.25$] weights+inputs+activations, $p_w{=}1$, {\color{colorbrewer5}$p_i{=}0.1$}, {\color{colorbrewer4}$p_a{=}0.5$} & 9.01 & 9.32 & 10.56 & 13.65\\
		\hline
	\end{tabularx}
\end{table*}

Following the evaluation on profiled bit errors outlined in \appref{subsec:supp-errors-profiled}, \tabref{tab:supp-randbet-generalization} shows complementary results for \Clipping[$0.05$], \Random[$0.05$], \PLClipping[$0.15$] and \PLRandom[$0.15$] trained with $p = 1.5\%$ and $p = 2\%$, respectively, on all profiled chips. Note that for particularly extreme cases, such as chip 3, \Clipping might perform slightly better than \Random, indicating a significantly different bit error distribution as assumed in the main paper. Nevertheless, \PLClipping as well as \PLRandom are able to cope even with particularly difficult bit error distributions. Overall, \PLRandom generalizes reasonably well, with very good results on chip 1 and chip 2. Note that, following \figref{fig:supp-errors}, the bit errors in chip 2 are strongly aligned along columns. Results on chip 3 are slightly worse. However, \PLRandom does not fail catastrophically with only a $\sim1\%$ increase in \RTE compared to chips 1 and 2.

In \tabref{tab:supp-randbet-baselines}, we follow the procedure of \appref{subsec:supp-errors-profiled} considering only stuck-at-$0$ and stuck-at-$1$ bit errors (\ie, where $p_{\text{1t0}}$ and $p_{\text{0t1}}$ are $1$). This is illustrated in \figref{fig:supp-errors} (right). Thus, the bit error rates deviate slightly from those reported in \tabref{tab:supp-randbet-generalization}, see the table in \appref{subsec:supp-errors-profiled} for details. Furthermore, We consider only one weight-to-SRAM mapping, \ie, without offset. \Pattern is trained and evaluated on the exact same bit error pattern, but potentially with different bit error rates $p$. Note that the bit errors for $p' < p$ are a subset of those for bit error rate $p$. Thus, it is surprising that, on both chips 1 and 2, \Pattern trained on higher bit error rates does not even generalize to lower bit error rates (\ie, higher voltage). This is problematic in practice as the DNN accelerator should not perform worse when increasing voltage.

\subsection{Guarantees from Prop. \ref{prop:bound}}
\label{subsec:experiments-stress}

Based on the bound derived in \secref{subsec:supp-bound}, we conduct experiments with $l = 1\text{Mio}$ random bit error patterns, such that $l \gg n$ where $n = \text{10k}$ is the number of test examples on \CifarT. Considering Prop. \ref{prop:bound}, this would guarantee a deviation in \RTE of at most $4.1\%$ with probability at least $99\%$. As shown in \tabref{tab:supp-stress}, the obtained \RTE with $1\text{Mio}$ random bit error patterns deviates insignificantly from the results in the main paper. Only standard deviation of \RTE increases slightly. These results emphasize that the results for \Clipping and \Random from the main paper generalize well.

\subsection{Summary Results}
\label{subsec:supp-experiments-summary}

\figref{fig:supp-summary} summarizes our results: In contrast to the main paper, we consider individual \Clipping and \Random models instead of focusing on the best results per bit error rate $p$. Additionally, we show our complete results for lower precisions, \ie, $m = 4,3,2$. Note that these results, in tabular form, are included \revision{at the end of this supplementary material}. Moderate \Clipping, \eg, using $\wmax = 0.15$ on \CifarT (in {\color{colorbrewer2}red} solid), has negligible impact on clean \TE (\ie, $p = 0$ on the x-axis) while improving robustness beyond $p = 0.1\%$ bit error rate. Generally, however, higher robustness is obtained at the cost of increased clean \TE, \eg, for $\wmax = 0.05$ (in {\color{colorbrewer2}blue} dotted). Here, it is important to note that in low-voltage operation, only \RTE matters -- clean \TE is only relevant for voltages higher than \Vmin. Per-layer weight clipping, \ie, \PLClipping, is able to avoid the increase in \TE in many cases, while preserving improved robustness. \Random further improves robustness, both on top of \Clipping and \PLClipping, for high bit error rates while continuing to increase (clean) \TE slightly. For example, \Random with $\wmax = 0.05$ and trained with $p = 2\%$ bit errors increases clean \TE to $5.42\%$ but is also able to keep \RTE below $7\%$ up to $p = 1\%$ bit error rate (in {\color{colorbrewer4}violet} dotted). While per-layer clipping, \ie, \PLClipping[$0.15$] (in {\color{colorbrewer6}yellow}), does not improve robustness compared to \Clipping[$0.05$], clean \TE is lowered. More importantly, using \PLRandom[$0.15$] (in {\color{colorbrewer7}brown}) clearly outperforms most other approaches, showing that \Random is particularly effective on top of \PLClipping, in contrast to ``just'' \Clipping.

The advantage of per-layer clipping, \ie, \PLClipping and \PLRandom, are pronounced when reducing precision. For example, using $m = 2$, \PLClipping not only boosts robustness significantly, but also avoids a significant increase in \TE. As result, using \PLRandom instead of \PLClipping is only necessary for high bit error rates, \eg, above $p = 1\%$. Similar trade-offs can be observed on \CifarH and \MNIST. On \CifarH, we see that task difficulty also reduces the bit error rate that is tolerable without significant increase it \RTE. Here, $p = 0.1\%$ increases \RTE by more than $3\%$, even with \Random (and weight clipping). Furthermore, \CifarH demonstrates that \Clipping and \Random are applicable to significantly larger architectures such as Wide ResNets without problems. On \MNIST, in contrast, bit error rates of up to $p = 20\%$ are easily possible. At such bit error rates, the benefit of \Random is extremely significant as even \Clipping[$0.025$] exhibits very high \RTE of $32.68\%$ at $p = 20\%$, \cf \tabref{tab:supp-summary-mnist}.

\revision{These observations can be confirmed on \MNIST and \CifarH. On TinyImageNet, however, it is more difficult to find a fixed \Random configuration that consistently outperforms \Clipping across multiple bit errors rates. While the main paper shows that \Random does improve consistently, this considers the \emph{best} hyper-parameters for each bit error rate. \figref{fig:supp-summary}, in contrast, shows that \Clipping might be easier to tune for multiple bit error rates. This is emphasized on low bit error rates $m < 8$ where we generally found that the benefit of \Random over \Clipping reduces slightly.}

\revision{
\subsection{Post-Training Quantization}
\label{subsec:supp-experiments-post}

So far, we applied quantization during training, \ie, we performed quantization-aware training \cite{JacobCVPR2018,KrishnamoorthiARXIV2018}. However, both (global and per-layer) weight clipping as well as (bit) error training can be applied in a post-training quantization setting. To this end, for \Random, bit errors are simulated through $L_0$ noise on weights. Specifically, with probability $p_{L_0}$ each weight $w_i$ is changed to a (uniformly) random value $\tilde{w}_i \in [-\wmax, \wmax]$. The same error model applies for per-layer weight clipping. Note that training with $L_0$ errors with probability $p_{L_0}$ simulates bit error training with $p = m\cdot p_{L_0}$, referred to as \LRandom. We apply our robust fixed-point quantization with $m = 8$ bits \emph{after} training to evaluate robustness to random bit errors. In \tabref{tab:supp-post-robustness}, we demonstrate that both \Clipping and \LRandom also provide robustness in a post-training quantization context. This allows to train robust models without knowing the exact quantization and precision used for deployment in advance.
}

\subsection{Bit Errors in Activations and Inputs}
\label{subsec:supp-experiments-activations-inputs}

\textbf{Setup:} We quantize inputs using $m = 8$ bits and $\qmin = 0$/$\qmax = 1$ in \eqnref{eq:supp-quantization} (using asymmetric quantization into unsigned integers). This does not introduce errors as images are typically provided using $8$ bit (\ie, $256$ distinct values) per pixel per color (\ie, channel). Random bit errors are injected once before the forward pass, this can easily be done per batch. As with random bit errors in weights, we consider $50$ samples of random bit errors per example per model. The bit error patterns are the same across models for comparison, \ie, in the notation of the main paper, we sample $u \sim U(0,1)^{D\times m}$ for each image and attempt. Here, $D$ is the input dimensionality ($D = 32\cdot32\cdot3$ on \CifarT). For bit error rate $p$, bit $i,j$ is flipped iff $u_{i,j} \leq p$. Again, this ensures that bit errors at probability $p' \leq p$ also occur at bit error rate $p$. Note that $u$ is fixed across models for comparability. Implementing bit error injection in input $x$ is straight-forward considering \secref{sec:supp-implementation}, \ie, computing $\tilde{x} = Q^{-1}(\text{BErr}_p(Q(x)))$.

For activations, we inject random bit errors after ``blocks'' consisting of convolution, normalization and ReLU. On \CifarT, using our SimpleNet architecture, this results in bit errors being injected after $13$ such blocks, as illustrated by a {\color{colorbrewer4}$\ast$} in \tabref{tab:supp-architectures}. We assume that these activations are temporally stored on the SRAM and thus subject to bit errors. As discussed in the main paper, this is a reasonable approximation of how the data flow could look on an actual accelerator. As with inputs, we pre-sample $u \in U(0,1)^{A_l\times m}$ for each block $l$ and for $50$ attempts, with $A_l$ being the dimensionality of activations, and flip bit $i,j$ iff $u_{i,j} \leq p$. Again, we use $m = 8$ bits. However, in contrast to inputs, quantization itself introduces an error and might increase (clean) \TE. Pre-sampling $u$ for each block independently essentially assumes that activations of different blocks are stored in different portions of the SRAM. This might be a simplification but ensures that bit errors are uncorrelated across blocks. Injecting bit errors in activations in PyTorch boils down to computing $\tilde{a_l} = Q^{-1}(\text{BErr}_p(Q(a_l)))$ after each block $l$ (marked by {\color{colorbrewer4}$\ast$} in \tabref{tab:supp-architectures}). When training with bit errors in activations, we use the straight-through estimator to compute gradients.

\begin{table}[t]
	\centering
	\caption{\revision{\textbf{Robustness of Post-Training Quantization.}
			\RTE for \PLClipping and \LRandom, \ie, our error training with $L_0$ errors on weights. Note that we train \emph{without} quantization, quantize the DNNs post-training with $m = 8$ or $4$ bits and evaluate robustness against $p = 1\%$ random bit errors. Weight clipping and $L_0$-based error training allow to train robust models without knowing the quantization scheme in advance. However, note that clean \TE generally increases slightly compared to quantization-aware training. * \TE reported for the $m = 8$ bit \PLRandom model.}}
	\label{tab:supp-post-robustness}
	\vspace*{-0.2cm}
	\begin{tabular}{|@{\hskip 3px}l@{\hskip 3px}|@{\hskip 3px}c@{\hskip 3px}|@{\hskip 3px}c@{\hskip 3px}|@{\hskip 3px}c@{\hskip 3px}|}
		\hline
		Model (\CifarT) & \multirow{2}{*}{\begin{tabular}{@{}c@{}}\TE\\in \%\end{tabular}} &\multicolumn{2}{c|}{\RTE in \%, $p = 1\%$}\\
		\cline{3-4}
		($p_{L_0}{=}$ $L_0$ error rate in train.) && 8bit & 4bit\\
		\hline
		\hline
		\PLClipping[$0.5$] & 4.61\hphantom{*} & 11.28 {\color{gray}\tiny ${\pm}$1.14} & 16.93 {\color{gray}\tiny ${\pm}$2.77}\\
		\PLClipping[$0.2$] & 5.08\hphantom{*} & 6.85 {\color{gray}\tiny ${\pm}$0.24} & 7.21 {\color{gray}\tiny ${\pm}$0.23}\\
		\hline
		\PLLRandom[$0.2$], $p_{L_0}{=}1$ & 5.01\hphantom{*} & 6.58 {\color{gray}\tiny ${\pm}$0.17} & 7.01 {\color{gray}\tiny ${\pm}$0.22}\\
		\PLLRandom[$0.2$], $p_{L_0}{=}4$ & 5.23\hphantom{*} & \bfseries 6.57 {\color{gray}\tiny ${\pm}$0.13} & \bfseries 6.89 {\color{gray}\tiny ${\pm}$0.13}\\
		\PLLRandom[$0.2$], $p_{L_0}{=}8$ & 5.49\hphantom{*} & 6.73 {\color{gray}\tiny ${\pm}$0.16} & 6.95 {\color{gray}\tiny ${\pm}$0.14}\\
		\hline
		\hline
		\PLRandom[$0.2$], $p{=}1$ & 4.92* & 6.29 {\color{gray}\tiny ${\pm}$0.14} & 6.60 {\color{gray}\tiny ${\pm}$0.18}\\
		\hline
	\end{tabular}
	\vspace*{-0.1cm}
\end{table}
\begin{table}[t]
	\centering
	\caption{\textbf{Required Bit Flips for BFA \cite{RakinICCV2019}.} We report the average number (and standard deviation) of required bit flips $\epsilon_{\text{BFA}}$ for BFA to increase \RTE to $90\%$ or above. This is complementary to the main paper, where we do not ask how many bit errors are required to reduce accuracy to random guessing but rather consider the \RTE given a limited amount of allowed bit errors $\epsilon$.}
	\label{tab:supp-bfa}
	\vspace*{-0.2cm}
	\hspace*{-0.25cm}
	\begin{tabular}{|@{\hskip 4px}l|@{\hskip 4px}c@{\hskip 4px}|}
		\hline
		\bfseries \MNIST & $\epsilon_{\text{BFA}}$\\
		\hline
		\hline
		\Quant & 98 $\pm$ 7.5\\
		\Clipping[$0.05$] & 437 $\pm$ 49.7\\
		\Random[$0.05$], $p{=}20$ & 754 $\pm$ 58.5\\
		\Adv[$0.05$], $\epsilon{=}160$ & 1403 $\pm$ 263.3\\
		\hline
		\hline
		\bfseries \CifarT & $\epsilon_{\text{BFA}}$\\
		\hline
		\hline
		\Quant \xspace{\color{colorbrewer1}BN} & 54 $\pm$ 8.5\\
		\Quant \emph{GN} & 385 $\pm$ 33.9\\
		\Clipping[$0.05$] {\color{colorbrewer1}BN} & 213 $\pm$ 20.9\\
		\Clipping[$0.05$] \emph{GN} & 1725 $\pm$ 60.8\\
		\Random[$0.05$], $p{=}2$ & 2253 $\pm$ 26\\
		\Adv[$0.05$], $\epsilon{=}160$ & 2187 $\pm$ 11.2\\
		\hline
	\end{tabular}
\end{table}
\begin{figure}[t]
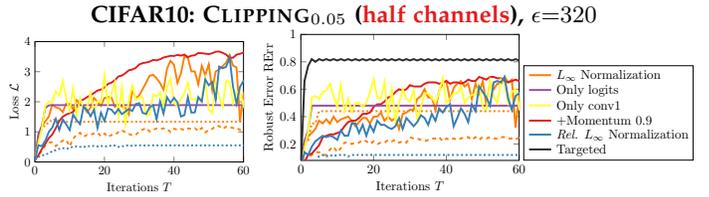

	\begin{minipage}{0.46\textwidth}
		\centering
		\small \bfseries\CifarT: \Clipping[$0.05$] ({\color{colorbrewer1}half channels}), $\epsilon{=}320$
	\end{minipage}
	\hspace*{-0.4cm}
	\begin{minipage}[t]{0.18\textwidth}
		\centering
		\includegraphics[height=2.15cm]{cifar10_005p_attack_error}
	\end{minipage}
	\begin{minipage}[t]{0.18\textwidth}
		\centering
		\includegraphics[height=2.25cm]{cifar10_005p_attack_success}
	\end{minipage}
	\vspace*{-6px}
	\caption{\textbf{Adversarial Bit Error Iterations.} We plot loss $\mathcal{L}$ and robust error \RTE against iterations, both measured on the $100$ held-out test examples used to find adversarial bit errors for the ``smaller'' model on \CifarT, corresponding to halved channels in \tabref{tab:supp-adversarial-ablation}. Thus, we also reduced $\epsilon$ compared to the main paper. Again, the targeted attack is most effective. For the untargeted one, first convolutional and logit layers are particularly vulnerable.}
	\label{fig:supp-adversarial-ablation}
\end{figure}
\begin{figure}[t]
	\vspace*{-0.1cm}
	\centering
	\small\bfseries \textbf{\CifarT: BFA \cite{RakinICCV2019}, $T = 192$ and $\epsilon = 5\cdot T = 960$}\\[-2px]
	
	\includegraphics[width=0.5\textwidth]{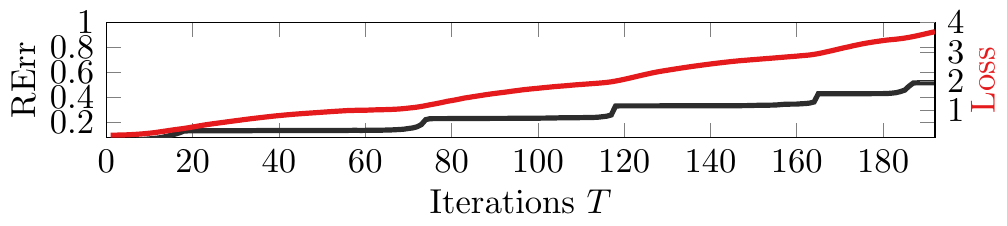}
	\vspace*{-10px}
	\caption{\revision{\textbf{BFA Iterations.} \RTE and loss (after bit errors, in {\color{colorbrewer1}red}) plotted against BFA iterations. We allow $5$ bit flips per iteration, totaling $960$ bit errors for $192$ iterations. \RTE tends to increase (roughly) in steps of $\sim 10\%$ by flipping predictions consecutively to a constant one for each class.}}
	\label{fig:supp-bfa-ablation}
	\vspace*{-0.1cm}
\end{figure}

\textbf{Bit Errors in Inputs:} \tabref{tab:supp-inputs} presents results against random bit errors in inputs ({\color{colorbrewer5}orange}). We report \RTE for bit error rates $p = 0.1\%$ and $p = 0.5\%$. We emphasize that in these experiments the weights are \emph{not} subject to random bit errors. Nevertheless, bit errors in the inputs can be devastating, resulting in \RTE above $20\%$ for \Clipping and \PLClipping with $p = 0.5\%$. While $0.5\%$ does not seem like much, it is important to remember that effectively $0.5 \cdot 8 = 4\%$ of pixels are affected. Also, more extreme clipping does result in higher \RTE, showing that robustness in weights and inputs might be contradictory to some extent. Similarly, \Random on bit errors in weights does \emph{not} improve robustness. \Random on bit errors in weights \emph{and} inputs, in contrast, can reduce \RTE considerably.

\begin{table*}[t]
	\centering
	\caption{\textbf{Adversarial Bit Error Ablation on \MNIST and \CifarT.} (Worst) \RTE for various models on \MNIST and \CifarT against our adversarial bit error attack with $\epsilon$ allowed bit errors. For \CifarT, we consider an \MNIST-like architecture (\cf \tabref{tab:supp-architectures}), a model with halved channels, and the model as used in the main paper. We report worst \RTE across all attacks and restarts as well as individually for: attacking all weights, attacking only \textbf{log}it layer, attacking only the first \textbf{conv}olutional layer, attacking only both logit and conv1 layer (``\textbf{l+c}'') and attacking all layers except logit and conv1 (``\textbf{*}''). In all cases, we consider targeted or untargeted attacks. Just because the first convolutional layer for the \MNIST-like model on \CifarT has $3\times$ more weights, \Clipping[$0.05$] is significantly more robust for $\epsilon{=}160$. This is emphasized when considering the slightly larger ``halved channels'' model and can be confirmed using a ResNet-20. Across all datasets, attacking logit or conv1 layer is highly effective. Only \Adv is able to reduce the vulnerability of these layers.}
	\label{tab:supp-adversarial-ablation}
	\vspace*{-0.2cm}
	\hspace*{-0.2cm}
	\begin{tabularx}{1.025\textwidth}{|@{\hskip 4px}X@{\hskip 4px}|@{\hskip 4px}c@{\hskip 4px}|@{\hskip 4px}c@{\hskip 4px}|@{\hskip 4px}c@{\hskip 4px}|@{\hskip 4px}c@{\hskip 4px}|@{\hskip 4px}c@{\hskip 4px}|@{\hskip 4px}c@{\hskip 4px}|@{\hskip 4px}c@{\hskip 4px}|@{\hskip 4px}c@{\hskip 4px}|@{\hskip 4px}c@{\hskip 4px}|@{\hskip 4px}c@{\hskip 4px}|@{\hskip 4px}c@{\hskip 4px}|@{\hskip 4px}c@{\hskip 4px}|@{\hskip 4px}c@{\hskip 4px}|}
			\hline
			\bfseries \MNIST && \multicolumn{12}{c|}{Worst \RTE in \%}\\
			\hline
			& \TE & $\epsilon{=}80$ & \multicolumn{11}{c|}{$\epsilon{=}160$}\\
			\cline{4-14}
			$W{\approx}1\text{Mio}$ & in \% & {\color{gray}\footnotesize(all U+T)} & {\color{gray}\footnotesize(all U+T)} & U {\color{gray}\footnotesize(all)} & T {\color{gray}\footnotesize(all)} & U {\color{gray}\footnotesize(log)} & T {\color{gray}\footnotesize(log)} & U {\color{gray}\footnotesize(conv)} & T {\color{gray}\footnotesize(conv)} & U {\color{gray}\footnotesize(l+c)} & T {\color{gray}\footnotesize(l+c)} & U {\color{gray}\footnotesize(*)} & T {\color{gray}\footnotesize(*)}\\
			\hline
			\hline
			\Quant & 0.37 & 91.08 & 91.08 & 0.48 & 0.51 & 89.86 & 91.08 & 85.42 & 84.27 & 89.86 & 91.08 & 0.48 & 0.48\\
			\Clipping[$0.05$] & 0.38 & 1.00 & 85.09 & 2.30 & 0.74 & 10.77 & 0.72 & 83.09 & 85.09 & 9.90 & 0.90 & 0.39 & 0.41\\
			\Adv[$0.05$], $\epsilon{=}160$ & 0.29 & 0.33 & 10.37 & 10.37 & 0.36 & 10.03 & 0.37 & 0.40 & 0.47 & 0.36 & 0.41 & 0.29 & 0.29\\
			\hline
			\hline
			
			\bfseries \CifarT && \multicolumn{12}{c|}{Worst \RTE in \%}\\
			\hline
			\emph{\color{colorbrewer1}like \MNIST} & \TE & $\epsilon{=}160$ & \multicolumn{11}{c|}{$\epsilon{=}320$}\\
			\cline{4-12}
			$W{\approx}1\text{Mio}$ & in \% & {\color{gray}\footnotesize(all U+T)} & {\color{gray}\footnotesize(all U+T)} & U {\color{gray}\footnotesize(all)} & T {\color{gray}\footnotesize(all)} & U {\color{gray}\footnotesize(log)} & T {\color{gray}\footnotesize(log)} & U {\color{gray}\footnotesize(conv)} & T {\color{gray}\footnotesize(conv)} & U {\color{gray}\footnotesize(l+c)} & T {\color{gray}\footnotesize(l+c)} & U {\color{gray}\footnotesize(*)} & T {\color{gray}\footnotesize(*)}\\
			\hline
			\hline
			\Quant & 6.63 & 91.49 & 91.49 & 61.34 & 91.49 & 90.42 & 91.49 & 86.91 & 86.46 & 90.42 & 91.49 & 37.77 & 45.56\\
			\Clipping[$0.05$] & 7.31 & 50.40 & 75.40 & 59.74 & 70.16 & 59.21 & 75.40 & 60.16 & 64.91 & 58.91 & 68.43 & 8.44 & 10.01\\
			\hline
			\hline
			
			\emph{\color{colorbrewer1}half channels} & \TE & $\epsilon{=}160$ & \multicolumn{11}{c|}{$\epsilon{=}320$}\\
			\cline{4-12}
			$W{\approx}1.3\text{Mio}$ & in \% & {\color{gray}\footnotesize(all U+T)} & {\color{gray}\footnotesize(all U+T)} & U {\color{gray}\footnotesize(all)} & T {\color{gray}\footnotesize(all)} & U {\color{gray}\footnotesize(log)} & T {\color{gray}\footnotesize(log)} & U {\color{gray}\footnotesize(conv)} & T {\color{gray}\footnotesize(conv)} & U {\color{gray}\footnotesize(l+c)} & T {\color{gray}\footnotesize(l+c)} & U {\color{gray}\footnotesize(*)} & T {\color{gray}\footnotesize(*)}\\
			\hline
			\hline
			\Quant & 6.67 & 91.58 & 91.58 & 46.96 & 91.58 & 90.44 & 91.58 & 87.46 & 87.78 & 90.44 & 91.58 & 53.96 & 52.46\\
			\Clipping[$0.05$] & 7.79 & 37.47 & 82.80 & 58.22 & 82.60 & 54.61 & 82.80 & 64.96 & 62.26 & 53.50 & 82.59 & 8.32 & 9.69\\
			\Clipping[$0.05$] (ResNet-20) & 7.30 & 38.91 & 64.72 & 35.19 & 35.57 & 43.96 & 62.37 & 47.94 & 55.60 & 45.53 & 64.72 & 8.19 & 8.94\\
			
			\hline
			\hline
			\emph{\bfseries full channels} & \TE & $\epsilon{=}160$ & \multicolumn{11}{c|}{$\epsilon{=}320$}\\
			\cline{4-12}
			$W{\approx}5.5\text{Mio}$ & in \% & {\color{gray}\footnotesize(all U+T)} & {\color{gray}\footnotesize(all U+T)} & U {\color{gray}\footnotesize(all)} & T {\color{gray}\footnotesize(all)} & U {\color{gray}\footnotesize(log)} & T {\color{gray}\footnotesize(log)} & U {\color{gray}\footnotesize(conv)} & T {\color{gray}\footnotesize(conv)} & U {\color{gray}\footnotesize(l+c)} & T {\color{gray}\footnotesize(l+c)} & U {\color{gray}\footnotesize(*)} & T {\color{gray}\footnotesize(*)}\\
			\hline
			\hline
			\Quant & 4.89 & 91.18 & 91.18 & 8.54 & 91.18 & 90.68 & 91.18 & 86.28 & 89.06 & 90.68 & 91.18 & 6.73 & 7.46\\
			\Clipping[$0.05$] & 5.34 & 20.48 & 60.76 & 24.04 & 35.20 & 23.67 & 35.86 & 52.57 & 60.76 & 25.27 & 35.86 & 5.80 & 5.61\\
			\Adv[$0.05$], $\epsilon{=}160$ & 5.54 & 15.20 & 26.22 & 10.01 & 20.20 & 25.70 & 26.22 & 8.84 & 12.33 & 8.91 & 25.47 & 5.82 & 6.18\\
			\hline
	\end{tabularx}
\end{table*}

\textbf{Bit Errors in Activations:} \tabref{tab:supp-activations} shows results for random bit errors in activations ({\color{colorbrewer4}violet}). As, by default, we do \emph{not} train with activation quantization, we report \TE w/ and w/o activation quantization (in $m = 8$ bits). As can be seen, our simple activation quantization leads to a slight increase of $0.2$ to $0.3\%$ in \TE. Bit errors in activations turn out to be difficult to handle, even for \Clipping and \PLClipping. In contrast to bit errors in inputs, weight clipping has a positive effect on robustness against bit errors in activations. However, the benefit is less pronounced than for bit errors in weights, \eg, \PLClipping does not improve over \Clipping. Similarly, \Random with bit errors in weights can improve robustness, but does not so consistently, see the variations in \RTE for \Random (weights only) in tabref{tab:supp-activations}. \Random with bit errors in activations, in contrast, has the expected effect of reducing \RTE considerably, allowing reasonably low \RTE for $p = 0.1\%$.

\subsection{Adversarial Bit Error Robustness}
\label{subsec:supp-experiments-adversarial}

\textbf{Bit Flip Attack (BFA):} \tabref{tab:supp-bfa} reports additional experimental results for the BFA attack proposed in \cite{RakinICCV2019}. Our experimental setup is described in detail in \secref{subsec:supp-experiments-setup}. We report the average number of bit flips required by BFA in order to increase \RTE to $90\%$ or above (\ie, reduce performance to random guessing, as also used in \cite{RakinICCV2019}).
Furthermore, we report the standard deviation across these $5$ restarts. As shown, BFA requires significantly more bit flips $\epsilon_{\text{BFA}}$ to break our models than reported in \cite{RakinICCV2019}. We believe that this is mainly due to \cite{RakinICCV2019} relying on batch normalization (BN) \cite{IoffeICML2015} in their ResNet models. In contrast, we use group normalization (GN) \cite{WuECCV2018}. On \CifarT, we demonstrate that using BN reduces robustness significantly, \ie, far fewer bit flips are required to achieve $\RTE \geq 90\%$. Furthermore, our \Clipping, \Random or \Adv model, specifically trained to be robust against random or adversarial bit errors, improve robustness significantly. On \MNIST, more than $1\text{k}$ and on \CifarT more than $2\text{k}$ bit flips are required.

\revision{
\figref{fig:supp-bfa-ablation} shows \RTE and loss over BFA iterations. While loss increases continuously, \RTE tends to increase roughly in steps of $10\%$. This is because BFA consecutively flips the labels for each class to a constant class, eventually arriving at $90\%$ \RTE which is equivalent to a random or constant classifier.
However, as the number of flipped bits is tied to iterations, each taking between 1 and 2 seconds, BFA is very expensive, especially for large numbers of bit errors $\epsilon$.
}

\begin{table*}[t]
	\centering
	\caption{\textbf{Adversarial Bit Error Robustness on \MNIST, \CifarT, \revision{and TinyImageNet}.} Complementing the results from the main paper, we report \RTE for \Quant, \Clipping \Random and \Adv. We evaluate adversarial bit error robustness for various $\epsilon$ and also train \Adv with different $\epsilon$. On \MNIST, we also present an ablation when training \Adv against attacks with or without momentum $0.9$ or with targeted attack (``T''). As targeted attacks are generally more effective, this also helps improve robustness using \Adv. On \CifarT, we found \Adv with $\epsilon > 160$ to reduce robustness, \revision{as on TinyImageNet}. We suspect that training gets more difficult and might require additional capacity or more sophisticated training schemes.}
	\label{tab:supp-advbet}
	\vspace*{-0.2cm}
	\begin{minipage}[t]{0.52\textwidth}
		\vspace*{0px}
		
		\begin{tabular}{|@{\hskip 3px}l@{\hskip 3px}|@{\hskip 3px}c@{\hskip 3px}|@{\hskip 3px}c@{\hskip 3px}|@{\hskip 3px}c@{\hskip 3px}|@{\hskip 3px}c@{\hskip 3px}|@{\hskip 3px}c@{\hskip 3px}|}
			\hline
			\bfseries \MNIST & \TE \% & \multicolumn{4}{c|}{\RTE in \%}\\
			\hline
			\hline
			&& $\epsilon{=}80$ & $\epsilon{=}160$ & $\epsilon{=}240$ & $\epsilon{=}320$\\
			\hline
			\Quant & 0.37 & 91.08 & 91.08 & 91.08 & 91.08\\
			\Clipping[$0.05$] & 0.38 & 85.09 & 88.81 & 90.11 & 90.26\\
			\Random[$0.05$], $p{=}20$ & 0.39 & 10.13 & 69.90 & 81.16 & 81.94\\
			\Adv[$0.05$], $\epsilon{=}160$ & 0.33 & 11.63 & 21.08 & 31.71 & 61.01\\
			\Adv[$0.05$], $\epsilon{=}240$ & 0.32 & 66.92 & 82.83 & 85.12 & 87.98\\
			\Adv[$0.05$], $\epsilon{=}320$ & 0.27 & 10.41 & 85.73 & 85.57 & 88.12\\
			\Adv[$0.05$] (mom $0.9$), $\epsilon{=}160$ & 0.29 & 10.37 & 86.12 & 84.83 & 87.28\\
			\Adv[$0.05$] (mom $0.9$), $\epsilon{=}240$ & 0.29 & 11.58 & 21.66 & 41.27 & 50.73\\
			\Adv[$0.05$] (mom $0.9$), $\epsilon{=}320$ & 0.30 & 10.38 & 21.76 & 39.92 & 69.61\\
			\Adv[$0.05$] T, $\epsilon{=}160$ & 0.39 & 10.54 & 21.41 & 29.72 & 49.77\\
			\Adv[$0.05$] T, $\epsilon{=}240$ & 0.36 & 10.10 & 19.44 & 31.23 & 51.01\\
			\Adv[$0.05$] T, $\epsilon{=}320$ & 0.36 & 10.51 & 19.72 & 31.28 & 50.77\\
			\hline
		\end{tabular}
	\end{minipage}
	\begin{minipage}[t]{0.46\textwidth}
		\vspace*{0px}
		
		\begin{tabular}{|@{\hskip 3px}l@{\hskip 3px}|@{\hskip 3px}c@{\hskip 3px}|@{\hskip 3px}c@{\hskip 3px}|@{\hskip 3px}c@{\hskip 3px}|@{\hskip 3px}c@{\hskip 3px}|@{\hskip 3px}c@{\hskip 3px}|}
			\hline
			\bfseries \CifarT & \TE \% & \multicolumn{4}{c|}{\RTE in \%}\\
			\hline
			\hline
			&& $\epsilon{=}160$ & $\epsilon{=}320$ & $\epsilon{=}480$ & $\epsilon{=}640$\\
			\hline
			\Quant & 4.89 & 91.18 & 91.18 & 91.18 & 91.18\\
			\Clipping[$0.05$] & 5.34 & 20.48 & 60.76 & 79.12 & 83.93\\
			\Random[$0.05$], $p{=}2$ & 5.42 & 14.66 & 33.86 & 54.24 & 80.36\\
			\Adv[$0.05$], $\epsilon{=}160$ & 5.54 & 15.20 & 26.22 & 55.06 & 77.43\\
			\Adv[$0.05$], $\epsilon{=}320$ & 5.78 & 15.27 & 42.79 & 70.56 & 91.13\\
			\Adv[$0.05$], $\epsilon{=}480$ & 5.99 & 15.41 & 44.66 & 83.39 & 91.47\\
			\revision{\Adv[$0.05$] T, $\epsilon{=}480$} & \revision{7.46} & \revision{21.20} & \revision{49.00} & \revision{70.29} & \revision{78.54}\\
			\hline
		\end{tabular}
		\vspace*{4px}
		
		\revision{
		\begin{tabular}{|@{\hskip 3px}l@{\hskip 3px}|@{\hskip 3px}c@{\hskip 3px}|@{\hskip 3px}c@{\hskip 3px}|@{\hskip 3px}c@{\hskip 3px}|@{\hskip 3px}c@{\hskip 3px}|@{\hskip 3px}c@{\hskip 3px}|}
			\hline
			\bfseries TinyImageNet & \TE \% & \multicolumn{4}{c|}{\RTE in \%}\\
			\hline
			\hline
			&& $\epsilon{=}80$ & $\epsilon{=}160$ & $\epsilon{=}240$ & $\epsilon{=}320$\\
			\hline
			\Quant & 36.77 & 99.70 & 99.78 & 99.78 & 99.78\\
			\Clipping[$0.1$] & 37.42 & 54.47 & 82.94 & 96.37 & 99.47\\
			\Random[$0.1$], $p{=}1$ & 42.30 & 58.11 & 76.74 & 99.94 & 99.51\\
			\Adv[$0.1$], $\epsilon{=}160$ & 37.83 & 52.91 & 61.06 & 97.73 & 99.58\\
			\Adv[$0.1$], $\epsilon{=}240$ & 38.5 & 43.59 & 72.56 & 97.37 & 99.34\\
			\Adv[$0.1$] T, $\epsilon{=}160$ & 37.87 & 55.79 & 77.48 & 98.49 & 99.66\\
			\hline
		\end{tabular}
		}
	\end{minipage}
\end{table*}

\textbf{Adversarial Bit Error Attack:} \tabref{tab:supp-adversarial-ablation} presents a comprehensive ablation study regarding our adversarial bit error attack. We report \RTE on \MNIST and \CifarT, considering an additional \MNIST-like architecture and an architecture with ``halved channels'' on \CifarT. Specifically, considering \tabref{tab:supp-architectures}, the \MNIST-like architecture is the same architecture as used for \MNIST, but using a larger first convolutional layer (conv1) due to the larger input dimensionality on \CifarT. Specifically, the number of weights in conv1 increase from $288$ to $864$ (factor $3$ due to $3$ input channels on \CifarT). Then, we consider the \CifarT architecture but with all channel widths halved (similar to \MNIST). These two architecture result in roughly $1\text{Mio}$ and $1.3\text{Mio}$ weights. While attacks on all weights (targeted or untargeted) are not very effective on \MNIST, compared to attacks on conv1 or logit layer, the models on \CifarT are more vulnerable in this regard, specially \Quant. However, even on \CifarT, the conv1 and logit layers are most susceptible to bit errors. Attacking all \emph{other} layers (\ie, all layers except conv1 and logit), in contrast, is not very fruitful in terms of increasing \RTE, especially considering \Clipping. Only \Adv is able to reduce the vulnerability of these layers.
Interestingly, \Clipping[$0.05$] is more robust on \CifarT, even when using the MNIST-like architecture. Here, for $\epsilon{=}160$, \RTE is $58.09\%$ on \MNIST but only $50.4\%$ on \CifarT. As the only difference between both models is the larger first convolutional layer conv1 on \CifarT, this further supports our experiments showing that conv1 is particularly vulnerable.
In all cases, targeted attacks are more successful.
This is also illustrated in \figref{fig:supp-adversarial-ablation} showing loss and \RTE over iterations for the half channels model on \CifarT. As in the main paper, proper $L_\infty$ gradient normalization and momentum make the attack more effective. However, the targeted attack is most effective and less prone to poor local minima (\ie, does rarely get stuck during optimization).

\textbf{Overall Results:} \tabref{tab:supp-advbet} summarizes our results against adversarial bit errors: We report \RTE against \Quant, \Clipping, \Random and \Adv. On \MNIST, we consider various variants of \Adv: training with or without using momentum ($0.9$) in the attack and training against targeted attacks (random target label in each iteration). As demonstrated in \tabref{tab:supp-adversarial-ablation}, targeted attacks are generally easier to optimize and more successful. Thus, \Adv with targeted attack outperforms training with untargeted attacks. On \CifarT, we present results for \Adv trained with larger $\epsilon$. Unfortunately, using larger $\epsilon$ does not increase robustness. We suspect that training with large $\epsilon$, \eg, $320$, is significantly more difficult. That is, our model might lack capacity or more sophisticated training schemes are necessary. As result, (clean) \TE increases, while robustness does not further improve or even decreases.
\revision{Moreover, in contrast to MNIST, training with a targeted attack does not help training or improve adversarial bit error robustness.}
\revision{Finally, on TinyImageNet, \Clipping and \Random also improve robustness against adversarial bit errors quite significantly. As a result, there is limited benefit in additionally using \Adv. As on \CifarT, this might also be due to training difficulties or insufficient hyper-parameter optimization.}

\clearpage
\begin{table*}
	\centering
	\caption{\textbf{Overall Robustness Results on \CifarT.} Tabular results corresponding to \figref{fig:supp-summary} for $m = 8, 4, 3$ and $2$ bits. We show \RTE for \Normal, \Clipping and \Random with various $\wmax$ and $p$ across a subset of evaluated bit error rates. Results with per-layer weight clipping can be found in \tabref{tab:supp-summary-cifar10-plc}.}
	\label{tab:supp-summary-cifar10}
	\vspace*{-0.2cm}
	\scriptsize
	\begin{tabular}{| c | l | c | c | c | c | c | c | c | c | c |}
		\hline
		\multicolumn{11}{|c|}{\bfseries \CifarT}\\
		\hline
		& Model & \multirow{2}{*}{\begin{tabular}{c}\TE\\in \%\end{tabular}} & \multicolumn{8}{c|}{\RTE in \%, $p$ in \%}\\
		\cline{4-11}
		&&& $0.01$ & $0.05$ & $0.1$ & $0.5$ & $1$ & $1.5$ & $2$ & $2.5$\\
		\hline
		\hline
		\multirow{30}{*}{\rotatebox{90}{$m = 8$ bit}} & \Normal & 4.36 & 4.82 & 5.51 & 6.37 & 24.76 & 72.65 & 87.40 & 89.76 & 90.15\\
		& \Quant & 4.32 & 4.60 & 5.10 & 5.54 & 11.28 & 32.05 & 68.65 & 85.28 & 89.01\\
		& \Clipping[$0.25$] & 4.58 & 4.84 & 5.29 & 5.71 & 10.52 & 27.95 & 62.46 & 82.61 & 88.08\\
		& \Clipping[$0.2$] & 4.63 & 4.91 & 5.28 & 5.62 & 8.27 & 18.00 & 53.74 & 82.02 & 88.27\\
		& \Clipping[$0.15$] & 4.42 & 4.66 & 5.01 & 5.31 & 7.81 & 13.08 & 23.85 & 42.12 & 61.20\\
		& \Clipping[$0.1$] & 4.82 & 5.04 & 5.33 & 5.58 & 6.95 & 8.93 & 12.22 & 17.80 & 27.02\\
		& \Clipping[$0.05$] & 5.44 & 5.59 & 5.76 & 5.90 & 6.53 & 7.18 & 7.92 & 8.70 & 9.56\\
		& \Clipping[$0.025$] & 7.10 & 7.20 & 7.32 & 7.40 & 7.82 & 8.18 & 8.43 & 8.74 & --\\
		\cline{2-11}
		& \Random[$1$] $p{=}0.01$ & 4.56 & 4.93 & 5.50 & 6.06 & 14.14 & 66.07 & 86.86 & 89.80 & 90.35\\
		& \Random[$1$] $p{=}0.1$ & 4.50 & 4.80 & 5.27 & 5.72 & 10.33 & 41.10 & 75.90 & 86.52 & 89.03\\
		& \Random[$1$] $p{=}1$ & 7.38 & 7.69 & 8.17 & 8.58 & 11.10 & 14.90 & 21.08 & 41.11 & 71.09\\
		& \Random[$0.2$] $p{=}0.01$ & 4.44 & 4.67 & 5.09 & 5.48 & 8.64 & 17.97 & 41.53 & 68.95 & 82.48\\
		& \Random[$0.2$] $p{=}0.1$ & 4.51 & 4.73 & 5.07 & 5.39 & 7.99 & 19.21 & 54.94 & 80.12 & 86.55\\
		& \Random[$0.2$] $p{=}1$ & 5.46 & 5.68 & 5.97 & 6.20 & 7.63 & 9.47 & 12.38 & 21.47 & 50.86\\
		& \Random[$0.15$] $p{=}0.01$ & 4.64 & 4.87 & 5.17 & 5.45 & 7.54 & 15.83 & 54.07 & 81.41 & 86.75\\
		& \Random[$0.15$] $p{=}0.1$ & 4.86 & 5.07 & 5.36 & 5.64 & 7.74 & 12.33 & 22.38 & 40.09 & 60.78\\
		& \Random[$0.15$] $p{=}1$ & 5.27 & 5.44 & 5.68 & 5.88 & 7.11 & 8.63 & 11.13 & 27.74 & 64.97\\
		& \Random[$0.1$] $p{=}0.01$ & 4.99 & 5.15 & 5.39 & 5.62 & 6.93 & 9.01 & 12.83 & 22.81 & 41.04\\
		& \Random[$0.1$] $p{=}0.1$ & 4.72 & 4.92 & 5.15 & 5.37 & 6.74 & 8.53 & 11.40 & 15.97 & 23.59\\
		& \Random[$0.1$] $p{=}1$ & 4.90 & 5.05 & 5.26 & 5.43 & 6.36 & 7.41 & 8.65 & 12.25 & 27.21\\
		& \Random[$0.1$] $p{=}1.5$ & 5.53 & 5.67 & 5.87 & 6.03 & 6.84 & 7.76 & 8.80 & 10.03 & 11.68\\
		& \Random[$0.1$] $p{=}2$ & 5.71 & 5.87 & 6.07 & 6.22 & 7.00 & 7.83 & 8.69 & 9.70 & 10.91\\
		& \Random[$0.05$] $p{=}0.1$ & 5.32 & 5.41 & 5.59 & 5.72 & 6.34 & 6.96 & 7.62 & 8.28 & 9.13\\
		& \Random[$0.05$] $p{=}1$ & 5.24 & 5.36 & 5.50 & 5.60 & 6.18 & 6.73 & 7.26 & 7.88 & 8.49\\
		& \Random[$0.05$] $p{=}1.5$ & 5.62 & 5.71 & 5.84 & 5.95 & 6.50 & 7.02 & 7.52 & 7.97 & 8.51\\
		& \Random[$0.05$] $p{=}2$ & 5.42 & 5.55 & 5.68 & 5.78 & 6.26 & 6.71 & 7.13 & 7.58 & 8.02\\
		& \Random[$0.025$] $p{=}1$ & 6.78 & 6.88 & 7.00 & 7.08 & 7.46 & 7.75 & 8.02 & 8.24 & 8.47\\
		& \Random[$0.025$] $p{=}1.5$ & 6.89 & 6.99 & 7.11 & 7.19 & 7.58 & 7.94 & 8.26 & 8.52 & 8.77\\
		& \Random[$0.025$] $p{=}2$ & 6.93 & 7.02 & 7.12 & 7.20 & 7.57 & 7.87 & 8.11 & 8.33 & 8.58\\
		& \Random[$0.025$] $p{=}2.5$ & 6.91 & 6.99 & 7.08 & 7.14 & 7.50 & 7.83 & 8.10 & 8.36 & 8.63\\
		\hline
		\hline
		\multirow{19}{*}{\rotatebox{90}{$m = 4$ bit}} & \Quant & 4.83 & 5.29 & 5.98 & 6.59 & 15.72 & 50.45 & 79.86 & 87.17 & 89.47\\
		& \Clipping[$0.25$] & 4.78 & 5.16 & 5.75 & 6.26 & 12.08 & 30.62 & 60.52 & 80.07 & 87.01\\
		& \Clipping[$0.2$] & 4.90 & 5.20 & 5.65 & 6.04 & 9.67 & 27.24 & 63.96 & 82.63 & 87.21\\
		& \Clipping[$0.15$] & 4.78 & 5.07 & 5.43 & 5.79 & 8.40 & 14.61 & 28.53 & 50.83 & 70.32\\
		& \Clipping[$0.1$] & 5.29 & 5.49 & 5.75 & 5.99 & 7.71 & 10.62 & 15.79 & 24.97 & 37.94\\
		& \Clipping[$0.05$] & 5.78 & 5.92 & 6.08 & 6.21 & 6.98 & 7.86 & 8.77 & 9.76 & 11.04\\
		\cline{2-11}
		& \Random[$0.2$] $p{=}0.01$ & 5.14 & 5.42 & 5.85 & 6.23 & 10.44 & 23.84 & 49.25 & 73.35 & 83.16\\
		& \Random[$0.2$] $p{=}0.1$ & 4.77 & 5.01 & 5.41 & 5.76 & 8.66 & 16.06 & 32.40 & 56.69 & 75.21\\
		& \Random[$0.2$] $p{=}1$ & 6.27 & 6.52 & 6.86 & 7.12 & 8.78 & 11.33 & 15.17 & 21.43 & 32.19\\
		& \Random[$0.15$] $p{=}0.01$ & 4.88 & 5.13 & 5.54 & 5.92 & 8.51 & 14.21 & 26.26 & 46.02 & 66.13\\
		& \Random[$0.15$] $p{=}0.1$ & 4.50 & 4.72 & 5.05 & 5.36 & 7.58 & 14.12 & 43.00 & 76.28 & 85.54\\
		& \Random[$0.15$] $p{=}1$ & 5.99 & 6.18 & 6.45 & 6.65 & 8.00 & 9.74 & 12.50 & 16.73 & 24.09\\
		& \Random[$0.1$] $p{=}0.01$ & 5.07 & 5.29 & 5.58 & 5.83 & 7.54 & 10.46 & 15.34 & 24.63 & 39.76\\
		& \Random[$0.1$] $p{=}0.1$ & 4.82 & 5.04 & 5.32 & 5.53 & 6.82 & 8.85 & 12.48 & 21.36 & 40.03\\
		& \Random[$0.1$] $p{=}1$ & 5.39 & 5.55 & 5.77 & 5.96 & 7.04 & 8.34 & 9.77 & 11.85 & 14.91\\
		& \Random[$0.05$] $p{=}0.1$ & 5.14 & 5.26 & 5.46 & 5.61 & 6.38 & 7.19 & 8.06 & 9.16 & 10.46\\
		& \Random[$0.05$] $p{=}1$ & 5.60 & 5.71 & 5.85 & 5.97 & 6.54 & 7.10 & 7.68 & 8.28 & 8.99\\
		& \Random[$0.05$] $p{=}1.5$ & 5.51 & 5.64 & 5.77 & 5.87 & 6.38 & 6.98 & 7.51 & 8.10 & 8.72\\
		& \Random[$0.05$] $p{=}2$ & 5.49 & 5.62 & 5.77 & 5.90 & 6.43 & 6.99 & 7.53 & 8.06 & 8.62\\
		\hline
		\hline
		\multirow{20}{*}{\rotatebox{90}{$m = 3$ bit}} & \Quant & 79.59 & 83.95 & 88.57 & 91.07 & 96.15 & 97.81 & 98.20 & 98.60 & 99.07\\
		& \Clipping[$0.25$] & 6.89 & 7.34 & 8.00 & 8.65 & 14.46 & 28.70 & 53.64 & 75.51 & 85.13\\
		& \Clipping[$0.2$] & 5.82 & 6.21 & 6.79 & 7.30 & 11.90 & 23.31 & 43.00 & 65.68 & 78.79\\
		& \Clipping[$0.15$] & 5.84 & 6.16 & 6.60 & 6.95 & 9.95 & 15.92 & 27.84 & 47.54 & 67.08\\
		& \Clipping[$0.1$] & 5.71 & 6.01 & 6.39 & 6.73 & 8.99 & 13.06 & 20.88 & 35.13 & 51.76\\
		& \Clipping[$0.05$] & 5.61 & 5.78 & 6.01 & 6.19 & 7.07 & 8.13 & 9.34 & 10.95 & 13.16\\
		\cline{2-11}
		& \Random[$0.2$] $p{=}0.01$ & 5.72 & 6.14 & 6.77 & 7.30 & 12.84 & 26.46 & 50.52 & 72.46 & 83.09\\
		& \Random[$0.2$] $p{=}0.1$ & 6.23 & 6.55 & 7.04 & 7.53 & 11.38 & 21.36 & 41.93 & 65.54 & 79.94\\
		& \Random[$0.2$] $p{=}1$ & 7.61 & 7.84 & 8.20 & 8.52 & 10.30 & 12.82 & 16.65 & 21.81 & 29.64\\
		& \Random[$0.15$] $p{=}0.01$ & 5.61 & 5.94 & 6.40 & 6.77 & 9.59 & 15.72 & 28.06 & 46.88 & 64.39\\
		& \Random[$0.15$] $p{=}0.1$ & 5.33 & 5.56 & 5.99 & 6.33 & 9.01 & 14.06 & 23.44 & 40.36 & 59.92\\
		& \Random[$0.15$] $p{=}1$ & 7.26 & 7.52 & 7.82 & 8.07 & 9.58 & 11.47 & 13.87 & 17.58 & 23.01\\
		& \Random[$0.1$] $p{=}0.01$ & 5.13 & 5.41 & 5.72 & 6.00 & 8.06 & 11.25 & 17.22 & 26.96 & 42.72\\
		& \Random[$0.1$] $p{=}0.1$ & 5.69 & 5.96 & 6.26 & 6.51 & 8.04 & 10.81 & 15.51 & 23.88 & 37.52\\
		& \Random[$0.1$] $p{=}1$ & 5.76 & 5.95 & 6.22 & 6.44 & 7.59 & 8.97 & 10.76 & 13.21 & 16.95\\
		& \Random[$0.05$] $p{=}0.01$ & 5.50 & 5.62 & 5.83 & 5.99 & 6.83 & 7.79 & 9.05 & 10.48 & 12.32\\
		& \Random[$0.05$] $p{=}0.1$ & 5.44 & 5.58 & 5.76 & 5.90 & 6.72 & 7.60 & 8.60 & 9.92 & 11.70\\
		& \Random[$0.05$] $p{=}1$ & 5.57 & 5.69 & 5.87 & 6.01 & 6.68 & 7.38 & 8.08 & 8.96 & 9.96\\
		\hline
		\hline
		\multirow{11}{*}{\rotatebox{90}{$m = 2$ bit}} & \Quant & 88.68 & 89.53 & 91.62 & 93.23 & 97.74 & 98.40 & 97.85 & 99.20 & 98.74\\
		& \Clipping[$0.25$] & 90.14 & 90.54 & 91.13 & 91.82 & 95.96 & 96.90 & 97.21 & 96.66 & 97.12\\
		& \Clipping[$0.2$] & 82.00 & 84.86 & 90.79 & 94.17 & 97.25 & 96.69 & 97.16 & 97.73 & 97.01\\
		& \Clipping[$0.15$] & 14.62 & 15.29 & 16.30 & 17.16 & 22.88 & 33.18 & 50.86 & 71.17 & 84.30\\
		& \Clipping[$0.1$] & 7.87 & 8.29 & 8.93 & 9.57 & 13.95 & 23.65 & 42.43 & 64.65 & 80.89\\
		& \Clipping[$0.05$] & 6.59 & 6.78 & 7.05 & 7.26 & 8.55 & 10.26 & 12.73 & 15.99 & 20.51\\
		& \Clipping[$0.025$] & 6.94 & 7.06 & 7.23 & 7.34 & 7.96 & 8.57 & 9.16 & 9.77 & 10.47\\
		& \Random[$0.05$] $p{=}0.01$ & 6.00 & 6.21 & 6.47 & 6.66 & 7.88 & 9.51 & 11.53 & 14.99 & 19.60\\
		& \Random[$0.05$] $p{=}0.1$ & 5.83 & 6.04 & 6.30 & 6.52 & 7.73 & 9.32 & 11.41 & 14.49 & 19.77\\
		& \Random[$0.025$] $p{=}0.01$ & 6.93 & 7.07 & 7.24 & 7.37 & 8.05 & 8.65 & 9.23 & 9.72 & 10.43\\
		& \Random[$0.025$] $p{=}0.1$ & 7.02 & 7.13 & 7.31 & 7.41 & 7.98 & 8.48 & 9.00 & 9.65 & 10.32\\
		& \Random[$0.025$] $p{=}1$ & 7.10 & 7.23 & 7.38 & 7.49 & 8.10 & 8.65 & 9.14 & 9.54 & 10.07\\
		\hline
	\end{tabular}
	\vspace*{-0.2cm}
\end{table*}
\begin{table*}
	\centering
	\caption{\textbf{Overall Robustness Results on \CifarT.} Tabular results corresponding to \figref{fig:supp-summary} for $m = 8$ and $4$ bits. We show \RTE for \PLClipping and \PLRandom with various $\wmax$ and $p$ across a subset of evaluated bit error rates. Results for $m = 3, 2$ in \tabref{tab:supp-summary-cifar10-plc-2}.}
	\label{tab:supp-summary-cifar10-plc}
	\vspace*{-0.2cm}
	\scriptsize
	\begin{tabular}{| c | l | c | c | c | c | c | c | c | c | c |}
		\hline
		\multicolumn{11}{|c|}{\bfseries \CifarT: results with per-layer weight clipping for $m = 8,4$ bit}\\
		\hline
		& Model & \multirow{2}{*}{\begin{tabular}{c}\TE\\in \%\end{tabular}} & \multicolumn{8}{c|}{\RTE in \%, $p$ in \%}\\
		\cline{4-11}
		&&& $0.01$ & $0.05$ & $0.1$ & $0.5$ & $1$ & $1.5$ & $2$ & $2.5$\\
		\hline
		\hline
		\multirow{30}{*}{\rotatebox{90}{$m = 8$ bit}} &
		\PLClipping[$1$] & 4.38 & 4.69 & 5.16 & 5.61 & 10.73 & 27.53 & 58.40 & 80.65 & 87.61\\
		& \PLClipping[$0.5$] & 4.61 & 4.89 & 5.21 & 5.48 & 7.39 & 10.90 & 20.05 & 38.24 & 60.01\\
		& \PLClipping[$0.25$] & 4.96 & -- & 5.25 & 5.39 & 6.21 & 7.04 & 8.14 & 9.36 & 10.88\\
		& \PLClipping[$0.2$] & 4.71 & 4.88 & 5.07 & 5.20 & 5.86 & 6.53 & 7.20 & 7.88 & 8.79\\
		& \PLClipping[$0.15$] & 5.27 & 5.38 & 5.51 & 5.60 & 6.09 & 6.52 & 6.99 & 7.43 & 7.94\\
		& \PLClipping[$0.1$] & 5.62 & 5.72 & 5.81 & 5.91 & 6.31 & 6.65 & 6.92 & 7.18 & 7.46\\
		& \PLClipping[$0.05$] & 7.23 & 7.32 & 7.43 & 7.50 & 7.88 & 8.13 & 8.35 & 8.58 & 8.75\\
		& \PLRandom[$0.75$] $p{=}0.01$ & 4.20 & 4.45 & 4.90 & 5.23 & 8.13 & 16.73 & 40.38 & 69.40 & 83.09\\
		& \PLRandom[$0.5$] $p{=}0.01$ & 4.59 & 4.85 & 5.19 & 5.46 & 7.37 & 11.06 & 17.77 & 29.32 & 45.72\\
		& \PLRandom[$0.4$] $p{=}0.01$ & 4.49 & 4.67 & 4.96 & 5.20 & 6.68 & 8.88 & 12.37 & 19.06 & 29.74\\
		& \PLRandom[$0.3$] $p{=}0.01$ & 4.52 & 4.66 & 4.86 & 5.06 & 6.15 & 7.49 & 9.28 & 11.87 & 15.43\\
		& \PLRandom[$0.25$] $p{=}0.01$ & 4.83 & 5.00 & 5.16 & 5.29 & 6.01 & 6.90 & 7.87 & 9.11 & 10.79\\
		& \PLRandom[$0.75$] $p{=}0.1$ & 4.57 & 4.78 & 5.15 & 5.47 & 8.02 & 13.39 & 26.01 & 49.54 & 71.45\\
		& \PLRandom[$0.5$] $p{=}0.1$ & 4.48 & 4.72 & 5.07 & 5.33 & 7.09 & 10.12 & 14.90 & 25.54 & 39.41\\
		& \PLRandom[$0.4$] $p{=}0.1$ & 4.43 & 4.64 & 4.93 & 5.18 & 6.48 & 8.40 & 11.39 & 16.36 & 24.36\\
		& \PLRandom[$0.3$] $p{=}0.1$ & 4.54 & 4.67 & 4.87 & 5.02 & 5.96 & 7.09 & 8.47 & 10.13 & 12.89\\
		& \PLRandom[$0.25$] $p{=}0.1$ & 4.49 & 4.62 & 4.82 & 4.98 & 5.80 & 6.65 & 7.59 & 8.72 & 10.36\\
		& \PLRandom[$0.2$] $p{=}0.1$ & 4.90 & 5.02 & 5.21 & 5.36 & 6.05 & 6.72 & 7.40 & 8.14 & 9.02\\
		& \PLRandom[$0.15$] $p{=}0.1$ & 5.10 & 5.21 & 5.35 & 5.46 & 5.97 & 6.43 & 6.84 & 7.23 & 7.67\\
		& \PLRandom[$0.5$] $p{=}1$ & 5.11 & 5.27 & 5.52 & 5.73 & 6.89 & 8.30 & 10.25 & 13.04 & 17.35\\
		& \PLRandom[$0.4$] $p{=}1$ & 4.80 & 4.94 & 5.15 & 5.31 & 6.24 & 7.31 & 8.61 & 10.51 & 13.78\\
		& \PLRandom[$0.3$] $p{=}1$ & 4.86 & 4.98 & 5.13 & 5.27 & 6.01 & 6.82 & 7.70 & 8.80 & 10.05\\
		& \PLRandom[$0.25$] $p{=}1$ & 4.62 & 4.73 & 4.90 & 5.02 & 5.62 & 6.36 & 7.02 & 7.79 & 8.76\\
		& \PLRandom[$0.2$] $p{=}1$ & 4.92 & 5.02 & 5.15 & 5.25 & 5.78 & 6.29 & 6.76 & 7.26 & 7.89\\
		& \PLRandom[$0.15$] $p{=}1$ & 5.10 & 5.19 & 5.31 & 5.41 & 5.87 & 6.28 & 6.70 & 7.02 & 7.45\\
		& \PLRandom[$0.1$] $p{=}1$ & 5.66 & 5.76 & 5.88 & 5.96 & 6.29 & 6.59 & 6.87 & 7.11 & 7.34\\
		& \PLRandom[$0.25$] $p{=}2$ & 4.94 & 5.06 & 5.22 & 5.33 & 5.92 & 6.48 & 7.04 & 7.59 & 8.25\\
		& \PLRandom[$0.2$] $p{=}2$ & 5.02 & 5.12 & 5.24 & 5.33 & 5.85 & 6.30 & 6.74 & 7.18 & 7.62\\
		& \PLRandom[$0.15$] $p{=}2$ & 4.99 & 5.08 & 5.20 & 5.29 & 5.74 & 6.12 & 6.49 & 6.80 & 7.15\\
		& \PLRandom[$0.1$] $p{=}2$ & 5.60 & 5.67 & 5.77 & 5.84 & 6.20 & 6.49 & 6.72 & 6.95 & 7.19\\
		& \PLRandom[$0.25$] $p{=}3$ & 5.46 & 5.56 & 5.69 & 5.77 & 6.29 & 6.84 & 7.26 & 7.74 & 8.25\\
		& \PLRandom[$0.2$] $p{=}3$ & 5.13 & 5.21 & 5.34 & 5.43 & 5.92 & 6.26 & 6.71 & 7.05 & 7.41\\
		& \PLRandom[$0.15$] $p{=}3$ & 5.21 & 5.29 & 5.39 & 5.47 & 5.87 & 6.18 & 6.49 & 6.75 & 7.06\\
		& \PLRandom[$0.1$] $p{=}3$ & 5.84 & 5.89 & 5.99 & 6.06 & 6.42 & 6.71 & 6.94 & 7.18 & 7.36\\
		& \PLRandom[$0.05$] $p{=}3$ & 7.57 & 7.63 & 7.73 & 7.80 & 8.13 & 8.38 & 8.61 & 8.78 & 8.98\\
		\hline
		\hline
		\multirow{30}{*}{\rotatebox{90}{$m = 4$ bit}} & \PLClipping[$1$] & 4.93 & 5.31 & 5.87 & 6.37 & 12.35 & 33.07 & 62.83 & 80.24 & 87.59\\
		& \PLClipping[$0.75$] & 4.66 & 4.96 & 5.43 & 5.85 & 9.59 & 20.57 & 44.09 & 69.93 & 82.21\\
		& \PLClipping[$0.5$] & 4.58 & 4.82 & 5.16 & 5.45 & 7.62 & 12.25 & 21.01 & 37.48 & 60.24\\
		& \PLClipping[$0.4$] & 4.87 & 5.04 & 5.35 & 5.59 & 7.18 & 9.67 & 14.06 & 22.08 & 35.51\\
		& \PLClipping[$0.3$] & 4.59 & 4.75 & 4.97 & 5.17 & 6.31 & 7.73 & 9.65 & 12.86 & 18.05\\
		& \PLClipping[$0.25$] & 4.63 & 4.79 & 5.01 & 5.20 & 6.15 & 7.34 & 8.70 & 10.70 & 13.49\\
		& \PLClipping[$0.1$] & 5.59 & 5.68 & 5.80 & 5.89 & 6.32 & 6.69 & 7.06 & 7.46 & 7.75\\
		& \PLClipping[$0.05$] & 7.63 & 7.74 & 7.84 & 7.94 & 8.30 & 8.62 & 8.88 & 9.16 & 9.40\\
		& \PLRandom[$0.75$] $p{=}0.01$ & 4.63 & 4.94 & 5.43 & 5.90 & 9.67 & 20.90 & 45.58 & 72.21 & 82.73\\
		& \PLRandom[$0.5$] $p{=}0.01$ & 4.69 & 4.94 & 5.31 & 5.60 & 7.63 & 12.50 & 23.41 & 42.24 & 63.73\\
		& \PLRandom[$0.4$] $p{=}0.01$ & 4.79 & 5.01 & 5.27 & 5.47 & 7.12 & 10.10 & 14.98 & 23.01 & 35.11\\
		& \PLRandom[$0.3$] $p{=}0.01$ & 4.87 & 5.02 & 5.26 & 5.45 & 6.64 & 8.14 & 10.11 & 13.11 & 17.71\\
		& \PLRandom[$0.25$] $p{=}0.01$ & 4.90 & 5.04 & 5.23 & 5.40 & 6.32 & 7.44 & 8.73 & 10.41 & 12.68\\
		& \PLRandom[$0.5$] $p{=}0.1$ & 5.29 & 5.52 & 5.90 & 6.22 & 8.47 & 12.65 & 20.36 & 33.56 & 51.59\\
		& \PLRandom[$0.4$] $p{=}0.1$ & 4.81 & 5.05 & 5.34 & 5.60 & 7.26 & 9.98 & 14.48 & 21.85 & 33.04\\
		& \PLRandom[$0.3$] $p{=}0.1$ & 4.84 & 5.00 & 5.25 & 5.44 & 6.54 & 7.88 & 9.86 & 12.47 & 16.31\\
		& \PLRandom[$0.25$] $p{=}0.1$ & 4.89 & 5.02 & 5.24 & 5.39 & 6.33 & 7.27 & 8.52 & 10.09 & 12.24\\
		& \PLRandom[$0.2$] $p{=}0.1$ & 4.69 & 4.80 & 4.95 & 5.07 & 5.72 & 6.41 & 7.18 & 8.07 & 9.09\\
		& \PLRandom[$0.25$] $p{=}1$ & 4.83 & 4.94 & 5.10 & 5.24 & 5.95 & 6.65 & 7.48 & 8.45 & 9.58\\
		& \PLRandom[$0.2$] $p{=}1$ & 4.96 & 5.10 & 5.24 & 5.35 & 5.97 & 6.60 & 7.17 & 7.83 & 8.64\\
		& \PLRandom[$0.15$] $p{=}1$ & 5.18 & 5.27 & 5.41 & 5.50 & 5.97 & 6.41 & 6.82 & 7.23 & 7.70\\
		& \PLRandom[$0.1$] $p{=}1$ & 5.81 & 5.89 & 6.00 & 6.10 & 6.52 & 6.88 & 7.20 & 7.52 & 7.86\\
		& \PLRandom[$0.2$] $p{=}2$ & 5.42 & 5.53 & 5.64 & 5.74 & 6.19 & 6.65 & 7.10 & 7.63 & 8.23\\
		& \PLRandom[$0.15$] $p{=}2$ & 5.31 & 5.43 & 5.57 & 5.68 & 6.14 & 6.54 & 6.91 & 7.23 & 7.67\\
		& \PLRandom[$0.1$] $p{=}2$ & 5.69 & 5.78 & 5.91 & 6.00 & 6.37 & 6.68 & 6.98 & 7.27 & 7.56\\
		& \PLRandom[$0.2$] $p{=}3$ & 5.20 & 5.32 & 5.44 & 5.55 & 6.00 & 6.41 & 6.82 & 7.22 & 7.70\\
		& \PLRandom[$0.15$] $p{=}3$ & 5.42 & 5.51 & 5.63 & 5.73 & 6.11 & 6.51 & 6.87 & 7.21 & 7.51\\
		& \PLRandom[$0.1$] $p{=}3$ & 6.04 & 6.11 & 6.23 & 6.31 & 6.67 & 6.97 & 7.26 & 7.53 & 7.81\\
		\hline
	\end{tabular}
	\vspace*{-0.2cm}
\end{table*}
\begin{table*}
	\centering
	\caption{\textbf{Overall Robustness Results on \CifarT.} Continued from \tabref{tab:supp-summary-cifar10-plc}. Tabular results corresponding to \figref{fig:supp-summary} for $m = 3$ and $2$ bits. We show \RTE for \PLClipping and \PLRandom with various $\wmax$ and $p$ across a subset of evaluated bit error rates.}
	\label{tab:supp-summary-cifar10-plc-2}
	\vspace*{-0.2cm}
	\scriptsize
	\begin{tabular}{| c | l | c | c | c | c | c | c | c | c | c |}
		\hline
		\multicolumn{11}{|c|}{\bfseries \CifarT: results with per-layer weight clipping for $m = 3,2$ bit}\\
		\hline
		& Model & \multirow{2}{*}{\begin{tabular}{c}\TE\\in \%\end{tabular}} & \multicolumn{8}{c|}{\RTE in \%, $p$ in \%}\\
		\cline{4-11}
		&&& $0.01$ & $0.05$ & $0.1$ & $0.5$ & $1$ & $1.5$ & $2$ & $2.5$\\
		\hline
		\hline
		\multirow{30}{*}{\rotatebox{90}{$m = 3$ bit}} & \PLClipping[$1$] & 6.34 & 6.75 & 7.40 & 8.02 & 13.31 & 26.54 & 49.60 & 72.07 & 83.24\\
		& \PLClipping[$0.75$] & 5.69 & 6.04 & 6.58 & 7.14 & 11.34 & 23.23 & 44.63 & 67.71 & 80.39\\
		& \PLClipping[$0.5$] & 5.21 & 5.46 & 5.87 & 6.21 & 8.66 & 13.42 & 21.82 & 36.08 & 54.85\\
		& \PLClipping[$0.25$] & 5.08 & 5.29 & 5.58 & 5.91 & 7.73 & 10.82 & 16.32 & 26.57 & 40.63\\
		& \PLClipping[$0.2$] & 4.81 & 4.98 & 5.17 & 5.37 & 6.72 & 10.14 & 19.80 & 40.39 & 64.79\\
		& \PLClipping[$0.15$] & 5.21 & 5.37 & 5.59 & 5.76 & 6.78 & 8.16 & 10.00 & 12.70 & 16.80\\
		& \PLClipping[$0.1$] & 6.02 & 6.10 & 6.22 & 6.31 & 6.76 & 7.20 & 7.61 & 8.01 & 8.46\\
		& \PLClipping[$0.05$] & 7.54 & 7.65 & 7.79 & 7.91 & 8.42 & 8.83 & 9.20 & 9.50 & 9.84\\
		& \PLRandom[$0.75$] $p{=}0.01$ & 6.38 & 6.77 & 7.30 & 7.88 & 12.26 & 22.93 & 41.96 & 64.96 & 78.82\\
		& \PLRandom[$0.5$] $p{=}0.01$ & 5.23 & 5.51 & 5.90 & 6.26 & 8.50 & 12.72 & 20.99 & 36.41 & 55.51\\
		& \PLRandom[$0.4$] $p{=}0.01$ & 5.07 & 5.30 & 5.62 & 5.94 & 7.88 & 11.48 & 17.38 & 28.41 & 42.84\\
		& \PLRandom[$0.3$] $p{=}0.01$ & 4.98 & 5.16 & 5.39 & 5.58 & 6.80 & 8.67 & 11.20 & 15.12 & 21.41\\
		& \PLRandom[$0.25$] $p{=}0.01$ & 4.94 & 5.09 & 5.30 & 5.48 & 6.49 & 7.76 & 9.47 & 11.97 & 16.02\\
		& \PLRandom[$0.5$] $p{=}0.1$ & 5.24 & 5.55 & 5.93 & 6.24 & 8.33 & 12.73 & 20.74 & 36.59 & 55.12\\
		& \PLRandom[$0.4$] $p{=}0.1$ & 5.12 & 5.34 & 5.63 & 5.89 & 7.62 & 10.66 & 16.12 & 25.76 & 39.06\\
		& \PLRandom[$0.3$] $p{=}0.1$ & 5.08 & 5.28 & 5.52 & 5.73 & 6.95 & 8.73 & 11.37 & 15.61 & 22.20\\
		& \PLRandom[$0.25$] $p{=}0.1$ & 5.12 & 5.28 & 5.52 & 5.71 & 6.70 & 7.98 & 9.61 & 11.93 & 15.23\\
		& \PLRandom[$0.2$] $p{=}0.1$ & 5.29 & 5.45 & 5.60 & 5.73 & 6.48 & 7.40 & 8.31 & 9.46 & 11.02\\
		& \PLRandom[$0.25$] $p{=}1$ & 5.22 & 5.39 & 5.59 & 5.75 & 6.55 & 7.46 & 8.38 & 9.47 & 10.96\\
		& \PLRandom[$0.2$] $p{=}1$ & 5.32 & 5.47 & 5.63 & 5.78 & 6.40 & 7.03 & 7.69 & 8.42 & 9.32\\
		& \PLRandom[$0.15$] $p{=}1$ & 5.28 & 5.37 & 5.51 & 5.63 & 6.19 & 6.76 & 7.29 & 7.89 & 8.46\\
		& \PLRandom[$0.1$] $p{=}1$ & 5.67 & 5.74 & 5.84 & 5.92 & 6.36 & 6.75 & 7.12 & 7.48 & 7.87\\
		& \PLRandom[$0.2$] $p{=}2$ & 5.32 & 5.45 & 5.57 & 5.69 & 6.24 & 6.83 & 7.36 & 8.00 & 8.68\\
		& \PLRandom[$0.15$] $p{=}2$ & 5.32 & 5.45 & 5.59 & 5.69 & 6.18 & 6.58 & 7.03 & 7.46 & 7.88\\
		& \PLRandom[$0.1$] $p{=}2$ & 5.97 & 6.04 & 6.13 & 6.22 & 6.65 & 6.99 & 7.31 & 7.69 & 8.06\\
		& \PLRandom[$0.05$] $p{=}2$ & 7.62 & 7.70 & 7.78 & 7.85 & 8.21 & 8.51 & 8.75 & 9.01 & 9.26\\
		& \PLRandom[$0.2$] $p{=}3$ & 5.63 & 5.74 & 5.85 & 5.94 & 6.40 & 6.83 & 7.25 & 7.72 & 8.20\\
		& \PLRandom[$0.15$] $p{=}3$ & 5.46 & 5.56 & 5.69 & 5.77 & 6.25 & 6.65 & 7.02 & 7.36 & 7.75\\
		& \PLRandom[$0.1$] $p{=}3$ & 6.14 & 6.25 & 6.38 & 6.48 & 6.86 & 7.20 & 7.47 & 7.79 & 8.07\\
		& \PLRandom[$0.05$] $p{=}3$ & 7.60 & 7.67 & 7.76 & 7.84 & 8.22 & 8.54 & 8.77 & 8.99 & 9.23\\
		\hline
		\hline
		\multirow{30}{*}{\rotatebox{90}{$m =2$ bit}} & \PLClipping[$1$] & 90.02 & 94.01 & 98.97 & 98.94 & 99.57 & 99.19 & 99.40 & 99.23 & 99.00\\
		& \PLClipping[$0.75$] & 66.16 & 67.94 & 71.39 & 74.31 & 85.34 & 91.21 & 93.56 & 95.35 & 95.92\\
		& \PLClipping[$0.5$] & 9.98 & 10.42 & 11.16 & 11.86 & 17.37 & 30.49 & 51.27 & 71.19 & 83.54\\
		& \PLClipping[$0.25$] & 8.10 & 8.54 & 9.17 & 9.74 & 14.12 & 21.93 & 36.88 & 55.20 & 73.57\\
		& \PLClipping[$0.2$] & 6.66 & 6.92 & 7.31 & 7.64 & 10.21 & 14.50 & 22.09 & 34.82 & 50.34\\
		& \PLClipping[$0.15$] & 6.08 & 6.35 & 6.66 & 6.96 & 8.65 & 11.43 & 15.87 & 22.66 & 33.32\\
		& \PLClipping[$0.1$] & 6.02 & 6.15 & 6.30 & 6.41 & 7.00 & 7.51 & 8.06 & 8.65 & 9.33\\
		& \PLClipping[$0.05$] & 7.52 & 7.62 & 7.76 & 7.87 & 8.45 & 8.95 & 9.41 & 9.87 & 10.35\\
		& \PLRandom[$0.75$] $p{=}0.01$ & 20.74 & 21.63 & 23.09 & 24.25 & 33.17 & 46.33 & 62.06 & 76.96 & 85.40\\
		& \PLRandom[$0.5$] $p{=}0.01$ & 10.22 & 10.79 & 11.56 & 12.21 & 17.29 & 28.67 & 48.03 & 70.85 & 83.98\\
		& \PLRandom[$0.4$] $p{=}0.01$ & 8.37 & 8.84 & 9.43 & 9.92 & 13.24 & 20.07 & 31.92 & 51.07 & 69.03\\
		& \PLRandom[$0.3$] $p{=}0.01$ & 6.48 & 6.81 & 7.24 & 7.61 & 10.04 & 14.53 & 22.43 & 37.15 & 53.81\\
		& \PLRandom[$0.25$] $p{=}0.01$ & 5.81 & 6.06 & 6.42 & 6.68 & 8.41 & 11.01 & 14.90 & 20.72 & 30.63\\
		& \PLRandom[$0.5$] $p{=}0.1$ & 11.03 & 11.50 & 12.23 & 12.86 & 16.98 & 25.83 & 40.03 & 60.23 & 77.68\\
		& \PLRandom[$0.4$] $p{=}0.1$ & 7.82 & 8.18 & 8.75 & 9.20 & 12.65 & 20.00 & 35.49 & 57.15 & 74.91\\
		& \PLRandom[$0.3$] $p{=}0.1$ & 6.17 & 6.42 & 6.80 & 7.15 & 9.28 & 13.15 & 19.52 & 30.75 & 49.51\\
		& \PLRandom[$0.25$] $p{=}0.1$ & 6.01 & 6.24 & 6.54 & 6.84 & 8.50 & 11.05 & 14.78 & 21.00 & 30.41\\
		& \PLRandom[$0.2$] $p{=}0.1$ & 5.70 & 5.88 & 6.13 & 6.36 & 7.44 & 8.71 & 10.45 & 13.06 & 16.44\\
		& \PLRandom[$0.25$] $p{=}1$ & 6.01 & 6.23 & 6.48 & 6.69 & 7.75 & 8.88 & 10.28 & 12.11 & 14.71\\
		& \PLRandom[$0.2$] $p{=}1$ & 5.82 & 5.98 & 6.19 & 6.35 & 7.13 & 8.02 & 9.00 & 10.05 & 11.47\\
		& \PLRandom[$0.15$] $p{=}1$ & 5.71 & 5.85 & 6.01 & 6.14 & 6.81 & 7.45 & 8.09 & 8.77 & 9.64\\
		& \PLRandom[$0.1$] $p{=}1$ & 6.34 & 6.43 & 6.55 & 6.65 & 7.19 & 7.67 & 8.12 & 8.55 & 9.07\\
		& \PLRandom[$0.2$] $p{=}2$ & 6.09 & 6.20 & 6.38 & 6.54 & 7.26 & 7.95 & 8.66 & 9.48 & 10.46\\
		& \PLRandom[$0.15$] $p{=}2$ & 6.22 & 6.37 & 6.55 & 6.70 & 7.29 & 7.73 & 8.17 & -- & --\\
		& \PLRandom[$0.1$] $p{=}2$ & 6.18 & 6.26 & 6.37 & 6.46 & 6.90 & 7.31 & 7.71 & 8.13 & 8.50\\
		& \PLRandom[$0.05$] $p{=}2$ & 7.79 & 7.89 & 8.02 & 8.13 & 8.61 & 8.97 & 9.32 & 9.65 & 10.04\\
		& \PLRandom[$0.05$] $p{=}3$ & 7.67 & 7.75 & 7.88 & 7.98 & 8.41 & 8.81 & 9.15 & 9.46 & 9.74\\
		\hline
	\end{tabular}
	\vspace*{-0.2cm}
\end{table*}
\begin{table*}
	\centering
	\caption{\textbf{Overall Robustness Results on \CifarH.} Tabular results corresponding to \figref{fig:supp-summary} for $m = 8$. We show \RTE for \Normal, \Clipping and \Random with various $\wmax$ and $p$ across a subset of test bit error rates.}
	\label{tab:supp-summary-cifar100}
	\vspace*{-0.2cm}
	\scriptsize
	\begin{tabular}{|l | c | c | c | c | c | c | c |}
		\hline
		\multicolumn{8}{|c|}{\bfseries \CifarH}\\
		\hline
		Model & \multirow{2}{*}{\begin{tabular}{c}\TE\\in \%\end{tabular}} & \multicolumn{6}{c|}{\RTE in \%, $p$ in \%}\\
		\cline{3-8}
		&& $0.005$ & $0.01$ & $0.05$ & $0.1$ & $0.5$ & $1$\\
		\hline
		\hline
		\Normal & 18.21 & 19.84 {} & 20.50 {} & 25.05 {} & 32.39 {} & 97.49 {} & 99.10 {}\\
		\Quant & 18.53 & 19.46 {} & 19.95 {} & 22.68 {} & 25.90 {} & 87.24 {} & 98.77 {}\\
		\Clipping[$0.25$] & 18.88 & 19.76 {} & 20.11 {} & 21.89 {} & 23.74 {} & 62.25 {} & 96.62 {}\\
		\Clipping[$0.2$] & 18.64 & 19.36 {} & 19.71 {} & 21.33 {} & 23.07 {} & 49.79 {} & 94.02 {}\\
		\Clipping[$0.15$] & 19.41 & 20.00 {} & 20.24 {} & 21.68 {} & 23.02 {} & 37.85 {} & 79.45 {}\\
		\Clipping[$0.1$] & 20.31 & 20.86 {} & 21.09 {} & 22.14 {} & 23.10 {} & 31.78 {} & 51.71 {}\\
		\Clipping[$0.05$] & 21.82 & 22.16 {} & 22.29 {} & 22.94 {} & 23.46 {} & 26.86 {} & 31.47 {}\\
		\hline
		\Random[$0.1$] $p{=}0.01$ & 19.68 & 20.21 {} & 20.46 {} & 21.52 {} & 22.56 {} & 30.59 {} & 48.93 {}\\
		\Random[$0.1$] $p{=}0.05$ & 19.94 & 20.47 {} & 20.69 {} & 21.72 {} & 22.60 {} & 29.93 {} & 46.76 {}\\
		\Random[$0.1$] $p{=}0.1$ & 19.18 & 19.67 {} & 19.86 {} & 20.87 {} & 21.69 {} & 28.03 {} & 41.29 {}\\
		\Random[$0.1$] $p{=}0.5$ & 19.90 & 20.24 {} & 20.41 {} & 21.17 {} & 21.83 {} & 25.66 {} & 31.55 {}\\
		\Random[$0.1$] $p{=}1$ & 21.08 & 21.43 {} & 21.59 {} & 22.24 {} & 22.76 {} & 25.73 {} & 29.31 {}\\
		\Random[$0.05$] $p{=}0.01$ & 21.86 & 22.17 {} & 22.31 {} & 23.00 {} & 23.57 {} & 26.84 {} & 31.33 {}\\
		\Random[$0.05$] $p{=}0.05$ & 20.97 & 21.30 {} & 21.44 {} & 22.12 {} & 22.72 {} & 25.95 {} & 30.14 {}\\
		\Random[$0.05$] $p{=}0.1$ & 21.22 & 21.53 {} & 21.66 {} & 22.29 {} & 22.81 {} & 25.88 {} & 29.93 {}\\
		\Random[$0.05$] $p{=}0.5$ & 21.29 & 21.55 {} & 21.65 {} & 22.13 {} & 22.60 {} & 25.01 {} & 27.70 {}\\
		\Random[$0.05$] $p{=}1$ & 20.83 & 21.08 {} & 21.20 {} & 21.73 {} & 22.16 {} & 24.33 {} & 26.49 {}\\
		\hline
		\hline
		\PLClipping[$1$] & 18.42 & 19.30 & 19.66 & 21.79 & 24.15 & 67.32 & 97.64\\
		\PLClipping[$0.75$] & 19.13 & 19.80 & 20.08 & 21.65 & 23.27 & 48.91 & 92.80\\
		\PLClipping[$0.5$] & 19.32 & 19.87 & 20.09 & 21.28 & 22.37 & 32.70 & 63.37\\
		\PLClipping[$0.4$] & 18.79 & 19.34 & 19.58 & 20.65 & 21.59 & 29.08 & 48.17\\
		\PLClipping[$0.3$] & 19.51 & 19.89 & 20.08 & 20.92 & 21.69 & 26.62 & 34.52\\
		\PLClipping[$0.25$] & 19.06 & 19.48 & 19.64 & 20.35 & 20.99 & 24.78 & 30.88\\
		\PLClipping[$0.2$] & 19.67 & 20.01 & 20.14 & 20.77 & 21.34 & 24.35 & 27.77\\
		\PLClipping[$0.15$] & 20.34 & 20.54 & 20.63 & 21.15 & 21.56 & 23.59 & 25.78\\
		\PLClipping[$0.1$] & 21.49 & 21.68 & 21.76 & 22.16 & 22.44 & 23.98 & 25.37\\
		\PLRandom[$0.4$] $p{=}0.001$ & 18.60 & 19.12 & 19.35 & 20.52 & 21.58 & 30.06 & 51.75\\
		\PLRandom[$0.3$] $p{=}0.001$ & 19.10 & 19.60 & 19.82 & 20.80 & 21.60 & 26.58 & 34.63\\
		\PLRandom[$0.25$] $p{=}0.001$ & 19.68 & 20.03 & 20.18 & 20.91 & 21.55 & 25.16 & 30.52\\
		\PLRandom[$0.2$] $p{=}1$ & 19.08 & 19.30 & 19.41 & 19.92 & 20.35 & 22.40 & 24.40\\
		\PLRandom[$0.15$] $p{=}1$ & 19.72 & 19.92 & 20.01 & 20.40 & 20.72 & 22.35 & 23.86\\
		\PLRandom[$0.1$] $p{=}1$ & 21.04 & 21.21 & 21.27 & 21.60 & 21.88 & 23.16 & 24.26\\
		\hline
	\end{tabular}
\end{table*}
\begin{table*}
	\centering
	\caption{\textbf{Overall Robustness Results on \MNIST.} Tabular results corresponding to \figref{fig:supp-summary} for $m = 8, 4, 2$ bits. We show \RTE for \Normal, \Clipping and \Random with various $\wmax$ and $p$ across a subset of test bit error rates. Results for per-layer weight clipping in \tabref{tab:supp-summary-mnist-plc}.}
	\label{tab:supp-summary-mnist}
	\vspace*{-0.2cm}
	\begin{tabular}{| c | l | c | c | c | c | c | c | c | c |}
		\hline
		\multicolumn{10}{|c|}{\bfseries MNIST}\\
		\hline
		& Model & \multirow{2}{*}{\begin{tabular}{c}\TE\\in \%\end{tabular}} & \multicolumn{7}{c|}{\RTE in \%, $p$ in \%}\\
		\cline{3-10}
		&&& $1$ & $5$ & $10$ & $12.5$ & $15$ & $17.5$ & $20$\\
		\hline
		\hline
		\multirow{11}{*}{\rotatebox{90}{$m =8$ bit}} & \Normal & 0.39 & 0.77 & 86.37 & 89.92 & 89.82 & 89.81 & 90.09 & 90.03\\
		& \Quant & 0.40 & 0.69 & 85.96 & 90.20 & 89.86 & 90.10 & 89.72 & 89.83\\
		& \Clipping[$0.1$] & 0.39 & 0.48 & 18.21 & 88.93 & 90.35 & 90.06 & 90.56 & 90.18\\
		& \Clipping[$0.05$] & 0.42 & 0.47 & 0.63 & 8.67 & 51.38 & 80.64 & 87.79 & 89.57\\
		& \Clipping[$0.025$] & 0.43 & 0.47 & 0.56 & 0.71 & 0.95 & 1.81 & 7.22 & 32.68\\
		\cline{2-10}
		& \Random[$0.1$] $p{=}1$ & 0.36 & 0.44 & 3.41 & 86.29 & 89.05 & 89.85 & 90.10 & 89.93\\
		& \Random[$0.05$] $p{=}1$ & 0.34 & 0.39 & 0.59 & 8.92 & 51.32 & 79.35 & 87.63 & 89.15\\
		& \Random[$0.05$] $p{=}5$ & 0.34 & 0.38 & 0.50 & 1.02 & 5.12 & 41.31 & 79.19 & 87.88\\
		& \Random[$0.05$] $p{=}10$ & 0.40 & 0.43 & 0.51 & 0.67 & 0.86 & 1.74 & 9.77 & 47.58\\
		& \Random[$0.05$] $p{=}15$ & 0.39 & 0.40 & 0.45 & 0.56 & 0.64 & 0.78 & 1.10 & 2.72\\
		& \Random[$0.05$] $p{=}20$ & 0.39 & 0.42 & 0.48 & 0.53 & 0.57 & 0.63 & 0.74 & 0.94\\
		\hline
		\hline
		\multirow{14}{*}{\rotatebox{90}{$m =4$ bit}} & \Quant & 0.36 & 0.72 & 87.21 & 90.23 & 90.01 & 89.88 & 89.97 & 89.67\\
		& \Clipping[$0.1$] & 0.38 & 0.51 & 38.75 & 88.33 & 89.47 & 89.57 & 90.10 & 89.67\\
		& \Clipping[$0.05$] & 0.31 & 0.39 & 0.78 & 44.15 & 78.64 & 87.32 & 89.03 & 89.71\\
		& \Clipping[$0.025$] & 0.37 & 0.41 & 0.50 & 0.67 & 0.99 & 4.63 & 29.46 & 67.21\\
		\cline{2-10}
		& \Random[$0.1$] $p{=}1$ & 0.38 & 0.48 & 13.29 & 87.43 & 89.70 & 89.63 & 89.41 & 90.02\\
		& \Random[$0.1$] $p{=}5$ & 0.38 & 0.48 & 0.78 & 24.73 & 74.88 & 87.04 & 88.72 & 89.55\\
		& \Random[$0.1$] $p{=}10$ & 0.40 & 0.47 & 0.64 & 1.22 & 2.62 & 16.72 & 64.33 & 83.80\\
		& \Random[$0.1$] $p{=}15$ & 0.56 & 0.59 & 0.73 & 1.03 & 1.28 & 1.87 & 3.71 & 14.39\\
		& \Random[$0.1$] $p{=}20$ & 0.56 & 9.48 & 14.29 & 7.39 & 6.07 & 5.80 & 6.10 & 8.12\\
		& \Random[$0.05$] $p{=}1$ & 0.37 & 0.43 & 0.67 & 36.99 & 77.12 & 85.97 & 88.62 & 89.94\\
		& \Random[$0.05$] $p{=}5$ & 0.38 & 0.42 & 0.53 & 1.38 & 12.90 & 60.73 & 83.69 & 88.75\\
		& \Random[$0.05$] $p{=}10$ & 0.34 & 0.39 & 0.47 & 0.65 & 0.91 & 2.11 & 19.15 & 71.25\\
		& \Random[$0.05$] $p{=}15$ & 0.37 & 0.39 & 0.43 & 0.52 & 0.63 & 0.79 & 1.17 & 3.16\\
		& \Random[$0.05$] $p{=}20$ & 0.44 & 0.48 & 0.53 & 0.60 & 0.65 & 0.72 & 0.86 & 1.04\\
		\hline
		\hline
		\multirow{6}{*}{\rotatebox{90}{$m =2$ bit}} & \Clipping[$0.1$] & 0.47 & 3.82 & 89.19 & 89.92 & 90.22 & 90.14 & \multicolumn{2}{c}{\hphantom{c}}\\
		& \Clipping[$0.05$] & 0.41 & 0.62 & 77.19 & 89.47 & 90.40 & 90.06 & \multicolumn{2}{c}{\hphantom{c}}\\
		\cline{2-8}
		& \Random[$0.05$] $p{=}3$ & 0.47 & 0.53 & 1.36 & 82.71 & 88.66 & 90.28 & \multicolumn{2}{c}{\hphantom{c}}\\
		& \Random[$0.05$] $p{=}5$ & 0.40 & 0.49 & 0.77 & 25.72 & 78.71 & 88.22 & \multicolumn{2}{c}{\hphantom{c}}\\
		& \Random[$0.05$] $p{=}10$ & 0.40 & 0.45 & 0.58 & 0.94 & 1.82 & 15.70 & \multicolumn{2}{c}{\hphantom{c}}\\
		& \Random[$0.05$] $p{=}15$ & 0.46 & 0.51 & 0.60 & 0.77 & 0.91 & 1.21 & \multicolumn{2}{c}{\hphantom{c}}\\
		\cline{1-8}
	\end{tabular}
\end{table*}
\begin{table*}
	\centering
	\caption{\textbf{Overall Robustness Results on \MNIST.} Tabular results corresponding to \figref{fig:supp-summary} for $m = 8, 4, 2$ bits for per-layer weight clipping, \ie, \PLClipping and \PLRandom with various $\wmax$ and $p$ across a subset of bit error rates.}
	\label{tab:supp-summary-mnist-plc}
	\vspace*{-0.2cm}
	\begin{tabular}{| c | l | c | c | c | c | c | c |}
		\hline
		\multicolumn{8}{|c|}{\bfseries MNIST: results with per-layer weight clipping}\\
		\hline
		& Model & \multirow{2}{*}{\begin{tabular}{c}\TE\\in \%\end{tabular}} & \multicolumn{5}{c|}{\RTE in \%, $p$ in \% }\\
		\cline{3-8}
		&&& $1$ & $5$ & $10$ & $15$ & $20$\\
		\hline
		\hline
		\multirow{24}{*}{\rotatebox{90}{$m =8$ bit}} & \PLClipping[$1$] & 0.40 & 0.88 & 88.09 & 90.19 &  90.21 &  90.20\\
		& \PLClipping[$0.5$] & 0.37 & 0.60 & 77.62 & 89.70 &  89.66 &  89.91\\
		& \PLClipping[$0.25$] & 0.36 & 0.44 & 3.87 & 80.29 &  89.30 &  89.52\\
		& \PLClipping[$0.2$] & 0.33 & 0.42 & 1.31 & 48.85 &  87.60 &  89.68\\
		& \PLClipping[$0.15$] & 0.39 & 0.44 & 0.68 & 6.31 &  67.90 &  88.08\\
		& \PLClipping[$0.1$] & 0.39 & 0.42 & 0.52 & 0.85 &  5.88 &  56.97\\
		& \PLClipping[$0.05$] & 0.38 & 0.41 & 0.45 & 0.54 &  0.77 &  3.52\\
		& \PLRandom[$0.2$] $p{=}1$ & 0.34 & 0.42 & 0.88 & 35.33 &  86.96 &  89.74\\
		& \PLRandom[$0.15$] $p{=}1$ & 0.38 & 0.43 & 0.65 & 3.23 &  58.71 &  87.13\\
		& \PLRandom[$0.1$] $p{=}1$ & 0.33 & 0.42 & 0.53 & 0.85 &  5.71 &  53.31\\
		& \PLRandom[$0.2$] $p{=}3$ & 0.37 & 0.45 & 0.71 & 9.03 &  77.97 &  89.59\\
		& \PLRandom[$0.15$] $p{=}3$ & 0.38 & 0.42 & 0.57 & 1.86 &  43.08 &  86.21\\
		& \PLRandom[$0.1$] $p{=}3$ & 0.37 & 0.40 & 0.49 & 0.70 &  2.60 &  42.77\\
		& \PLRandom[$0.2$] $p{=}5$ & 0.36 & 0.40 & 0.59 & 2.83 &  62.93 &  87.39\\
		& \PLRandom[$0.15$] $p{=}5$ & 0.39 & 0.43 & 0.57 & 1.27 &  25.75 &  82.66\\
		& \PLRandom[$0.1$] $p{=}5$ & 0.40 & 0.44 & 0.53 & 0.72 &  2.00 &  37.44\\
		& \PLRandom[$0.05$] $p{=}5$ & 0.36 & 0.39 & 0.44 & 0.52 &  0.68 &  1.59\\
		& \PLRandom[$0.15$] $p{=}10$ & 0.36 & 0.41 & 0.48 & 0.67 &  1.85 &  38.64\\
		& \PLRandom[$0.1$] $p{=}10$ & 0.39 & 0.41 & 0.47 & 0.59 &  0.96 &  8.96\\
		& \PLRandom[$0.05$] $p{=}10$ & 0.38 & 0.39 & 0.43 & 0.47 &  0.57 &  0.98\\
		& \PLRandom[$0.1$] $p{=}15$ & 0.36 & 0.39 & 0.43 & 0.49 &  0.65 &  1.42\\
		& \PLRandom[$0.05$] $p{=}15$ & 0.36 & 0.38 & 0.42 & 0.47 &  0.52 &  0.67\\
		& \PLRandom[$0.1$] $p{=}20$ & 0.38 & 0.40 & 0.44 & 0.48 &  0.56 &  0.76\\
		& \PLRandom[$0.05$] $p{=}20$ & 0.33 & 0.34 & 0.38 & 0.42 &  0.47 &  0.56\\
		\hline
		\hline
		\multirow{25}{*}{\rotatebox{90}{$m =4$ bit}} & \PLClipping[$1$] & 0.44 & 1.16 & 88.78 & 89.89 & 90.05 & 89.78\\
		& \PLClipping[$0.5$] & 0.44 & 0.68 & 78.09 & 89.89 & 89.66 & 90.49\\
		& \PLClipping[$0.25$] & 0.39 & 0.52 & 7.51 & 82.87 & 89.75 & 89.55\\
		& \PLClipping[$0.2$] & 0.40 & 0.49 & 1.93 & 66.75 & 88.17 & 89.83\\
		& \PLClipping[$0.15$] & 0.40 & 0.48 & 0.81 & 12.92 & 78.62 & 89.15\\
		& \PLClipping[$0.1$] & 0.39 & 0.45 & 0.56 & 1.23 & 23.15 & 80.47\\
		& \PLClipping[$0.05$] & 0.36 & 0.39 & 0.46 & 0.57 & 1.03 & 13.81\\
		& \PLRandom[$0.25$] $p{=}1$ & 0.34 & 0.47 & 3.44 & 80.84 & 89.82 & 89.83\\
		& \PLRandom[$0.2$] $p{=}1$ & 0.34 & 0.44 & 1.27 & 59.95 & 88.07 & 90.14\\
		& \PLRandom[$0.15$] $p{=}1$ & 0.38 & 0.44 & 0.73 & 9.22 & 78.16 & 89.18\\
		& \PLRandom[$0.1$] $p{=}1$ & 0.34 & 0.41 & 0.55 & 1.17 & 20.42 & 79.29\\
		& \PLRandom[$0.25$] $p{=}3$ & 0.39 & 0.47 & 1.04 & 35.72 & 87.54 & 90.17\\
		& \PLRandom[$0.2$] $p{=}3$ & 0.31 & 0.39 & 0.80 & 22.57 & 85.08 & 89.74\\
		& \PLRandom[$0.15$] $p{=}3$ & 0.33 & 0.40 & 0.62 & 3.37 & 65.18 & 88.82\\
		& \PLRandom[$0.1$] $p{=}3$ & 0.40 & 0.44 & 0.55 & 0.92 & 10.87 & 78.08\\
		& \PLRandom[$0.2$] $p{=}5$ & 0.36 & 0.45 & 0.67 & 3.79 & 66.59 & 88.48\\
		& \PLRandom[$0.15$] $p{=}5$ & 0.38 & 0.44 & 0.59 & 1.56 & 36.84 & 86.87\\
		& \PLRandom[$0.1$] $p{=}5$ & 0.46 & 0.49 & 0.58 & 0.82 & 5.05 & 70.44\\
		& \PLRandom[$0.05$] $p{=}5$ & 0.31 & 0.35 & 0.41 & 0.51 & 0.78 & 6.44\\
		& \PLRandom[$0.15$] $p{=}10$ & 0.38 & 0.41 & 0.49 & 0.72 & 2.68 & 54.87\\
		& \PLRandom[$0.1$] $p{=}10$ & 0.44 & 0.50 & 0.55 & 0.66 & 1.10 & 21.86\\
		& \PLRandom[$0.05$] $p{=}10$ & 0.30 & 0.33 & 0.38 & 0.47 & 0.63 & 1.50\\
		& \PLRandom[$0.1$] $p{=}15$ & 0.40 & 0.42 & 0.45 & 0.53 & 0.69 & 1.68\\
		& \PLRandom[$0.05$] $p{=}15$ & 0.40 & 0.43 & 0.46 & 0.50 & 0.58 & 0.79\\
		& \PLRandom[$0.1$] $p{=}20$ & 0.36 & 0.39 & 0.43 & 0.47 & 0.57 & 0.82\\
		\hline
		\hline
		\multirow{25}{*}{\rotatebox{90}{$m =2$ bit}} & \PLClipping[$1$] & 1.01 & 65.76 & 90.02 & 90.04 & 90.10 & 89.85\\
		& \PLClipping[$0.5$] & 0.52 & 27.23 & 89.24 & 89.97 & 89.88 & 90.34\\
		& \PLClipping[$0.25$] & 0.50 & 3.84 & 87.87 & 90.33 & 89.88 & 90.10\\
		& \PLClipping[$0.2$] & 0.49 & 0.91 & 63.98 & 89.79 & 90.51 & 90.32\\
		& \PLClipping[$0.15$] & 0.48 & 0.62 & 30.61 & 87.62 & 89.72 & 89.77\\
		& \PLClipping[$0.1$] & 0.34 & 0.43 & 0.91 & 38.84 & 87.94 & 90.09\\
		& \PLClipping[$0.05$] & 0.42 & 0.48 & 0.61 & 1.42 & 42.80 & 88.61\\
		& \PLRandom[$0.25$] $p{=}1$ & 0.42 & 0.67 & 67.48 & 89.64 & 90.03 & 89.97\\
		& \PLRandom[$0.2$] $p{=}1$ & 0.32 & 0.53 & 19.68 & 88.73 & 90.11 & 89.91\\
		& \PLRandom[$0.15$] $p{=}1$ & 0.38 & 0.52 & 6.07 & 85.03 & 90.04 & 90.18\\
		& \PLRandom[$0.1$] $p{=}1$ & 0.42 & 0.51 & 1.04 & 42.66 & 88.81 & 89.93\\
		& \PLRandom[$0.25$] $p{=}3$ & 0.42 & 0.56 & 5.06 & 86.20 & 90.21 & 90.05\\
		& \PLRandom[$0.2$] $p{=}3$ & 0.33 & 0.48 & 2.44 & 81.45 & 89.73 & 89.97\\
		& \PLRandom[$0.15$] $p{=}3$ & 0.33 & 0.41 & 1.18 & 60.65 & 89.23 & 90.07\\
		& \PLRandom[$0.1$] $p{=}3$ & 0.31 & 0.39 & 0.63 & 12.21 & 85.13 & 90.17\\
		& \PLRandom[$0.2$] $p{=}5$ & 0.41 & 0.52 & 1.00 & 32.74 & 88.28 & 90.04\\
		& \PLRandom[$0.15$] $p{=}5$ & 0.37 & 0.44 & 0.77 & 21.81 & 86.35 & 90.02\\
		& \PLRandom[$0.1$] $p{=}5$ & 0.44 & 0.50 & 0.65 & 2.72 & 74.74 & 89.42\\
		& \PLRandom[$0.05$] $p{=}5$ & 0.40 & 0.43 & 0.52 & 0.76 & 6.26 & 84.26\\
		& \PLRandom[$0.1$] $p{=}10$ & 0.43 & 0.47 & 0.57 & 0.87 & 6.36 & 82.09\\
		& \PLRandom[$0.05$] $p{=}10$ & 0.34 & 0.39 & 0.46 & 0.55 & 0.93 & 61.55\\
		& \PLRandom[$0.1$] $p{=}15$ & 0.42 & 0.44 & 0.51 & 0.62 & 1.15 & 13.14\\
		& \PLRandom[$0.05$] $p{=}15$ & 0.36 & 0.40 & 0.45 & 0.51 & 0.68 & 4.77\\
		& \PLRandom[$0.1$] $p{=}20$ & 0.49 & 0.53 & 0.62 & 0.70 & 0.86 & 1.75\\
		& \PLRandom[$0.05$] $p{=}20$ & 0.44 & 0.48 & 0.53 & 0.60 & 0.72 & 1.38\\
		\cline{1-8}
	\end{tabular}
\end{table*}
\begin{table*}
	\centering
	\caption{\revision{\textbf{Overall Robustness Results on TinyImageNet.} Tabular results corresponding to \figref{fig:supp-summary} for $m = 8$ bits with various $\wmax$ and $p$ across a subset of bit error rates. \figref{tab:supp-summary-tinyimagenet-4} to \ref{tab:supp-summary-tinyimagenet-2} contain results for $m = 4,3$ and $2$ bits.}}
	\label{tab:supp-summary-tinyimagenet-8}
	\vspace*{-0.2cm}
	\scriptsize
	\begin{tabular}{| l | c | c | c | c | c | c | c | c |}
		\hline
		\multicolumn{9}{|c|}{\bfseries TinyImageNet}\\
		\hline
		Model & \multirow{2}{*}{\begin{tabular}{c}\TE\\in \%\end{tabular}} & \multicolumn{7}{c|}{\RTE in \%, $p$ in \%}\\
		\hline
		&& $0.001$ & $0.005$ & $0.01$ & $0.05$ & $0.1$ & $0.5$ & $1$\\
		\hline
		\hline
		\Normal & 36.23 & 43.77 & 44.03 & 44.42 & 47.67 & 52.00 & 89.58 & 98.86\\
		\Quant & 36.77 & 37.27 & 37.98 & 38.51 & 41.35 & 44.02 & 69.88 & 95.77\\
		\Clipping[$0.5$] & 35.89 & 36.45 & 37.21 & 37.82 & 40.90 & 43.76 & 71.52 & 96.58\\
		\Clipping[$0.25$] & 36.02 & 36.45 & 37.10 & 37.59 & 39.96 & 42.17 & 61.21 & 89.51\\
		\Clipping[$0.2$] & 36.82 & 37.25 & 37.78 & 38.23 & 40.39 & 42.41 & 58.60 & 83.64\\
		\Clipping[$0.15$] & 37.42 & 37.73 & 38.22 & 38.59 & 40.40 & 41.94 & 52.55 & 69.39\\
		\Clipping[$0.1$] & 37.52 & 37.83 & 38.25 & 38.56 & 39.92 & 40.99 & 47.47 & 55.92\\
		\Clipping[$0.05$] & 40.02 & 40.23 & 40.46 & 40.63 & 41.37 & 41.92 & 44.70 & 47.72\\
		\Clipping[$0.025$] & 44.98 & 45.07 & 45.20 & 45.30 & 45.72 & 46.03 & 47.47 & 48.78\\
		\PLClipping[$1$] & 35.98 & 36.52 & 37.19 & 37.73 & 40.52 & 43.00 & 66.52 & 93.72\\
		\PLClipping[$0.5$] & 36.36 & 36.74 & 37.18 & 37.52 & 39.12 & 40.57 & 50.03 & 64.13\\
		\PLClipping[$0.25$] & 37.33 & 37.49 & 37.73 & 37.92 & 38.85 & 39.60 & 43.69 & 47.97\\
		\PLClipping[$0.2$] & 38.12 & 38.27 & 38.47 & 38.63 & 39.40 & 40.05 & 43.15 & 46.08\\
		\PLClipping[$0.15$] & 38.13 & 38.27 & 38.44 & 38.58 & 39.21 & 39.65 & 42.05 & 44.18\\
		\PLClipping[$0.1$] & 40.72 & 40.81 & 40.94 & 41.04 & 41.52 & 41.88 & 43.47 & 44.86\\
		\PLClipping[$0.05$] & 46.19 & 46.25 & 46.36 & 46.45 & 46.83 & 47.09 & 48.35 & 49.40\\
		\PLClipping[$0.025$] & 55.84 & 55.89 & 55.98 & 56.06 & 56.41 & 56.68 & 57.79 & 58.69\\
		\hline
		\Random[$1$] $p{=}0.01$ & 36.02 & 36.58 & 37.28 & 37.83 & 40.69 & 43.47 & 68.99 & 95.82\\
		\Random[$0.5$] $p{=}0.01$ & 36.32 & 36.81 & 37.49 & 38.03 & 40.83 & 43.42 & 68.03 & 95.21\\
		\Random[$0.25$] $p{=}0.01$ & 36.06 & 36.47 & 37.05 & 37.50 & 39.73 & 41.72 & 57.23 & 81.51\\
		\Random[$0.2$] $p{=}0.01$ & 36.37 & 36.74 & 37.29 & 37.70 & 39.79 & 41.62 & 55.43 & 77.68\\
		\Random[$0.15$] $p{=}0.01$ & 36.87 & 37.25 & 37.73 & 38.10 & 39.82 & 41.31 & 51.63 & 68.02\\
		\Random[$0.1$] $p{=}0.01$ & 38.44 & 38.66 & 39.01 & 39.27 & 40.52 & 41.54 & 47.89 & 55.97\\
		\Random[$0.05$] $p{=}0.01$ & 40.50 & 40.68 & 40.90 & 41.02 & 41.67 & 42.20 & 45.04 & 47.78\\
		\Random[$1$] $p{=}0.1$ & 36.63 & 37.15 & 37.75 & 38.28 & 40.92 & 43.38 & 64.32 & 93.55\\
		\Random[$0.5$] $p{=}0.1$ & 36.88 & 37.32 & 37.93 & 38.42 & 40.67 & 42.92 & 61.81 & 91.36\\
		\Random[$0.25$] $p{=}0.1$ & 37.13 & 37.54 & 38.04 & 38.42 & 40.41 & 42.25 & 56.21 & 79.18\\
		\Random[$0.2$] $p{=}0.1$ & 37.49 & 37.85 & 38.31 & 38.68 & 40.53 & 42.21 & 54.63 & 75.19\\
		\Random[$0.15$] $p{=}0.1$ & 37.17 & 37.44 & 37.80 & 38.13 & 39.67 & 41.04 & 49.91 & 63.11\\
		\Random[$0.1$] $p{=}0.1$ & 38.19 & 38.37 & 38.64 & 38.88 & 39.98 & 40.95 & 46.53 & 53.42\\
		\Random[$0.05$] $p{=}0.1$ & 40.28 & 40.43 & 40.64 & 40.79 & 41.51 & 42.10 & 44.95 & 47.70\\
		\Random[$0.025$] $p{=}0.1$ & 44.77 & 44.85 & 44.96 & 45.07 & 45.50 & 45.82 & 47.35 & 48.72\\
		\Random[$0.2$] $p{=}0.5$ & 41.33 & 41.61 & 41.99 & 42.32 & 43.69 & 44.82 & 51.85 & 61.95\\
		\Random[$0.15$] $p{=}0.5$ & 39.96 & 40.20 & 40.53 & 40.79 & 42.03 & 43.06 & 49.08 & 56.62\\
		\Random[$0.1$] $p{=}0.5$ & 39.91 & 40.11 & 40.39 & 40.62 & 41.63 & 42.47 & 46.70 & 51.75\\
		\Random[$0.05$] $p{=}0.5$ & 40.58 & 40.69 & 40.88 & 41.03 & 41.74 & 42.22 & 44.70 & 47.13\\
		\Random[$0.025$] $p{=}0.5$ & 45.34 & 45.42 & 45.53 & 45.60 & 45.98 & 46.27 & 47.70 & 48.96\\
		\Random[$0.5$] $p{=}1$ & 50.31 & 50.62 & 51.07 & 51.44 & 53.03 & 54.36 & 61.71 & 69.88\\
		\Random[$0.25$] $p{=}1$ & 46.63 & 46.86 & 47.21 & 47.48 & 48.81 & 49.93 & 56.27 & 63.16\\
		\Random[$0.2$] $p{=}1$ & 45.89 & 46.16 & 46.47 & 46.71 & 47.79 & 48.73 & 54.17 & 60.08\\
		\Random[$0.15$] $p{=}1$ & 43.77 & 43.98 & 44.28 & 44.50 & 45.55 & 46.44 & 50.92 & 55.95\\
		\Random[$0.1$] $p{=}1$ & 42.30 & 42.43 & 42.66 & 42.85 & 43.60 & 44.29 & 47.87 & 51.50\\
		\Random[$0.05$] $p{=}1$ & 41.13 & 41.28 & 41.41 & 41.51 & 42.04 & 42.46 & 44.69 & 46.86\\
		\Random[$0.025$] $p{=}1$ & 45.91 & 46.02 & 46.13 & 46.21 & 46.59 & 46.88 & 48.28 & 49.45\\
		\Random[$0.2$] $p{=}1.5$ & 51.53 & 51.75 & 52.05 & 52.28 & 53.34 & 54.23 & 58.62 & 63.24\\
		\Random[$0.15$] $p{=}1.5$ & 48.06 & 48.29 & 48.56 & 48.79 & 49.78 & 50.52 & 54.43 & 58.27\\
		\Random[$0.1$] $p{=}1.5$ & 43.83 & 44.06 & 44.28 & 44.46 & 45.21 & 45.82 & 48.95 & 52.18\\
		\Random[$0.05$] $p{=}1.5$ & 42.00 & 42.12 & 42.27 & 42.38 & 42.93 & 43.39 & 45.48 & 47.44\\
		\Random[$0.025$] $p{=}1.5$ & 45.90 & 45.99 & 46.11 & 46.20 & 46.59 & 46.88 & 48.32 & 49.36\\
		\hline
		\PLRandom[$1$] $p{=}0.01$ & 35.27 & 35.69 & 36.36 & 36.89 & 39.80 & 42.21 & 62.62 & 90.24\\
		\PLRandom[$0.5$] $p{=}0.01$ & 36.44 & 36.78 & 37.21 & 37.60 & 39.29 & 40.69 & 50.22 & 64.53\\
		\PLRandom[$0.25$] $p{=}0.01$ & 37.01 & 37.17 & 37.42 & 37.64 & 38.61 & 39.38 & 43.45 & 47.76\\
		\PLRandom[$0.2$] $p{=}0.01$ & 37.73 & 37.91 & 38.11 & 38.25 & 39.01 & 39.66 & 42.82 & 45.78\\
		\PLRandom[$0.15$] $p{=}0.01$ & 38.92 & 39.04 & 39.22 & 39.36 & 39.94 & 40.43 & 42.69 & 44.94\\
		\PLRandom[$0.1$] $p{=}0.01$ & 40.47 & 40.57 & 40.71 & 40.81 & 41.28 & 41.63 & 43.27 & 44.72\\
		\PLRandom[$0.05$] $p{=}0.01$ & 45.88 & 45.96 & 46.05 & 46.12 & 46.50 & 46.81 & 48.02 & 49.10\\
		\PLRandom[$0.5$] $p{=}0.1$ & 36.53 & 36.82 & 37.22 & 37.60 & 39.24 & 40.56 & 49.15 & 61.36\\
		\PLRandom[$0.25$] $p{=}0.1$ & 36.99 & 37.18 & 37.43 & 37.66 & 38.64 & 39.43 & 43.25 & 47.11\\
		\PLRandom[$0.2$] $p{=}0.1$ & 37.49 & 37.66 & 37.88 & 38.07 & 38.85 & 39.45 & 42.50 & 45.51\\
		\PLRandom[$0.15$] $p{=}0.1$ & 38.91 & 39.03 & 39.18 & 39.31 & 39.89 & 40.33 & 42.62 & 44.70\\
		\PLRandom[$0.1$] $p{=}0.1$ & 40.40 & 40.52 & 40.69 & 40.81 & 41.33 & 41.69 & 43.38 & 44.78\\
		\PLRandom[$0.05$] $p{=}0.1$ & 46.41 & 46.49 & 46.58 & 46.66 & 46.98 & 47.24 & 48.46 & 49.48\\
		\PLRandom[$0.025$] $p{=}0.1$ & 55.82 & 55.91 & 56.04 & 56.12 & 56.46 & 56.74 & 57.91 & 58.84\\
		\PLRandom[$0.2$] $p{=}0.5$ & 38.31 & 38.45 & 38.65 & 38.79 & 39.53 & 40.10 & 42.82 & 45.44\\
		\PLRandom[$0.15$] $p{=}0.5$ & 38.18 & 38.30 & 38.46 & 38.59 & 39.13 & 39.57 & 41.67 & 43.58\\
		\PLRandom[$0.1$] $p{=}0.5$ & 40.56 & 40.66 & 40.77 & 40.86 & 41.28 & 41.58 & 43.09 & 44.50\\
		\PLRandom[$0.05$] $p{=}0.5$ & 46.42 & 46.51 & 46.60 & 46.65 & 46.96 & 47.18 & 48.35 & 49.24\\
		\PLRandom[$0.025$] $p{=}0.5$ & 55.50 & 55.57 & 55.66 & 55.72 & 56.00 & 56.22 & 57.28 & 58.17\\
		\PLRandom[$0.5$] $p{=}1$ & 40.50 & 40.71 & 41.03 & 41.27 & 42.46 & 43.40 & 48.54 & 54.00\\
		\PLRandom[$0.25$] $p{=}1$ & 39.18 & 39.32 & 39.54 & 39.69 & 40.31 & 40.82 & 43.65 & 46.35\\
		\PLRandom[$0.2$] $p{=}1$ & 39.04 & 39.15 & 39.34 & 39.48 & 40.08 & 40.54 & 42.86 & 44.98\\
		\PLRandom[$0.15$] $p{=}1$ & 39.88 & 39.99 & 40.12 & 40.22 & 40.74 & 41.13 & 43.05 & 44.70\\
		\PLRandom[$0.1$] $p{=}1$ & 41.02 & 41.11 & 41.22 & 41.30 & 41.74 & 42.08 & 43.57 & 44.81\\
		\PLRandom[$0.05$] $p{=}1$ & 47.39 & 47.43 & 47.50 & 47.57 & 47.91 & 48.19 & 49.40 & 50.37\\
		\PLRandom[$0.025$] $p{=}1$ & 55.40 & 55.46 & 55.56 & 55.62 & 55.91 & 56.13 & 57.18 & 58.03\\
		\PLRandom[$0.2$] $p{=}1.5$ & 39.07 & 39.17 & 39.30 & 39.45 & 40.06 & 40.57 & 42.82 & 44.92\\
		\PLRandom[$0.15$] $p{=}1.5$ & 39.41 & 39.49 & 39.65 & 39.75 & 40.30 & 40.69 & 42.45 & 44.01\\
		\PLRandom[$0.1$] $p{=}1.5$ & 40.83 & 40.91 & 41.01 & 41.09 & 41.51 & 41.83 & 43.35 & 44.60\\
		\PLRandom[$0.05$] $p{=}1.5$ & 46.67 & 46.74 & 46.83 & 46.92 & 47.26 & 47.51 & 48.64 & 49.57\\
		\PLRandom[$0.025$] $p{=}1.5$ & 55.80 & 55.85 & 55.94 & 56.00 & 56.31 & 56.52 & 57.52 & 58.29\\
		\hline
	\end{tabular}
\end{table*}
\begin{table*}
	\centering
	\caption{\revision{\textbf{Overall Robustness Results on TinyImageNet.} Tabular results corresponding to \figref{fig:supp-summary} for $m = 4$ bits with various $\wmax$ and $p$ across a subset of bit error rates.}}
	\label{tab:supp-summary-tinyimagenet-4}
	\vspace*{-0.2cm}
	\scriptsize
	\begin{tabular}{| l | c | c | c | c | c | c | c | c |}
		\hline
		\multicolumn{9}{|c|}{\bfseries TinyImageNet}\\
		\hline
		Model & \multirow{2}{*}{\begin{tabular}{c}\TE\\in \%\end{tabular}} & \multicolumn{7}{c|}{\RTE in \%, $p$ in \%}\\
		\hline
		&& $0.001$ & $0.005$ & $0.01$ & $0.05$ & $0.1$ & $0.5$ & $1$\\
		\hline
		\hline
		\Quant & 38.02 & 38.51 & 39.28 & 39.93 & 43.18 & 46.64 & 78.29 & 97.87\\
		\Clipping[$0.5$] & 38.27 & 38.81 & 39.55 & 40.12 & 43.12 & 46.12 & 71.40 & 95.41\\
		\Clipping[$0.25$] & 38.02 & 38.50 & 39.14 & 39.64 & 42.14 & 44.48 & 63.22 & 89.22\\
		\Clipping[$0.2$] & 37.92 & 38.33 & 38.92 & 39.41 & 41.53 & 43.47 & 57.80 & 82.42\\
		\Clipping[$0.15$] & 37.89 & 38.21 & 38.64 & 38.93 & 40.54 & 42.03 & 52.60 & 68.99\\
		\Clipping[$0.1$] & 38.47 & 38.78 & 39.19 & 39.53 & 40.89 & 42.09 & 49.32 & 59.30\\
		\Clipping[$0.05$] & 40.29 & 40.44 & 40.65 & 40.82 & 41.57 & 42.22 & 45.38 & 48.83\\
		\Clipping[$0.025$] & 45.62 & 45.72 & 45.86 & 45.97 & 46.43 & 46.80 & 48.51 & 50.03\\
		\PLClipping[$1$] & 37.67 & 38.16 & 38.86 & 39.45 & 42.43 & 45.44 & 72.74 & 96.41\\
		\PLClipping[$0.5$] & 37.66 & 38.00 & 38.47 & 38.86 & 40.59 & 42.14 & 52.20 & 68.11\\
		\PLClipping[$0.25$] & 36.97 & 37.16 & 37.44 & 37.68 & 38.60 & 39.42 & 43.82 & 48.77\\
		\PLClipping[$0.2$] & 37.86 & 38.08 & 38.33 & 38.54 & 39.44 & 40.13 & 43.54 & 46.98\\
		\PLClipping[$0.15$] & 39.13 & 39.28 & 39.47 & 39.62 & 40.23 & 40.74 & 43.06 & 45.39\\
		\PLClipping[$0.1$] & 40.96 & 41.07 & 41.24 & 41.37 & 41.83 & 42.17 & 43.90 & 45.38\\
		\PLClipping[$0.05$] & 46.69 & 46.77 & 46.89 & 46.98 & 47.35 & 47.64 & 48.91 & 50.04\\
		\PLClipping[$0.025$] & 56.09 & 56.17 & 56.25 & 56.33 & 56.68 & 56.97 & 58.14 & 59.17\\
		\hline
		\Random[$1$] $p{=}0.01$ & 37.43 & 37.96 & 38.72 & 39.35 & 42.33 & 45.29 & 70.23 & 94.61\\
		\Random[$0.25$] $p{=}0.01$ & 37.93 & 38.38 & 39.05 & 39.61 & 42.05 & 44.35 & 63.85 & 89.83\\
		\Random[$0.2$] $p{=}0.01$ & 38.04 & 38.47 & 39.04 & 39.48 & 41.53 & 43.48 & 58.21 & 80.65\\
		\Random[$0.15$] $p{=}0.01$ & 38.58 & 38.89 & 39.33 & 39.72 & 41.36 & 42.86 & 53.49 & 70.29\\
		\Random[$0.1$] $p{=}0.01$ & 38.73 & 39.02 & 39.36 & 39.62 & 40.88 & 42.05 & 49.11 & 58.87\\
		\Random[$0.05$] $p{=}0.01$ & 40.20 & 40.41 & 40.63 & 40.81 & 41.57 & 42.20 & 45.24 & 48.34\\
		\Random[$1$] $p{=}0.1$ & 38.69 & 39.13 & 39.80 & 40.35 & 43.10 & 45.57 & 65.75 & 91.82\\
		\Random[$0.25$] $p{=}0.1$ & 38.80 & 39.13 & 39.67 & 40.10 & 42.15 & 44.05 & 58.40 & 81.53\\
		\Random[$0.2$] $p{=}0.1$ & 39.50 & 39.82 & 40.28 & 40.64 & 42.31 & 43.83 & 55.59 & 75.32\\
		\Random[$0.15$] $p{=}0.1$ & 38.40 & 38.75 & 39.19 & 39.51 & 41.12 & 42.58 & 51.74 & 66.26\\
		\Random[$0.1$] $p{=}0.1$ & 39.23 & 39.47 & 39.79 & 40.05 & 41.20 & 42.21 & 48.51 & 56.92\\
		\Random[$0.05$] $p{=}0.1$ & 40.58 & 40.77 & 40.99 & 41.16 & 41.86 & 42.49 & 45.66 & 48.78\\
		\Random[$0.025$] $p{=}0.1$ & 45.29 & 45.41 & 45.55 & 45.66 & 46.05 & 46.42 & 48.09 & 49.65\\
		\Random[$0.2$] $p{=}0.5$ & 42.91 & 43.20 & 43.62 & 43.93 & 45.35 & 46.64 & 53.60 & 63.18\\
		\Random[$0.15$] $p{=}0.5$ & 41.90 & 42.24 & 42.61 & 42.88 & 44.12 & 45.12 & 51.25 & 58.63\\
		\Random[$0.1$] $p{=}0.5$ & 41.30 & 41.57 & 41.89 & 42.14 & 43.22 & 44.12 & 48.75 & 54.28\\
		\Random[$0.05$] $p{=}0.5$ & 41.09 & 41.19 & 41.36 & 41.50 & 42.16 & 42.71 & 45.46 & 48.28\\
		\Random[$0.025$] $p{=}0.5$ & 45.91 & 46.01 & 46.13 & 46.22 & 46.66 & 47.04 & 48.76 & 50.20\\
		\Random[$0.25$] $p{=}1$ & 52.60 & 52.85 & 53.17 & 53.48 & 54.80 & 55.86 & 61.79 & 68.07\\
		\Random[$0.2$] $p{=}1$ & 48.56 & 48.82 & 49.16 & 49.44 & 50.67 & 51.64 & 57.14 & 63.59\\
		\Random[$0.15$] $p{=}1$ & 46.61 & 46.87 & 47.17 & 47.44 & 48.55 & 49.47 & 54.27 & 59.46\\
		\Random[$0.1$] $p{=}1$ & 43.12 & 43.34 & 43.60 & 43.80 & 44.69 & 45.45 & 49.38 & 53.30\\
		\Random[$0.05$] $p{=}1$ & 41.94 & 42.14 & 42.35 & 42.51 & 43.15 & 43.62 & 46.12 & 48.40\\
		\Random[$0.025$] $p{=}1$ & 45.96 & 46.05 & 46.21 & 46.33 & 46.78 & 47.10 & 48.67 & 50.05\\
		\Random[$0.2$] $p{=}1.5$ & 56.26 & 56.47 & 56.80 & 57.05 & 58.22 & 59.19 & 63.92 & 68.31\\
		\Random[$0.15$] $p{=}1.5$ & 51.90 & 52.15 & 52.45 & 52.71 & 53.72 & 54.45 & 58.46 & 62.65\\
		\Random[$0.1$] $p{=}1.5$ & 46.42 & 46.60 & 46.82 & 47.00 & 47.78 & 48.42 & 51.73 & 55.13\\
		\Random[$0.05$] $p{=}1.5$ & 43.46 & 43.60 & 43.78 & 43.91 & 44.44 & 44.88 & 47.09 & 49.19\\
		\Random[$0.025$] $p{=}1.5$ & 46.49 & 46.59 & 46.73 & 46.83 & 47.26 & 47.57 & 49.03 & 50.24\\
		\hline
		\PLRandom[$1$] $p{=}0.01$ & 37.31 & 37.77 & 38.39 & 38.98 & 41.81 & 44.50 & 66.66 & 92.72\\
		\PLRandom[$0.25$] $p{=}0.01$ & 37.83 & 38.01 & 38.31 & 38.54 & 39.55 & 40.40 & 44.67 & 49.54\\
		\PLRandom[$0.2$] $p{=}0.01$ & 37.82 & 38.00 & 38.21 & 38.38 & 39.24 & 39.88 & 43.15 & 46.54\\
		\PLRandom[$0.15$] $p{=}0.01$ & 38.50 & 38.66 & 38.85 & 39.00 & 39.68 & 40.24 & 42.80 & 45.12\\
		\PLRandom[$0.1$] $p{=}0.01$ & 40.56 & 40.64 & 40.77 & 40.88 & 41.33 & 41.69 & 43.40 & 44.97\\
		\PLRandom[$0.05$] $p{=}0.01$ & 46.48 & 46.56 & 46.65 & 46.74 & 47.15 & 47.48 & 48.81 & 49.98\\
		\PLRandom[$0.25$] $p{=}0.1$ & 37.33 & 37.54 & 37.82 & 38.05 & 39.03 & 39.84 & 43.78 & 48.28\\
		\PLRandom[$0.2$] $p{=}0.1$ & 37.91 & 38.12 & 38.38 & 38.57 & 39.41 & 40.05 & 43.27 & 46.51\\
		\PLRandom[$0.15$] $p{=}0.1$ & 38.88 & 39.01 & 39.19 & 39.34 & 39.94 & 40.43 & 42.89 & 45.25\\
		\PLRandom[$0.1$] $p{=}0.1$ & 40.46 & 40.55 & 40.69 & 40.80 & 41.26 & 41.63 & 43.39 & 45.06\\
		\PLRandom[$0.05$] $p{=}0.1$ & 45.97 & 46.04 & 46.16 & 46.25 & 46.63 & 46.95 & 48.31 & 49.45\\
		\PLRandom[$0.025$] $p{=}0.1$ & 55.19 & 55.25 & 55.34 & 55.43 & 55.80 & 56.08 & 57.31 & 58.24\\
		\PLRandom[$0.2$] $p{=}0.5$ & 38.77 & 38.91 & 39.11 & 39.28 & 40.00 & 40.62 & 43.58 & 46.32\\
		\PLRandom[$0.15$] $p{=}0.5$ & 38.81 & 38.96 & 39.13 & 39.27 & 39.84 & 40.34 & 42.69 & 44.96\\
		\PLRandom[$0.1$] $p{=}0.5$ & 40.71 & 40.79 & 40.94 & 41.05 & 41.54 & 41.92 & 43.62 & 45.12\\
		\PLRandom[$0.05$] $p{=}0.5$ & 46.14 & 46.23 & 46.31 & 46.40 & 46.72 & 46.96 & 48.24 & 49.35\\
		\PLRandom[$0.025$] $p{=}0.5$ & 55.52 & 55.61 & 55.73 & 55.82 & 56.15 & 56.42 & 57.60 & 58.58\\
		\PLRandom[$0.25$] $p{=}1$ & 39.21 & 39.37 & 39.56 & 39.73 & 40.45 & 41.05 & 43.99 & 46.97\\
		\PLRandom[$0.2$] $p{=}1$ & 39.38 & 39.53 & 39.74 & 39.89 & 40.52 & 41.06 & 43.57 & 46.00\\
		\PLRandom[$0.15$] $p{=}1$ & 39.61 & 39.71 & 39.82 & 39.96 & 40.51 & 40.99 & 43.17 & 45.07\\
		\PLRandom[$0.1$] $p{=}1$ & 41.42 & 41.52 & 41.67 & 41.78 & 42.23 & 42.58 & 44.11 & 45.49\\
		\PLRandom[$0.05$] $p{=}1$ & 46.47 & 46.56 & 46.66 & 46.75 & 47.10 & 47.38 & 48.64 & 49.73\\
		\PLRandom[$0.025$] $p{=}1$ & 55.64 & 55.72 & 55.84 & 55.92 & 56.25 & 56.53 & 57.68 & 58.52\\
		\PLRandom[$0.2$] $p{=}1.5$ & 39.70 & 39.82 & 39.99 & 40.16 & 40.84 & 41.36 & 43.71 & 45.96\\
		\PLRandom[$0.15$] $p{=}1.5$ & 39.88 & 39.99 & 40.17 & 40.29 & 40.83 & 41.28 & 43.24 & 45.00\\
		\PLRandom[$0.1$] $p{=}1.5$ & 41.49 & 41.57 & 41.69 & 41.81 & 42.27 & 42.66 & 44.29 & 45.63\\
		\PLRandom[$0.05$] $p{=}1.5$ & 47.83 & 47.92 & 48.04 & 48.12 & 48.51 & 48.82 & 50.03 & 51.05\\
		\PLRandom[$0.025$] $p{=}1.5$ & 56.27 & 56.36 & 56.47 & 56.55 & 56.92 & 57.21 & 58.38 & 59.29\\
		\hline
	\end{tabular}
\end{table*}
\begin{table*}
	\centering
	\caption{\revision{\textbf{Overall Robustness Results on TinyImageNet.} Tabular results corresponding to \figref{fig:supp-summary} for $m = 3$ bits with various $\wmax$ and $p$ across a subset of bit error rates.}}
	\label{tab:supp-summary-tinyimagenet-3}
	\vspace*{-0.2cm}
	\scriptsize
	\begin{tabular}{| l | c | c | c | c | c | c | c | c |}
		\hline
		\multicolumn{9}{|c|}{\bfseries TinyImageNet}\\
		\hline
		Model & \multirow{2}{*}{\begin{tabular}{c}\TE\\in \%\end{tabular}} & \multicolumn{7}{c|}{\RTE in \%, $p$ in \%}\\
		\hline
		&& $0.001$ & $0.005$ & $0.01$ & $0.05$ & $0.1$ & $0.5$ & $1$\\
		\hline
		\hline
		\Quant & 45.27 & 45.80 & 46.61 & 47.27 & 50.47 & 53.43 & 75.79 & 94.22\\
		\Clipping[$0.5$] & 48.31 & 48.87 & 49.62 & 50.25 & 53.41 & 56.24 & 82.20 & 98.11\\
		\Clipping[$0.25$] & 44.96 & 45.41 & 45.95 & 46.35 & 48.49 & 50.50 & 64.87 & 83.79\\
		\Clipping[$0.2$] & 43.16 & 43.53 & 44.06 & 44.49 & 46.38 & 47.94 & 58.80 & 74.64\\
		\Clipping[$0.15$] & 42.43 & 42.76 & 43.25 & 43.61 & 45.34 & 46.94 & 57.10 & 72.27\\
		\Clipping[$0.1$] & 40.41 & 40.73 & 41.15 & 41.47 & 42.88 & 44.18 & 51.94 & 63.77\\
		\Clipping[$0.05$] & 41.48 & 41.67 & 41.91 & 42.09 & 42.84 & 43.47 & 46.93 & 50.83\\
		\Clipping[$0.025$] & 46.40 & 46.51 & 46.67 & 46.79 & 47.24 & 47.59 & 49.42 & 51.19\\
		\PLClipping[$1$] & 45.41 & 45.85 & 46.49 & 47.01 & 49.54 & 51.87 & 67.76 & 86.09\\
		\PLClipping[$0.5$] & 39.71 & 40.06 & 40.50 & 40.88 & 42.62 & 44.19 & 54.23 & 68.95\\
		\PLClipping[$0.25$] & 38.52 & 38.71 & 39.02 & 39.23 & 40.21 & 41.07 & 45.53 & 50.71\\
		\PLClipping[$0.2$] & 38.66 & 38.84 & 39.05 & 39.25 & 40.04 & 40.67 & 44.31 & 48.22\\
		\PLClipping[$0.15$] & 39.19 & 39.33 & 39.50 & 39.65 & 40.42 & 41.02 & 43.89 & 46.55\\
		\PLClipping[$0.1$] & 40.21 & 40.29 & 40.45 & 40.58 & 41.10 & 41.54 & 43.65 & 45.54\\
		\PLClipping[$0.05$] & 46.06 & 46.12 & 46.27 & 46.38 & 46.84 & 47.19 & 48.70 & 50.00\\
		\PLClipping[$0.025$] & 56.83 & 56.92 & 57.05 & 57.15 & 57.51 & 57.80 & 59.06 & 60.26\\
		\hline
		\Random[$1$] $p{=}0.01$ & 43.16 & 43.57 & 44.24 & 44.77 & 47.34 & 49.70 & 66.58 & 86.88\\
		\Random[$0.25$] $p{=}0.01$ & 43.77 & 44.14 & 44.71 & 45.19 & 47.37 & 49.28 & 62.06 & 79.70\\
		\Random[$0.2$] $p{=}0.01$ & 43.01 & 43.40 & 43.96 & 44.42 & 46.53 & 48.27 & 63.21 & 84.45\\
		\Random[$0.15$] $p{=}0.01$ & 41.09 & 41.36 & 41.81 & 42.17 & 43.75 & 45.11 & 53.80 & 66.14\\
		\Random[$0.1$] $p{=}0.01$ & 40.14 & 40.47 & 40.84 & 41.15 & 42.66 & 44.03 & 52.66 & 65.34\\
		\Random[$0.05$] $p{=}0.01$ & 40.97 & 41.12 & 41.32 & 41.48 & 42.22 & 42.90 & 46.52 & 50.32\\
		\Random[$1$] $p{=}0.1$ & 50.58 & 51.16 & 51.93 & 52.58 & 55.58 & 58.34 & 77.45 & 94.14\\
		\Random[$0.25$] $p{=}0.1$ & 44.57 & 45.06 & 45.69 & 46.15 & 48.48 & 50.57 & 65.73 & 85.01\\
		\Random[$0.2$] $p{=}0.1$ & 43.04 & 43.39 & 43.81 & 44.13 & 45.74 & 47.24 & 57.10 & 71.58\\
		\Random[$0.15$] $p{=}0.1$ & 42.33 & 42.61 & 42.94 & 43.25 & 44.72 & 46.02 & 53.87 & 65.05\\
		\Random[$0.1$] $p{=}0.1$ & 40.88 & 41.09 & 41.43 & 41.69 & 42.99 & 44.17 & 51.03 & 60.63\\
		\Random[$0.05$] $p{=}0.1$ & 41.06 & 41.24 & 41.48 & 41.68 & 42.52 & 43.20 & 46.54 & 50.29\\
		\Random[$0.025$] $p{=}0.1$ & 45.20 & 45.33 & 45.51 & 45.63 & 46.11 & 46.52 & 48.41 & 50.19\\
		\Random[$0.2$] $p{=}0.5$ & 49.99 & 50.27 & 50.65 & 50.97 & 52.34 & 53.44 & 60.22 & 68.01\\
		\Random[$0.15$] $p{=}0.5$ & 45.63 & 45.87 & 46.25 & 46.52 & 47.91 & 49.04 & 55.21 & 62.93\\
		\Random[$0.1$] $p{=}0.5$ & 42.81 & 43.06 & 43.39 & 43.64 & 44.69 & 45.54 & 50.00 & 55.43\\
		\Random[$0.05$] $p{=}0.5$ & 42.08 & 42.22 & 42.44 & 42.62 & 43.34 & 43.91 & 46.71 & 49.49\\
		\Random[$0.025$] $p{=}0.5$ & 45.57 & 45.68 & 45.83 & 45.93 & 46.41 & 46.75 & 48.58 & 50.20\\
		\Random[$0.25$] $p{=}1$ & 62.68 & 62.92 & 63.26 & 63.55 & 64.89 & 65.98 & 71.30 & 76.29\\
		\Random[$0.2$] $p{=}1$ & 55.94 & 56.20 & 56.54 & 56.81 & 58.06 & 59.12 & 64.39 & 69.83\\
		\Random[$0.15$] $p{=}1$ & 50.59 & 50.84 & 51.13 & 51.36 & 52.39 & 53.21 & 57.57 & 62.18\\
		\Random[$0.1$] $p{=}1$ & 46.22 & 46.43 & 46.66 & 46.86 & 47.79 & 48.59 & 52.52 & 56.88\\
		\Random[$0.05$] $p{=}1$ & 43.83 & 43.95 & 44.13 & 44.26 & 44.90 & 45.44 & 47.85 & 50.32\\
		\Random[$0.025$] $p{=}1$ & 46.72 & 46.79 & 46.92 & 47.04 & 47.53 & 47.93 & 49.63 & 51.22\\
		\Random[$0.2$] $p{=}1.5$ & 62.20 & 62.45 & 62.73 & 62.96 & 63.97 & 64.80 & 68.94 & 72.77\\
		\Random[$0.15$] $p{=}1.5$ & 56.30 & 56.49 & 56.80 & 57.03 & 57.95 & 58.74 & 62.71 & 66.85\\
		\Random[$0.1$] $p{=}1.5$ & 50.19 & 50.36 & 50.62 & 50.83 & 51.70 & 52.39 & 55.87 & 59.43\\
		\Random[$0.05$] $p{=}1.5$ & 44.87 & 45.00 & 45.19 & 45.33 & 45.94 & 46.42 & 48.70 & 50.91\\
		\Random[$0.025$] $p{=}1.5$ & 47.39 & 47.48 & 47.63 & 47.74 & 48.21 & 48.56 & 50.21 & 51.60\\
		\hline
		\PLRandom[$1$] $p{=}0.01$ & 46.46 & 46.94 & 47.59 & 48.14 & 50.90 & 53.47 & 71.01 & 90.51\\
		\PLRandom[$0.5$] $p{=}0.01$ & 40.76 & 41.14 & 41.63 & 42.07 & 43.96 & 45.59 & 56.41 & 72.60\\
		\PLRandom[$0.25$] $p{=}0.01$ & 38.63 & 38.84 & 39.12 & 39.33 & 40.21 & 40.97 & 45.53 & 51.00\\
		\PLRandom[$0.2$] $p{=}0.01$ & 38.84 & 39.02 & 39.27 & 39.47 & 40.34 & 41.08 & 44.80 & 48.69\\
		\PLRandom[$0.15$] $p{=}0.01$ & 39.61 & 39.76 & 39.94 & 40.11 & 40.73 & 41.23 & 43.94 & 46.60\\
		\PLRandom[$0.1$] $p{=}0.01$ & 40.60 & 40.73 & 40.92 & 41.05 & 41.59 & 42.03 & 44.01 & 45.85\\
		\PLRandom[$0.05$] $p{=}0.01$ & 46.88 & 46.97 & 47.10 & 47.21 & 47.60 & 47.94 & 49.38 & 50.60\\
		\PLRandom[$0.5$] $p{=}0.1$ & 41.20 & 41.56 & 42.07 & 42.50 & 44.27 & 45.76 & 55.54 & 69.14\\
		\PLRandom[$0.25$] $p{=}0.1$ & 38.87 & 39.07 & 39.34 & 39.56 & 40.50 & 41.30 & 45.79 & 50.81\\
		\PLRandom[$0.2$] $p{=}0.1$ & 39.41 & 39.58 & 39.85 & 40.05 & 40.86 & 41.59 & 45.15 & 48.89\\
		\PLRandom[$0.1$] $p{=}0.1$ & 40.79 & 40.90 & 41.06 & 41.19 & 41.68 & 42.11 & 44.14 & 45.97\\
		\PLRandom[$0.05$] $p{=}0.1$ & 46.57 & 46.68 & 46.83 & 46.96 & 47.38 & 47.77 & 49.35 & 50.60\\
		\PLRandom[$0.025$] $p{=}0.1$ & 55.76 & 55.85 & 55.98 & 56.08 & 56.48 & 56.79 & 58.14 & 59.29\\
		\PLRandom[$0.2$] $p{=}0.5$ & 38.84 & 39.04 & 39.28 & 39.47 & 40.28 & 40.94 & 44.31 & 47.67\\
		\PLRandom[$0.15$] $p{=}0.5$ & 39.57 & 39.71 & 39.91 & 40.07 & 40.77 & 41.33 & 44.11 & 46.67\\
		\PLRandom[$0.1$] $p{=}0.5$ & 40.86 & 40.95 & 41.09 & 41.22 & 41.76 & 42.18 & 44.18 & 46.03\\
		\PLRandom[$0.05$] $p{=}0.5$ & 46.70 & 46.77 & 46.89 & 46.99 & 47.41 & 47.75 & 49.25 & 50.50\\
		\PLRandom[$0.025$] $p{=}0.5$ & 55.80 & 55.87 & 55.98 & 56.08 & 56.49 & 56.76 & 58.08 & 59.21\\
		\PLRandom[$0.5$] $p{=}1$ & 49.41 & 49.61 & 49.97 & 50.23 & 51.47 & 52.51 & 57.90 & 63.49\\
		\PLRandom[$0.25$] $p{=}1$ & 40.38 & 40.54 & 40.78 & 40.97 & 41.84 & 42.56 & 45.99 & 49.39\\
		\PLRandom[$0.2$] $p{=}1$ & 39.29 & 39.45 & 39.65 & 39.83 & 40.54 & 41.18 & 43.99 & 46.73\\
		\PLRandom[$0.15$] $p{=}1$ & 40.31 & 40.43 & 40.59 & 40.74 & 41.33 & 41.79 & 44.00 & 46.09\\
		\PLRandom[$0.1$] $p{=}1$ & 40.82 & 40.93 & 41.07 & 41.19 & 41.70 & 42.10 & 44.13 & 45.80\\
		\PLRandom[$0.05$] $p{=}1$ & 46.80 & 46.87 & 46.98 & 47.06 & 47.43 & 47.77 & 49.11 & 50.35\\
		\PLRandom[$0.025$] $p{=}1$ & 55.89 & 55.97 & 56.10 & 56.21 & 56.56 & 56.82 & 58.13 & 59.21\\
		\PLRandom[$0.2$] $p{=}1.5$ & 41.23 & 41.39 & 41.58 & 41.74 & 42.42 & 42.99 & 45.64 & 48.11\\
		\PLRandom[$0.05$] $p{=}1.5$ & 46.82 & 46.87 & 46.98 & 47.07 & 47.44 & 47.76 & 49.21 & 50.48\\
		\PLRandom[$0.025$] $p{=}1.5$ & 55.56 & 55.64 & 55.76 & 55.84 & 56.20 & 56.51 & 57.73 & 58.69\\
		\hline
	\end{tabular}
\end{table*}
\begin{table*}
	\centering
	\caption{\revision{\textbf{Overall Robustness Results on TinyImageNet.} Tabular results corresponding to \figref{fig:supp-summary} for $m = 2$ bits with various $\wmax$ and $p$ across a subset of bit error rates.}}
	\label{tab:supp-summary-tinyimagenet-2}
	\vspace*{-0.2cm}
	\footnotesize
	\begin{tabular}{| l | c | c | c | c | c | c | c | c |}
		\hline
		\multicolumn{9}{|c|}{\bfseries TinyImageNet}\\
		\hline
		Model & \multirow{2}{*}{\begin{tabular}{c}\TE\\in \%\end{tabular}} & \multicolumn{7}{c|}{\RTE in \%, $p$ in \%}\\
		\hline
		&& $0.001$ & $0.005$ & $0.01$ & $0.05$ & $0.1$ & $0.5$ & $1$\\
		\hline
		\hline
		\Clipping[$0.05$] & 45.63 & 45.90 & 46.21 & 46.42 & 47.49 & 48.33 & 53.27 & 59.42\\
		\PLClipping[$0.25$] & 44.16 & 44.45 & 44.79 & 45.07 & 46.31 & 47.42 & 53.68 & 61.66\\
		\PLClipping[$0.2$] & 42.99 & 43.19 & 43.47 & 43.70 & 44.75 & 45.62 & 50.38 & 55.58\\
		\PLClipping[$0.15$] & 41.64 & 41.84 & 42.10 & 42.26 & 43.15 & 43.83 & 47.34 & 50.81\\
		\PLClipping[$0.1$] & 41.36 & 41.52 & 41.70 & 41.86 & 42.57 & 43.13 & 45.77 & 48.18\\
		\PLClipping[$0.05$] & 46.26 & 46.35 & 46.48 & 46.58 & 47.07 & 47.50 & 49.43 & 51.08\\
		\hline
		\Random[$0.05$] $p{=}0.01$ & 44.66 & 44.91 & 45.22 & 45.44 & 46.49 & 47.41 & 52.48 & 58.67\\
		\Random[$0.05$] $p{=}0.1$ & 45.88 & 46.08 & 46.34 & 46.55 & 47.52 & 48.33 & 52.81 & 57.91\\
		\Random[$0.025$] $p{=}0.1$ & 46.94 & 47.09 & 47.27 & 47.41 & 48.05 & 48.51 & 50.87 & 53.03\\
		\Random[$0.05$] $p{=}0.5$ & 46.93 & 47.11 & 47.35 & 47.54 & 48.40 & 49.04 & 52.48 & 56.02\\
		\Random[$0.025$] $p{=}0.5$ & 47.82 & 47.96 & 48.12 & 48.26 & 48.88 & 49.31 & 51.46 & 53.41\\
		\Random[$0.05$] $p{=}1$ & 47.62 & 47.78 & 47.99 & 48.13 & 48.91 & 49.54 & 52.52 & 55.53\\
		\Random[$0.025$] $p{=}1$ & 48.84 & 48.98 & 49.14 & 49.28 & 49.84 & 50.28 & 52.23 & 54.02\\
		\Random[$0.05$] $p{=}1.5$ & 49.17 & 49.36 & 49.60 & 49.80 & 50.57 & 51.12 & 53.90 & 56.60\\
		\Random[$0.025$] $p{=}1.5$ & 49.18 & 49.29 & 49.43 & 49.53 & 50.04 & 50.45 & 52.37 & 54.06\\
		\hline
		
		\hline
	\end{tabular}
\end{table*}
}


\vfill


\end{document}